\documentclass{nsr}

\usepackage{amsmath,graphicx,array}
\usepackage{dcolumn,soul}%

\usepackage[figuresright]{rotating}%
\usepackage{algorithm, algorithmicx, algpseudocode}
\usepackage{listings}%
\usepackage{hyperref}
\usepackage{makecell}
\usepackage{subfigure}
\usepackage{booktabs}
\usepackage{cite,multirow,bm,multicol,booktabs,threeparttable,extpfeil}
\usepackage{bbding}
\usepackage{makecell}

\makeatletter
\def\uns{\ifmmode\,\else$\,$\fi}%

\newtheorem{definition}{Definition}

\makeatother

\jvol{}
\jnum{}
\jyear{2026}
\doi{10.1093/nsr/XXXX}
\received{XX XX 2026}
\revised{XX XX 2026}
\accepted{XX XX 2026}

\begin{document}

\dhead{Review}

\subhead{Information Sciences}

\title{EEG-FM-Compass: Progress, Benchmarking, and Future Directions for EEG Foundation Models}

\author{Dingkun Liu$^{1,2,\dagger}$}

\author{Yuheng Chen$^{1,\dagger}$}

\author{Zhu Chen$^{1,\dagger}$}

\author{Zhenyao Cui$^1$}

\author{Yaozhi Wen$^{2,3}$}

\author{Jiayu An$^1$}

\author{Jingwei Luo$^1$}

\author{Dongrui Wu$^{1,2,*}$}

\affil{$^1$Ministry of Education Key Laboratory of Image Processing and Intelligent Control, School of Artificial Intelligence and Automation, Huazhong University of Science and Technology, Wuhan, 430074 China}

\affil{$^2$Zhongguancun Academy, Beijing, 100094, China}

\affil{$^3$State Key Laboratory of Brain Cognition and Brain-inspired Intelligence Technology, Institute of Automation, Chinese Academy of Sciences, Beijing, 100190, China}

\authornote{\textbf{Corresponding author.} Email: drwu09@gmail.com}
\authornote{Equal contribution}

\abstract[ABSTRACT]{Electroencephalography (EEG) foundation models (FMs) have recently emerged as a promising paradigm for brain-computer interfaces, aiming to learn transferable neural representations from large-scale heterogeneous recordings. Despite rapid progress, a fair and comprehensive comparison of existing EEG FMs is still lacking, owing to inconsistent pre-training objectives, preprocessing choices, and downstream evaluation protocols. To fill this gap, we present EEG-FM-Compass. We first review 55 representative models and organize their design choices into a unified taxonomic framework including data standardization, model architectures, and self-supervised pre-training strategies. We then evaluate 12 open source FMs and competitive specialist baselines across 13 EEG datasets spanning nine brain-computer interface paradigms. Emphasizing real-world deployments, we consider both cross-subject generalization under a leave-one-subject-out protocol and rapid calibration under a within-subject few-shot setting. We further compare full-parameter fine-tuning with linear probing to assess the transferability of pre-trained representations, and examine the relationship between model scale and downstream performance. Our results indicate that: 1) linear probing is frequently insufficient; 2) specialist models trained from scratch remain competitive across many tasks; and 3) larger FMs do not necessarily yield better generalization performance under current data regimes and training practices.}



\maketitle

\section{INTRODUCTION}\label{intro}

Brain-computer interfaces (BCIs) establish a direct communication pathway between the brain and external devices, by decoding various brain signals~\cite{nicolas2012brain}. They can be diagnostic and therapeutic tools for a wide range of neurological and psychiatric diseases, e.g., epilepsy \cite{drwuNSR2025}, disorder of consciousness \cite{Li2016}, and mood disorders \cite{drwuPIEEE2023}, and can support communications and interactions for individuals with severe motor or speech impairments~\cite{patel2021artificial} caused by amyotrophic lateral sclerosis, brainstem stroke, high level spinal cord injury, etc.

BCI systems are commonly categorized into invasive, non-invasive, and partially invasive systems. This paper focuses on non-invasive BCIs, which do not require implanting sensors inside the brain. Electroencephalography (EEG), recorded by electrodes placed on the scalp, is one of the most widely used signal modalities in non-invasive BCIs because of its millisecond-level temporal resolution, portability, and relatively low acquisition cost~\cite{kumar2020comparative}.

Classical EEG-based BCI paradigms include motor imagery (MI), steady state visual evoked potentials (SSVEP), event related potentials (ERP), epilepsy recognition, and so on. Recent years have witnessed some emerging paradigms that target higher level cognitive states, such as mental workload~\cite{dehais2020neuroergonomics} and imagined speech~\cite{proix2022imagined,drwuMEG2025}. Together, these developments underscore the potential of EEG as a versatile interface to motor, perceptual, cognitive, and clinical brain states.

Deep learning has driven substantial progress in EEG decoding over the past decade. Early convolutional architectures that remain widely used, such as ShallowConvNet and DeepConvNet~\cite{schirrmeister2017shallowdeep} and EEGNet~\cite{lawhern2018eegnet}, employ compact spatio-temporal filters to learn discriminative EEG patterns from limited labeled data. Beyond these generic backbones, a large body of paradigm-specific decoders has been developed to encode task-dependent inductive biases. For motor imagery, representative models range from filter-bank and spatial-filter-inspired networks, such as MSFBCNN~\cite{wu2019msfbcnn} and FBCNet~\cite{mane2020fbcnet}, to convolutional-transformer hybrids such as EEG Conformer~\cite{song2022conformer} that couple local temporal filtering with global dependency modeling. For ERP, P300, and RSVP target detection, architectures such as HDNN and CNN4EEG~\cite{zhang2018hdnn}, EEG-Inception~\cite{santamaria2020inception}, and MOCNN~\cite{jin2024mocnn} exploit hierarchical convolutions, inception-style temporal filters, and multi-scale feature extraction to improve single-trial event decoding. For emotion recognition and cognitive-state decoding, graph- and attention-based models, including DGCNN~\cite{song2018dgcnn}, RGNN~\cite{zhong2020rgnn}, and LGGNet~\cite{ding2023lggnet}, capture functional relations among EEG channels and brain regions. In clinical EEG analysis, specialized architectures have been tailored to sleep staging, seizure detection, and abnormality assessment, with representative examples including DeepSleepNet~\cite{supratak2017deepsleepnet} for sleep staging and SPaRCNet~\cite{jing2023sparcnet} for seizure and rhythmic / periodic pattern classification. These methods generally outperform classical pipelines that rely on handcrafted features.

Despite these advances, real-world deployment of deep learning models remains challenging. First, most approaches require large amounts of labeled data, whereas EEG acquisition and expert annotation are costly and time consuming. Second, EEG devices differ widely in channel counts and electrode layouts, and conventional architectures often fail to accommodate heterogeneous inputs~\cite{liu2025sdda,drwuAFPM2026}. Third, many existing models are trained for a single task with limited capacity and transferability, which restricts their generalization to new BCI paradigms.

\begin{figure*}[htpb]
\includegraphics[width=\linewidth,clip]{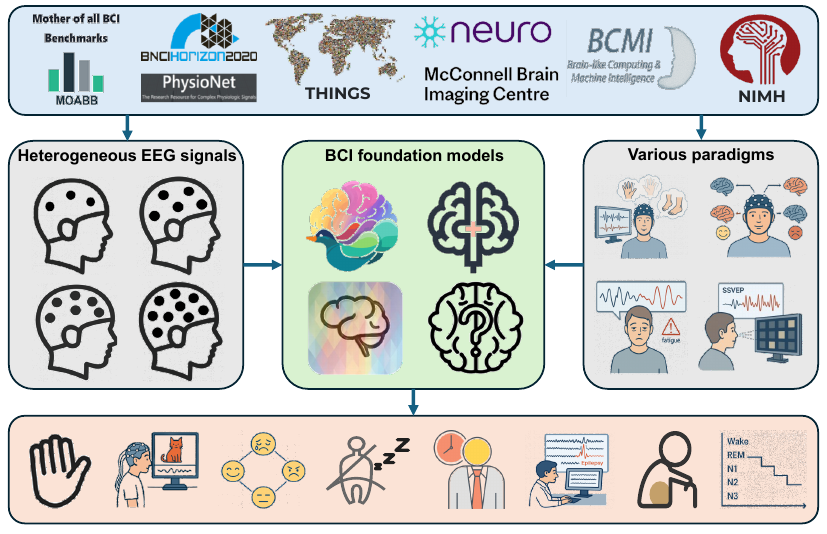}
\caption{Overview of EEG FMs. Models are pre-trained on large scale heterogeneous EEG data collected from devices with diverse electrode configurations across various BCI paradigms. Through self-supervised pre-training, the learned representations may generalize to a wide range of downstream tasks.} \label{fig:FM_scheme}
\end{figure*}

Built on recent progress on large language models (LLMs)~\cite{naveed2025comprehensive} and vision large models (VLMs)~\cite{liu2025toward}, EEG FMs~\cite{yao2025foundation}, as illustrated in Fig.~\ref{fig:FM_scheme}, have emerged as a promising direction for addressing these challenges. The core premise is that a model pre-trained on large-scale heterogeneous EEG data can learn general-purpose representations that transfer effectively to diverse downstream tasks with minimal task-specific adaptation. This paradigm offers a principled solution to the data scarcity problem, as self-supervised pre-training can leverage vast amounts of unlabeled recordings that would otherwise remain unutilized. Furthermore, FMs can be designed to accommodate diverse electrode configurations, enabling a single pre-trained model to generalize across heterogeneous devices.

Numerous EEG FMs were proposed in the past two years, with diverse pre-training objectives, architectures, and target applications. Some models focus on general-purpose representation learning across multiple paradigms, whereas others focus on specific clinical or cognitive applications. Pre-training strategies range from masked signal reconstruction and contrastive learning to codebook-based discrete modeling and autoregressive sequence prediction. Architecturally, these models have evolved from Transformer-based encoders to Mamba-based designs that offer improved efficiency for long sequences. The scale of pre-training data has also increased substantially, with some recent models leveraging thousands of hours of EEG recordings from dozens of public datasets.

Unfortunately, different studies evaluated their proposed models on different datasets using inconsistent protocols, making direct comparison difficult. Moreover, several fundamental questions regarding the capabilities and limitations of these models have not been rigorously examined. These considerations motivate the following three research questions in this paper:

\textbf{Q1.} Can EEG FMs extract generalizable EEG representations, so that they can be easily adapted to various different downstream tasks?

\textbf{Q2.} Do EEG FMs consistently and significantly outperform traditional and deep learning methods trained from scratch using only the fine-tuning data?

\textbf{Q3.} Do EEG FMs exhibit behavior consistent with scaling laws? Specifically, do larger model sizes and greater volumes of pre-training data lead to better generalization performance on downstream BCI tasks?

Our main contributions are:\\
\textbf{1) A comprehensive overview of existing EEG FMs.}
\begin{itemize}
  \item We survey 55 EEG FMs, constituting the most comprehensive collection to date.
  \item We provide a detailed and structured comparison of their technical designs, encompassing basic information, pre-training data scale, preprocessing pipelines, pre-training strategies, and architectural choices.
  \item We propose a unified taxonomic framework for EEG FMs that organizes existing work into a coherent design space.
\end{itemize}

\textbf{2) Fair and comprehensive benchmarking for open source EEG FMs.}
\begin{itemize}
  \item We systematically compare ``full-parameter fine-tuning" with ``classification head fine-tuning" across various models and tasks to assess whether pre-trained encoders provide broadly transferable EEG representations. Beyond the commonly used leave-one-subject-out (LOSO) scenario, we introduce a within-subject few-shot adaptation scenario in which the fine-tuning data volume is approximately 1/20 $\sim$ 1/100 of that typically used in LOSO protocols, as illustrated in Fig.~\ref{fig:method_bench}b and c.
  \item We comprehensively compare traditional machine learning methods, CNN-based models, and Transformer-based models trained from scratch against fine-tuned EEG FMs to evaluate whether conventional approaches remain competitive.
  \item We evaluate EEG FMs of varying parameter sizes pre-trained on diverse datasets to investigate whether a larger model necessarily leads to better generalization performance.
\end{itemize}

The remainder of this paper is organized as follows. Section `Overview of EEG FMs' reviews 55 different EEG FMs. Section `Benchmark of EEG FMs' presents the benchmark. Section `Discussion' points out future research directions. Finally, Section `Conclusion' draws conclusions.

\section{OVERVIEW OF EEG FMs} \label{sect:overview}

\begin{figure*}[!t]
\noindent\hspace*{-\extramargin}%
\begin{minipage}{\typewidth}
\includegraphics[width=\typewidth,clip]{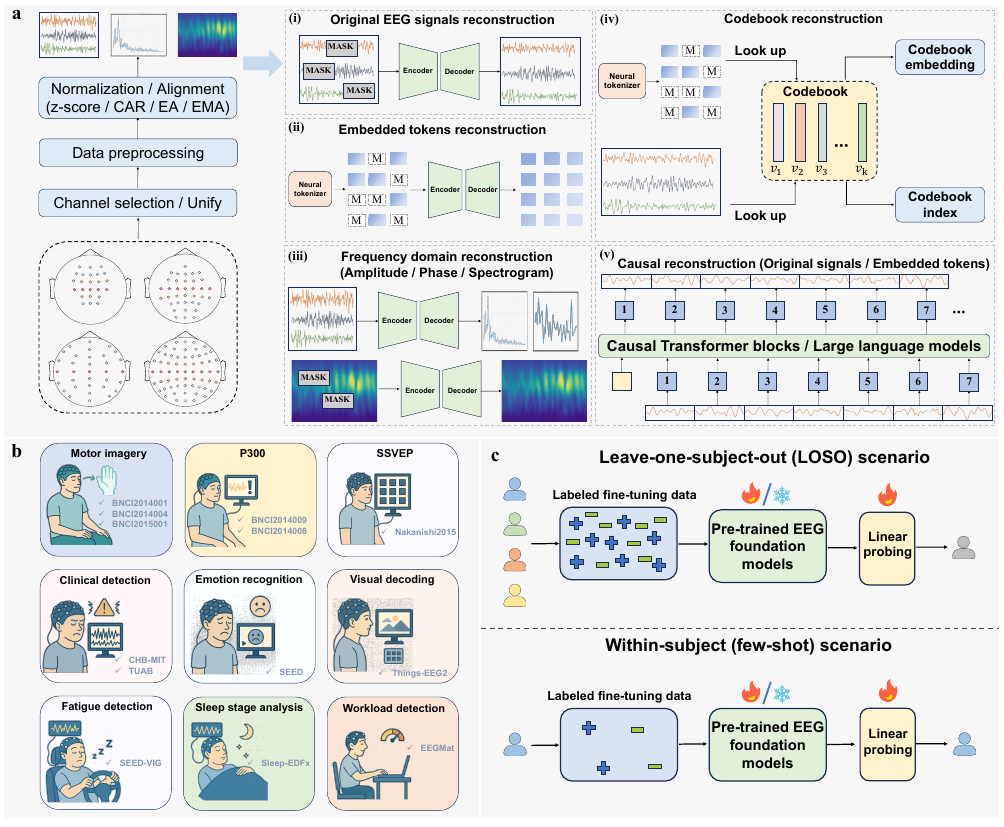}
\caption{Overview of the EEG FM pre-training pipeline and benchmarking scenarios. (a) Raw EEG trials are first standardized through channel selection or unification, followed by dataset-dependent preprocessing and normalization/alignment. The standardized signals are then used for self-supervised pre-training with representative objectives: (i) masked reconstruction of raw EEG signals in the time domain; (ii) masked reconstruction of embedded tokens after tokenization; (iii) frequency-domain reconstruction, where the target can be the spectrogram, spectral amplitude, or phase-related representation; (iv) codebook-based reconstruction, where a tokenizer maps the signal to discrete codebook indices or embeddings and the model learns to predict the corresponding discrete units; and (v) autoregressive or causal reconstruction using causal masking, implemented with causal Transformer blocks or large language models. (b) The 13 downstream datasets spanning 9 representative BCI paradigms, including MI, P300, SSVEP, clinical detection, emotion recognition, visual decoding, fatigue detection, sleep stage analysis, and workload detection. (c) Illustration of the two evaluation scenarios: the LOSO scenario, which aggregates labeled data from multiple subjects for fine-tuning and evaluates on a held-out subject, and the within-subject few-shot scenario, which uses only a small amount of labeled data from the target subject for adaptation.}\label{fig:method_bench}
\end{minipage}
\end{figure*}

This section introduces the conceptual framework of EEG FMs, provides a comprehensive summary of existing approaches, and organizes prevalent pre-training strategies into a unified taxonomic framework, as shown in Fig.~\ref{fig:method_bench}a.


\subsection{Advances and Trends of EEG FMs}

\begin{figure*}[t]
\includegraphics[width=\linewidth,clip]{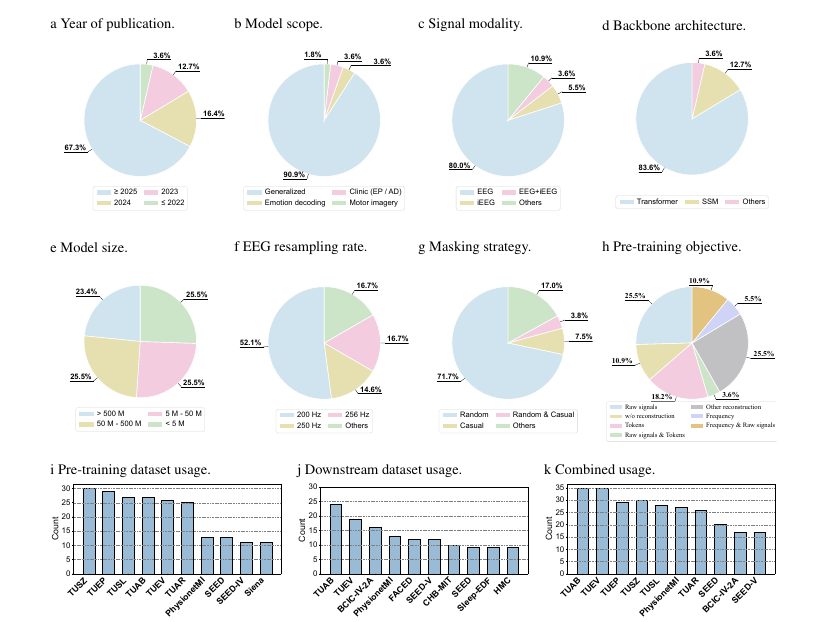}
\caption{Statistical overview of 55 existing EEG FMs. (a-h) Distribution of key model characteristics: year of publication, model scope (generalized vs. paradigm-specific), signal modality, backbone architecture, model size, EEG resampling rate, masking strategy, and pre-training objective. (i-k) Dataset usage statistics: (i) frequency of datasets used during pre-training, (j) frequency of datasets used for downstream evaluation, and (k) combined frequency across both pre-training and downstream stages.} \label{fig:pie_data}
\end{figure*}

Fig.~\ref{fig:pie_data} presents an overview of the 55 existing EEG FMs. As shown in Fig.~\ref{fig:pie_data}a, 16.4\% of the surveyed studies were published in 2024 and 67.3\% in 2025 or 2026, indicating a clear surge in research activity. This accelerated progress is accompanied by increasing diversity in model scope, signal modalities, backbone architectures, and training methodologies. Table~\ref{tab:basic_info} summarizes the 55 surveyed models in chronological order, reporting the affiliation of the first author, publication date, targeted modality, pre-training data scale, computational cost, and parameter size (bold represents open source).

\begin{table*}[htbp]
\centering
\caption{Comparative overview of EEG FMs. `---' indicates that the corresponding information was not reported in the original paper.}
\label{tab:basic_info}
\renewcommand{\arraystretch}{1.3}
\setlength{\tabcolsep}{9pt}
\scalebox{0.65}{
\begin{tabular}{cccccc}
\toprule
Foundation & Author & Signal & Pre-training & Computational & Number of \\
Model & Affiliation & Modalities & Data Size & Cost & Parameters \\
\midrule
BENDR~\cite{kostas2021bendr} & UofT & EEG & 1.5TB & --- & \textbf{4.0M} \\
BrainBERT~\cite{wang2023brainbert} & MIT & iEEG & 43.7h & --- & \textbf{43.2M} \\
MBrain~\cite{cai2023mbrain} & ZJU & iEEG & 470h & 4 $\times$ 3090 & --- \\
BIOT~\cite{yang2023biot} & MIT & EEG + ECG & 58,021h & 8 $\times$ A6000 & \textbf{3.2M} \\
Brant~\cite{zhang2023brant} & ZJU & iEEG & 2,528h & 4 $\times$ A100, 67.2h & 68M / 104M / 249M / \textbf{506M} \\
\addlinespace
LaBraM~\cite{jiang2024labram} & SJTU & EEG & 2,500h & 8 $\times$ A800 & \textbf{5.8M} / 46M / 369M \\
Mentality~\cite{panchavati2025mentality} & UCLA & EEG & --- & --- & --- \\
Neuro-GPT~\cite{cui2024neurogpt} & USC & EEG & 5,656h & --- & \textbf{0.16M} \\
MEET~\cite{shi2023meet} & NPU & EEG (Emotion) & --- & 1 $\times$ 3090 & 30M / 61M / 215M \\
EEGFormer~\cite{chen2024eegformer} & MSR & EEG & 1.7TB & --- & 1.9M / 2.3M / 3.2M / 5.8M \\
\addlinespace
BrainWave~\cite{yuan2024brainwave} & ZJU & EEG + iEEG & 40,907h & 4 $\times$ A100, 100h & 115M / 204M / 459M / 1065M \\
NeuroLM~\cite{jiang2024neurolm} & SJTU & EEG & 25,000h & 8 $\times$ A100 & \textbf{254M} / \textbf{500M} / \textbf{1696M} \\
Brant-X~\cite{zhang2024brantx} & ZJU & EEG + EXG & 4TB & 2 $\times$ A100 & $>$1B \\
FoME~\cite{shi2024fome} & NPU & EEG + iEEG & 26,000h & 6 $\times$ 4090, 350h & 476M / 745M \\
EEGPT~\cite{wang2024eegpt} & HIT & EEG & --- & 8 $\times$ 3090 & 4.7M / \textbf{25M} \\
\addlinespace
BrainGPT~\cite{yue2024braingpt} & CAS & EEG & 37,500K trials & 8 $\times$ A800, 20h & 1.5M / 11.3M / 184M / 1.1B \\
GEFM~\cite{wang2025gefm} & UTokyo & EEG & --- & --- & --- \\
CBraMod~\cite{wang2025cbramod} & ZJU & EEG & 27,062h & 4 $\times$ A5000, 120h & \textbf{4.0M} \\
CEReBrO~\cite{dimofte2025cerebro} & UZH & EEG & $>$20,000h & 4 $\times$ 2080 Ti & 3.58M / 39.95M / 85.15M \\
LEAD~\cite{wang2025lead} & UNCC & EEG (AD) & 730.48h & 4 $\times$ A5000 & --- \\
\addlinespace
FEMBA~\cite{tegon2025femba} & POLIMI & EEG & 21,000h & --- & 7.9M / 47.7M / 77.8M / 389M \\
LCM~\cite{chen2025lcm} & Neuro Industry, Inc. & EEG & --- & --- & 33.9M \\
TFM~\cite{pradeepkumar2025TFM} & UIUC & EEG & $\approx$1,900h & --- & \textbf{1.9M} \\
Gram~\cite{li2025gram} & SJTU & EEG & 7,291h & 8 $\times$ A800, 240h & 6.0M / 47.19M / \textbf{251.28M} \\
ALFEE~\cite{xiong2025alfee} & TJU (Tongji) & EEG & 25,000h & 8 $\times$ A100 & 16M / 44M / 120M / 300M / 540M \\
\addlinespace
BrainOmni~\cite{xiao2025brainomni} & AI Lab & EEG + MEG & 2,653h & 16 $\times$ A100, 18h & \textbf{8.4M} / \textbf{33M} \\
E3GT~\cite{ogg2025e3gt} & JHU & EEG & 26,496h & --- & 96.4M \\
CodeBrain~\cite{ma2025codebrain} & NUS & EEG & 9,246h & A100 & 3.9M -- 146.8M \\
DIVER-0~\cite{han2025diver0} & SNU & EEG & $\approx$3,000h & 4 $\times$ A100 & --- \\
UniMind~\cite{lu2025unimind} & AI Lab & EEG & 929K trials & 8 $\times$ A800, 21.78h & 0.5B / 1.8B / 7B \\
\addlinespace
CSBrain~\cite{zhou2025csbrain} & SHA AI Lab & EEG & 9,000h & 4 $\times$ A100, 101h & 4.9M \\
DMAE-EEG~\cite{zhang2025dmaeeeg} & NUDT & EEG & --- & 8 $\times$ 4090 & --- \\
EEGMamba~\cite{wang2025eegmamba} & ZJU & EEG & 16,724h & 1 $\times$ A5000, 120h & \textbf{3.3M} \\
MIRepNet~\cite{liu2026mirepnet} & HUST & EEG (MI) & 50,355 trials & 1 $\times$ 3090, 3h & \textbf{5.2M} \\
PSGFM~\cite{coon2025psgfm} & JHUAPL & EEG + EXG (Sleep) & 482,270 trials & --- & 97.1M \\
\addlinespace
EEGDM~\cite{puah2025eegdm} & XMUM & EEG & --- & 8 $\times$ 4090 & \textbf{12.8M} \\
CoMET~\cite{li2025comet} & UCAS & EEG & $>$1,000K trials & 4 $\times$ A100 & 5M / 19M / 151M \\
EpilepsyFM~\cite{li2025epilepsyfm} & NPU & EEG (Epilepsy) & --- & 6 $\times$ 4090, 58h & 6.3M \\
SingLEM~\cite{sukhbaatar2025singlem} & TUAT & EEG & 357,000h & 4 $\times$ A100 & \textbf{3.3M} \\
BrainPro~\cite{ding2025brainpro} & NTU & EEG & 2,180h & 5 $\times$ A800 & 7.69M \\
\addlinespace
Uni-NTFM~\cite{chen2025unintfm} & UCAS & EEG & 28,000h & 32 $\times$ A100 & 57M / 427M / 912M / 1.9B \\
ELASTIQ~\cite{jiang2025elastiq} & NTU & EEG & 1,153h & 4 $\times$ H100 & 26.42M \\
BioCodec~\cite{avramidis2025bioCodec} & USC & EEG + EMG & $>$1,000h & 4 $\times$ A100 & 0.3M -- 2.6M \\
HEAR~\cite{chen2025hear} & HKPU & EEG & 8,782h & 8 $\times$ A6000 & 3.1M / 6.0M \\
NeuroRVQ~\cite{barmpas2025neurorvq} & ICL & EEG & --- & 4 $\times$ V100 & 5.9M \\
\addlinespace
NeurIPT~\cite{fang2025neuript} & XMUM & EEG & $>$2,100h & 8 $\times$ 4090 & 73.5M \\
REVE~\cite{el2025reve} & LAB-STICC & EEG & 61,415h & 1 $\times$ A100, 260h & 12M / 69M / 408M \\
mdJPT~\cite{zhang2025mdjpt} & SUSTech & EEG (Emotion) & --- & 1 $\times$ 3090 & 1.0M \\
LUNA~\cite{doner2025luna} & ETH & EEG & 21,928h & 8 $\times$ A100 & \textbf{7.0M} / \textbf{43M} / \textbf{311.4M} \\
THD-BAR~\cite{yang2025thdbar} & BUAA & EEG & 2,123h & 8 $\times$ L40S & 124M / 354M / 1555M \\
\addlinespace
EEG-X~\cite{foumani2025eegx} & Emotiv & EEG & 1,267h & --- & --- \\
SAMBA~\cite{hong2025samba} & Emotiv & EEG & $>$1,000h & 2 $\times$ A6000 Ada & 1.0M \\
DeeperBrain~\cite{wang2026deeperbrain} & ZJU & EEG & $>$17,200h & 1 $\times$ A5000, 7h & --- \\
EEGMoE~\cite{gao2026eegmoe} & UCAS & EEG & 115h & A800 & 1.68M \\
Brain-OF~\cite{guo2026brainof} & FZJ & EEG + MEG + fMRI & 5,870,984 trials & 224 $\times$ A100, 144h & 47.5M / 331M / 1.7B \\
\bottomrule
\end{tabular}}
\end{table*}

Model scope has begun to bifurcate. As shown in Fig.~\ref{fig:pie_data}b, while most studies aim to develop generalized EEG FMs, a nontrivial subset focuses on paradigm-specific FMs. In practical BCI deployment, the target paradigm is often known prior to downstream data collection. Motivated by this observation, paradigm-specific models are pre-trained exclusively on data from a single paradigm to prioritize domain-aligned representation learning, potentially at the expense of cross-paradigm generalizability.

The distribution of signal modalities further reflects this diversity. Fig.~\ref{fig:pie_data}c shows that non-invasive scalp EEG remains the dominant modality, largely because it does not require surgical implantation and is substantially easier to collect at scale than invasive recordings. However, scalp EEG signals are attenuated and spatially blurred due to volume conduction through the scalp and skull, which limits both signal strength and spatial resolution~\cite{burle2015spatial}. To improve robustness and enrich the supervisory signal, several studies incorporate auxiliary physiological modalities such as electrocardiogram (ECG)~\cite{yang2023biot}, electromyogram (EMG)~\cite{zhang2024brantx}, or magnetoencephalography (MEG)~\cite{xiao2025brainomni}, suggesting that representation learning can benefit from correlated biosignals.

From an architectural perspective, Fig.~\ref{fig:pie_data}d indicates that Transformer-based backbones dominate current EEG FMs. Model capacity and training resources, however, exhibit substantial variability rather than a monotonic scaling trend. Fig.~\ref{fig:pie_data}e reveals a wide distribution of parameter scales, and Table~\ref{tab:basic_info} confirms that model sizes range from fewer than one million parameters to several billion parameters.

In summary, EEG FMs have entered a phase of rapid exploration characterized by diverse model scopes, modalities, architectures, and pre-training strategies. However, existing models employed heterogeneous pre-training objectives and were evaluated under diverse downstream scenarios and fine-tuning protocols, which complicates the derivation of consistent conclusions on the factors that truly drive generalization. This observation motivates the unified framework illustrated in Fig.~\ref{fig:method_bench}, which standardizes the major design axes of EEG FMs. The need for systematic comparison further calls for a comprehensive benchmark, which is presented in Section `Benchmark of EEG FMs'.

\subsection{Definition of EEG FMs}

EEG FMs aim to learn transferable and generalizable neural representations from large scale EEG data. In contrast to conventional BCI pipelines that are optimized for a single task and dataset, often through hand crafted features and supervised training from scratch, EEG FMs are typically pre-trained on heterogeneous EEG data collected under different devices and paradigms. After pre-training, they can be adapted to downstream BCI tasks using fine-tuning or prompting, with the expectation of improved generalization and reduced dependence on task specific labels. Due to their versatility, FMs have become a prominent research direction in the BCI community.

A commonly used definition of an AI FM is~\cite{schneider2024foundation}:

\begin{definition}[FM]
A foundation model is any model that is trained on broad data and can be adapted to a wide range of downstream tasks.
\end{definition}

While this definition captures the essence of general purpose FMs, the design of EEG FMs exhibits several domain-specific characteristics:
\begin{enumerate}
    \item[1)] EEG FMs are pre-trained on large scale EEG data, including both scalp EEG and intracranial EEG (iEEG). Additionally, other physiological signals such as electrocardiogram (ECG) and electromyogram (EMG) may serve as auxiliary data during pre-training.

    \item[2)] General-purpose FMs are expected to handle highly heterogeneous downstream tasks, such as semantic segmentation, object detection, long form question answering, and video generation~\cite{awais2025foundation}. In contrast, current EEG FMs primarily target classification tasks across various electrode configurations and BCI paradigms.

    \item[3)] EEG data acquisition is resource-intensive, requiring participant recruitment, paradigm design, and stringent environmental control. As a result, EEG corpora are typically smaller than text or image corpora, and current EEG FMs are often trained with considerably fewer parameters and lower computational budgets.
\end{enumerate}

Based on these domain specific considerations, we propose the following definition for EEG FMs:

\begin{definition}[EEG FM]
An EEG foundation model is pre-trained on large scale electrophysiological data, and can be adapted through fine-tuning or prompting to heterogeneous EEG devices and downstream BCI tasks (categories).
\end{definition}

\subsection{Problem Definition} \label{sect:definition}

Let the pre-training EEG corpus be $\mathcal D_{\mathrm{pre}} = \{ \mathcal D^{(m)} \}_{m=1}^{M}$, where $\mathcal D^{(m)} = \{ X_{i}^{(m)} \}_{i=1}^{N_m}$, $M$ is the number of EEG datasets, and $N_m$ is the number of EEG trials in dataset $m$. Each raw EEG trial is a multi-channel time series $X_{i}^{(m)} \in \mathbb{R}^{C_m \times T_m}$, where $C_m$ is the number of channels and $T_m$ is the number of sampled time points.

Let the downstream task be $\mathcal T=\{\tau_j\}_{j=1}^{J}$, where each task $\tau_j$ specifies a paradigm, a device configuration, and a label space of size $\mathcal C_j$. For each task $\tau_j$, we define a labeled dataset:
\begin{multline}
\label{eq:down_task}
\mathcal D(\tau_j)=\mathcal D_{\mathrm{task}}^{(j)}=\{(X_{k}^{(j)},y_{k}^{(j)})\}_{k=1}^{N_j},\\
 X_{k}^{(j)}\in\mathbb R^{C_j\times T_j}, y_{k}^{(j)}\in\{1,2,\ldots,\mathcal C_j\}.
\end{multline}
We further denote the corpus of all downstream task datasets as
\begin{align}
\mathcal D_{\mathrm{down}}=\{\mathcal D_{\mathrm{task}}^{(j)}\}_{j=1}^{J}.
\label{eq:down_set}
\end{align}

An EEG FM is expected to be pre-trained on $\mathcal D_{\mathrm{pre}}$ and then adapted to each downstream task in $\mathcal T$. Let 
$f_{\Theta}$ denote a pre-trained model with parameters $\Theta$, 
which maps an input trial to a predictive distribution over 
$\mathcal{C}_j$ classes for task $\tau_j$. Let $\mathcal{L}_{\mathrm{pre}}$ 
denote a self-supervised pre-training objective defined on the 
pre-training corpus $\mathcal{D}_{\mathrm{pre}}$. The pre-training stage estimates
\begin{align}
\Theta^\star=\arg\min_{\Theta}\sum_{m=1}^{M}\sum_{i=1}^{N_m}\mathcal L_{\mathrm{pre}}\!\left(X_{i}^{(m)};\Theta\right).
\label{eq:pretrain_obj}
\end{align}

For each downstream task $\tau_j$, we split its labeled dataset into a fine-tuning set and a test set:
\begin{align}
\mathcal D_{\mathrm{task}}^{(j)}=\mathcal D_{\mathrm{ft}}^{(j)}\cup\mathcal D_{\mathrm{te}}^{(j)},
\label{eq:data_split_1}
\end{align}
\begin{align}
\mathcal D_{\mathrm{ft}}^{(j)}\cap\mathcal D_{\mathrm{te}}^{(j)}=\emptyset,
\label{eq:data_split_2}
\end{align}
with
\begin{align}
\mathcal D_{\mathrm{ft}}^{(j)} &=\{(X_{k}^{(j)},y_{k}^{(j)})\}_{k=1}^{n_j}, \\
\mathcal D_{\mathrm{te}}^{(j)} &=\{(X_{k}^{(j)},y_{k}^{(j)})\}_{k=n_j+1}^{N_j}.
\label{eq:data_ft_te}
\end{align}
We expect a pre-trained model could achieve strong generalization performance on $\mathcal D_{\mathrm{te}}^{(j)}$ after fine-tuning on $\mathcal D_{\mathrm{ft}}^{(j)}$.

\subsubsection{Data collection and curation}

Constructing EEG FMs begins with data collection, where the primary sources include public datasets and self-collected recordings. Fig.~\ref{fig:pie_data} presents the frequency of dataset usage in pre-training and downstream evaluation across existing EEG FMs. Most existing approaches aim to develop general-purpose models that accommodate various paradigms, and therefore aggregate large volumes of unlabeled data spanning diverse tasks for pre-training.

However, directly downloaded EEG data often exhibit inconsistent quality, including recordings from poor-performing subjects, corrupted channels, and various artifacts. Curating a large-scale, high-quality dataset is therefore critical for effective FM training~\cite{liu2025clean, wang2025cbramod}. Common curation strategies include subject-level screening to exclude participants with abnormally low task performance or excessive noise, and channel-level screening to remove channels exhibiting persistent artifacts or disconnections. Filtering out such noisy data facilitates more stable optimization and improves the quality of learned representations.

\subsubsection{Data preprocessing} \label{sect:preprocessing}

EEG signals exhibit substantial variability across subjects and devices. Variations in electrode placement, impedance, environmental noise, and physiological state can induce considerable distribution shifts, which may impair large scale pre-training and downstream transfer. Therefore, EEG FM pipelines incorporate preprocessing and normalization to reduce nuisance variability and stabilize optimization. Detailed preprocessing configurations are summarized in Section~A1 of the supplementary information.

To maintain notational consistency throughout this section, we denote a raw EEG trial by $X\in\mathbb R^{C\times T}$ and use the symbol $\widetilde X$ for the model input after preprocessing. Specifically, we define a preprocessing operator
\begin{align}
\widetilde X=\mathcal G(X),
\label{eq:preprocess}
\end{align}
where $\mathcal G$ represents a specific alignment or normalization strategy.

\textit{\textbf{Channel Unification}.}
EEG signals exhibit substantial spatial heterogeneity because different devices adopt different electrode layouts, channel counts, reference schemes, and recording protocols. This heterogeneity makes it difficult to directly reuse conventional task-specific models across datasets. Transformer-based and sequence-based backbones can in principle process variable-length token sequences, but EEG channel tokens are not exchangeable: their meaning depends on electrode location, montage, and neurophysiological context. 
Among the surveyed EEG FMs, channel unification strategies can be summarized into the following categories.

\begin{enumerate}
\item[1)] \textit{Fixed or common-montage modeling.}
A straightforward strategy is to restrict pre-training to a fixed set of channels or to datasets that share a common montage, typically using standard 10--20 clinical EEG configurations. This simplifies cross-dataset alignment but discards recordings or channels that do not match the selected montage and limits transfer to unseen electrode layouts. Representative models include BENDR~\cite{kostas2021bendr}, Mentality~\cite{panchavati2025mentality}, CBraMod~\cite{wang2025cbramod}, CodeBrain~\cite{ma2025codebrain}, and DIVER-0~\cite{han2025diver0}.

\item[2)] \textit{Canonical template or region-level remapping.}
Another strategy maps heterogeneous channels into a predefined template space. This can be implemented by selecting channels that match a canonical montage, interpolating or aligning channels to a standard layout, or grouping electrodes into anatomically motivated brain regions. Compared with fixed-montage modeling, this strategy provides more explicit spatial normalization, but the resulting representation remains constrained by the chosen template or region definition. Representative models include LEAD~\cite{wang2025lead}, MIRepNet~\cite{liu2026mirepnet}, ELASTIQ~\cite{jiang2025elastiq}, DMAE-EEG~\cite{zhang2025dmaeeeg}, EpilepsyFM~\cite{li2025epilepsyfm}, CSBrain~\cite{zhou2025csbrain}, BrainPro~\cite{ding2025brainpro}, and Uni-NTFM~\cite{chen2025unintfm}.

\item[3)] \textit{Channel-wise or single-channel modeling.}
Some models avoid strict montage unification by processing each channel independently or by tokenizing single-channel signals before cross-channel aggregation. This design is naturally robust to missing channels and varying channel counts, because the model can operate on arbitrary subsets of electrodes. 
However, it may require additional mechanisms to recover inter-channel spatial interactions. Representative models include BIOT~\cite{yang2023biot}, BrainGPT~\cite{yue2024braingpt}, SingLEM~\cite{sukhbaatar2025singlem}, TFM~\cite{pradeepkumar2025TFM}, and BioCodec~\cite{avramidis2025bioCodec}.

\item[4)] \textit{Learnable channel embeddings and local spatial encodings.}
Several EEG FMs inject channel identity or local spatial information through learnable channel embeddings, electrode IDs, local electrode convolutions, or positional encodings. This strategy preserves channel-specific information and is easy to integrate with Transformer backbones. However, if the embedding is tied to a fixed channel vocabulary or electrode index, its ability to generalize to unseen layouts is limited. Representative models include LaBraM~\cite{jiang2024labram}, NeuroLM~\cite{jiang2024neurolm}, EEGPT~\cite{wang2024eegpt}, CBraMod~\cite{wang2025cbramod}, and DIVER-0~\cite{han2025diver0}.

\item[5)] \textit{Coordinate-aware spatial encoding.}
A more flexible strategy encodes electrode coordinates or sensor metadata directly. Instead of relying only on channel names or indices, these models use physical electrode positions, pairwise distances, sensor type, or 3D geometry to construct spatially aware representations. This design better supports heterogeneous and previously unseen electrode configurations. Representative models include HEAR~\cite{chen2025hear}, REVE~\cite{el2025reve}, EEG-X~\cite{foumani2025eegx}, SAMBA~\cite{hong2025samba}, DeeperBrain~\cite{wang2026deeperbrain}, NeurIPT~\cite{fang2025neuript}, LUNA~\cite{doner2025luna}, and BrainOmni~\cite{xiao2025brainomni}.

\item[6)] \textit{Projection to a unified latent or source space.}
The most explicit form of channel unification maps heterogeneous channel sets into a fixed latent space before the main backbone. This can be achieved with learnable queries, cross-attention, sensor encoders, retrieval-based spatial modules, or channel encoders. The backbone then operates on a unified representation whose size is decoupled from the original number of electrodes. Representative models include LUNA~\cite{doner2025luna}, BrainOmni~\cite{xiao2025brainomni}, HEAR~\cite{chen2025hear}, ALFEE~\cite{xiong2025alfee}, and BrainPro~\cite{ding2025brainpro}.
\end{enumerate}

\textit{\textbf{Resampling and Bandpass Filtering}.}
Resampling and filtering are widely adopted to standardize temporal resolution and suppress nuisance components in EEG signals. Fig.~\ref{fig:pie_data}f summarizes the resampling choices across the surveyed studies. Specifically, 50.0\% of the studies resample signals to 200~Hz, while 34.1\% resample to 250~Hz or 256~Hz. Bandpass filtering is commonly applied to attenuate slow drifts and high-frequency noise, and notch filters at 50~Hz or 60~Hz are frequently employed to reduce power-line interference.

\textit{\textbf{Normalization and Marginal Alignment}.}
Below we describe several widely adopted normalization or data alignment approaches used in preprocessing.

\begin{enumerate}

\item[1)] \textit{$z$-score Normalization.}
$z$-score normalization rescales each channel to zero-mean and unit variance, ensuring comparable magnitude across channels. This technique is widely employed in the preprocessing stage of EEG FMs. For a trial $X$, the channel-wise statistics are defined as
\begin{align}
\mu_c &=\frac{1}{T}\sum_{t=1}^{T}X_{c,t},\\
\sigma_c &=\sqrt{\frac{1}{T}\sum_{t=1}^{T}(X_{c,t}-\mu_c)^2+\epsilon},
\label{eq:zscore_stat}
\end{align}
and the normalized signal is given by
\begin{align}
\mathcal G_{\mathrm{z}}(X)_{c,t}=\frac{X_{c,t}-\mu_c}{\sigma_c},
\label{eq:zscore_norm}
\end{align}
where $\epsilon>0$ is a small constant for numerical stability. Depending on the protocol, the statistics can be computed per trial, per session, or over the entire training set. Representative FMs that adopt $z$-score normalization include BrainBERT~\cite{wang2023brainbert}, Neuro-GPT~\cite{cui2024neurogpt}, Brant-X~\cite{zhang2024brantx}, EEGPT~\cite{wang2024eegpt}, BrainGPT~\cite{yue2024braingpt}, LEAD~\cite{wang2025lead}, ALFEE~\cite{xiong2025alfee}, BrainOmni~\cite{xiao2025brainomni}, UniMind~\cite{lu2025unimind}, EpilepsyFM~\cite{li2025epilepsyfm}, BioCodec~\cite{avramidis2025bioCodec}, REVE~\cite{el2025reve}, and LUNA~\cite{doner2025luna}.

\item[2)] \textit{Common Average Reference (CAR).}
Another standard preprocessing technique is CAR, which suppresses common mode activity shared across all channels. Let $\mathbf 1_C\in\mathbb R^{C}$ denote an all-ones vector. CAR transforms $X$ as follows:
\begin{align}
\mathcal G_{\mathrm{car}}(X)=X-\frac{1}{C}\mathbf 1_C\mathbf 1_C^\top X.
\label{eq:car}
\end{align}
The underlying assumption of CAR is that signals recorded at all electrodes contain a common noise component, such as reference electrode drift or environmental interference. By subtracting the instantaneous average across all electrodes from each individual electrode, CAR effectively attenuates this common mode component while preserving spatially localized neural activity. Representative FMs employing CAR include EEGPT~\cite{wang2024eegpt}, E3GT~\cite{ogg2025e3gt}, HEAR~\cite{chen2025hear}, and mdJPT~\cite{zhang2025mdjpt}.

\item[3)] \textit{Euclidean Alignment (EA).}
EA \cite{he2019transfer,drwuEA2025} performs subject-wise or session-wise whitening to reduce covariance shifts and improve cross-subject consistency. Assume a subject contains $n$ trials $\{X_i\}_{i=1}^{n}$, where each $X_i\in\mathbb R^{C\times T}$. EA first computes the mean covariance matrix as
\begin{align}
\bar R=\frac{1}{n}\sum_{i=1}^{n}X_iX_i^\top,
\label{eq:ea_cov}
\end{align}
and then applies the whitening transformation
\begin{align}
\mathcal G_{\mathrm{ea}}(X_i)=\bar R^{-1/2}X_i.
\label{eq:ea_trans}
\end{align}
After this transformation, the mean covariance of the aligned trials becomes the identity matrix, thereby reducing discrepancies in second-order statistics across subjects. Representative FMs adopting EA include EEGPT~\cite{wang2024eegpt} and MIRepNet~\cite{liu2026mirepnet}.

\item[4)] \textit{Exponential Moving Average (EMA) Normalization.}
To handle gradual drift in long recordings, some approaches adopt exponential moving average normalization, in which normalization statistics are updated sequentially. Let $x_t\in\mathbb R^{C}$ denote the multichannel sample at time $t$. EMA maintains exponentially decaying estimates of the first and second moments as follows:
\begin{align}
m_t &= \alpha m_{t-1} + (1-\alpha) x_t, \label{eq:ema_mean}\\
s_t &= \alpha s_{t-1} + (1-\alpha) x_t \odot x_t, \label{eq:ema_second}
\end{align}
where $\alpha\in(0,1)$ is the decay factor and $\odot$ denotes the element-wise product. The variance estimate is given by $v_t = s_t - m_t \odot m_t$, and the normalized sample is computed as
\begin{align}
\mathcal G_{\mathrm{ema}}(x)_t = \frac{x_t - m_t}{\sqrt{v_t + \epsilon}}.
\label{eq:ema_norm}
\end{align}
EMA normalization is particularly suitable for online or streaming settings, as it does not require precomputed global statistics and can adapt to non-stationarities in the signal. A representative FM employing EMA normalization is FoME~\cite{shi2024fome}.

\item[5)] \textit{Summary.}
$z$-score normalization, CAR, and EMA normalization are widely adopted as generic preprocessing components, whereas EA provides an explicit mechanism to reduce session-level covariance shifts. In practice, $\mathcal G$ can be instantiated as a composition of these operations. For notational simplicity, we use the unified notation $\widetilde X=\mathcal G(X)$ to denote the preprocessed model input throughout the remainder of this section.
\end{enumerate}

\textit{\textbf{Data Augmentation}.} 
Although data augmentation is widely adopted in supervised EEG learning, its role in EEG FM pre-training remains relatively under-explored. Among the surveyed models, only a few explicitly incorporate augmentation-like mechanisms. Specifically, BIOT~\cite{yang2023biot}, LEAD~\cite{wang2025lead}, and CoMET~\cite{li2025comet} adopt contrastive learning with perturbed views; PSGFM~\cite{coon2025psgfm} enlarges the effective pre-training corpus through overlapping temporal windows; Uni-NTFM~\cite{chen2025unintfm} applies signal-level perturbations including Gaussian noise, channel dropout, and temporal shifts; and REVE~\cite{el2025reve} injects Gaussian noise into electrode coordinates to improve robustness against spatial variability. These observations indicate that augmentation strategies for EEG FMs remain sparse and heterogeneous, and that the systematic design of physiologically grounded augmentation policies constitutes a promising direction for future research. Such policies could leverage neurophysiological priors specific to EEG signals, including oscillatory structure, spatial topography, and inter-trial variability, to introduce meaningful invariances while preserving task-relevant information.

\subsubsection{Model pre-training}

Most EEG FMs are pre-trained with self-supervised objectives that remove or corrupt part of the input and require the model to recover the masked information. Section~A1 of the supplementary information summarizes the pre-training strategies of 55 FMs, and Fig.~\ref{fig:pie_data} highlights eight empirical trends across these models. Several insights can be drawn from this analysis. First, a Transformer-based backbone is adopted by approximately $83.6\%$ of the models. Second, masked reconstruction constitutes the dominant pre-training paradigm; among the masking strategies, random masking is the most prevalent, accounting for approximately $71.7\%$, while causal masking and mixed masking together occupy a smaller portion. Third, regarding the reconstruction target, raw-signal reconstruction is the most frequent strategy, accounting for approximately $25.5\%$, while token reconstruction and hybrid approaches combining raw and token reconstruction together constitute a comparable fraction. Codebook-based and frequency-domain objectives appear less frequently as standalone targets, but are often employed as auxiliary supervision in multi-target designs. Based on these observations, we organize the mainstream pre-training strategies into five categories: masked reconstruction of raw signals, masked reconstruction of embedded tokens, frequency-domain reconstruction, codebook-based objectives, and autoregressive pre-training.

Given a raw EEG trial $X\in\mathbb R^{C\times T}$, we denote the model input after preprocessing by $\widetilde X=\mathcal G(X)$. Since EEG is typically sampled at a high temporal resolution, most EEG FMs further aggregate time steps into patches to reduce the sequence length and capture local temporal structure. Let the patch length be $M$ and the stride be $S$, with overlap $O=M-S$. We segment $\widetilde X$ along the temporal axis into $N_p$ patches, where
\begin{align}
N_p=\left\lfloor \frac{T-M}{S}\right\rfloor +1,
\label{eq:patch_num}
\end{align}
and denote the resulting patch tensor by
\begin{align}
P=\mathcal S(\widetilde X)\in\mathbb R^{N_p\times C\times M},
\label{eq:patch_tensor}
\end{align}
where $\mathcal S$ denotes the patching operator. The $k$-th patch of channel $c$ is denoted by $p_{k,c}\in\mathbb R^{M}$.

For notational clarity, the superscripts used in Section~A1 of the supplementary information indicate the target space: raw signals ($rs$), embedded tokens ($et$ or $emb$), spectrograms ($spec$), spectral amplitude ($sa$), band power ($bp$), codebook indices ($ci$), denoised raw signals ($nrs$), and diffusion velocity ($v$).

\paragraph{Masked Reconstruction of Raw Signals}
Raw signal reconstruction is the most prevalent pre-training strategy, employed by representative models such as Brant~\cite{zhang2023brant}, FoME~\cite{shi2024fome}, CBraMod~\cite{wang2025cbramod}, CSBrain~\cite{zhou2025csbrain}, EEGMamba~\cite{wang2025eegmamba}, BrainPro~\cite{ding2025brainpro}, REVE~\cite{el2025reve}, and EEG-X~\cite{foumani2025eegx}. In this strategy, masking is applied directly to the patched raw signal,
\begin{align}
P_{\mathrm{msk}}=\mathcal M_x(P),
\label{eq:mask_raw}
\end{align}
and the encoder consumes the masked patches. The model is trained to reconstruct the original patches using the mean squared error (MSE) loss:
\begin{align}
\widehat{\widetilde X} &=\mathcal D_{\phi}\!\left(\mathcal E_{\theta}(P_{\mathrm{msk}})\right), \\
\mathcal L_{\mathrm{mse}}^{\mathrm{rs}} &=\bigl\lVert \widehat{\widetilde X}-\widetilde X \bigr\rVert_2^2.
\label{eq:loss_raw}
\end{align}
This approach directly constrains the encoder to preserve the waveform structure and the cross-channel dependencies that are critical for event-related components and oscillatory bursts. However, non-invasive EEG typically exhibits a low signal-to-noise ratio and contains substantial nuisance variability arising from artifacts, impedance fluctuations, and background activity. When the pre-training target is the waveform itself, a model with sufficient capacity may expend representational power on reconstructing idiosyncratic noise patterns that are not predictive for downstream tasks.

To mitigate this issue, several models replace or augment the $\mathcal L_{\mathrm{mse}}^{\mathrm{rs}}$ with more robust variants. For example, FEMBA employs the smooth $\ell_1$ loss $\mathcal L_{s\text{-}l1}^{\mathrm{rs}}$, BrainPro uses a weighted variant $\mathcal L_{w\text{-}\mathrm{mse}}^{\mathrm{rs}}$ together with a decomposition loss $\mathcal L_{\mathrm{dec}}$, and REVE incorporates an auxiliary loss $\mathcal L_{\mathrm{aux}}$. EEG-X further emphasizes denoising by reconstructing noise-removed signals via $\mathcal L_{\mathrm{mse}}^{\mathrm{nrs}}$ and incorporates teacher-student distillation $\mathcal L_{\mathrm{kd}}^{\mathrm{tea}\text{-}\mathrm{stu}}$, which aligns with the practical need to suppress artifacts rather than reproduce them. Denoising variants like EEG-X replace the noisy target $p_{k,c}$ with a cleaned or artifact-suppressed target $p_{k,c}^{den}$:
\begin{equation}
\mathcal L_{\mathrm{mse}}^{nrs}
=
\frac{1}{|\Omega|}
\sum_{(k,c)\in\Omega}
\left\|
\widehat p_{k,c}-p_{k,c}^{den}
\right\|_2^2 .
\label{eq:loss_denoised}
\end{equation}
These robust and denoising variants are particularly relevant for non-invasive EEG, where direct waveform regression may encourage the model to reproduce artifacts and recording-specific noise rather than task-relevant neural activity.

In summary, raw-signal reconstruction serves as a strong baseline when the data quality is controlled and the masking is sufficiently challenging, but it may benefit from explicit regularization that discourages the memorization of noise.

\paragraph{Masked Reconstruction of Embedded Tokens}
Instead of reconstructing waveforms, another group of EEG FMs reconstructs latent representations produced by a tokenizer, convolutional embedder, or patch embedding module. This strategy is adopted by BENDR~\cite{kostas2021bendr}, BIOT~\cite{yang2023biot}, GEFM~\cite{wang2025gefm}, CEReBrO~\cite{dimofte2025cerebro}, LCM~\cite{chen2025lcm}, LUNA~\cite{doner2025luna}, ELASTIQ~\cite{jiang2025elastiq}, MIRepNet~\cite{liu2026mirepnet}, and EEGMoE~\cite{gao2026eegmoe}.

For token-based pre-training, each patch is first mapped to an embedding through a tokenizer:
\begin{align}
Z=\mathcal T_{\psi}(P),
\qquad
Z\in\mathbb R^{N_p\times d},
\label{eq:tokenizer}
\end{align}
where $d$ denotes the embedding dimension. Masking is then applied in the token space:
\begin{align}
Z_{\mathrm{msk}}=\mathcal M_z(Z).
\label{eq:mask_token}
\end{align}
The model learns to predict the original token embeddings:
\begin{align}
\widehat Z=\mathcal D_{\phi}\!\left(\mathcal E_{\theta}(Z_{\mathrm{msk}})\right),
\qquad
\mathcal L_{\mathrm{mse}}^{\mathrm{et}}=\bigl\lVert \widehat Z-Z \bigr\rVert_2^2.
\label{eq:loss_token}
\end{align}

This embedded-token reconstruction loss, denoted as $\mathcal L_{\mathrm{mse}}^{et}$ or $\mathcal L_{\mathrm{mse}}^{emb}$ in Section~A1 of the supplementary information, is used by CEReBrO~\cite{dimofte2025cerebro} for masked token reconstruction, by LCM~\cite{chen2025lcm} as a reconstruction term combined with contrastive learning, and by MIRepNet~\cite{liu2026mirepnet} as part of its hybrid pre-training objective together with a supervised MI classification loss. EEGMoE~\cite{gao2026eegmoe} adopts an $\ell_1$ variant, denoted as $\mathcal L_{\ell_1}^{et}$, to learn domain-decoupled token representations with additional MoE regularization.

Compared to raw signal reconstruction, token-level objectives aim to reduce sensitivity to amplitude scaling and local waveform noise, as the tokenizer compresses the input into a representation that can be designed to emphasize spatio-temporal structure. This design often improves optimization stability for large encoders and naturally supports patch-based processing, which is consistent with the high prevalence of patching observed in Fig.~\ref{fig:pie_data}.

Contrastive learning constitutes a complementary latent-space objective, in which a representation is contrasted against semantically related and unrelated counterparts, which is used by BENDR~\cite{kostas2021bendr}, BIOT~\cite{yang2023biot}, GEFM~\cite{wang2025gefm}, LCM~\cite{chen2025lcm}, Brant-X~\cite{zhang2024brantx}, LEAD~\cite{wang2025lead}, MBrain~\cite{cai2023mbrain}, and CoMET~\cite{li2025comet}. Given an anchor representation $h_i$, a positive representation $h_i^+$, a set of negatives $\mathcal N_i$, a similarity function $\operatorname{sim}(\cdot,\cdot)$, and a temperature $\tau$, a generic InfoNCE-style loss is
\begin{equation}
s_i^+ = \frac{\operatorname{sim}(h_i,h_i^+)}{\tau},
\qquad
s_{ij}^- = \frac{\operatorname{sim}(h_i,h_j)}{\tau}.
\label{eq:contrastive_scores}
\end{equation}

\begin{equation}
\mathcal L_{\mathrm{cl}}
=
-\frac{1}{B}
\sum_{i=1}^{B}
\log
\frac{
\exp(s_i^+)
}{
\exp(s_i^+)
+
\sum_{h_j\in\mathcal N_i}
\exp(s_{ij}^-)
}.
\label{eq:loss_infonce}
\end{equation}

The construction of positive--negative pairs varies considerably across models. BENDR and GEFM apply contrastive prediction over latent EEG tokens, denoted $\mathcal L_{\mathrm{cl}}^{\mathrm{et}}$; BIOT contrasts channel-wise biosignal tokens; and CoMET augments masked reconstruction with a global contrastive term $\mathcal L_{\mathrm{cl}}^{\mathrm{glob}}$ to reinforce global discriminative structure. Brant-X performs two-level contrastive alignment, denoted $\mathcal L_{\mathrm{cl}}^{\mathrm{coarse}}$ and $\mathcal L_{\mathrm{cl}}^{\mathrm{fine}}$, to align EEG with auxiliary physiological signals at different semantic scales, whereas MBrain combines multi-channel contrastive predictive coding with time-shift prediction to capture spatial and temporal dependencies. LEAD employs contrastive pre-training to learn AD-relevant EEG representations.

Compared with raw-signal reconstruction, latent-token reconstruction and contrastive learning reduce direct dependence on local waveform amplitude and can improve robustness to nuisance variability. However, their effectiveness depends strongly on the tokenizer, augmentation scheme, and positive-negative pair construction. If the tokenizer is too shallow, the latent reconstruction target may remain close to waveform regression; if it is too coarse, transient or task-relevant EEG patterns may be discarded. 

Several models address this tension by combining token-level objectives with auxiliary terms. CEReBrO incorporates $\mathcal L_{\mathrm{aux}}$ to enrich the learning signal. LCM combines contrastive learning $\mathcal L_{\mathrm{cl}}^{\mathrm{emb}}$ with a weighted reconstruction term $\lambda\mathcal L_{\mathrm{mse}}^{\mathrm{et}}$. MIRepNet integrates $\mathcal L_{\mathrm{mse}}^{\mathrm{et}}$ with a classification loss $\mathcal L_{\mathrm{cls}}$ to bias the representation toward discriminative structure relevant to its target paradigm. These examples suggest that token reconstruction often benefits from complementary objectives that encourage global semantic learning rather than purely local fidelity.

\paragraph{Frequency-Domain Reconstruction}
Frequency-domain objectives supervise EEG FMs with spectral targets rather than, or in addition to, time-domain waveforms. 
This strategy is used by BrainBERT and BrainWave for spectrogram or time-frequency reconstruction, Mentality for joint raw-signal and spectrogram reconstruction, EEGFormer for spectral-amplitude reconstruction, Uni-NTFM for frequency / band-power supervision, Gram for spectral mimic learning with codebook classification, E3GT and PSGFM for spectrogram-derived pseudo-label prediction, HEAR for codebook-embedding and frequency reconstruction, BioCodec for STFT-based codec reconstruction, and Brain-OF for joint raw-signal and spectral-amplitude reconstruction.

Let $\mathcal F_r$ denote a spectral transform and
\begin{equation}
S^r=\mathcal F_r(P)
\end{equation}
be the spectral target of EEG patches $P$, where $r$ denotes the target type. 
For regression-based spectral supervision, the generic masked spectral reconstruction loss is
\begin{equation}
\begin{aligned}
\mathcal L_{\mathrm{mse}}^{r}
&=
\frac{1}{|\Omega|}
\sum_{(k,c)\in\Omega}
\left\|
\widehat S_{k,c}^{r}-S_{k,c}^{r}
\right\|_2^2, \\
&\qquad
r\in\{\mathrm{spec},\mathrm{sa},\mathrm{bp},\mathrm{freq}\}.
\end{aligned}
\label{eq:loss_frequency_mse}
\end{equation}
Here, $\mathcal L_{\mathrm{mse}}^{spec}$ is used by BrainBERT and Mentality; $\mathcal L_{\mathrm{mse}}^{sa}$ is used by EEGFormer; $\mathcal L_{\mathrm{mse}}^{bp}$ is used by Uni-NTFM; and $\mathcal L_{\mathrm{mse}}^{freq}$ is used by HEAR and SAMBA as a frequency-domain reconstruction term. 
Brain-OF uses a robust smooth-$\ell_1$ variant for spectral-amplitude reconstruction, denoted by $\mathcal L_{s\text{-}l1}^{sa}$.

Some models use discrete spectral pseudo-labels instead of continuous spectral regression. 
E3GT and PSGFM follow a HuBERT-style design, where spectrogram-derived $k$-means labels are treated as prediction targets. 
If $q_{k,c}$ denotes the spectral k-means label of patch $(k,c)$ and $\widehat{\pi}_{k,c}$ denotes the predicted categorical distribution, the corresponding loss is
\begin{equation}
\mathcal L_{\mathrm{cls}}^{skl}
=
-\frac{1}{|\Omega|}
\sum_{(k,c)\in\Omega}
\log \widehat{\pi}_{k,c,q_{k,c}},
\label{eq:loss_spectral_kmeans}
\end{equation}
where $\mathcal L_{\mathrm{cls}}^{skl}$ refers to spectral pseudo-label classification, as used by E3GT and PSGFM.

Several models combine spectral supervision with other reconstruction or auxiliary terms. 
EEGFormer combines spectral-amplitude reconstruction with codebook-embedding reconstruction, denoted as $\mathcal L_{\mathrm{mse}}^{sa}+\mathcal L_{\mathrm{mse}}^{cbe}$. 
Gram combines spectral-amplitude mimic learning with codebook-index classification, denoted as $\mathcal L_{\mathrm{mse}}^{sa}+\mathcal L_{\mathrm{cls}}^{ci}$. 
HEAR combines codebook-embedding reconstruction and frequency-domain reconstruction, denoted as $\mathcal L_{\mathrm{mse}}^{ce}+\mathcal L_{\mathrm{mse}}^{freq}$, where the superscript $ce$ refers to codebook embedding rather than cross-entropy. 
BioCodec combines a Huber-style raw-signal reconstruction term with multi-scale STFT losses and an auxiliary codec regularizer, summarized as $\mathcal L_{huber}^{rs}+\mathcal L_{\ell_1}^{stft}+\mathcal L_{\ell_2}^{stft}+\mathcal L_{aux}$. 
SAMBA combines time-domain and frequency-domain reconstruction, denoted as $\mathcal L_{\mathrm{mae}}^{os}+\mathcal L_{\mathrm{mse}}^{freq}$. 
ALFEE combines raw-signal reconstruction, PSD-enhanced reconstruction, and downstream-task prediction, denoted as $\mathcal L_{\mathrm{mse}}^{rs}+\mathcal L_{\mathrm{mse}}^{rs\,\oplus\,PSD}+\mathcal L_{\mathrm{cls}}^{dt}$. 
These designs indicate that frequency-domain supervision is usually used to emphasize rhythmic EEG structure, while time-domain or auxiliary losses are often retained to preserve waveform fidelity and task-relevant information.

\paragraph{Codebook-Based Objectives}

Codebook-based pre-training discretizes continuous EEG representations into neural tokens, which can then be predicted as code indices or reconstructed as codebook embeddings. This strategy is adopted by LaBraM~\cite{jiang2024labram}, BrainOmni~\cite{xiao2025brainomni}, CodeBrain~\cite{ma2025codebrain}, EpilepsyFM~\cite{li2025epilepsyfm}, NeuroRVQ~\cite{barmpas2025neurorvq}, NeuroLM~\cite{jiang2024neurolm}, and THD-BAR~\cite{yang2025thdbar}. 

Let a quantizer map embeddings to discrete indices:
\begin{equation}
I=\mathcal Q(Z),
\qquad
I\in\{1,2,\ldots,K\}^{L},
\label{eq:quantizer}
\end{equation}
where $K$ denotes the codebook size and $L$ denotes the sequence length. For masked code prediction, the model estimates a categorical distribution $\widehat P\in[0,1]^{L\times K}$ over the codebook and minimizes the cross-entropy loss:
\begin{equation}
\mathcal L_{\mathrm{cls}}^{ci}=\sum_{\ell=1}^{L}\mathcal L_{\mathrm{ce}}\!\left(I_{\ell},\widehat P_{\ell}\right).
\label{eq:loss_codebook_cls}
\end{equation}
Autoregressive index modeling employs negative log-likelihood objectives such as $\mathcal L_{\mathrm{nll}}^{ci}$, which is adopted by NeuroLM and THD-BAR. An alternative branch aligns predicted embeddings with codebook embeddings, represented by $\mathcal{L}_{mse}^{cbe}$ or related terms, as employed by HEAR~\cite{chen2025hear}.

The advantage of codebook-based objectives is that discretization provides a compact symbolic representation of EEG, which can reduce the influence of low-level waveform noise and enable large-scale sequence modeling. It also facilitates causal generation and prompt-style adaptation when combined with decoder-only architectures. However, codebook learning introduces additional design choices, including codebook size, commitment regularization, residual quantization depth, and update schedules. If these factors are not carefully controlled, the learned codebook may collapse or show imbalanced usage, thereby weakening the learned representations.

\paragraph{Autoregressive Pre-training}
Autoregressive pre-training imposes a causal factorization over an ordered EEG token sequence. It is used by Neuro-GPT~\cite{cui2024neurogpt}, BrainGPT~\cite{yue2024braingpt}, NeuroLM~\cite{jiang2024neurolm}, and THD-BAR~\cite{yang2025thdbar}. Given an ordered token sequence $u_{1:L}$, where $u_{\ell}$ can represent an EEG chunk, an embedded token, or a codebook index, the generic autoregressive loss is
\begin{equation}
\mathcal L_{\mathrm{ar}}
=
-\sum_{\ell=1}^{L}
\log p_{\Theta}
\left(
u_{\ell}\mid u_{1:\ell-1}
\right).
\label{eq:autoregressive}
\end{equation}
For continuous EEG tokens, as in Neuro-GPT and BrainGPT, this objective is typically implemented through regression or next-token prediction over EEG chunks or embeddings. 
For discrete neural codes, as in NeuroLM, Eq.~\ref{eq:autoregressive} corresponds to the negative log-likelihood objective $\mathcal L_{\mathrm{nll}}^{ci}$.

Autoregressive modeling is attractive because it aligns naturally with decoder-only Transformer architectures and supports sequence continuation, generative modeling, and prompt-based adaptation; it also provides a principled means of modeling temporal dependencies in EEG. Nevertheless, causal objectives can be more difficult to optimize than bidirectional masked reconstruction, particularly when the EEG tokens are high-dimensional or when their ordering does not faithfully reflect the underlying neurophysiological structure. A purely temporal ordering may further bias the model toward local continuity rather than task-relevant global structure, whereas topology-aware designs, exemplified by THD-BAR, instead seek to jointly capture spatial hierarchy and temporal evolution.

\paragraph{Hybrid Objectives and Practical Selection}
Several models adopt hybrid designs that combine two or more complementary targets. Examples include Mentality, which combines $\mathcal L_{\mathrm{mse}}^{rs}$ and $\mathcal L_{\mathrm{mse}}^{spec}$; EEGPT~\cite{wang2024eegpt}, which combines $\mathcal L_{\mathrm{mse}}^{rs}$ with an embedding reconstruction term $\mathcal L_{\mathrm{mse}}^{emb}$; DMAE-EEG~\cite{zhang2025dmaeeeg}, which combines $\mathcal L_{\mathrm{mse}}^{rs}$ and $\mathcal L_{\mathrm{mse}}^{et}$; and SAMBA~\cite{hong2025samba}, which combines time-domain and frequency-domain reconstruction. These hybrid formulations should be interpreted as deliberate design choices to constrain the representation from complementary perspectives, rather than an indiscriminate aggregation of all available losses.

In practice, the choice of pre-training objective should be guided by the intended deployment setting. When robustness to cross-dataset heterogeneity and low signal-to-noise ratio is critical, token-level or codebook-based objectives can provide stronger invariance than direct waveform regression. When rhythmic neural activity is central to the downstream task, frequency-domain supervision can be beneficial, especially when combined with a time-domain constraint. When causal generation, prompt-based adaptation, or LLM-style modeling is desired, autoregressive objectives become a natural choice, but require careful tokenization and ordering design.

\paragraph{Summary}
Existing EEG FMs largely adhere to a masked prediction paradigm, but differ substantially in their target space and implied inductive biases. Table 2 of the supplementary information summarizes these design choices through the reconstruction objective and loss function columns, encompassing $\mathcal{L}_{mse}^{rs}$ and its robust variants, $\mathcal{L}_{mse}^{et}$ and $\mathcal{L}_{cl}^{et}$, spectral losses such as $\mathcal{L}_{mse}^{spec}$ and $\mathcal{L}_{mse}^{sa}$, codebook index losses such as $\mathcal{L}_{cls}^{ci}$ and $\mathcal{L}_{nll}^{ci}$, as well as additional auxiliary terms that refine the learning signal. This taxonomy provides a consistent framework for comparing pre-training strategies under heterogeneous EEG settings and clarifies why different approaches may be preferable for specific BCI paradigms and deployment constraints. Detailed experimental analysis is presented in the following section.

\subsubsection{Downstream generalization}

After pre-training on large-scale EEG corpora, downstream evaluation is required to assess whether the learned representations transfer effectively to practical BCI tasks. Fig.~\ref{fig:pie_data}j summarizes the most frequently used downstream datasets. These datasets span multiple representative paradigms. For example, TUAB, TUEV, and CHB-MIT are clinical EEG datasets; BCIC-IV-2A and PhysioNetMI are MI datasets; FACED, SEED, and SEED-V are emotion recognition datasets; and Sleep-EDF is a sleep-related dataset. This distribution indicates that downstream evaluation in existing work is largely concentrated on clinical applications, MI, affective decoding, and sleep analysis.

Following the problem definition, each downstream task $\tau_j$ is associated with a dataset $\mathcal D_{\mathrm{task}}^{(j)}$, which is partitioned into a fine-tuning set $\mathcal D_{\mathrm{ft}}^{(j)}$ and a held-out test set $\mathcal D_{\mathrm{te}}^{(j)}$. Given a pre-trained model $f_{\Theta^\star}$, task-specific fine-tuning estimates
\begin{equation}
\Theta_{j}^\star=\arg\min_{\Theta}\sum_{(X,y)\in\mathcal D_{\mathrm{ft}}^{(j)}}\mathcal L_{\mathrm{cls}}^{(j)}\!\left(y,f_{\Theta}(X)\right),
\label{eq:eval_ft}
\end{equation}
where $\mathcal L_{\mathrm{cls}}^{(j)}$ denotes a supervised loss for task $\tau_j$. The fine-tuned model $f_{\Theta_{j}^\star}$ is then evaluated on the held-out test set. Taking classification accuracy as an example, the evaluation is formulated as follows:
\begin{equation}
\widehat y=\arg\max_{c\in\{1,2,\ldots,\mathcal C_j\}}\left[f_{\Theta_{j}^\star}(X)\right]_c,
\label{eq:eval_te}
\end{equation}
\begin{equation}
\mathrm{Acc}(\tau_j)=\frac{1}{|\mathcal D_{\mathrm{te}}^{(j)}|}\sum_{(X,y)\in\mathcal D_{\mathrm{te}}^{(j)}}\mathbf 1\!\left(\widehat y=y\right),
\label{eq:eval_acc}
\end{equation}
where $\mathbf 1(\cdot)$ denotes the indicator function.

Regarding evaluation scenarios, most existing studies adopt a leave-one-subject-out (LOSO) setting or perform subject-disjoint splits into training, validation, and test sets. Such protocols are valuable for assessing cross-subject generalization, as the test subject remains unseen during fine-tuning. However, these protocols typically require a substantial amount of labeled data from the same device and paradigm for fine-tuning, since data from multiple training subjects are aggregated for adaptation. This requirement can be restrictive in practical deployment scenarios where only limited calibration data may be available for a new user.

In principle, a FM is expected to reduce dependence on task-matched labeled data and enable effective adaptation with minimal calibration. This motivates the need for a more comprehensive evaluation suite that encompasses both data-rich cross-subject transfer and data-limited calibration regimes under consistent protocols. Section `Benchmark of EEG FMs' presents our benchmark design, which is constructed to address these complementary requirements.

\section{BENCHMARK OF EEG FMs} \label{sect:bench}

This section presents a comprehensive benchmark for EEG FMs. We evaluate 12 open source FMs and 8 specialist baselines, including conventional machine learning and deep learning methods, on 13 datasets spanning 9 representative BCI paradigms. To assess model generalization under realistic deployment constraints, we design a set of comprehensive and fair evaluation scenarios. An overview of the datasets and evaluation scenarios is illustrated in Fig.~\ref{fig:method_bench}.

\subsection{Datasets}
Fig.~\ref{fig:pie_data}j summarizes the downstream datasets most frequently adopted in prior studies, where clinical EEG, MI, emotion recognition, and sleep staging emerge as the dominant evaluation paradigms. However, this concentration on a limited set of paradigms may not fully reflect the generalization capability of FMs across diverse BCI applications. To address this limitation, we selected 13 datasets spanning 9 BCI paradigms for downstream evaluation, providing broader coverage of representative BCI scenarios. Specifically, the selected datasets include BNCI2014001~\cite{tangermann2012BCICIV2A}, BNCI2014004~\cite{leeb2007BCICIV2B}, BNCI2015001~\cite{faller2012_15001}, BNCI2014009~\cite{riccio2013_14009}, BNCI2014008~\cite{riccio2013_14009}, CHB-MIT~\cite{Shoeb2009chbmit}, TUAB~\cite{lopez2015tuab}, Sleep-EDFx~\cite{kemp2000sleepedfx}, SEED~\cite{zheng2015seed}, Nakanishi2015~\cite{nakanishi2015nakanishi}, EEGMat~\cite{zyma2019eegmat}, Things-EEG2~\cite{gifford2022things2}, and SEED-VIG~\cite{zheng2017multimodal}. For paradigms with widespread practical applications, such as MI, P300, and clinical diagnosis, we included two to three datasets to ensure comprehensive evaluation. The dataset characteristics are summarized in Section~A3 of the supplementary information.

\subsection{Evaluation Scenarios}

Most existing EEG FM studies evaluated downstream transfer under a LOSO scenario or subject-disjoint splits into training, validation, and test sets. While this setting is widely adopted, it remains unclear whether it provides a comprehensive assessment of FM generalization and whether it aligns with practical deployment requirements.

\paragraph{LOSO Scenario}
The LOSO scenario evaluates cross-subject generalization within the same task and headset configuration. Concretely, a model is fine-tuned using labeled data from a subset of subjects recorded with the same EEG device and paradigm, and is subsequently evaluated on held-out subjects. The primary advantage of LOSO is that the target subject is evaluated in a zero-calibration manner, as no labeled data from the test subject are used during fine-tuning. However, LOSO has two important limitations. First, it typically requires a substantial amount of labeled data from multiple subjects, which increases the fine-tuning cost. Second, it implicitly assumes the availability of a corpus collected with the same device configuration and task as the target deployment setting, which may not hold in practice, particularly for new devices or customized paradigms.

In our benchmark, LOSO fine-tuning followed the common practice of using all available trials from the fine-tuning subjects for most datasets. However, for the MI and P300 datasets (BNCI2014001, BNCI2014004, BNCI2015001, BNCI2014009, and BNCI2014008), we used only a single session from each subject for fine-tuning. For CHB-MIT, we used the ictal segments together with the 10-minute pre-ictal segments of each seizure for model adaptation and evaluation. For TUAB, we used the first 3 minutes of each recording as input segments. For Things-EEG2, we used three out of ten images per class. For Sleep-EDFx and TUAB, which contain a large number of subjects, we adopted a ten-fold subject split, where subjects were evenly partitioned into ten folds for cross-validation.

\paragraph{Within-Subject Few-Shot Scenario}
To better reflect deployment settings where only limited calibration data are available for a new user, we designed a within-subject few-shot evaluation protocol. In this setting, the model was fine-tuned using a small labeled subset from the target subject and evaluated on the remaining data of the same subject. The within-subject scenario offers two advantages. First, it substantially reduces the amount of fine-tuning data and lowers the adaptation cost. Second, it does not require an external training corpus that matches the target device and paradigm, thereby supporting rapid personalization for new devices, tasks, and users. The primary limitation is that it requires collecting labeled calibration data from the target user during deployment, which may be inconvenient in certain applications.

The fine-tuning ratio adopted for each dataset is determined by dataset-specific constraints, including the total number of trials available per subject, the trial length, and the class distribution of the paradigm. Detailed protocols are summarized in Table~12 of the supplementary information.

Specifically, for the MI datasets (BNCI2014001, BNCI2014004, and BNCI2015001), we fine-tuned using 30\% of one session from the target subject, with fewer than 30 trials per class. For P300 datasets, we fine-tuned using 10\% of one session for BNCI2014009 and 5\% of one session for BNCI2014008. For CHB-MIT, we fine-tuned using the first seizure's ictal segments from the target subject together with its 10-minute pre-ictal segment, and evaluated on the remaining seizures' ictal segments with their corresponding pre-ictal segments. For Sleep-EDFx, we used 10\% of the target subject's data for fine-tuning. For SEED, we used one video per class for fine-tuning. For Nakanishi2015 and EEGMat, we fine-tuned using 80\% and 60\% of the target subject's data, respectively. For Things-EEG2, we used three out of ten images per class for each subject. For SEED-VIG, we fine-tuned using 10\% of the data from a single subject. We did not include a within-subject few-shot setting for TUAB, as each subject is associated with a single diagnostic label (normal or abnormal).

\paragraph{Fine-Tuning Strategies}
For both LOSO and within-subject few-shot scenarios, we compared two fine-tuning strategies to assess the quality of pre-trained representations. The first strategy, \textit{full-parameter fine-tuning}, updated all model parameters during fine-tuning, allowing the entire network to be optimized for the downstream task. The second strategy, \textit{linear probing}, froze the pre-trained encoder and trained only the classification head, which directly evaluated the transferability of the learned representations without task-specific feature adaptation. By comparing these two strategies, we aimed to disentangle the contributions of pre-trained representations from those of end-to-end fine-tuning.

\paragraph{Summary}
The LOSO and within-subject few-shot protocols evaluate complementary aspects of generalization. LOSO measures cross-subject transfer under a fixed paradigm and device configuration without test-subject calibration, whereas within-subject few-shot evaluates rapid personalization with limited calibration data. Furthermore, the comparison between full-parameter fine-tuning and linear probing provides insights into the quality and transferability of pre-trained representations. By reporting results under both scenarios with both fine-tuning strategies, our benchmark provides a more comprehensive assessment of EEG FM generalization and better reflects practical deployment constraints.

\subsection{Evaluated Approaches}

In this benchmark, we compared traditional machine learning baselines, 7 deep learning models (including 4 CNN-based and 3 Transformer-based architectures), and 12 EEG FMs. The following provides a detailed description of each category.

\paragraph{Traditional Machine Learning Baselines}
For each dataset, we selected a paradigm-specific traditional machine learning algorithm as the baseline, which remains competitive against deep learning methods in the respective domain. CSP+LDA (linear discriminant analysis)~\cite{blankertz2007csplda} was used for BNCI2014001, BNCI2014004, BNCI2015001, TUAB, SEED, and EEGMat. xDAWN+LDA~\cite{rivet2009xdawn} was employed for the P300 datasets BNCI2014009 and BNCI2014008. PSD+SVM (Power Spectral Density features and Support Vector Machine classifier)~\cite{zuo2024psdsvm} was used for CHB-MIT. PSD+LDA~\cite{lal2023psdlda} was applied to Sleep-EDFx. TRCA (task-related component analysis)~\cite{nakanishi2017trca} was used for Nakanishi2015. PSD+Ridge (ridge regression)~\cite{wu2020psdridge} was employed for SEED-VIG.

\paragraph{Deep Learning Baselines}
We evaluated 7 task-specific deep learning models trained from scratch. For CNN-based architectures, we included EEGNet~\cite{lawhern2018eegnet}, ShallowConvNet~\cite{schirrmeister2017shallowdeep}, LMDA-Net~\cite{miao2023lmda}, and TSception~\cite{ding2022tsception}. For Transformer-based architectures, we included CNN-Transformer~\cite{peh2022cnntransformer}, Deformer~\cite{ding2024deformer}, and Conformer~\cite{song2022conformer}. 

These models are widely adopted in EEG decoding and serve as strong baselines for comparison with FMs. We selected general-purpose models that are applicable across paradigms, rather than algorithms designed around the neurophysiological principles of a specific paradigm. Consequently, the chosen baselines are competitive but not necessarily the state-of-the-art for every individual paradigm. Our aim is to contrast EEG FMs against strong, representative task-specific models, rather than to quantify the full gap between FMs and the best-performing specialized method for each task.

\paragraph{EEG FMs}
We evaluated all EEG FMs with publicly available code and pre-trained weights, including BENDR~\cite{kostas2021bendr}, BIOT~\cite{yang2023biot}, LaBraM~\cite{jiang2024labram}, Neuro-GPT~\cite{cui2024neurogpt}, EEGPT~\cite{wang2024eegpt}, CBraMod~\cite{wang2025cbramod}, TFM~\cite{pradeepkumar2025TFM}, BrainOmni~\cite{xiao2025brainomni}, EEGMamba~\cite{wang2025eegmamba}, MIRepNet~\cite{liu2026mirepnet}, SingLEM~\cite{sukhbaatar2025singlem}, and LUNA~\cite{doner2025luna}. Among these, MIRepNet is a paradigm-specific FM designed exclusively for MI tasks, while the remaining 11 models are general-purpose EEG FMs intended to support multiple BCI paradigms.

\subsection{Main Results}

We performed all benchmarking experiments on 24 NVIDIA A800 GPUs and 8 NVIDIA A100 GPUs. All results are averaged over 3 random seeds, and we reported the performance on the test set at the final epoch of fine-tuning. For fair comparison, the fine-tuning hyperparameters of each model were individually tuned rather than fixed to a single shared configuration. 

For evaluation metrics, we reported balanced classification accuracy (BCA) as the primary metric for all classification tasks, 2-way accuracy for the Things-EEG2 dataset, and root mean squared error (RMSE) for the SEED-VIG regression task. Several settings specific to individual models and datasets are also noted here. BrainOmni was excluded from the CHB-MIT and Sleep-EDFx datasets, as its implementation does not support the bi-polar montage used in these datasets. In addition, BIOT includes three variants, BIOT-1D, BIOT-2D, and BIOT-6D, pre-trained on 1, 2, and 6 datasets, respectively. We reported results for BIOT-6D, the variant trained on the largest data scale, in the main tables, while results for BIOT-1D and BIOT-2D are provided in Section~A4 of the supplementary information. The main results are summarized in Tables~\ref{tab:bench_1} and~\ref{tab:bench_2}, and comprehensive per-subject results are provided in Section~A4 of the supplementary information.

The detailed ranking results are provided in Tables~49 and~50 of the supplementary information. To assess the potential influence of pre-training--downstream data overlap on these rankings, we further reported a leakage-aware ranking in Table~51 of the supplementary information. Specifically, for each FM, we excluded downstream datasets utilized in its pre-training corpus (marked in Tables~\ref{tab:bench_1} and~\ref{tab:bench_2}) before computing its average rank, yielding a more conservative estimate of generalization to unseen downstream distributions. The leakage-aware ranking preserves the relative ordering of the top-ranked models, with CBraMod and Neuro-GPT retaining their positions among the top FMs. This suggests that the main findings reported in this section are not primarily driven by such overlap.

\begin{table*}[!t]
    \centering
    \renewcommand{\arraystretch}{0.9}
    \fontsize{9}{12}\selectfont
    \caption{Benchmark performance. The best metrics are marked in bold, and the second best by an underline. $\bm{\ddagger}$ indicates that the corresponding dataset was used during pre-training of the model.}
    \label{tab:bench_1}
    \setlength{\tabcolsep}{5pt}
    \scalebox{0.65}{
    \begin{tabular}{l*{7}{w{c}{2.35cm}}}
        \toprule
        Approach & BNCI2014001 & BNCI2014004 & BNCI2015001 & BNCI2014009 & BNCI2014008 & CHB-MIT & TUAB \\
        \midrule
 
        \multicolumn{8}{l}{\textbf{Cross-subject (LOSO)\,---\,Full fine-tuning}} \\
        \addlinespace[2pt]
        \multicolumn{8}{l}{\hspace{0.8em}\textit{Specialist models}} \\
        Traditional ML & 38.19 & 73.24 & 56.42 & 65.14 & 58.69 & 69.09$_{\pm0.04}$ & 66.03$_{\pm0.25}$ \\
        EEGNet & 44.97$_{\pm0.57}$ & 76.38$_{\pm0.59}$ & 63.40$_{\pm1.32}$ & \underline{78.39}$_{\pm 0.44}$ & \textbf{72.29}$_{\pm 0.17}$ & 77.36$_{\pm0.54}$ & 77.03$_{\pm0.67}$ \\
        ShallowConv & 44.80$_{\pm0.50}$ & 74.23$_{\pm0.39}$ & 63.89$_{\pm0.80}$ & 78.05$_{\pm 0.62}$ & 69.92$_{\pm 0.05}$ & \textbf{80.18}$_{\pm0.22}$ & 79.81$_{\pm0.12}$ \\
        LMDA & 46.80$_{\pm0.31}$ & 74.88$_{\pm0.67}$ & 61.40$_{\pm 0.94}$ & \textbf{78.45}$_{\pm 0.46}$ & \underline{71.74}$_{\pm 0.12}$ & 77.47$_{\pm0.41}$ & 62.69$_{\pm1.52}$ \\
        CNN-T & 39.15$_{\pm0.56}$ & 71.20$_{\pm0.42}$ & 59.64$_{\pm1.41}$ & 61.50$_{\pm 1.71}$ & 51.93$_{\pm 0.16}$ & 76.34$_{\pm0.61}$ & 73.20$_{\pm0.73}$ \\
        Deformer & 41.53$_{\pm0.67}$ & 74.86$_{\pm0.82}$ & 63.06$_{\pm1.17}$ & 76.89$_{\pm 0.14}$ & 57.75$_{\pm 1.21}$ & \underline{79.77}$_{\pm0.14}$ & \underline{81.48}$_{\pm0.21}$ \\
        Conformer & 41.64$_{\pm1.23}$ & 74.17$_{\pm0.49}$ & 59.10$_{\pm 2.14}$ & 76.78$_{\pm 0.33}$ & 67.87$_{\pm 0.11}$ & 78.58$_{\pm0.56}$ & 77.78$_{\pm0.04}$ \\
        TSception & 41.58$_{\pm0.63}$ & 69.51$_{\pm0.99}$ & 57.65$_{\pm1.51}$ & 51.77$_{\pm 0.70}$ & 50.40$_{\pm 0.09}$ & 74.37$_{\pm0.53}$ & 79.26$_{\pm0.49}$ \\
        \addlinespace[4pt]
        \multicolumn{8}{l}{\hspace{0.8em}\textit{Foundation models}} \\
        BENDR & \underline{51.11}$_{\pm0.25}$ & 73.35$_{\pm0.06}$ & 62.68$_{\pm 0.49}$ & 73.46$_{\pm 0.28}$ & 65.01$_{\pm 0.18}$ & 75.50$_{\pm0.93}$ & $^{\bm{\ddagger}}$79.09$_{\pm0.16}$ \\
        BIOT-6D & 34.27$_{\pm0.93}$ & 70.22$_{\pm1.32}$ & \underline{63.94}$_{\pm 1.20}$ & 58.14$_{\pm 0.33}$ & 54.77$_{\pm 0.21}$ & $^{\bm{\ddagger}}$74.85$_{\pm0.33}$ & $^{\bm{\ddagger}}$77.90$_{\pm0.14}$ \\
        LaBraM & 46.93$_{\pm1.43}$ & \underline{76.97}$_{\pm1.08}$ & \textbf{64.14}$_{\pm1.03}$ & 70.31$_{\pm 0.24}$ & 63.07$_{\pm 0.97}$ & 70.87$_{\pm0.59}$ & 76.23$_{\pm0.27}$ \\
        Neuro-GPT & 46.97$_{\pm0.71}$ & \textbf{77.70}$_{\pm0.70}$ & 60.62$_{\pm1.63}$ & 75.97$_{\pm 0.53}$ & 68.59$_{\pm 0.25}$ & 73.27$_{\pm0.27}$ & $^{\bm{\ddagger}}$79.50$_{\pm0.17}$ \\
        EEGPT & 32.24$_{\pm1.45}$ & 71.37$_{\pm0.16}$ & 59.88$_{\pm1.39}$ & 62.77$_{\pm 1.85}$ & 58.24$_{\pm 0.40}$ & 66.91$_{\pm2.89}$ & 77.67$_{\pm0.17}$ \\
        CBraMod & \textbf{53.03}$_{\pm0.22}$ & 75.45$_{\pm0.35}$ & 63.47$_{\pm0.36}$ & 77.30$_{\pm 0.28}$ & 69.91$_{\pm 0.06}$ & 74.23$_{\pm0.19}$ & $^{\bm{\ddagger}}$79.98$_{\pm0.11}$ \\
        TFM & 32.02$_{\pm0.66}$ & 60.12$_{\pm3.00}$ & 55.35$_{\pm1.46}$ & 53.10$_{\pm0.48}$ & 52.99$_{\pm0.29}$ & $^{\bm{\ddagger}}$63.46$_{\pm0.60}$ & $^{\bm{\ddagger}}$75.65$_{\pm0.07}$ \\
        BrainOmni-Tiny & 41.58$_{\pm0.80}$ & 70.13$_{\pm0.89}$ & 61.88$_{\pm0.30}$ & 70.48$_{\pm 0.23}$ & 61.40$_{\pm 0.17}$ & --- & 72.92$_{\pm0.25}$ \\
        BrainOmni-Base & 40.93$_{\pm0.83}$ & 69.30$_{\pm0.89}$ & 60.64$_{\pm0.39}$ & 70.87$_{\pm0.29}$ & 59.31$_{\pm0.06}$ & --- & 80.49$_{\pm0.29}$ \\
        EEGMamba & 45.72$_{\pm0.54}$ & 73.30$_{\pm0.57}$ & 61.53$_{\pm0.50}$ & 76.01$_{\pm0.05}$ & 68.18$_{\pm0.05}$ & 75.92$_{\pm0.21}$ & $^{\bm{\ddagger}}$80.90$_{\pm0.10}$ \\
        SingLEM & 30.57$_{\pm0.10}$ & 67.31$_{\pm0.18}$ & 54.47$_{\pm0.62}$ & 71.98$_{\pm0.56}$ & 63.42$_{\pm0.13}$ & 60.78$_{\pm1.82}$ & 50.80$_{\pm1.02}$ \\
        LUNA-Base & 28.86$_{\pm0.50}$ & 56.17$_{\pm3.14}$ & 55.71$_{\pm1.37}$ & 51.67$_{\pm0.54}$ & 50.00$_{\pm0.04}$ & 78.12$_{\pm0.61}$ & $^{\bm{\ddagger}}$\textbf{81.92}$_{\pm0.07}$ \\
        \midrule
 
        \multicolumn{8}{l}{\textbf{Cross-subject (LOSO)\,---\,Linear probing}} \\
        \addlinespace[2pt]
        \multicolumn{8}{l}{\hspace{0.8em}\textit{Foundation models}} \\
        BENDR & 32.18$_{\pm 0.41}$ & 60.46$_{\pm 1.06}$ & 53.85$_{\pm 1.11}$ & 61.24$_{\pm 2.07}$ & 67.11$_{\pm 0.76}$ & 53.22$_{\pm0.68}$ & $^{\bm{\ddagger}}$58.33$_{\pm0.66}$ \\
        BIOT-6D & 30.88$_{\pm 1.00}$ & 61.35$_{\pm 2.11}$ & 63.43$_{\pm 0.63}$ & 51.04$_{\pm 0.12}$ & 53.19$_{\pm 0.66}$ & $^{\bm{\ddagger}}$69.79$_{\pm0.62}$ & $^{\bm{\ddagger}}$73.46$_{\pm0.25}$ \\
        LaBraM & 42.59$_{\pm 0.27}$ & 65.05$_{\pm 0.23}$ & 61.97$_{\pm 0.17}$ & 67.75$_{\pm 0.32}$ & 56.82$_{\pm 0.60}$ & 68.84$_{\pm0.36}$ & 78.32$_{\pm0.11}$ \\
        Neuro-GPT & 48.24$_{\pm 1.04}$ & 75.57$_{\pm 1.26}$ & 61.24$_{\pm 0.50}$ & 58.70$_{\pm 0.26}$ & 50.08$_{\pm 0.04}$ & 70.45$_{\pm0.24}$ & $^{\bm{\ddagger}}$79.64$_{\pm0.15}$ \\
        EEGPT & 37.37$_{\pm 1.25}$ & 72.08$_{\pm 2.07}$ & 63.00$_{\pm 2.89}$ & 66.53$_{\pm 0.05}$ & 57.26$_{\pm 0.17}$ & 70.94$_{\pm0.69}$ & 77.51$_{\pm0.09}$ \\
        CBraMod & 41.45$_{\pm0.50}$ & 69.27$_{\pm 0.55}$ & 59.93$_{\pm 0.29}$ & 58.63$_{\pm 0.54}$ & 53.31$_{\pm 0.14}$ & 75.21$_{\pm0.52}$ & $^{\bm{\ddagger}}$78.04$_{\pm0.04}$ \\
        TFM & 28.34$_{\pm0.30}$ & 50.97$_{\pm1.13}$ & 53.43$_{\pm0.53}$ & 51.56$_{\pm0.51}$ & 51.93$_{\pm0.63}$ & $^{\bm{\ddagger}}$56.83$_{\pm0.75}$ & $^{\bm{\ddagger}}$68.25$_{\pm0.09}$ \\
        BrainOmni-Tiny & 39.78$_{\pm 0.36}$ & 66.05$_{\pm 0.53}$ & 61.54$_{\pm 0.54}$ & 61.34$_{\pm 0.42}$ & 51.33$_{\pm 0.14}$ & --- & 77.03$_{\pm0.37}$ \\
        BrainOmni-Base & 39.63$_{\pm0.58}$ & 67.48$_{\pm0.25}$ & 60.38$_{\pm0.19}$ & 69.50$_{\pm0.69}$ & 58.93$_{\pm0.43}$ & --- & 79.32$_{\pm0.30}$ \\
        EEGMamba & 34.32$_{\pm0.20}$ & 61.94$_{\pm0.31}$ & 56.38$_{\pm0.27}$ & 74.13$_{\pm0.30}$ & 63.65$_{\pm0.11}$ & 71.81$_{\pm0.26}$ & $^{\bm{\ddagger}}$78.24$_{\pm0.04}$ \\
        SingLEM & 33.42$_{\pm0.22}$ & 63.69$_{\pm0.36}$ & 54.08$_{\pm0.20}$ & 72.52$_{\pm0.05}$ & 61.89$_{\pm0.03}$ & 50.50$_{\pm0.11}$ & 51.53$_{\pm0.13}$ \\
        LUNA-Base & 29.21$_{\pm0.57}$ & 51.10$_{\pm0.65}$ & 56.72$_{\pm0.41}$ & 52.16$_{\pm0.50}$ & 49.99$_{\pm0.02}$ & 67.43$_{\pm0.45}$ & $^{\bm{\ddagger}}$75.88$_{\pm0.17}$ \\
        \midrule
 
        \multicolumn{8}{l}{\textbf{Within-subject (Few-shot)\,---\,Full fine-tuning}} \\
        \addlinespace[2pt]
        \multicolumn{8}{l}{\hspace{0.8em}\textit{Specialist models}} \\
        Traditional ML & \textbf{60.62} & 79.20 & \underline{75.89} & 54.45 & 56.65 & 77.71$_{\pm0.35}$ & --- \\
        EEGNet & 50.22$_{\pm1.14}$ & 75.53$_{\pm3.67}$ & 72.08$_{\pm0.39}$ & \textbf{68.99}$_{\pm 0.51}$ & \textbf{70.91}$_{\pm 1.17}$ & 88.45$_{\pm0.48}$ & --- \\
        ShallowConv & 52.87$_{\pm0.88}$ & 74.52$_{\pm 0.71}$ & 73.79$_{\pm0.66}$ & 57.13$_{\pm 0.93}$ & 55.97$_{\pm 0.20}$ & 85.81$_{\pm0.44}$ & --- \\
        LMDA & 51.13$_{\pm0.76}$ & 75.11$_{\pm1.63}$ & 73.79$_{\pm1.50}$ & 60.28$_{\pm 0.82}$ & 63.58$_{\pm 0.30}$ & 86.80$_{\pm0.55}$ & --- \\
        CNN-T & 51.27$_{\pm1.10}$ & 75.77$_{\pm0.38}$ & 71.63$_{\pm1.36}$ & 50.32$_{\pm 0.52}$ & 50.23$_{\pm 0.29}$ & 87.66$_{\pm0.33}$ & --- \\
        Deformer & 42.21$_{\pm0.73}$ & 73.02$_{\pm2.13}$ & 70.32$_{\pm1.28}$ & 65.38$_{\pm 0.32}$ & \underline{64.36}$_{\pm 0.93}$ & 86.88$_{\pm0.33}$ & --- \\
        Conformer & 57.19$_{\pm1.32}$ & \underline{80.17}$_{\pm0.25}$ & \textbf{77.10}$_{\pm1.49}$ & 65.43$_{\pm 0.90}$ & 58.34$_{\pm 0.23}$ & \textbf{91.58}$_{\pm0.34}$ & --- \\
        TSception & \underline{58.68}$_{\pm0.86}$ & \textbf{80.36}$_{\pm0.65}$ & 74.35$_{\pm1.12}$ & 63.99$_{\pm 0.43}$ & 51.73$_{\pm0.30}$ & \underline{90.92}$_{\pm 0.74}$ & --- \\
        \addlinespace[4pt]
        \multicolumn{8}{l}{\hspace{0.8em}\textit{Foundation models}} \\
        BENDR & 44.90$_{\pm1.31}$ & 71.70$_{\pm1.46}$ & 56.59$_{\pm0.25}$ & 59.27$_{\pm 1.03}$ & 58.33$_{\pm 1.00}$ & 54.14$_{\pm0.33}$ & --- \\
        BIOT-6D & 48.60$_{\pm0.72}$ & 68.43$_{\pm0.42}$ & 68.43$_{\pm1.48}$ & 52.19$_{\pm 0.16}$ & 52.12$_{\pm 0.27}$ & $^{\bm{\ddagger}}$79.60$_{\pm0.51}$ & --- \\
        LaBraM & 37.13$_{\pm0.92}$ & 69.76$_{\pm1.23}$ & 61.39$_{\pm0.65}$ & 60.06$_{\pm 0.58}$ & 54.67$_{\pm 0.67}$ & 71.31$_{\pm0.29}$ & --- \\
        Neuro-GPT & 46.39$_{\pm1.92}$ & 75.25$_{\pm1.58}$ & 61.90$_{\pm0.90}$ & 55.02$_{\pm 0.97}$ & 61.61$_{\pm 0.14}$ & 83.09$_{\pm0.59}$ & --- \\
        EEGPT & 34.99$_{\pm0.25}$ & 62.07$_{\pm0.74}$ & 59.17$_{\pm0.87}$ & 52.02$_{\pm 1.14}$ & 51.93$_{\pm 0.41}$ & 63.74$_{\pm0.28}$ & --- \\
        CBraMod & 50.34$_{\pm1.18}$ & 77.39$_{\pm0.35}$ & 70.30$_{\pm0.72}$ & 56.85$_{\pm 1.03}$ & 58.00$_{\pm 0.99}$ & 88.54$_{\pm0.16}$ & --- \\
        TFM & 33.30$_{\pm0.46}$ & 58.92$_{\pm2.41}$ & 55.34$_{\pm0.28}$ & 51.42$_{\pm0.87}$ & 51.21$_{\pm0.28}$ & $^{\bm{\ddagger}}$59.55$_{\pm0.57}$ & --- \\
        BrainOmni-Tiny & 39.00$_{\pm0.19}$ & 63.30$_{\pm0.98}$ & 61.59$_{\pm1.54}$ & 57.02$_{\pm 0.36}$ & 56.34$_{\pm 0.33}$ & --- & --- \\
        BrainOmni-Base & 37.60$_{\pm0.27}$ & 61.84$_{\pm0.76}$ & 59.86$_{\pm0.83}$ & 59.75$_{\pm1.14}$ & 56.18$_{\pm0.14}$ & --- & --- \\
        EEGMamba & 38.04$_{\pm0.45}$ & 65.07$_{\pm1.86}$ & 60.24$_{\pm0.48}$ & 65.50$_{\pm0.46}$ & 61.12$_{\pm0.35}$ & 79.50$_{\pm0.42}$ & --- \\
        SingLEM & 28.94$_{\pm0.73}$ & 56.75$_{\pm1.66}$ & 52.12$_{\pm0.63}$ & 57.70$_{\pm0.41}$ & 56.38$_{\pm0.28}$ & 56.70$_{\pm1.49}$ & --- \\
        LUNA-Base & 31.94$_{\pm1.24}$ & 56.50$_{\pm0.74}$ & 60.36$_{\pm1.69}$ & 51.66$_{\pm0.99}$ & 50.29$_{\pm0.12}$ & 72.30$_{\pm0.26}$ & --- \\
        \midrule
 
        \multicolumn{8}{l}{\textbf{Within-subject (Few-shot)\,---\,Linear probing}} \\
        \addlinespace[2pt]
        \multicolumn{8}{l}{\hspace{0.8em}\textit{Foundation models}} \\
        BENDR & 31.43$_{\pm 0.85}$ & 55.27$_{\pm 2.40}$ & 50.95$_{\pm 0.90}$ & 52.03$_{\pm 0.18}$ & 51.75$_{\pm 0.70}$ & 50.07$_{\pm0.18}$ & --- \\
        BIOT-6D & 47.95$_{\pm 0.53}$ & 67.15$_{\pm 1.64}$ & 67.80$_{\pm 1.21}$ & 53.03$_{\pm 0.78}$ & 51.62$_{\pm 0.47}$ & $^{\bm{\ddagger}}$73.41$_{\pm1.06}$ & --- \\
        LaBraM & 35.38$_{\pm 0.45}$ & 59.83$_{\pm 1.22}$ & 59.74$_{\pm 0.86}$ & 55.15$_{\pm 0.21}$ & 52.96$_{\pm 0.24}$ & 66.93$_{\pm0.88}$ & --- \\
        Neuro-GPT & 49.82$_{\pm 1.55}$ & 76.69$_{\pm 1.35}$ & 65.60$_{\pm 0.86}$ & 50.65$_{\pm 0.31}$ & 50.88$_{\pm 0.02}$ & 71.68$_{\pm1.57}$ & --- \\
        EEGPT & 35.86$_{\pm 0.54}$ & 65.88$_{\pm 2.18}$ & 60.87$_{\pm 2.36}$ & 53.18$_{\pm 0.97}$ & 51.34$_{\pm 0.18}$ & 66.71$_{\pm1.20}$ & --- \\
        CBraMod & 40.74$_{\pm 1.06}$ & 70.38$_{\pm 0.90}$ & 60.81$_{\pm 0.06}$ & 50.18$_{\pm 0.15}$ & 51.43$_{\pm 0.09}$ & 87.77$_{\pm0.73}$ & --- \\
        TFM & 27.78$_{\pm1.05}$ & 51.30$_{\pm1.08}$ & 53.73$_{\pm1.18}$ & 51.12$_{\pm0.83}$ & 50.93$_{\pm0.36}$ & $^{\bm{\ddagger}}$53.32$_{\pm0.37}$ & --- \\
        BrainOmni-Tiny & 40.61$_{\pm 0.31}$ & 59.79$_{\pm 0.60}$ & 63.21$_{\pm 0.76}$ & 56.33$_{\pm 0.16}$ & 52.96$_{\pm 0.05}$ & --- & --- \\
        BrainOmni-Base & 38.71$_{\pm0.36}$ & 59.10$_{\pm0.30}$ & 60.89$_{\pm1.07}$ & 58.37$_{\pm0.35}$ & 54.33$_{\pm0.30}$ & --- & --- \\
        EEGMamba & 33.84$_{\pm0.40}$ & 59.02$_{\pm0.34}$ & 54.25$_{\pm0.65}$ & \underline{65.79}$_{\pm0.16}$ & 61.45$_{\pm0.09}$ & 82.18$_{\pm0.17}$ & --- \\
        SingLEM & 29.87$_{\pm0.18}$ & 56.51$_{\pm1.27}$ & 53.75$_{\pm0.32}$ & 63.66$_{\pm0.96}$ & 58.27$_{\pm0.40}$ & 51.07$_{\pm0.11}$ & --- \\
        LUNA-Base & 34.73$_{\pm0.05}$ & 55.61$_{\pm0.89}$ & 57.56$_{\pm0.18}$ & 51.42$_{\pm0.19}$ & 50.79$_{\pm0.15}$ & 71.11$_{\pm0.20}$ & --- \\
        \bottomrule
    \end{tabular}
    }
\end{table*}

\begin{table*}[!t]
    \centering
    \renewcommand{\arraystretch}{0.9}
    \fontsize{9}{12}\selectfont
    \caption{Benchmark performance. The best metrics are marked in bold, and the second best by an underline. $\bm{\ddagger}$ indicates that the corresponding dataset was used during pre-training of the model (continued).}
    \label{tab:bench_2}
    \setlength{\tabcolsep}{5pt}
    \scalebox{0.66}{
    \begin{tabular}{l*{6}{w{c}{2.7cm}}}
        \toprule
        Approach & Sleep-EDFx & SEED & Nakanishi2015 & EEGMat & Things-EEG2 & SEED-VIG \\
        \midrule
 
        \multicolumn{7}{l}{\textbf{Cross-subject (LOSO)\,---\,Full fine-tuning}} \\
        \addlinespace[2pt]
        \multicolumn{7}{l}{\hspace{0.8em}\textit{Specialist models}} \\
        Traditional ML & 51.78$_{\pm0.22}$ & 48.91 & 94.07 & 67.41 & --- & 0.2489 \\
        EEGNet & 73.75$_{\pm0.19}$ & 48.57$_{\pm 1.34}$ & \underline{95.88}$_{\pm 0.18}$ & 66.60$_{\pm 0.63}$ & 74.42$_{\pm3.67}$ & 0.2561$_{\pm 0.0092}$ \\
        ShallowConv & 74.86$_{\pm0.42}$ & \underline{53.41}$_{\pm 0.12}$ & 69.61$_{\pm 0.86}$ & 72.22$_{\pm 1.24}$ & 72.03$_{\pm0.38}$ & 0.2290$_{\pm 0.0029}$ \\
        LMDA & 74.58$_{\pm0.34}$ & 50.12$_{\pm 0.43}$ & 85.12$_{\pm 0.94}$ & 67.47$_{\pm 1.21}$ & \underline{78.72}$_{\pm0.86}$ & 0.2389$_{\pm 0.0029}$ \\
        CNN-T & 75.74$_{\pm0.48}$ & 44.56$_{\pm 1.40}$ & 46.34$_{\pm 0.34}$ & 70.77$_{\pm 2.49}$ & 59.05$_{\pm2.13}$ & 0.2556$_{\pm 0.0140}$ \\
        Deformer & \textbf{78.73}$_{\pm0.09}$ & 51.05$_{\pm 0.97}$ & \textbf{97.18}$_{\pm 0.13}$ & 71.73$_{\pm 0.37}$ & 78.47$_{\pm0.65}$ & 0.2512$_{\pm 0.0053}$ \\
        Conformer & 68.40$_{\pm2.87}$ & 48.76$_{\pm 1.23}$ & 91.34$_{\pm 1.34}$ & 70.49$_{\pm 0.93}$ & 64.00$_{\pm0.91}$ & 0.2405$_{\pm 0.0044}$ \\
        TSception & \underline{76.41}$_{\pm0.43}$ & 41.82$_{\pm 0.79}$ & 60.51$_{\pm0.88}$ & \textbf{74.10}$_{\pm 0.04}$ & 63.33$_{\pm1.47}$ & 0.2302$_{\pm 0.0019}$ \\
        \addlinespace[4pt]
        \multicolumn{7}{l}{\hspace{0.8em}\textit{Foundation models}} \\
        BENDR & 71.45$_{\pm0.43}$ & 52.50$_{\pm 0.85}$ & 92.94$_{\pm 0.30}$ & 54.32$_{\pm 0.71}$ & 71.47$_{\pm0.39}$ & 0.2412$_{\pm 0.0025}$ \\
        BIOT-6D & 66.35$_{\pm0.23}$ & 49.04$_{\pm 0.90}$ & 72.35$_{\pm 3.04}$ & 70.77$_{\pm 1.49}$ & 50.57$_{\pm0.22}$ & 0.2374$_{\pm0.0033}$ \\
        LaBraM & 63.56$_{\pm0.03}$ & $^{\bm{\ddagger}}$52.23$_{\pm 0.92}$ & 79.18$_{\pm 0.73}$ & 65.74$_{\pm 1.61}$ & 75.15$_{\pm0.93}$ & \textbf{0.2281}$_{\pm 0.0035}$ \\
        Neuro-GPT & 59.09$_{\pm0.25}$ & 49.67$_{\pm 0.38}$ & 87.35$_{\pm 0.40}$ & \underline{72.62}$_{\pm1.35}$ & \textbf{80.28}$_{\pm1.08}$ & 0.2509$_{\pm0.0055}$ \\
        EEGPT & 62.93$_{\pm1.06}$ & $^{\bm{\ddagger}}$48.74$_{\pm 3.41}$ & 88.31$_{\pm 3.30}$ & 58.02$_{\pm1.02}$ & 74.90$_{\pm1.35}$ & 0.2402$_{\pm0.0025}$ \\
        CBraMod & 72.30$_{\pm0.18}$ & \textbf{53.61}$_{\pm 0.61}$ & 85.39$_{\pm 0.60}$ & 68.43$_{\pm 0.72}$ & 75.88$_{\pm0.89}$ & 0.2718$_{\pm 0.0019}$ \\
        TFM & 67.38$_{\pm0.25}$ & 36.66$_{\pm0.26}$ & 12.84$_{\pm0.68}$ & 63.02$_{\pm1.95}$ & 50.48$_{\pm0.45}$ & \underline{0.2283}$_{\pm0.0026}$ \\
        BrainOmni-Tiny & --- & 38.02$_{\pm 0.03}$ & 78.33$_{\pm 0.23}$ & 57.50$_{\pm0.80}$ & 66.83$_{\pm1.16}$ & 0.2434$_{\pm0.0003}$ \\
        BrainOmni-Base & --- & 44.72$_{\pm0.18}$ & 50.88$_{\pm1.99}$ & 51.51$_{\pm0.76}$ & 67.13$_{\pm0.39}$ & 0.2549$_{\pm0.0055}$ \\
        EEGMamba & 66.68$_{\pm0.42}$ & 52.19$_{\pm0.04}$ & 70.08$_{\pm0.94}$ & 49.97$_{\pm0.04}$ & 74.48$_{\pm0.12}$ & 0.2297$_{\pm0.0010}$ \\
        SingLEM & 67.23$_{\pm0.48}$ & 50.16$_{\pm1.48}$ & 33.54$_{\pm1.23}$ & 49.85$_{\pm0.16}$ & 61.97$_{\pm4.80}$ & 0.2349$_{\pm0.0004}$ \\
        LUNA-Base & 71.03$_{\pm0.36}$ & 49.96$_{\pm0.35}$ & 8.52$_{\pm0.22}$ & 50.00$_{\pm0.00}$ & 63.68$_{\pm0.55}$ & 0.2342$_{\pm0.0043}$ \\
        \midrule
 
        \multicolumn{7}{l}{\textbf{Cross-subject (LOSO)\,---\,Linear probing}} \\
        \addlinespace[2pt]
        \multicolumn{7}{l}{\hspace{0.8em}\textit{Foundation models}} \\
        BENDR & 54.78$_{\pm0.12}$ & 34.69$_{\pm 0.10}$ & 57.61$_{\pm 4.18}$ & 51.51$_{\pm 1.57}$ & 54.87$_{\pm0.75}$ & 0.3068$_{\pm 0.0017}$ \\
        BIOT-6D & 60.59$_{\pm0.19}$ & 51.08$_{\pm 0.77}$ & 68.31$_{\pm 3.34}$ & 66.45$_{\pm 1.24}$ & 48.35$_{\pm3.15}$ & 0.2349$_{\pm0.0011}$ \\
        LaBraM & 62.73$_{\pm0.24}$ & $^{\bm{\ddagger}}$52.46$_{\pm 0.05}$ & 53.27$_{\pm 0.83}$ & 64.48$_{\pm 0.56}$ & 59.65$_{\pm1.31}$ & 0.2334$_{\pm 0.0010}$ \\
        Neuro-GPT & 57.95$_{\pm0.58}$ & 49.67$_{\pm0.38}$ & 72.39$_{\pm1.75}$ & 68.58$_{\pm1.89}$ & 66.20$_{\pm0.99}$ & 0.2443$_{\pm0.0035}$ \\
        EEGPT & 55.67$_{\pm0.87}$ & $^{\bm{\ddagger}}$50.04$_{\pm1.03}$ & 85.47$_{\pm4.26}$ & 59.38$_{\pm0.53}$ & 66.42$_{\pm0.49}$ & 0.2335$_{\pm0.0043}$ \\
        CBraMod & 52.56$_{\pm0.30}$ & 51.83$_{\pm 0.07}$ & 17.41$_{\pm 0.28}$ & 60.34$_{\pm 0.27}$ & 75.80$_{\pm1.19}$ & 0.2765$_{\pm 0.0020}$ \\
        TFM & 56.95$_{\pm0.46}$ & 35.28$_{\pm0.04}$ & 10.88$_{\pm0.58}$ & 62.44$_{\pm0.09}$ & 49.43$_{\pm0.95}$ & 0.2417$_{\pm0.0004}$ \\
        BrainOmni-Tiny & --- & 44.27$_{\pm0.11}$ & 73.77$_{\pm0.66}$ & 62.50$_{\pm0.50}$ & 60.75$_{\pm0.25}$ & 0.2447$_{\pm0.0012}$ \\
        BrainOmni-Base & --- & 44.06$_{\pm0.26}$ & 24.94$_{\pm0.15}$ & 50.40$_{\pm0.69}$ & 63.85$_{\pm0.95}$ & 0.2360$_{\pm0.0032}$ \\
        EEGMamba & 51.96$_{\pm0.13}$ & 51.90$_{\pm0.15}$ & 17.82$_{\pm0.03}$ & 50.06$_{\pm0.09}$ & 73.40$_{\pm0.29}$ & 0.2343$_{\pm0.0002}$ \\
        SingLEM & 34.77$_{\pm0.13}$ & 35.69$_{\pm0.04}$ & 17.90$_{\pm0.22}$ & 50.40$_{\pm0.43}$ & 63.12$_{\pm0.48}$ & 0.2632$_{\pm0.0006}$ \\
        LUNA-Base & 58.14$_{\pm0.22}$ & 42.77$_{\pm0.22}$ & 9.34$_{\pm0.34}$ & 49.97$_{\pm0.39}$ & 49.20$_{\pm0.32}$ & 0.2367$_{\pm0.0013}$ \\
        \midrule
 
        \multicolumn{7}{l}{\textbf{Within-subject (Few-shot)\,---\,Full fine-tuning}} \\
        \addlinespace[2pt]
        \multicolumn{7}{l}{\hspace{0.8em}\textit{Specialist models}} \\
        Traditional ML & 59.00 & 53.38 & \textbf{98.77} & \textbf{95.60} & --- & 0.1764$_{\pm0.0013}$ \\
        EEGNet & 48.29$_{\pm0.54}$ & 52.12$_{\pm 0.47}$ & 66.67$_{\pm 4.00}$ & 60.57$_{\pm 1.29}$ & \textbf{89.25}$_{\pm0.60}$ & 0.2082$_{\pm 0.0045}$ \\
        ShallowConv & 55.29$_{\pm0.49}$ & 51.97$_{\pm 0.26}$ & 51.23$_{\pm 2.02}$ & 69.37$_{\pm 1.43}$ & 64.52$_{\pm0.56}$ & 0.3839$_{\pm 0.0063}$ \\
        LMDA & 46.75$_{\pm2.12}$ & 53.20$_{\pm 1.39}$ & 49.49$_{\pm 1.96}$ & 54.78$_{\pm 0.22}$ & 84.53$_{\pm0.94}$ & 0.2027$_{\pm 0.0110}$ \\
        CNN-T & \underline{64.77}$_{\pm0.61}$ & 51.95$_{\pm 1.06}$ & 61.63$_{\pm 1.19}$ & 51.08$_{\pm 0.95}$ & 57.50$_{\pm0.94}$ & \underline{0.1538}$_{\pm 0.0100}$ \\
        Deformer & 52.26$_{\pm0.38}$ & 52.19$_{\pm 0.30}$ & 71.60$_{\pm 1.90}$ & 52.01$_{\pm 0.39}$ & 82.95$_{\pm0.72}$ & 0.2902$_{\pm 0.0167}$ \\
        Conformer & 63.31$_{\pm0.48}$ & 55.67$_{\pm 1.63}$ & 41.36$_{\pm 3.07}$ & 68.33$_{\pm 1.71}$ & 59.90$_{\pm0.19}$ & \textbf{0.1421}$_{\pm 0.0034}$ \\
        TSception & 60.31$_{\pm0.29}$ & 52.70$_{\pm1.21}$ & 72.74$_{\pm 2.79}$ & \underline{89.12}$_{\pm 0.95}$ & 63.23$_{\pm0.86}$ & 0.1851$_{\pm 0.0063}$ \\
        \addlinespace[4pt]
        \multicolumn{7}{l}{\hspace{0.8em}\textit{Foundation models}} \\
        BENDR & 37.34$_{\pm0.24}$ & 41.03$_{\pm 0.54}$ & \underline{85.91}$_{\pm 0.29}$ & 52.16$_{\pm 0.22}$ & 65.85$_{\pm0.49}$ & 0.2436$_{\pm0.0023}$ \\
        BIOT-6D & 61.75$_{\pm0.48}$ & 48.41$_{\pm 0.56}$ & 84.67$_{\pm 1.48}$ & 86.11$_{\pm 3.16}$ & 49.22$_{\pm2.46}$ & 0.2230$_{\pm0.0414}$ \\
        LaBraM & 35.99$_{\pm0.23}$ & $^{\bm{\ddagger}}$47.00$_{\pm 0.92}$ & 77.88$_{\pm 2.14}$ & 65.74$_{\pm 1.50}$ & 83.75$_{\pm0.74}$ & 0.1956$_{\pm0.0017}$ \\
        Neuro-GPT & 54.50$_{\pm0.29}$ & \textbf{55.90}$_{\pm 0.09}$ & 76.44$_{\pm 1.48}$ & 71.22$_{\pm3.69}$ & 81.02$_{\pm1.10}$ & 0.1880$_{\pm0.0035}$ \\
        EEGPT & 56.38$_{\pm0.16}$ & $^{\bm{\ddagger}}$40.48$_{\pm 0.44}$ & 75.72$_{\pm 8.11}$ & 65.59$_{\pm2.51}$ & 64.35$_{\pm1.28}$ & 0.1990$_{\pm0.0007}$ \\
        CBraMod & 57.72$_{\pm0.08}$ & \underline{55.82}$_{\pm 0.39}$ & 62.35$_{\pm 1.15}$ & 79.32$_{\pm 1.15}$ & 84.83$_{\pm0.43}$ & 0.2051$_{\pm0.0036}$ \\
        TFM & 58.82$_{\pm0.14}$ & 35.40$_{\pm0.15}$ & 11.73$_{\pm1.26}$ & 77.55$_{\pm0.95}$ & 50.40$_{\pm0.55}$ & 0.2208$_{\pm0.0058}$ \\
        BrainOmni-Tiny & --- & 46.74$_{\pm 0.39}$ & 82.00$_{\pm 0.52}$ & 58.72$_{\pm0.79}$ & 67.95$_{\pm0.25}$ & 0.2146$_{\pm0.0089}$ \\
        BrainOmni-Base & --- & 44.41$_{\pm0.12}$ & 52.88$_{\pm1.29}$ & 51.08$_{\pm1.86}$ & 67.70$_{\pm0.21}$ & 0.1923$_{\pm0.0062}$ \\
        EEGMamba & 61.50$_{\pm0.62}$ & 51.33$_{\pm0.27}$ & 44.03$_{\pm0.15}$ & 50.08$_{\pm0.11}$ & \underline{86.12}$_{\pm0.65}$ & 0.1774$_{\pm0.0018}$ \\
        SingLEM & 29.71$_{\pm1.75}$ & 47.01$_{\pm2.01}$ & 33.02$_{\pm2.63}$ & 50.46$_{\pm0.19}$ & 52.00$_{\pm1.27}$ & 0.2271$_{\pm0.0021}$ \\
        LUNA-Base & \textbf{66.02}$_{\pm0.06}$ & 46.06$_{\pm0.68}$ & 8.95$_{\pm0.50}$ & 50.62$_{\pm0.44}$ & 51.87$_{\pm0.80}$ & 0.1676$_{\pm0.0034}$ \\
        \midrule
 
        \multicolumn{7}{l}{\textbf{Within-subject (Few-shot)\,---\,Linear probing}} \\
        \addlinespace[2pt]
        \multicolumn{7}{l}{\hspace{0.8em}\textit{Foundation models}} \\
        BENDR & 21.58$_{\pm0.06}$ & 34.00$_{\pm 0.16}$ & 29.12$_{\pm 2.45}$ & 49.46$_{\pm 1.52}$ & 56.87$_{\pm1.17}$ & 0.2271$_{\pm0.0015}$ \\
        BIOT-6D & 59.42$_{\pm0.10}$ & 48.10$_{\pm 0.65}$ & 78.19$_{\pm 5.09}$ & 77.08$_{\pm 3.95}$ & 51.83$_{\pm0.53}$ & 0.2937$_{\pm0.0419}$ \\
        LaBraM & 49.20$_{\pm0.10}$ & $^{\bm{\ddagger}}$49.46$_{\pm 0.14}$ & 66.98$_{\pm 0.50}$ & 70.22$_{\pm 2.94}$ & 60.67$_{\pm0.47}$ & 0.1935$_{\pm0.0020}$ \\
        Neuro-GPT & 49.69$_{\pm0.62}$ & 52.03$_{\pm0.10}$ & 52.67$_{\pm1.24}$ & 70.60$_{\pm1.43}$ & 65.37$_{\pm0.33}$ & 0.2015$_{\pm0.0051}$ \\
        EEGPT & 61.99$_{\pm0.47}$ & $^{\bm{\ddagger}}$42.90$_{\pm0.42}$ & 77.78$_{\pm3.81}$ & 67.05$_{\pm1.22}$ & 66.30$_{\pm1.97}$ & 0.2004$_{\pm0.0062}$ \\
        CBraMod & 41.33$_{\pm0.26}$ & 52.32$_{\pm 0.39}$ & 24.28$_{\pm 1.77}$ & 60.49$_{\pm 2.25}$ & 81.67$_{\pm0.38}$ & 0.1895$_{\pm0.0075}$ \\
        TFM & 49.56$_{\pm0.26}$ & 34.28$_{\pm0.44}$ & 10.08$_{\pm1.72}$ & 58.26$_{\pm6.15}$ & 49.77$_{\pm1.25}$ & 0.2208$_{\pm0.0058}$ \\
        BrainOmni-Tiny & --- & 43.54$_{\pm0.14}$ & 51.95$_{\pm1.39}$ & 59.41$_{\pm1.22}$ & 63.77$_{\pm0.50}$ & 0.2202$_{\pm0.0035}$ \\
        BrainOmni-Base & --- & 43.62$_{\pm0.15}$ & 23.56$_{\pm2.52}$ & 52.16$_{\pm1.47}$ & 63.27$_{\pm0.72}$ & 0.1861$_{\pm0.0031}$ \\
        EEGMamba & 46.30$_{\pm0.18}$ & 49.36$_{\pm0.03}$ & 20.99$_{\pm0.50}$ & 50.15$_{\pm0.22}$ & 78.50$_{\pm0.33}$ & 0.1712$_{\pm0.0014}$ \\
        SingLEM & 22.51$_{\pm0.24}$ & 34.62$_{\pm0.05}$ & 19.14$_{\pm0.91}$ & 49.69$_{\pm1.42}$ & 75.90$_{\pm0.53}$ & 0.2752$_{\pm0.0075}$ \\
        LUNA-Base & 61.99$_{\pm0.07}$ & 42.27$_{\pm0.19}$ & 9.47$_{\pm1.72}$ & 50.08$_{\pm0.11}$ & 50.43$_{\pm0.56}$ & 0.1654$_{\pm0.0009}$ \\
        \bottomrule
    \end{tabular}
    }
\end{table*}

\subsubsection{Can FMs learn generalized representations?}

Tables~\ref{tab:bench_1} and \ref{tab:bench_2} present the performance of 12 EEG FMs across 13 downstream datasets. A notable observation is that head-only fine-tuning (linear probing) consistently yielded inferior, and in many cases substantially lower, performance compared to full-parameter fine-tuning for most FMs. This finding suggested that adapting FMs to diverse downstream tasks cannot rely solely on features extracted by pre-trained encoders; task-specific fine-tuning of the encoder parameters remains essential. While several models such as EEGPT exhibited superior head-only performance relative to full fine-tuning, neither adaptation strategy achieved consistently strong results across the benchmark.

EEG FMs also exhibited considerable variability in their task-specific performance. For instance, CBraMod demonstrated competitive results across most tasks, achieving first and second place on the SEED dataset under LOSO and few-shot scenarios, respectively. However, it yielded the highest RMSE among all evaluated methods on SEED-VIG in the LOSO scenario. Similarly, LUNA attained state-of-the-art performance on TUAB but failed to generalize effectively to paradigms beyond clinical applications, a limitation likely attributable to its pre-training exclusively on TUEG and Siena datasets.

An encouraging finding emerged from the Nakanishi2015 dataset, an SSVEP paradigm with extremely limited fine-tuning data (12 trials per class). Despite this constraint, several FMs, including BENDR, EEGPT, and Neuro-GPT, achieved strong performance. Since the SSVEP paradigm relies on decoding neural responses to target stimuli flickering at distinct frequencies, the resulting signals exhibit pronounced periodicity and temporal structure. Consequently, the masked reconstruction objectives employed during pre-training may endow these models with enhanced capability to capture temporal dynamics of EEG signals.

\begin{figure*}[t]
\includegraphics[width=\linewidth,clip]{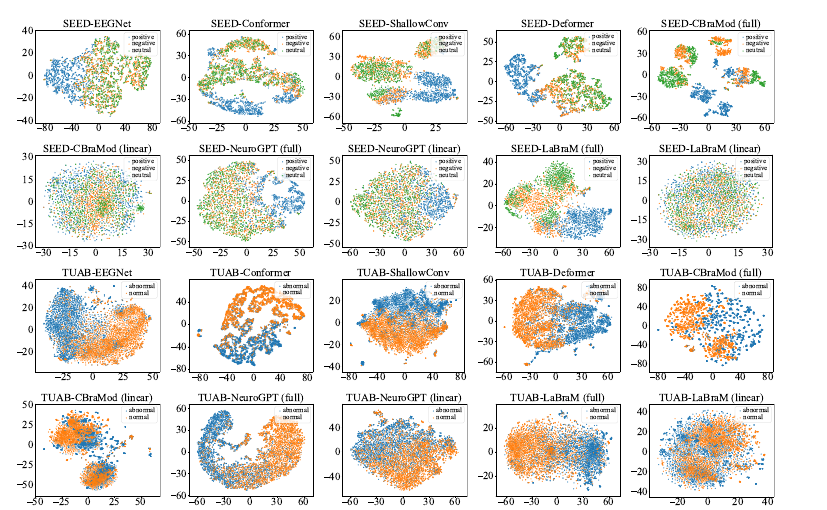}
\caption{$t$-SNE visualization of the SEED and TUAB datasets.} \label{fig:tsne_rep}
\end{figure*}

To qualitatively examine the separability of the learned representations, we visualized the feature embeddings using $t$-SNE. We considered three representative datasets, SEED, TUAB, and BNCI2014008, corresponding to emotion recognition, clinical abnormality detection, and ERP/RSVP decoding, respectively. For compactness, Fig.~\ref{fig:tsne_rep} presents a representative subset, namely four top-performing specialist models (EEGNet, Conformer, ShallowConv, and Deformer) and three EEG FMs (CBraMod, Neuro-GPT, and LaBraM) on SEED and TUAB, with both full-parameter fine-tuning and linear probing shown for each FM. The complete visualizations are provided in Section~A5 of the supplementary information. Full-parameter fine-tuning generally yields more discriminative structures than linear probing, with clearer between-class separation and more compact within-class clusters. This qualitative evidence is consistent with our quantitative results, indicating that the frozen representations of current EEG FMs are not always sufficiently separable for downstream decoding.

We further conducted a scenario-level paired Wilcoxon signed-rank analysis to quantify whether full-parameter adaptation consistently improves over linear probing, with detailed results reported in Table~55 of the supplementary information. The results show that full fine-tuning provides a statistically significant advantage for the majority of EEG FMs. Specifically, 9 out of 12 FMs exhibited significant positive median $\Delta$ values, indicating that full fine-tuning outperformed linear probing across dataset--scenario pairs. Several models showed particularly strong directional consistency: BENDR and CBraMod favored full fine-tuning in 23 out of 25 scenarios, while TFM favored full fine-tuning in 24 scenarios with one tie. In contrast, EEGPT was the only model for which linear probing showed a significant advantage, whereas BrainOmni-Tiny and SingLEM did not exhibit significant differences between the two adaptation strategies. These findings reinforce our conclusion that current EEG FMs have not yet learned sufficiently universal representations for robust downstream decoding. Although pre-training provides useful initialization, task-specific full-parameter adaptation remains important for most EEG FMs.

In summary, pre-trained EEG FMs demonstrated a capacity to extract transferable representations to a certain extent. However, this generalization capability remained insufficiently robust: the majority of models required full-parameter fine-tuning on downstream tasks and could not directly leverage pre-trained encoders to obtain features for effective decoding. Even the best-performing FMs in our evaluation exhibited notable performance degradation on specific tasks, indicating that achieving truly universal EEG representations remains an open challenge.

\subsubsection{Can FMs consistently outperform specialist models?}

\begin{figure*}[t]
\includegraphics[width=\linewidth,clip]{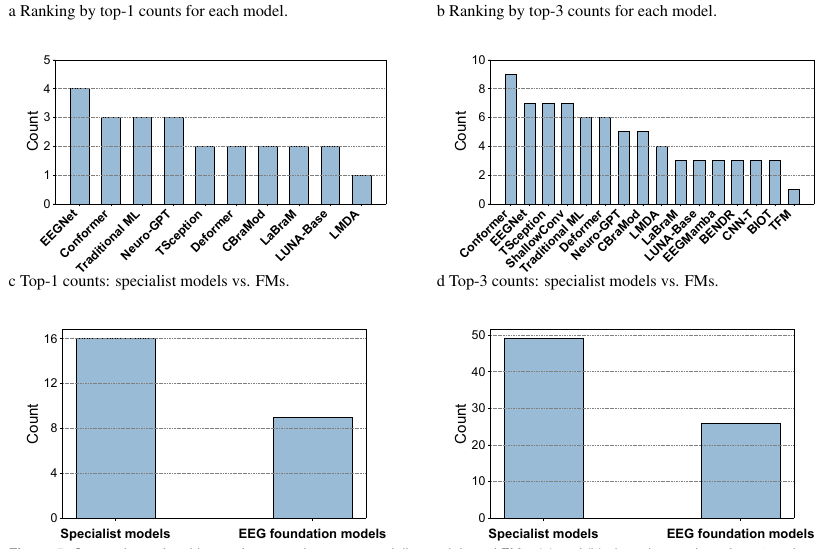}
\caption{Comparison of ranking performance between specialist models and FMs. (a) and (b) show the number of top-1 and top-3 placements for individual models across all tasks and scenarios. (c) and (d) compare the aggregate top-1 and top-3 counts between specialist models and the seven best-performing FMs.} \label{fig:rank1}
\end{figure*}

With the proliferation of EEG FMs, whether traditional deep learning methods remain competitive is a question worth investigating. We compared existing FMs against eight specialist models, including one traditional machine learning method, four CNN-based methods, and three Transformer-based methods. These specialist models were trained from scratch using the same fine-tuning data as the FMs and evaluated on identical test sets. Fig.~\ref{fig:rank1}a and b present the ranking of each model based on the number of top-1 and top-3 placements across all tasks and scenarios. Notably, EEGNet achieved the highest number of top-1 placements, demonstrating remarkable performance despite having only 2K parameters. ShallowConv obtained the highest number of top-3 placements. Among the top five models in both Fig.~\ref{fig:rank1}a and b, four were specialist models. Specifically, EEGNet, Conformer, Traditional ML, and Deformer ranked 1st, 2nd, 4th, and 5th in top-1 counts, respectively, while ShallowConv, EEGNet, Conformer, and Deformer ranked 1st to 4th in top-3 counts.

Furthermore, Fig.~\ref{fig:rank1}c and d compare the total number of top-1 and top-3 placements achieved by specialist models versus EEG FMs. To ensure a fair comparison given the larger number of FMs, we selected the seven FMs with the highest top-3 counts for this analysis. Specialist models achieved 16 first-place finishes and 47 top-3 placements, outperforming the selected FMs.

Overall, specialist models achieved higher average decoding accuracy than EEG FMs. A closer examination of the two evaluation scenarios reveals a more nuanced pattern. In the within-subject few-shot scenario, specialist models obtained the majority of top-1 placements, whereas in the LOSO scenario the two groups performed comparably. This contrast suggests that current EEG FMs rely on a relatively large amount of fine-tuning data to realize their advantage, and that their benefit diminishes when only limited calibration data are available. We emphasize, however, that both scenarios correspond to realistic deployment settings: the LOSO scenario reflects applications with access to a labeled multi-subject corpus, while the few-shot scenario reflects rapid personalization for new users with minimal calibration. 

Beyond accuracy, the two groups also differ markedly in computational cost. As measured in Section~A4 of the supplementary information, the fine-tuning cost of EEG FMs is orders of magnitude higher than that of lightweight specialist models. For example, under the 4s, 10-channel setting, EEGPT requires approximately 361~GFLOPs per forward pass, more than three orders of magnitude larger than the 0.35~GFLOPs of EEGNet, and its fine-tuning throughput is about 46 times lower. This substantial overhead may limit the practicality of large EEG FMs in resource-constrained or real-time BCI applications. Taken together with the accuracy comparison, these results indicate that lightweight specialist architectures remain highly competitive in the era of FMs, particularly when computational efficiency is a critical deployment consideration.

To statistically verify whether EEG FMs consistently outperform specialist models, we performed a rank-based analysis using the Friedman test followed by the Nemenyi post-hoc test, with the critical difference (CD) diagram~\cite{demvsar2006statistical} reported in Figure~1 of the supplementary information. The Friedman test revealed significant overall differences among the model rankings, confirming that the observed ranking structure was not due to random variation. Although CBraMod attained the best average rank among all compared models, the Nemenyi post-hoc test showed that its difference from several top-ranked specialist models, including EEGNet, ShallowConv, Conformer, Deformer, and LMDA, was not statistically significant; these specialist architectures belonged to the same non-significant group as the best-performing FM. This provides statistical support for our observation that current EEG FMs do not consistently and significantly outperform competitive specialist models trained from scratch, and that specialist architectures remain highly competitive under the present data regimes and evaluation scenarios.





\subsubsection{Do EEG FMs exhibit behavior consistent with scaling laws?}

\begin{figure*}[htpb]
\centering
\includegraphics[width=\linewidth,clip]{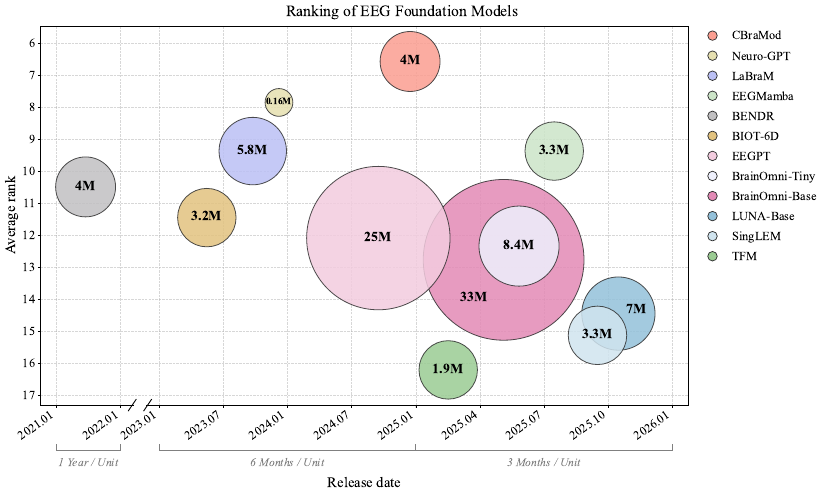}
\caption{Overall ranking of EEG FMs with respect to release date and model size (bubble size indicates parameter count; lower rank is better).}
\label{fig:bubble}
\end{figure*}

Fig.~\ref{fig:bubble} presents the average rankings of EEG FMs across the 13 datasets under both the LOSO and few-shot scenarios. Detailed ranking results are provided in Table~49 of the supplementary information. CBraMod achieved the highest overall ranking with an average rank of 6.56 and 4.0M parameters. EEGNet, a widely adopted lightweight EEG decoding backbone, attained the second position with only 2K parameters. Among the EEG FMs, the top six overall, CBraMod (4.0M), Neuro-GPT (0.16M), LaBraM (5.8M), EEGMamba (3.3M), BENDR (4.0M), and BIOT-6D (3.2M), exhibit no clear correlation between model size and downstream performance.

These observations indicate that, under current data regimes and training practices, increasing the parameter count of an EEG FM does not yet guarantee better downstream generalization, and the scaling behavior commonly observed in vision and language foundation models has not been clearly established for EEG. This may be attributed to two factors. First, EEG data acquisition is costly in terms of time, labor, and resources, resulting in limited data availability and substantial measurement noise, with a notable shortage of large-scale, high-quality corpora~\cite{liu2025clean} that may be insufficient to elicit emergent capabilities. Second, the pre-training objectives currently adopted by EEG FMs may not effectively translate increased model capacity into richer representations, so that the performance gains from enlarging the model are weaker and less consistent than the steady improvements predicted by scaling laws. Constructing large-scale, high-quality EEG corpora and developing pre-training strategies that can better exploit larger model capacity therefore remain essential directions for future research, and whether scaling laws can emerge for EEG FMs is an open question worth further investigation.

\subsection{Comparison of Different Fine-Tuning Ratios}

\begin{figure*}[htpb]
\centering
\includegraphics[width=\linewidth,clip]{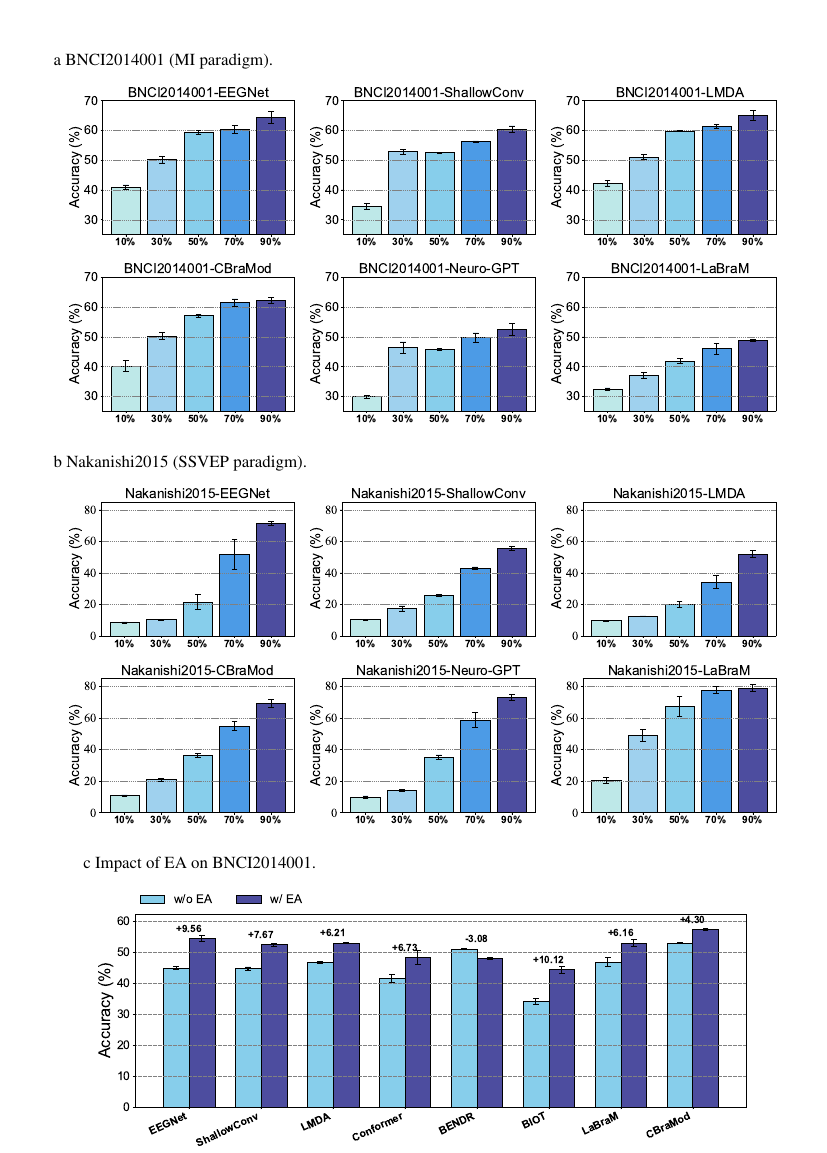}
\caption{Performance comparison across different fine-tuning data ratios and the impact of EA. (a) and (b) Accuracy comparison on the BNCI2014001 dataset (MI paradigm) and the Nakanishi2015 dataset (SSVEP paradigm) with fine-tuning ratios varying from 10\% to 90\%. (c) Impact of EA on model performance on the BNCI2014001 dataset.} \label{fig:fint_ratio}
\end{figure*}

In the within-subject few-shot scenario, a pre-trained model is expected to achieve satisfactory performance with minimal calibration data. However, many models exhibited suboptimal performance under this setting. To investigate whether this limitation is primarily attributable to insufficient fine-tuning data, we conducted an analysis across different fine-tuning data ratios, varying from 10\% to 90\% in increments of 20\%. We selected three specialist models (EEGNet, ShallowConv, and LMDA) and the top three EEG FMs (CBraMod, Neuro-GPT, and LaBraM). The results are presented in Fig.~\ref{fig:fint_ratio}, with detailed results for all models across various datasets provided in Section~A4 of the supplementary information. Most models exhibited consistent performance improvements as the amount of fine-tuning data increased, which aligns with intuitive expectations. However, minimal calibration or even calibration-free adaptation remains a critical requirement for practical deployment. Developing models that can rapidly adapt to downstream tasks with limited data remains an open and pressing challenge.







\section{DISCUSSION} \label{sect:discussion}

This section presents additional discussions.

\subsection{Pre-training Objectives vs. Downstream Performance}

A natural question following the taxonomy in Section `Overview OF EEG FMs' and the benchmark results in Section `Benchmark of EEG FMs' is whether specific categories of pre-training objectives systematically yield superior downstream performance, and whether they exhibit paradigm-specific advantages. To address this question, we organize the 12 evaluated FMs by their pre-training objectives in Table 50 of the supplementary information, and analyze their performance across the 13 downstream datasets.

Several observations can be drawn from this analysis. First, no single category of pre-training objective dominates the overall ranking. Models adopting raw-signal reconstruction (CBraMod, EEGMamba, EEGPT) span ranks from 1 to 15, while models adopting embedded-token reconstruction or prediction (Neuro-GPT, BENDR, BIOT-6D, LUNA-Base) span ranks from 7 to 18. Codebook-based objectives, including LaBraM, BrainOmni-Tiny, BrainOmni-Base, and TFM, span ranks from 9 to 20. This dispersion suggests that the choice of reconstruction or prediction target alone is not a decisive factor for downstream transferability; rather, the combined effect of pre-training corpus scale, data curation, masking strategy, and architectural inductive bias appears to be more influential. Second, the top-ranked FM in each major objective category, namely CBraMod (raw signals, 4.0M), Neuro-GPT (embedded tokens, 0.16M), and LaBraM (codebook index, 5.8M), achieves a competitive overall rank. This indicates that different objective families can yield useful transferable representations when paired with appropriate design choices. Third, within the codebook-based family, LaBraM, BrainOmni-Tiny, BrainOmni-Base, and TFM obtain average ranks of 9.36, 12.33, 12.76, and 16.20, respectively, which do not exhibit a monotonic trend with respect to model size. This is consistent with the broader observation in Section `Benchmark of EEG FMs' that scaling alone does not currently guarantee improved generalization.

A further question of practical interest is whether specific pre-training objectives are systematically advantageous for specific BCI paradigms. The per-subject results in Tables~13--48 of the supplementary information do not reveal a consistent pairing between objective category and paradigm. For instance, although SSVEP signals are inherently frequency-coded, FMs employing explicit frequency-domain supervision are not among the top performers on Nakanishi2015. Instead, BENDR, EEGPT, and Neuro-GPT, which adopt time-domain or token-level objectives, achieve the strongest performance on this dataset. Similarly, on clinical paradigms such as TUAB and CHB-MIT, both raw-signal and embedded-token objectives can attain competitive performance, with no single family demonstrating a consistent advantage. These findings suggest that, under current data regimes and training practices, the alignment between pre-training objective and paradigm-specific signal structure is weaker than might be expected.

In summary, the reconstruction objective adopted by existing EEG FMs is not a decisive factor for downstream performance. Developing more reliable pre-training strategies that can extract universal and transferable EEG representations is therefore of greater importance than the choice of reconstruction target itself, and remains an essential direction for future research.

\subsection{Paradigm-Specific FMs}

In real-world applications, the paradigm for a downstream task is generally determined prior to data collection. For example, stroke patients requiring exoskeleton-assisted rehabilitation are naturally suited to MI-based systems~\cite{li2024survey}, while epilepsy monitoring demands epilepsy-specific approaches~\cite{hermann2017paradigm}. Therefore, when user information such as patient demographics and target applications is available, employing a paradigm-specific FM for direct adaptation represents a practical and effective strategy.

In recent years, several researchers have attempted to develop paradigm-specific FMs tailored to particular tasks, such as MEET~\cite{shi2023meet} for emotion recognition, MIRepNet~\cite{liu2026mirepnet} for MI, PSGFM~\cite{coon2025psgfm} for sleep staging, and EpilepsyFM~\cite{li2025epilepsyfm} for epilepsy detection. Among these, we compared MIRepNet, which provides open source pre-trained weights, against existing general-purpose FMs on MI tasks. Section~A4 of the supplementary information presents detailed results on three MI datasets: BNCI2014001, BNCI2014004, and BNCI2015001. MIRepNet achieved state-of-the-art performance in terms of both subject-averaged accuracy and Cohen's Kappa. This superior performance may be attributed to the fact that paradigm-specific FMs are pre-trained exclusively on datasets from the target task and incorporate neurophysiological principles relevant to that paradigm in their pre-training strategies. Consequently, the pre-trained encoder is capable of extracting task-specific representations that facilitate rapid adaptation to downstream applications.

Given that the required paradigm is typically known before data acquisition in practical BCI deployment, developing paradigm-specific pre-trained FMs represents a viable and promising research direction. Furthermore, whether auxiliary data from other paradigms can enhance pre-training for a target paradigm remains an open question worthy of further investigation.

\subsection{Effectiveness of EA}

As mentioned in Section `Data preprocessing', EA~\cite{he2019transfer,drwuEA2025} aligns the marginal distributions across EEG trials. We compared model performance with and without EA on the BNCI2014001 dataset. As shown in Fig.~\ref{fig:fint_ratio}, incorporating EA during training or fine-tuning improved generalization performance for the majority of models.

EA normalizes the average trial covariance of each subject to the identity matrix, thereby reducing inter-subject discrepancies in second-order statistics. From a Riemannian geometry perspective, the covariance matrices of different subjects occupy distinct regions on the symmetric positive-definite (SPD) manifold. EA re-centers each subject's covariance distribution to the identity matrix, which serves as the canonical reference point on the SPD manifold, so that the aligned trials of all subjects share a common distribution after being mapped back to the Euclidean space.

To ensure a fair evaluation, EA was applied only to the FMs whose original implementations adopt it, namely EEGPT and MIRepNet; the remaining models were evaluated without EA. This is because applying EA during fine-tuning while it was absent during pre-training, or vice versa, would introduce a mismatch between the aligned and unaligned representation spaces, thereby confounding the comparison.

Despite its effectiveness, EA has potential limitations. As a whitening transformation, EA mixes the original channels into linear combinations, which may distort the spatial structure of the electrode layout. This could be detrimental to models that explicitly rely on spatial information, such as those employing electrode coordinate encoding or graph-based representations, and may also reduce the interpretability of the learned spatial patterns. The trade-off between the cross-subject alignment provided by EA and the preservation of spatial structure therefore warrants careful consideration when applying EA to spatially aware EEG FMs.

\subsection{Future Research Directions}

\subsubsection{Large-scale high-quality data construction}
Non-invasive EEG signals inherently suffer from low signal-to-noise ratios due to hardware limitations, environmental interference, and variations in subject attention during acquisition. Several existing approaches attempt to reconstruct raw signals during pre-training, which may inadvertently encourage models to fit noise patterns rather than learn generalizable representations beneficial for downstream tasks. Furthermore, current EEG FMs do not exhibit scaling law behavior, potentially due to the lack of large-scale, high-quality EEG datasets. Liu et al.~\cite{liu2025clean} demonstrated the importance of data quality in the MI paradigm by performing channel selection based on neurophysiological principles and removing low-quality subjects. Models trained on this cleaner and smaller dataset achieved superior performance compared to those trained on uncleaned data. Therefore, constructing large-scale, high-quality EEG corpus through systematic data collection and rigorous cleaning procedures across diverse paradigms represents a critical direction for future research.

\subsubsection{Paradigm-specific FMs}
As discussed above, the target paradigm is typically known prior to downstream data acquisition, making paradigm-specific FMs a practical and well-motivated approach. Recent works have explored this direction, including MEET~\cite{shi2023meet} for emotion recognition, MIRepNet~\cite{liu2026mirepnet} for MI, PSGFM~\cite{coon2025psgfm} for sleep staging, and EpilepsyFM~\cite{li2025epilepsyfm} for epilepsy detection. The results reported in Section~A4 of the supplementary information for MIRepNet demonstrate its superior performance on MI tasks. This advantage may stem from pre-training exclusively on paradigm-specific data while incorporating neurophysiological principles into the pre-training pipeline, enabling the model to learn more effective representations for the target paradigm. These findings support the validity, feasibility, and practicality of paradigm-specific FMs as a promising research direction.

\subsubsection{Efficient pre-training strategies}
Most existing approaches adopt masked reconstruction as the primary pre-training objective, targeting raw signals, frequency-domain representations, or embedded tokens. However, no single model has demonstrated consistently strong performance across all tasks. Future research should address the following challenges: 1) developing more effective solutions for cross-device heterogeneity; 2) designing pre-training strategies that enable models to learn truly universal and transferable representations; and 3) exploring efficient fine-tuning strategies that facilitate rapid adaptation to new tasks, including methods to achieve competitive performance with less calibration data and techniques to accelerate distribution alignment between pre-trained models and downstream data. As a preliminary step in the latter direction, we conducted a parameter-efficient fine-tuning experiment, reported in Section~A4 of the supplementary information.

\section{CONCLUSION} \label{sect:conclusion}

This paper has presented EEG-FM-Compass, a comprehensive benchmark for EEG FMs in BCIs. We reviewed 55 studies and distilled their common pipeline components and pre-training objectives into a unified framework that enables structured comparison across heterogeneous devices and paradigms. Based on this analysis, we established a benchmark that evaluates 12 open source EEG FMs alongside competitive specialist baselines across 13 datasets spanning 9 representative BCI paradigms, under both cross-subject LOSO and within-subject few-shot evaluation protocols.

The experimental results indicate that current EEG FMs have not yet achieved universally transferable representations. Specifically, full-parameter fine-tuning consistently provides substantial advantages over linear probing, suggesting that pre-trained encoders cannot be directly employed as fixed feature extractors across diverse downstream tasks. Furthermore, specialist models trained from scratch remain highly competitive, and increasing model size alone does not guarantee improved generalization. These findings highlight the need for future research on advancing pre-training strategies, as well as enhancing robustness to noise and cross-task heterogeneity. 

\section{DATA AVAILABILITY}

The source code will be available on GitHub \href{https://github.com/Dingkun0817/EEG-FM-Benchmark}{https://github.com/Dingkun0817/EEG-FM-Benchmark}.

\section{FUNDING}\label{funding}

This work was supported by National Natural Science Foundation of China (62525305), Major Science and Technology Project of Industry-university-research Cooperation for Colleges and Universities in Hebei Province in Shijiazhuang City (2511303107A), and Zhongguancun Academy (02012401).

\section{AUTHOR CONTRIBUTIONS}\label{contri}

D.L. conceived the study, conducted the analysis, and wrote the manuscript. Y.C. and Z.C. performed the experiments and contributed to the analysis. Z.C. and Y.W. contributed to the literature review and data collection. J.A. and J.L. revised the manuscript. D.W. supervised the study and revised the manuscript.

\vspace{\baselineskip}
\noindent\textbf{\textit{Conflict of interest statement.}} None declared.

\clearpage







\begin{appendix}

\section{SUPPLEMENTARY INFORMATION}

\subsection{SUMMARY OF EEG FMs} \label{app1_summary}

Table~\ref{tab:prepro_info} summarizes the preprocessing details adopted by the EEG FMs, while Table~\ref{tab:pretrain_info} provides an overview of their pre-training strategies and backbone architectural configurations.

\begin{table*}[htbp]
\centering
\renewcommand{\arraystretch}{1.46}
\setlength{\tabcolsep}{3.8pt}
\caption{Preprocessing of EEG FMs.}
\label{tab:prepro_info}
\scalebox{0.64}{
\begin{tabular}{c|c|c|c|c|c|c|c}
\toprule
Model & Resampling (Hz) & Filtering & Alignment & Channel Mapping & Patch & Stride & Overlap \\
\midrule
BENDR & 256 & $\leq$120 Hz (P300) & Norm $\rightarrow$ [-1,1] & 19 & 375ms & 375ms & \XSolidBrush \\
BrainBERT & --- & $\geq$0.1 Hz + 60 Hz notch & STFT + $z$-score & --- & 200ms & 25ms & \Checkmark \\
MBrain & --- & --- & Norm $\rightarrow \mathcal{N}(0, 1)$ & 19 / --- (EEG / iEEG) & 1s & 1s & \XSolidBrush \\
BIOT & 200 / 500 (EEG / ECG) & --- & Norm $\rightarrow \frac{x}{P_{95}(|x|)}$ & 16 / 12 (EEG / ECG) & 1s & 0.5s & \Checkmark \\
Brant & 250 & --- & --- & --- & 6 s & --- & --- \\
LaBraM & 200 & 0.1-75 Hz + 50 Hz notch & Norm $\rightarrow$ [-1,1] & --- & 1s & 1s & \XSolidBrush \\
Mentality & 200 & 60\&120 Hz notch & --- & 19 & 10s & 10s & \XSolidBrush \\
Neuro-GPT & 250 & 0.5-100 Hz + 60 Hz notch & $z$-score & 22 & 2s & 1.8s & \Checkmark \\
MEET & 200 & 1-50 Hz & --- & 32$\times$32 Map & --- & --- & \XSolidBrush \\
EEGFormer & 250 & --- & Instance Norm & --- & --- & --- & --- \\
BrainWave & $>$1000 $\rightarrow$ 1000 & 0.01-$\frac{f_s}{3}$ Hz + 50/60 Hz notch & --- & --- & 1 s & 1 s & \XSolidBrush \\
NeuroLM & 200 & 0.1-75 Hz + 50 Hz notch & Norm $\rightarrow$ [-1,1] & --- & 1s & 1s & \XSolidBrush \\
Brant-X & --- & $\leq$45 Hz (EM) & $z$-score (only EM) & --- & --- & --- & \XSolidBrush \\
FoME & 250 & 0.5-100.5 Hz + 50 / 60 Hz notch & EMA & --- & 6s & 6s & \XSolidBrush \\
EEGPT & 256 & $\leq$38 Hz / $\leq$30 Hz / $\leq$120 Hz & EA / CAR / $z$-score & --- & 250ms & 250ms & \XSolidBrush \\
BrainGPT & 256 & 0.1-100 Hz & $z$-score & --- & 1s & 125ms & \Checkmark \\
GEFM & 256 & --- & --- & 19 & --- & --- & --- \\
CBraMod & 200 & 0.3-75 Hz + 60 Hz notch & Norm $\rightarrow$ [-1,1] & 19 & 1s & 1s & \XSolidBrush \\
CEReBrO & --- & --- & --- & 64 & L = 64 & S = 64 & \XSolidBrush \\
LEAD & 128 / 64 / 32 & 0.5-45 Hz & $z$-score & 19 & L = 128 & S = 64 & \Checkmark \\
FEMBA & 250 & --- & Quantile Norm & 22 & 128 ms & --- & --- \\
LCM & 256 & 0-38 Hz (MI) & CAR & --- & --- & --- & --- \\
TFM & 200 & 0.1-75 Hz + 50 Hz notch (TUH) & --- & 16 & 1s & 0.5s & \Checkmark \\
Gram & 200 & 0.1-75 Hz + 50 / 60 Hz notch & Norm $\rightarrow$ [-1,1] & 130 & 1s & 1s & \XSolidBrush \\
ALFEE & 256 & 50 / 60 Hz notch & $z$-score & 90 & 1s & 1s & \XSolidBrush \\
BrainOmni & 256 & 0.1-96 Hz + 50 / 60 Hz notch & $z$-score & 16 & 2s  & 1s & \Checkmark \\
E3GT & 125 & 0.1-50 Hz & CAR & 8 & 4s & 1s & \Checkmark \\
CodeBrain & 200 & 0.3-75 Hz + 60 Hz notch & Norm $\rightarrow \frac{x}{100}$ & 19 & 1s & 1s & \XSolidBrush \\
DIVER-0 & 200 & 0.3-75 Hz + 60 Hz notch & --- & 19 & 1s & 1s & \XSolidBrush \\
UniMind & 200 & 0.1-75 Hz + 50 / 60 Hz notch & $z$-score & --- & --- & --- & \XSolidBrush \\
CSBrain & 200 & 0.3-75 Hz + 60 Hz notch & Norm $\rightarrow$ [-1,1] & 19 & NA & NA & NA \\
DMAE-EEG & --- & --- & --- & 10 Regions  & --- & --- & \XSolidBrush \\
EEGMamba & 200 & 0.3-75 Hz + 50 / 60 Hz notch & Norm $\rightarrow$ [-1,1] & --- & 1s & 1s & \XSolidBrush \\
MIRepNet & 250 & 8-32 Hz & Zero-mean + EA & 45 & NA & NA & NA \\
PSGFM & 100 & --- & IQR Scaling & 1 & 30s & 30s & \XSolidBrush \\
EEGDM & 200 & 0.1-75 Hz + 50 Hz notch & Norm $\rightarrow$ [-1,1] & --- & NA & NA & NA \\
CoMET & 200 & 0.5-70 Hz & Norm $\rightarrow$ [-1,1] & 62 & 250ms & 250ms & \XSolidBrush \\
EpilepsyFM & 200 & 0.5-70 Hz + 50 Hz notch & $z$-score & 8 Regions & 1s & 1s & \XSolidBrush \\
SingLEM & 128 & 0.5-50 Hz + 50 Hz notch & Norm $\rightarrow$ (-1,1) & 1 & 1s & 0.75s & \Checkmark \\
BrainPro & 200 & --- & Norm $\rightarrow$ [-1,1] & 60 & 0.1s & 0.1s & \XSolidBrush \\
Uni-NTFM & 200 & 0.5-50 Hz + 50 Hz notch & --- & 5 Regions & NA & NA & NA \\
ELASTIQ & 200 & 0.3-40 Hz (MI) / 0.3-70 Hz & --- & 65 & 0.5s & 0.5s & \XSolidBrush \\
BioCodec & 250 / 1000 (EEG / EMG) & 0.5-100 Hz & $z$-score & 1 & NA & NA & NA \\
HEAR & 200 & 1-75 Hz & CAR & --- & --- & --- & \XSolidBrush \\
NeuroRVQ & 200 & --- & --- & --- & 1s & --- & --- \\
NeurIPT & 64 & 0.1-30 Hz + 50 Hz notch & A-Law Companding & 20 & 4s & 4s & \XSolidBrush \\
REVE & 200 & 0.5-99.5 Hz & $z$-score & --- & 1s & 0.1s & \Checkmark \\
mdJPT & 125 & 0.5-47 Hz & ICA + CAR & 60 & 5s & 2s & \Checkmark \\
LUNA & 256 & 0.1-75 Hz + 50 / 60 Hz notch & $z$-score & --- & L = 40 & L = 40 & \XSolidBrush \\
THD-BAR & 200 & 0.1-75 Hz + 50 / 60 Hz notch & IQR Scaling & --- & 1s & 1s & \XSolidBrush \\
EEG-X & 128 & --- & --- & --- & 1s & 0.75s & \Checkmark \\
SAMBA & --- & --- & --- & --- & --- & --- & --- \\
DeeperBrain & 200 & 0.3-75 Hz + 50 / 60 Hz notch & Norm $\rightarrow$ [-1,1] & --- & 1 s & 1 s & \XSolidBrush \\
EEGMoE & --- & 1-45 Hz & Gaussian Filter / $z$-score & 9$\times$9 Map & --- & --- & \XSolidBrush \\
Brain-OF & 200 & 1-80 Hz + 50 / 60 Hz notch & $z$-score + CAR & --- & L = 64 & L = 64 & \XSolidBrush \\
\bottomrule
\end{tabular}
}
\end{table*}

\begin{table*}[htbp]
\centering
\renewcommand{\arraystretch}{1.85}
\setlength{\tabcolsep}{1.7pt}
\caption{Pre-training strategy of EEG FMs.}
\label{tab:pretrain_info}
\scalebox{0.5}{
\begin{tabular}{c|c|c|c|c|c|c|c}
\toprule
Model & Masking strategy & Reconstruction objective & Loss function & Encoder depth & Attn-head & d\_model & FFN \\
\midrule
BENDR & Random Mask & Embedded Tokens & $\mathcal{L}_{cl}^{et}$ & 8 & 8 & 1536 & 3076 \\
BrainBERT & Random Mask & Spectrogram & $\mathcal{L}_{mse}^{spec}$ & 6 & 12 & 768 & --- \\
MBrain & Random Mask & NA & NA & NA & NA & NA & NA \\
BIOT & Random Mask & Embedded Tokens & $\mathcal{L}_{cl}^{et}$ & 4 & 8 & 256 & 1024 \\
Brant & Random Mask & Raw Signals & $\mathcal{L}_{mse}^{rs}$ & 17 & 16 & 2048 & 3072 \\
LaBraM & Random Mask & EEG Codebook Index & $\mathcal{L}_{cls}^{ci}$ & 12 / 24 / 48 & 10 / 16 / 16 & 200 / 400 / 800 & 800 / 1600 / 3200 \\
Mentality & --- & Raw Signals & $\mathcal{L}_{mse}^{rs}$ + $\mathcal{L}_{mse}^{spec}$ & NA & NA & NA & NA \\
Neuro-GPT & Causal Mask & Embedded Tokens & $\mathcal{L}_{mse}^{et}$ & 6 & --- & 1080 & --- \\
MEET & NA & NA & $\mathcal{L}_{cls}$ & 3 / 6 / 12 & 3 / 12 / 16 & 768 / 768 / 1024 & 3072 / 3072 / 4096 \\
EEGFormer & --- & Spectral Amplitude & $\mathcal{L}_{mse}^{sa}$ + $\mathcal{L}_{mse}^{cbe}$ & 6 / 8 / 12 & --- & 128 & --- \\
BrainWave & Random Mask & Spectrogram & --- & 10 & 16 & 768 & 2048 \\
NeuroLM & Causal Mask & Codebook Index & $\mathcal{L}_{nll}^{ci}$ & 12 / 24 / 48 & 12 / 16 / 25 & 768 / 1024 / 1600 & 3072 / 4096 / 6400 \\
Brant-X & NA & NA & $\mathcal{L}_{cl}^{coarse}$ + $\mathcal{L}_{cl}^{fine}$ & --- & --- & --- & ---  \\
FoME & Random Mask & Raw Signals & $\mathcal{L}_{mse}^{rs}$ & 16 & --- & --- & 3072 / 7168 \\
EEGPT & Random Mask & Raw Signals & $\mathcal{L}_{mse}^{rs}$ + $\mathcal{L}_{mse}^{emb}$ & --- & --- & --- & ---  \\
BrainGPT & Causal Mask & Raw Signals & $\mathcal{L}_{mse}^{rs}$ & 3 / 9 / 12 / 20 & 4 / 8 / 14 / 28 & 128 / 256 / 896 / 1792 & 512 / 1024 / 3584 / 7168 \\
GEFM & Random Mask & Embedded Tokens & $\mathcal{L}_{cl}^{et}$ & 8 & 8 & 1536 & 3076 \\
CBraMod & Random Mask & Raw Signals & $\mathcal{L}_{mse}^{rs}$ & 12 & 8 & 200 & 800 \\
CEReBrO & Random Mask & Embedded Tokens & $\mathcal{L}_{mse}^{et}$ + $\mathcal{L}_{aux}$ & 8 / 10 / 12 & 12 & 192 / 576 / 768 & 768 / 2304 / 3072 \\
LEAD & NA & NA & $\mathcal{L}_{cl}^{emb}$ + $\mathcal{L}_{cls}$ & 12 & 8 & 128 & 256 \\
FEMBA & Random Mask & Raw Signals & $\mathcal{L}_{s\text{-}l1}^{rs}$ & NA & NA & NA & NA \\
LCM & Random Mask & Embedded Tokens & $\mathcal{L}_{cl}^{emb}$ + $\lambda\mathcal{L}_{mse}^{et}$ & --- & --- & --- & --- \\
TFM & Random Mask & Codebook Index & $\mathcal{L}_{cls}^{ci}$ & 4 & 8 & 64 & --- \\
Gram & Random Mask & Spectral Amp.+ Codebook Index & $\mathcal{L}^{sa}_{mse} + \mathcal{L}^{ci}_{cls}$ & 12 / 24 / 32 & 10 / 16 / 16 & 200 / 400 / 800 & 800 / 1600 / 3200 \\
ALFEE & Random \& Causal Mask & Raw Signals + PSD & $\mathcal{L}_{mse}^{rs}$ + $\mathcal{L}_{mse}^{rs\,\oplus\,PSD}$ + $\mathcal{L}_{cls}^{dt}$ & 6 / 8 / 16 / 18 / 22 & 4 / 4 / 8 / 8 / 12 & 384 / 512 / 640 / 896 / 1152 & 256 / 512 / 512 / 768 / 768 \\
BrainOmni & Random Mask & Codebook Index & $\mathcal{L}_{cls}^{ci}$ & 12 & 8 / 16 & 256 / 512 & 1024 / 2048 \\
E3GT & Random Mask & SpecKMeansLabels & $\mathcal{L}_{cls}^{skl}$ & 12 & 12 & 768 & 3072 \\
CodeBrain & Random Mask & Codebook Index & $\mathcal{L}_{cls}^{ci}$ & NA & NA & NA & NA \\
DIVER-0 & Random Mask & Raw Signals & --- & 12 & 10 & 200 & 800 \\
UniMind & NA & NA & $\mathcal{L}_{cce}$ & 12 & 10 & 1152 & --- \\
CSBrain & Random Mask & Raw Signals & $\mathcal{L}_{mse}^{rs}$ & 12 & 8 & 200 & 800 \\
DMAE-EEG & Random Mask & Raw Signals + Embedded Tokens & $\mathcal{L}_{mse}^{rs}$ + $\mathcal{L}_{mse}^{et}$ & --- & --- & --- & --- \\
EEGMamba & Random Mask & Raw Signals & $\mathcal{L}_{mse}^{rs}$ & NA & NA & NA & NA \\
MIRepNet & Random Mask & Embedded Tokens & $\mathcal{L}_{mse}^{et}$ + $\mathcal{L}_{cls}$ & 6 & 8 & 256 & 1024 \\
PSGFM & Random Mask & SpecKMeansLabels & $\mathcal{L}_{cls}^{skl}$ & 12 & 12 & 768 & 3072 \\
EEGDM & NA & Velocity & $\mathcal{L}_{mse}^{v}$ & NA & NA & NA & NA \\
CoMET & Random Mask & Raw Signals & $\mathcal{L}_{mse}^{rs}$ + $\mathcal{L}_{cl}^{glob}$ & 6 / 6 / 12 & 4 / 8 / 16 & 256 / 512 / 1024 & 1024 / 2048 / 4096 \\
EpilepsyFM & Random Mask & EEG Codebook Index & $\mathcal{L}_{cls}^{ci}$ & 12 & 10 & 200 & 800 \\
SingLEM & Random Mask & Raw Signals + Frequency & $\mathcal{L}_{huber}^{ms}$ + $\mathcal{L}_{huber}^{ums}$ + $\mathcal{L}_{\ell_2}^{freq}$ & 4 + 12 & 4 + 8 & 128 & --- \\
BrainPro & Random Mask & Raw Signals & $\mathcal{L}_{w-mse}^{rs}$ + $\mathcal{L}_{dec}$ & 4 & 32 & 32 & 64 \\
Uni-NTFM & Random Mask & Time + Band Power & $\mathcal{L}_{mse}^{emb}$ + $\mathcal{L}_{mse}^{bp}$ + $\mathcal{L}_{aux}$ & 12 / 12 / 26 / 24 & --- & 256 / 512 / 512 / 768 & --- \\
ELASTIQ & Random \& Causal Mask & Embedded Tokens & $\mathcal{L}_{mse}^{et}$ & 12 & 8 & 256 & 1024 \\
BioCodec & NA & Time + Frequency & $\mathcal{L}_{huber}^{rs}$ + $\mathcal{L}_{\ell_1}^{stft}$ + $\mathcal{L}_{\ell_2}^{stft}$ + $\mathcal{L}_{aux}$ & 2 & 8 & 128 & --- \\
HEAR & NA & Codebook + Frequency & $\mathcal{L}_{mse}^{cbe}$ + $\mathcal{L}_{mse}^{freq}$ & 6 / 12 & 4 / 8 & --- & --- \\
NeuroRVQ & Random Mask & EEG Codebook Index & $\mathcal{L}_{cls}^{ci}$ & 12 & 10 & 200 & 800 \\
NeurIPT & Amplitude-Aware Mask & Raw Signals & $\mathcal{L}^{rs}_{\ell_p} + \mathcal{L}_{aux}$ & 6 & 8 & 768 & 512 / 768 \\
REVE & Random Mask & Raw Signals & $\mathcal{L}_{mae}^{rs}$ + $\mathcal{L}_{aux}$ & 4 / 22 / 22 & 8 / 8 / 19 & 512 / 512 / 1250 & 1365 / 1365 / 3333 \\
mdJPT & NA & NA & $\mathcal{L}_{CDA}$ + $\mathcal{L}_{ISA}$ & 2 & 8 & 128 & --- \\
LUNA & Random Mask & Embedded Tokens & $\mathcal{L}_{mse}^{et}$ & 8 / 10 / 24 & 8 / 12 / 16 & 256 / 576 / 1024 & 1024 / 2304 / 4096 \\
THD-BAR & Causal Mask & Codebook Index & $\mathcal{L}_{cls}^{ci}$ & 12 / 24 / 48 & 12 / 16 / 25 & 768 / 1024 / 1600 & 3072 / 4096 / 6400 \\
EEG-X & Random Mask & Noised-Removed Signals & $\mathcal{L}_{mse}^{nrs}$ + $\mathcal{L}_{kd}^{tea-stu}$ + $\mathcal{L}_{aux}$ & 4 & 8 & 16 & 64 \\
SAMBA & Random Mask & Time + Frequency & $\mathcal{L}_{mae}^{os}$ + $\mathcal{L}_{mse}^{freq}$ & NA & NA & NA & NA \\
DeeperBrain & Random Mask & Raw Signals + Neuro Info & $\mathcal{L}_{huber}^{rs}$ + $\mathcal{L}_{huber}^{ni}$ & 12 & 8 & 200 & 800 \\
EEGMoE & Random Mask & Embedded Tokens & $\mathcal{L}_{\ell_1}^{et}$ + $\mathcal{L}_{aux}$ & 3 & 4 & 512 & --- \\
Brain-OF & Random Mask & Raw Signals + Spectral Amp. & $\mathcal{L}_{s\text{-}l1}^{rs}$ + $\mathcal{L}_{s\text{-}l1}^{sa}$ & 12 / 24 / 36 & 4 / 8 / 12 & 256 / 512 / 768 & --- \\
\bottomrule
\end{tabular}
}
\end{table*}

\subsection{PRE-TRAINING AND DOWNSTREAM DATASETS}
\label{appendix:data_use}

Tables~\ref{tab:datasets1}--\ref{tab:datasets8} summarize the pre-training and downstream datasets used by existing EEG FMs. For each model, these tables document both the datasets used during pre-training and those adopted for downstream evaluation, providing a systematic overview of the data resources underlying current EEG FMs.

\begin{table}[!t]
\centering
\renewcommand{\arraystretch}{1.7}
\caption{Summary of the pre-trained and downstream datasets utilized in EEG FMs.}
\label{tab:datasets1}
\scalebox{0.5}{
\begin{tabular}{c|cc|cc}
\toprule
\multirow{2}{*}{Models} &
\multicolumn{2}{c|}{Pre-training} &
\multicolumn{2}{c}{Downstream} \\
\cmidrule(lr){2-3}\cmidrule(lr){4-5}
 & \makebox[1.3cm]{Datasets} & \makebox[1.3cm]{Paradigms}
 & \makebox[1.3cm]{Datasets} & \makebox[1.3cm]{Paradigms} \\
\midrule
\multirow{5}{*}{BENDR}
& TUEG & Clinic & PhysionetMI & MI / ME \\
&  &  & BCIC-IV-2A & MI / ME \\
&  &  & Margaux2012 & ERN / ERP \\
&  &  & Citi2010 & ERN / ERP \\
&  &  &  Sleep-EDF & Sleep \\
\midrule
\multirow{1}{*}{BrainBERT}
& Brain TreeBank & Clinic &  Brain TreeBank & Clinic \\
\bottomrule
\end{tabular}}
\end{table}

\begin{table}[t]
\centering
\renewcommand{\arraystretch}{1.6}
\caption{Summary of the pre-trained and downstream datasets utilized in EEG FMs (continued).}
\label{tab:datasets2}
\scalebox{0.5}{
\begin{tabular}{c|cc|cc}
\toprule
\multirow{2}{*}{Models} &
\multicolumn{2}{c|}{Pre-training} &
\multicolumn{2}{c}{Downstream} \\
\cmidrule(lr){2-3}\cmidrule(lr){4-5}
 & \makebox[1.3cm]{Datasets} & \makebox[1.3cm]{Paradigms}
 & \makebox[1.3cm]{Datasets} & \makebox[1.1cm]{Paradigms} \\
\midrule
\multirow{16}{*}{NeuroLM}
& TUEG & Clinic & TUAB & Clinic \\
& Siena & Clinic & TUEV & Clinic \\
& BCIC IV-1 & MI / ME & TUSL & Clinic \\
& Grasp and Lift & MI / ME & SEED & Emotion \\
& PhysionetMI & MI / ME & HMC & Sleep \\
& SEED-FRA & Emotion & EEGMat & Workload \\
\bottomrule
\end{tabular}}
\end{table}

\begin{table}[htpb]
\centering
\renewcommand{\arraystretch}{1.6}
\caption{Summary of the pre-trained and downstream datasets utilized in EEG FMs (continued).}
\label{tab:datasets3}
\scalebox{0.5}{
\begin{tabular}{c|cc|cc}
\toprule
\multirow{2}{*}{Models} &
\multicolumn{2}{c|}{Pre-training} &
\multicolumn{2}{c}{Downstream} \\
\cmidrule(lr){2-3}\cmidrule(lr){4-5}
 & \makebox[1.3cm]{Datasets} & \makebox[1.3cm]{Paradigms}
 & \makebox[1.3cm]{Datasets} & \makebox[1.1cm]{Paradigms} \\
\midrule
\multirow{4}{*}{CEReBrO}
& TUEG & Clinic & TUAB & Clinic \\
&  &  & Neonate & Clinic \\
&  &  & MoBI & MI / ME \\
&  &  & SEED & Emotion \\
\midrule
\multirow{12}{*}{LEAD}
& BrainLat & Clinic & ADFTD & Clinic \\
& P-ADIC & Clinic & CNBPM & Clinic \\
\bottomrule
\end{tabular}}
\end{table}

\begin{table}[htpb]
\centering
\renewcommand{\arraystretch}{1.50}
\caption{Summary of the pre-trained and downstream datasets utilized in EEG FMs (continued).}
\label{tab:datasets4}
\scalebox{0.5}{
\begin{tabular}{c|cc|cc}
\toprule
\multirow{2}{*}{Models} &
\multicolumn{2}{c|}{Pre-training} &
\multicolumn{2}{c}{Downstream} \\
\cmidrule(lr){2-3}\cmidrule(lr){4-5}
 & \makebox[1.3cm]{Datasets} & \makebox[1.3cm]{Paradigms}
 & \makebox[1.3cm]{Datasets} & \makebox[1.1cm]{Paradigms} \\
\midrule
\multirow{22}{*}{BrainOmni}
& MusicEEG & Emotion & AD65 & Clinic \\
& HFO & Sleep & MDD & Clinic \\
& Awakening & Sleep & PD31 & Clinic \\
& Go-Nogo & Visual & TUAB & Clinic \\
& Features-EEG & Visual & TUEV & Clinic \\
& SRM & Resting & WBCIC\_SHU & MI / ME \\
\bottomrule
\end{tabular}}
\end{table}

\begin{table}[htpb]
\centering
\renewcommand{\arraystretch}{1.52}
\caption{Summary of the pre-trained and downstream datasets utilized in EEG FMs (continued).}
\label{tab:datasets5}
\scalebox{0.5}{
\begin{tabular}{c|cc|cc}
\toprule

\multirow{2}{*}{Models} &
\multicolumn{2}{c|}{Pre-training} &
\multicolumn{2}{c}{Downstream} \\
\cmidrule(lr){2-3}\cmidrule(lr){4-5}
 & \makebox[1.3cm]{Datasets} & \makebox[1.3cm]{Paradigms}
 & \makebox[1.3cm]{Datasets} & \makebox[1.1cm]{Paradigms} \\
\midrule
\multirow{6}{*}{EEGMamba}
& TUEG & Clinic & CHB-MIT & Clinic \\
& Siena & Clinic & PhysionetMI & MI / ME \\
& Physionet & Sleep & FACED & Emotion \\
& B-SNIP1 & Resting & ISRUC & Sleep \\
& RAW & --- & BCIC2020–3 & Imagined Speech \\
&  &  & MODMA & MDD Diagnosis \\
\bottomrule
\end{tabular}}
\end{table}

\begin{table}[htpb]
\centering
\renewcommand{\arraystretch}{1.45}
\caption{Summary of the pre-trained and downstream datasets utilized in EEG FMs (continued).}
\label{tab:datasets6}
\scalebox{0.5}{
\begin{tabular}{c|cc|cc}
\toprule

\multirow{2}{*}{Models} &
\multicolumn{2}{c|}{Pre-training} &
\multicolumn{2}{c}{Downstream} \\
\cmidrule(lr){2-3}\cmidrule(lr){4-5}
 & \makebox[1.3cm]{Datasets} & \makebox[1.3cm]{Paradigms}
 & \makebox[1.3cm]{Datasets} & \makebox[1.1cm]{Paradigms} \\
\midrule
\multirow{12}{*}{Uni-NTFM}
& CAUEEG & Clinic & TUAB & Clinic \\
& TUEG & Clinic & TUEV & Clinic \\
& Siena & Clinic & TUSL & Clinic \\
& BCIC IV-1 & MI / ME & BCIC-IV-2A & MI / ME \\
& Emobrain & Emotion & SEED & Emotion \\
& SEED-IV & Emotion & HMC & Sleep \\
\bottomrule
\end{tabular}}
\end{table}

\begin{table}[htpb]
\centering
\renewcommand{\arraystretch}{1.5}
\caption{Summary of the pre-trained and downstream datasets utilized in EEG FMs (continued).}
\label{tab:datasets7}
\scalebox{0.5}{
\begin{tabular}{c|cc|cc}
\toprule
\multirow{2}{*}{Models} &
\multicolumn{2}{c|}{Pre-training} &
\multicolumn{2}{c}{Downstream} \\
\cmidrule(lr){2-3}\cmidrule(lr){4-5}
 & \makebox[1.3cm]{Datasets} & \makebox[1.3cm]{Paradigms}
 & \makebox[1.3cm]{Datasets} & \makebox[1.1cm]{Paradigms} \\
\midrule
\multirow{17}{*}{NeurIPT}
& TUAR & Clinic & TUAB & Clinic \\
& TUEP & Clinic & TUEV & Clinic \\
& TUSZ & Clinic & BCIC-IV-2A & MI / ME \\
& TUSL & Clinic & SEED-V & Emotion \\
& Siena & Clinic & Sleep-EDFx & Sleep \\
& PhysionetMI & MI / ME & PhysionetP300 & ERN/ERP \\
\bottomrule
\end{tabular}}
\end{table}

\begin{table}[htpb]
\centering
\renewcommand{\arraystretch}{1.5}
\caption{Summary of the pre-trained and downstream datasets utilized in EEG FMs (continued).}
\label{tab:datasets8}
\scalebox{0.5}{
\begin{tabular}{c|cc|cc}
\toprule
\multirow{2}{*}{Models} &
\multicolumn{2}{c|}{Pre-training} &
\multicolumn{2}{c}{Downstream} \\
\cmidrule(lr){2-3}\cmidrule(lr){4-5}
 & \makebox[1.3cm]{Datasets} & \makebox[1.3cm]{Paradigms}
 & \makebox[1.3cm]{Datasets} & \makebox[1.1cm]{Paradigms} \\
\midrule
\multirow{14}{*}{DeeperBrain}
& TUEG & Clinic & CHB-MIT & Clinic \\
& Siena & Clinic & PhysionetMI & MI / ME \\
& PhysioNet2018 & Sleep &  BCIC-IV-2A & MI / ME \\
& ds006171 & Visual & SHU-MI & MI / ME \\
& ds006547 & Visual & FACED & Emotion \\
& ds006480 & Sleep & SEED-V & Emotion \\
& ds006525 & Sleep & SEED-VII & Emotion \\
& ds006317 & Imagined Speech & ISRUC & Sleep \\
& RAW & --- & SEED-VIG & Vigilance \\
& ds006367 & --- & BCIC2020-3 & Imagined Speech \\
& ds006370 & --- & MODMA & MDD Diagnosis \\
& ds006437 & --- & MentalArithmetic & Mental Stress \\
& ds006446 & --- &  &  \\
& ds006466 & --- &  &  \\
\midrule
\multirow{6}{*}{EEGMoE}
& BCIC-IV-1 & MI / ME & BCIC-IV-2A & MI / ME \\
& EEGMMIDB & MI / ME & DEAP & Emotion \\
& DREAMER & Emotion & STEW & Workload \\
& SEED & Emotion &  &  \\
& MAHNOB-HCI & Emotion &  &  \\
& EEGMat & Workload &  &  \\
\midrule
\multirow{13}{*}{Brain-OF}
& TUEG & Clinic & TUAB & Clinic \\
& PhysionetMI & MI / ME & CHB-MIT & Clinic \\
& WBCIC-MI & MI / ME & SEED-V & Emotion \\
& SEED & Emotion & LEMON-EEG & Aging \\
& SEED-DV & Emotion &  &  \\
& SEED-SD & Emotion &  &  \\
& SEED-VII & Emotion &  &  \\
& FACED & Emotion &  &  \\
& REEG-BACA & Resting &  &  \\
& TDBRAIN & Resting &  &  \\
& PREDICT & Resting &  &  \\
& SEED-VIG & Vigilance &  &  \\
& SEED-VLA/VRW & Vigilance &  &  \\
& $\cdots$ & (MEG) &  &  \\
& $\cdots$ & (fMRI) &  &  \\
\bottomrule
\end{tabular}}
\end{table}

\subsection{DATASET DESCRIPTIONS} \label{appendix:data_intro}

Tables~\ref{tab:datasets} and~\ref{tab:fewshot_protocol} summarize 
the characteristics of the 13 benchmark datasets and the corresponding 
within-subject few-shot fine-tuning protocols, respectively. The 13 datasets used in this benchmark are summarized below.

\begin{enumerate}
\item \textbf{BNCI2014001} contains EEG data from 9 subjects performing four MI tasks: left hand, right hand, both feet, and tongue. Each subject participated in two sessions, with each session consisting of 6 runs, yielding a total of 288 trials per session.

\item \textbf{BNCI2015001} contains EEG data from 12 subjects performing sustained MI of the right hand and both feet. The data were recorded at 512 Hz using 13 electrodes, with a bandpass filter between 0.5 and 100 Hz and a notch filter at 50 Hz.

\item \textbf{BNCI2014004} contains EEG data from 9 right-handed subjects performing two MI tasks: left hand and right hand. Each subject participated in five sessions, with the first two sessions for screening without feedback and the remaining three sessions with feedback. The data were recorded using three bipolar EEG channels (C3, Cz, C4) at 250 Hz, with 120 trials per subject for each MI class.

\item \textbf{BNCI2014009} contains P300 evoked potentials from 10 healthy subjects performing a $6 \times 6$ matrix speller task under overt attention conditions. EEG was recorded from 16 channels (Fz, FCz, Cz, CPz, Pz, Oz, F3, F4, C3, C4, CP3, CP4, P3, P4, PO7, PO8) at 256 Hz with 0.1--20 Hz bandpass filtering. Each subject completed four sessions with three runs per session.

\item \textbf{BNCI2014008} contains P300 evoked potentials from 8 subjects with amyotrophic lateral sclerosis (ALS) performing a $6 \times 6$ matrix speller task. EEG was recorded from 8 channels (Fz, Cz, Pz, Oz, P3, P4, PO7, PO8) at 256 Hz with 0.1--30 Hz bandpass filtering. Each subject completed seven runs of five-character spelling, yielding 35 trials in total.

\item \textbf{CHB-MIT} contains EEG recordings from 23 pediatric subjects with intractable seizures, recorded using a 16-channel bipolar montage at 256 Hz. The dataset is used for seizure detection, a binary classification task to identify the presence of epileptic seizures from EEG signals.

\item \textbf{TUAB} (Temple University Hospital Abnormal) is a large-scale clinical EEG dataset from the TUH EEG Corpus, containing recordings from 2,383 adult patients with over 1,000 hours of data in total. The dataset is used for abnormal EEG detection, a binary classification task to distinguish pathological brain activity from normal recordings.

\item \textbf{Sleep-EDFx} (Sleep-EDF Expanded) is a polysomnographic dataset containing 197 whole-night recordings from 78 healthy subjects. Each recording includes EEG signals from Fpz--Cz and Pz--Oz derivations, annotated into five sleep stages: Wake, N1, N2, N3, and REM. The dataset serves as a standard benchmark for automatic sleep stage classification.

\item \textbf{SEED} (SJTU Emotion EEG Dataset) is a benchmark dataset for EEG-based emotion recognition, containing recordings from 15 subjects who watched 15 film clips across three sessions spaced one week apart. The 62-channel EEG was recorded at 1,000 Hz using an ESI NeuroScan system, with each clip labeled as positive, neutral, or negative.

\item \textbf{Nakanishi2015} is an SSVEP benchmark dataset for multi-class target identification. It contains EEG recordings from 9 subjects responding to 12 visual stimuli with frequencies ranging from 9.25 to 14.75 Hz. Each subject completed 15 blocks of 12 trials, yielding 180 trials per subject. EEG was recorded at 256 Hz using 8 occipital channels.

\item \textbf{EEGMat} is a cognitive workload dataset collected from 36 subjects during mental arithmetic tasks. EEG was recorded using 19 channels at 500 Hz following the international 10--20 system. Subjects were categorized into good and poor performers based on task accuracy, enabling analysis of individual differences in workload-related brain activity.

\item \textbf{Things-EEG2} is a large-scale dataset for visual object decoding, containing EEG recordings from 10 participants viewing natural object images. The dataset comprises 1,854 object classes from the THINGS image collection, supporting research on neural representations of visual semantics.

\item \textbf{SEED-VIG} is a dataset for EEG-based vigilance estimation collected during simulated driving. Vigilance levels are quantified using the PERCLOS (percentage of eye closure) metric derived from eye-tracking data. EEG was recorded at 200 Hz using 17 channels and segmented into 8-second epochs, supporting continuous vigilance prediction as a regression task.
\end{enumerate}

\begin{table*}[htpb]
\centering
\renewcommand{\arraystretch}{1.6}
\caption{Summary of the EEG datasets in benchmarking.}
\label{tab:datasets}
\scalebox{0.6}{
\begin{tabular}{c|c|c|c|c|c|c|c|c}
\toprule
BCI&\multirow{2}{*}{Dataset} & Number of & Number of & Sampling & Trial Length & Number of & \multirow{2}{*}{Tasks} & \multirow{2}{*}{Labels} \\
Paradigm & & Subjects & Channels & Rate (Hz) & (seconds) & Trials &  &  \\
\midrule
\multirow{3}{*}{MI}
& BNCI2014001~\cite{tangermann2012BCICIV2A} & 9 & 22 & 250 & 4  & 2,592 & Classification & left / right hand, feet, tongue \\
& BNCI2014004~\cite{leeb2007BCICIV2B} & 9 & 3 & 250 & 4.5 & 1,400 & Classification & left hand, right hand \\
& BNCI2015001~\cite{faller2012_15001} & 12 & 13 & 512 & 5 & 2,400 & Classification & right hand, both feet \\
\midrule
\multirow{2}{*}{P300}
& BNCI2014009~\cite{riccio2013_14009} & 10 & 16 & 256 & 0.8 & 5,760 & Classification & target, non-target \\
& BNCI2014008~\cite{riccio2013_14009} & 8 & 8 & 256 & 1 & 33,600 & Classification & target, non-target  \\
\midrule
\multirow{2}{*}{Clinic}
& CHB-MIT~\cite{Shoeb2009chbmit} & 23 & 18 & 256 & 4 & 29,840 & Classification & interictal, ictal \\
& TUAB~\cite{lopez2015tuab} & 2,383 & 21 & 250 & 10 & 53,604 & Classification & normal, abnormal  \\
\midrule
Sleep & Sleep-EDFx~\cite{kemp2000sleepedfx} & 78 & 2 & 100 & 30 & 414,961 & Classification & W, N1, N2, N3, REM \\
\midrule
Emotion & SEED~\cite{zheng2015seed} & 15 & 62 & 200 & 1 & 50,910 & Classification & positive, neutral, negative \\
\midrule
SSVEP & Nakanishi2015~\cite{nakanishi2015nakanishi} & 9 & 8 & 256 & 4 & 1,620 & Classification & 9.25--14.75 Hz (0.5 Hz interval) \\
\midrule
Workload & EEGMat~\cite{zyma2019eegmat} & 36 & 19 & 500 & 4 & 1,080 & Classification & low, high \\
\midrule
Visual Decoding & Things-EEG2~\cite{gifford2022things2} & 10 & 63 & 1000 & 1 & 18,540 & Retrieve & 200 images matching \\
\midrule
Fatigue & SEED-VIG~\cite{zheng2017multimodal} & 21 & 17 & 200 & 8 & 18,585 & Regression & PERCLOS \\
\bottomrule
\end{tabular}
}
\end{table*}

\begin{table*}[htpb]
\centering
\renewcommand{\arraystretch}{1.6}
\caption{Summary of the within-subject few-shot fine-tuning protocol across the evaluated datasets.}
\label{tab:fewshot_protocol}
\scalebox{0.6}{
\begin{tabular}{c|c|c|c|c|c}
\toprule
BCI & \multirow{2}{*}{Dataset} & \multirow{2}{*}{Few-shot Protocol} & Total Fine-tuning & Fine-tuning Trials & Trial Length \\
Paradigm & & & Trials & per Class & (seconds) \\
\midrule
\multirow{3}{*}{MI}
& BNCI2014001 & 30\% of one session per subject & 84 & 21 & 4 \\
& BNCI2014004 & 30\% of one session per subject & 36--48 & 18--24 & 4 \\
& BNCI2015001 & 30\% of one session per subject & 60 & 30 & 4 \\
\midrule
\multirow{2}{*}{P300}
& BNCI2014008 & 5\% of one session per subject & 210 & 35 / 175 & 1 \\
& BNCI2014009 & 10\% of one session per subject & 57 & 9 / 48 & 0.8$^\dagger$ \\
\midrule
Clinic & CHB-MIT & First seizure with 10-min pre-ictal segment$^\ddagger$ & 152--192 & 150 / 2--42 & 4 \\
\midrule
Sleep & Sleep-EDFx & 10\% of the target subject & 223--574 & 59--413 / 3--88 / 49--128 / 0--63 / 6--61$^\S$ & 30 \\
\midrule
Emotion & SEED & One video per class & 674 & 206 / 233 / 235 & 1 \\
\midrule
SSVEP & Nakanishi2015 & 80\% of the target subject & 144 & 12 & 4 \\
\midrule
Workload & EEGMat & 60\% of the target subject & 18 & 9 & 4 \\
\midrule
Visual Decoding & Things-EEG2 & 3 of 10 images per class & 4,962 & 3 & 1 \\
\midrule
Fatigue & SEED-VIG & 10\% of the target subject & 88 & --- & 8 \\
\bottomrule
\end{tabular}
}

\vspace{2pt}
\begin{flushleft}
\footnotesize
$^\dagger$ The original epoch length is 0.8s; signals are extended to 1s by prepending the 0.2s segment preceding each epoch, in order to satisfy the minimum input-length requirement of the evaluated EEG FMs. \\
$^\ddagger$ For each subject, the model is fine-tuned on the ictal segments of the first seizure together with its 10-min pre-ictal segment, and evaluated on the remaining seizures with their corresponding pre-ictal segments. \\
$^\S$ Trials per class correspond to the five sleep stages (W / N1 / N2 / N3 / REM). \\
SEED-VIG is a regression task (continuous PERCLOS values), and therefore the trials-per-class column is not applicable.
\end{flushleft}
\end{table*}

\subsection{BENCHMARK RESULTS} \label{appendix:bench_result}

\subsubsection{Main Results} \label{appendix:bench_main}

The detailed benchmark results for each subject are presented in Tables~\ref{tab:14001_acc} -~\ref{tab:VIG_cc2}.


\begin{table*}[!t]
    \centering
    \renewcommand{\arraystretch}{0.97}
    \vspace{-0em}
    \fontsize{9}{12}\selectfont
    \caption{Accuracies (\%) on BNCI2014001. The best accuracies are marked in bold, and the second best by an underline.}
    \label{tab:14001_acc}
    \scalebox{0.66}{
\begin{tabular}{c|c|w{c}{1.5cm}|w{c}{2.0cm}|*{9}{w{c}{0.7cm}}|w{c}{1.6cm}}
        \toprule
        Scenario & Tuning & Model Type & Approach & S0 & S1 & S2 & S3 & S4 & S5 & S6 & S7 & S8 & Avg. \\
        \midrule
    \multirow{40}{*}{\makecell{Cross-subject \\ (LOSO)}}%
    & \multirow{25}{*}{\makecell{Full \\ Fine-tuning}}%
    & \multirow{10}{*}{\makecell{Specialist \\ Models}}%
 & CSP+LDA & 45.83 & 27.43 & 52.78 & 30.21 & 30.21 & 23.61 & 36.81 & 57.29 & 39.58 & 38.19 \\
 &  &  & EEGNet & 61.23 & 26.85 & 69.56 & 35.07 & 25.12 & 28.47 & 32.18 & 60.88 & \textbf{65.39} & 44.97$_{\pm0.57}$ \\
 &  &  & ShallowConv & 68.17 & 31.13 & 52.89 & 37.62 & 27.43 & 27.78 & 31.25 & 65.74 & 61.23 & 44.80$_{\pm0.50}$ \\
 &  &  & LMDA & 66.55 & 32.64 & 66.09 & 35.65 & 27.78 & 31.02 & 29.51 & \underline{67.36} & \underline{64.58} & 46.80$_{\pm0.31}$ \\
 &  &  & CNN-T & 56.83 & 31.48 & 50.93 & 32.29 & 26.04 & 28.24 & 27.20 & 51.04 & 48.26 & 39.15$_{\pm0.56}$ \\
 &  &  & Deformer & 56.48 & 24.54 & 60.53 & 34.26 & 26.62 & 29.28 & 31.02 & 55.90 & 55.09 & 41.53$_{\pm0.67}$ \\
 &  &  & Conformer & 60.88 & 26.27 & 57.29 & 31.48 & 26.85 & 30.56 & 23.84 & 63.77 & 53.82 & 41.64$_{\pm1.23}$ \\
\bottomrule
    \end{tabular}
    }
    \vspace{-0em}
\end{table*}
\clearpage

\begin{table*}[!t]
    \centering
    \renewcommand{\arraystretch}{0.97}
    \vspace{-0em}
    \fontsize{9}{12}\selectfont
    \caption{Cohen's Kappa (\%) on BNCI2014001. The best metrics are marked in bold, and the second best by an underline.}
    \label{tab:14001_kappa}
    \scalebox{0.66}{
    \begin{tabular}{c|c|w{c}{1.5cm}|w{c}{2.0cm}|*{9}{w{c}{0.7cm}}|w{c}{1.6cm}}
        \toprule
        Scenario & Tuning & Model Type & Approach & S0 & S1 & S2 & S3 & S4 & S5 & S6 & S7 & S8 & Avg. \\
        \midrule
    \multirow{40}{*}{\makecell{Cross-subject \\ (LOSO)}}%
    & \multirow{25}{*}{\makecell{Full \\ Fine-tuning}}%
    & \multirow{10}{*}{\makecell{Specialist \\ Models}}%
 & CSP+LDA & 27.78 & 3.24 & 37.04 & 6.94 & 6.94 & -1.85 & 15.74 & 43.06 & 19.44 & 17.59 \\
 &  &  & EEGNet & 48.30 & 2.47 & 59.41 & 13.42 & 0.92 & 4.63 & 9.57 & 47.84 & \textbf{53.86} & 26.63$_{\pm 0.75}$ \\
 &  &  & ShallowConv & 57.56 & 8.18 & 37.19 & 16.82 & 3.24 & 3.70 & 8.33 & 54.32 & 48.30 & 26.40$_{\pm 0.67}$ \\
 &  &  & LMDA & 55.40 & 10.19 & 54.78 & 14.20 & 3.70 & 8.02 & 6.02 & \underline{56.48} & \underline{52.78} & 29.06$_{\pm 0.40}$ \\
 &  &  & CNN-T & 42.44 & 8.64 & 34.57 & 9.72 & 1.39 & 4.32 & 2.93 & 34.72 & 31.02 & 18.86$_{\pm 0.74}$ \\
 &  &  & Deformer & 41.98 & 0.15 & 47.38 & 12.34 & 2.16 & 5.71 & 8.02 & 41.20 & 40.12 & 22.03$_{\pm 0.89}$ \\
 &  &  & Conformer & 47.84 & 1.85 & 43.06 & 8.64 & 2.47 & 7.41 & 0.46 & 51.70 & 38.43 & 22.19$_{\pm 1.64}$ \\
\bottomrule
    \end{tabular}
    }
    \vspace{-0em}
\end{table*}
\clearpage


\begin{table*}[!t]
    \centering
    \renewcommand{\arraystretch}{0.97}
    \vspace{-0em}
    \fontsize{9}{12}\selectfont
    \caption{Accuracies (\%) on BNCI2014004. The best accuracies are marked in bold, and the second best by an underline.}
    \label{tab:14004_acc}
    \scalebox{0.66}{
    \begin{tabular}{c|c|w{c}{1.5cm}|w{c}{2.0cm}|*{9}{w{c}{0.7cm}}|w{c}{1.6cm}}
        \toprule
        Scenario & Tuning & Model Type & Approach & S0 & S1 & S2 & S3 & S4 & S5 & S6 & S7 & S8 & Avg. \\
        \midrule
    \multirow{40}{*}{\makecell{Cross-subject \\ (LOSO)}}%
    & \multirow{25}{*}{\makecell{Full \\ Fine-tuning}}%
    & \multirow{10}{*}{\makecell{Specialist \\ Models}}%
 & CSP+LDA & 77.50 & 54.17 & 53.75 & 91.25 & 85.62 & \underline{83.12} & 63.12 & 68.75 & 81.88 & 73.24 \\
 &  &  & EEGNet & 73.33 & 59.72 & 55.42 & 94.58 & 82.92 & \textbf{83.96} & 73.54 & 76.67 & 87.29 & 76.38$_{\pm0.59}$ \\
 &  &  & ShallowConv & 72.71 & 60.56 & 55.83 & 91.46 & 77.50 & 70.83 & 74.38 & 77.92 & 86.88 & 74.23$_{\pm0.39}$ \\
 &  &  & LMDA & 75.42 & 54.17 & 52.71 & 96.46 & 81.67 & 82.08 & 73.12 & 73.96 & 84.38 & 74.88$_{\pm0.67}$ \\
 &  &  & CNN-T & 66.04 & 53.06 & 51.04 & \underline{97.08} & 78.12 & 76.88 & 68.12 & 70.83 & 79.58 & 71.20$_{\pm0.42}$ \\
 &  &  & Deformer & 68.12 & 56.67 & 57.50 & 97.50 & 80.62 & 82.71 & 70.42 & 75.42 & 84.79 & 74.86$_{\pm0.82}$ \\
 &  &  & Conformer & 73.54 & 51.11 & 54.17 & 97.50 & 87.92 & 76.25 & 72.29 & 74.17 & 80.62 & 74.17$_{\pm0.49}$ \\
\bottomrule
    \end{tabular}
    }
    \vspace{-0em}
\end{table*}
\clearpage

\begin{table*}[!t]
    \centering
    \renewcommand{\arraystretch}{0.97}
    \vspace{-0em}
    \fontsize{9}{12}\selectfont
    \caption{Cohen's Kappa (\%) on BNCI2014004. The best metrics are marked in bold, and the second best by an underline.}
    \label{tab:14004_kappa}
    \scalebox{0.66}{
    \begin{tabular}{c|c|w{c}{1.5cm}|w{c}{2.0cm}|*{9}{w{c}{0.7cm}}|w{c}{1.6cm}}
        \toprule
        Scenario & Tuning & Model Type & Approach & S0 & S1 & S2 & S3 & S4 & S5 & S6 & S7 & S8 & Avg. \\
        \midrule
    \multirow{40}{*}{\makecell{Cross-subject \\ (LOSO)}}%
    & \multirow{25}{*}{\makecell{Full \\ Fine-tuning}}%
    & \multirow{10}{*}{\makecell{Specialist \\ Models}}%
 & CSP+LDA & 55.00 & 8.33 & 7.50 & 82.50 & 71.25 & \underline{66.25} & 26.25 & 37.50 & 63.75 & 46.48 \\
 &  &  & EEGNet & 46.67 & 19.44 & 10.83 & 89.17 & 65.83 & \textbf{67.92} & 47.08 & 53.33 & 74.58 & 52.76$_{\pm 1.17}$ \\
 &  &  & ShallowConv & 45.42 & 21.11 & 11.67 & 82.92 & 55.00 & 41.67 & 48.75 & 55.83 & 73.75 & 48.46$_{\pm 0.78}$ \\
 &  &  & LMDA & 50.83 & 8.33 & 5.42 & 92.92 & 63.33 & 64.17 & 46.25 & 47.92 & 68.75 & 49.77$_{\pm 1.34}$ \\
 &  &  & CNN-T & 32.08 & 6.11 & 2.08 & \underline{94.17} & 56.25 & 53.75 & 36.25 & 41.67 & 59.17 & 42.39$_{\pm 0.84}$ \\
 &  &  & Deformer & 36.25 & 13.33 & 15.00 & \textbf{95.00} & 61.25 & 65.42 & 40.83 & 50.83 & 69.58 & 49.72$_{\pm 1.64}$ \\
 &  &  & Conformer & 47.08 & 2.22 & 8.33 & \textbf{95.00} & 75.83 & 52.50 & 44.58 & 48.33 & 61.25 & 48.35$_{\pm 0.97}$ \\
\bottomrule
    \end{tabular}
    }
    \vspace{-0em}
\end{table*}
\clearpage


\begin{table*}[!t]
    \centering
    \renewcommand{\arraystretch}{1.03}
    \vspace{-0em}
    \fontsize{9}{12}\selectfont
    \caption{Accuracies (\%) on BNCI2015001. The best accuracies are marked in bold, and the second best by an underline.}
    \label{tab:15001_acc}
    \scalebox{0.63}{
    \begin{tabular}{c|c|w{c}{1.25cm}|w{c}{1.8cm}|*{12}{w{c}{0.6cm}}|w{c}{1.4cm}}
        \toprule
        Scenario & Tuning & Model Type & Approach & S0 & S1 & S2 & S3 & S4 & S5 & S6 & S7 & S8 & S9 & S10 & S11 & Avg. \\
        \midrule
    \multirow{40}{*}{\makecell{Cross-subject \\ (LOSO)}}%
    & \multirow{25}{*}{\makecell{Full \\ Fine-tuning}}%
    & \multirow{10}{*}{\makecell{Specialist \\ Models}}%
 & CSP+LDA & 58.00 & 55.50 & 51.50 & 82.00 & 64.00 & 58.00 & 49.00 & 50.00 & 57.00 & 50.00 & 50.00 & 52.00 & 56.42 \\
 &  &  & EEGNet & 91.50 & \textbf{96.50} & 72.00 & 72.00 & 60.17 & 54.67 & 61.50 & 51.33 & 49.67 & 49.67 & 50.50 & 51.33 & 63.40$_{\pm1.32}$ \\
 &  &  & ShallowConv & 94.33 & 82.00 & 62.67 & 70.17 & 61.00 & 63.00 & 68.83 & 55.83 & 52.00 & 55.50 & 51.50 & 49.83 & 63.89$_{\pm0.80}$ \\
 &  &  & LMDA & 65.67 & 89.50 & 56.00 & 73.67 & 55.33 & 63.17 & 66.50 & 57.83 & 51.00 & 57.67 & 50.83 & 49.67 & 61.40$_{\pm 0.94}$ \\
 &  &  & CNN-T & 53.00 & 87.33 & 62.67 & 69.67 & 69.00 & 58.50 & 54.83 & 54.00 & 53.67 & 51.00 & 51.17 & 50.83 & 59.64$_{\pm1.41}$ \\
 &  &  & Deformer & 82.67 & 78.17 & 68.17 & 67.50 & 57.33 & 63.67 & \underline{72.33} & 55.83 & 54.33 & 56.17 & 50.00 & 50.50 & 63.06$_{\pm1.17}$ \\
 &  &  & Conformer & 52.83 & 64.83 & 52.00 & 67.50 & 73.17 & 57.50 & 65.00 & 53.33 & 56.83 & 55.50 & \textbf{59.00} & 51.67 & 59.10$_{\pm 2.14}$ \\
\bottomrule
    \end{tabular}
    }
    \vspace{-0em}
\end{table*}
\clearpage

\begin{table*}[!t]
    \centering
    \renewcommand{\arraystretch}{1.03}
    \vspace{-0em}
    \fontsize{9}{12}\selectfont
    \caption{Cohen's Kappa (\%) on BNCI2015001. The best metrics are marked in bold, and the second best by an underline.}
    \label{tab:15001_kappa}
    \scalebox{0.63}{
    \begin{tabular}{c|c|w{c}{1.25cm}|w{c}{1.8cm}|*{12}{w{c}{0.6cm}}|w{c}{1.4cm}}
        \toprule
        Scenario & Tuning & Model Type & Approach & S0 & S1 & S2 & S3 & S4 & S5 & S6 & S7 & S8 & S9 & S10 & S11 & Avg. \\
        \midrule
    \multirow{40}{*}{\makecell{Cross-subject \\ (LOSO)}}%
    & \multirow{25}{*}{\makecell{Full \\ Fine-tuning}}%
    & \multirow{10}{*}{\makecell{Specialist \\ Models}}%
 & CSP+LDA & 16.00 & 11.00 & 3.00 & 64.00 & 28.00 & 16.00 & -2.00 & 0.00 & 14.00 & 0.00 & 0.00 & 4.00 & 12.83 \\
 &  &  & EEGNet & 83.00 & \textbf{93.00} & 44.00 & 44.00 & 20.33 & 9.33 & 23.00 & 2.67 & 0.00 & 0.33 & 1.00 & 2.67 & 26.80$_{\pm 2.63}$ \\
 &  &  & ShallowConv & 88.67 & 64.00 & 25.33 & 40.33 & 22.00 & 26.00 & 37.67 & 11.67 & 4.00 & 11.00 & 3.00 & 0.67 & 27.78$_{\pm 1.59}$ \\
 &  &  & LMDA & 31.33 & 79.00 & 12.00 & 47.33 & 10.67 & 26.33 & 33.00 & 15.67 & 2.00 & 15.67 & 1.67 & 1.00 & 22.81$_{\pm 1.88}$ \\
 &  &  & CNN-T & 6.00 & 74.67 & 25.33 & 39.33 & 38.00 & 17.00 & 9.67 & 8.00 & 7.33 & 2.33 & 2.33 & 2.00 & 19.28$_{\pm 2.82}$ \\
 &  &  & Deformer & 65.33 & 56.33 & 36.33 & 35.00 & 14.67 & 27.33 & \underline{44.67} & 11.67 & 8.67 & 12.67 & 0.00 & 3.00 & 26.11$_{\pm 2.35}$ \\
 &  &  & Conformer & 5.67 & 29.67 & 4.00 & 35.00 & 46.33 & 15.00 & 30.00 & 6.67 & 13.67 & 11.33 & \textbf{18.00} & 3.33 & 18.19$_{\pm 4.28}$ \\
\bottomrule
    \end{tabular}
    }
    \vspace{-0em}
\end{table*}
\clearpage


\begin{table*}[!t]
    \centering
    \renewcommand{\arraystretch}{1.13}
    \vspace{-0em}
    \fontsize{9}{12}\selectfont
    \caption{Classification AUCs (\%) on BNCI2014009. The best metrics are marked in bold, and the second best by an underline.}
    \label{tab:14009_auc}
    \scalebox{0.63}{
    \begin{tabular}{c|c|w{c}{1.5cm}|w{c}{1.9cm}|*{10}{w{c}{0.7cm}}|w{c}{1.5cm}}
        \toprule
        Scenario & Tuning & Model Type & Approach & S0 & S1 & S2 & S3 & S4 & S5 & S6 & S7 & S8 & S9 & Avg. \\
        \midrule
    \multirow{36}{*}{\makecell{Cross-subject \\ (LOSO)}}%
    & \multirow{22}{*}{\makecell{Full \\ Fine-tuning}}%
    & \multirow{8}{*}{\makecell{Specialist \\ Models}}%
 & xDAWN+LDA & 72.41 & 78.53 & 68.27 & 72.34 & 84.86 & 71.73 & 69.54 & 78.35 & 88.25 & 90.35 & 77.46 \\
 &  &  & EEGNet & 81.81 & 88.18 & 73.04 & 86.03 & 92.46 & \textbf{85.42} & 84.73 & 85.83 & \underline{96.47} & 90.22 & 86.42$_{\pm 0.22}$ \\
 &  &  & ShallowConv & \textbf{85.92} & 86.56 & \textbf{74.76} & 83.29 & \textbf{93.78} & 82.81 & \underline{85.49} & 85.70 & 95.04 & 92.26 & 86.56$_{\pm 0.15}$ \\
 &  &  & LMDA & 80.62 & 89.16 & \underline{73.42} & \underline{88.89} & 92.10 & 84.28 & 82.66 & 86.18 & 96.37 & 93.13 & 86.68$_{\pm0.34}$ \\
 &  &  & CNN-T & 58.56 & 71.51 & 60.30 & 64.32 & 76.97 & 65.45 & 61.77 & 68.54 & 67.77 & 62.83 & 65.80$_{\pm2.39}$ \\
 &  &  & Deformer & \underline{82.99} & 88.38 & 73.35 & 85.65 & 92.20 & \underline{84.76} & \textbf{85.98} & \underline{86.37} & \textbf{97.36} & \underline{93.55} & \textbf{87.06}$_{\pm 0.14}$ \\
 &  &  & Conformer & 79.48 & 85.20 & 73.07 & 83.14 & \underline{92.62} & 78.09 & 77.34 & 81.27 & 95.82 & 92.63 & 83.87$_{\pm0.31}$ \\
\bottomrule
    \end{tabular}
    }
    \vspace{-0em}
\end{table*}
\clearpage

\begin{table*}[!t]
    \centering
    \renewcommand{\arraystretch}{1.13}
    \vspace{-0em}
    \fontsize{9}{12}\selectfont
    \caption{Balanced classification accuracies (\%) on BNCI2014009. The best BCAs are marked in bold, and the second best by an underline.}
    \label{tab:14009_bca}
    \scalebox{0.63}{
    \begin{tabular}{c|c|w{c}{1.5cm}|w{c}{1.9cm}|*{10}{w{c}{0.7cm}}|w{c}{1.5cm}}
        \toprule
        Scenario & Tuning & Model Type & Approach & S0 & S1 & S2 & S3 & S4 & S5 & S6 & S7 & S8 & S9 & Avg. \\
        \midrule
    \multirow{36}{*}{\makecell{Cross-subject \\ (LOSO)}}%
    & \multirow{22}{*}{\makecell{Full \\ Fine-tuning}}%
    & \multirow{8}{*}{\makecell{Specialist \\ Models}}%
 & xDAWN+LDA & 59.27 & 69.79 & 62.71 & 59.17 & 74.27 & 54.48 & 55.31 & 67.29 & 76.04 & 73.02 & 65.14 \\
 &  &  & EEGNet & \textbf{73.96} & 79.17 & 68.09 & 77.60 & 83.89 & \textbf{77.12} & \textbf{74.90} & 75.69 & 90.76 & 82.74 & \underline{78.39}$_{\pm 0.44}$ \\
 &  &  & ShallowConv & 73.51 & \underline{79.83} & 68.02 & 76.32 & 84.24 & \underline{76.84} & 71.91 & 75.83 & 90.56 & 83.44 & 78.05$_{\pm 0.62}$ \\
 &  &  & LMDA & 70.14 & 79.13 & \textbf{69.72} & \underline{81.56} & 84.72 & 75.62 & 70.24 & \textbf{78.58} & 88.72 & \underline{86.11} & \textbf{78.45}$_{\pm 0.46}$ \\
 &  &  & CNN-T & 57.33 & 66.67 & 56.77 & 60.49 & 67.12 & 59.62 & 58.75 & 63.58 & 63.78 & 60.90 & 61.50$_{\pm 1.71}$ \\
 &  &  & Deformer & 71.22 & 77.47 & 67.36 & 73.19 & 84.55 & 72.43 & 68.92 & 75.42 & \textbf{91.81} & \textbf{86.56} & 76.89$_{\pm 0.14}$ \\
 &  &  & Conformer & 56.25 & 56.56 & 60.21 & 61.77 & 72.22 & 65.45 & 58.65 & 58.89 & 70.14 & 62.05 & 62.22$_{\pm 0.97}$ \\
\bottomrule
    \end{tabular}
    }
    \vspace{-0em}
\end{table*}
\clearpage


\begin{table*}[!t]
    \centering
    \renewcommand{\arraystretch}{1.08}
    \vspace{-0em}
    \fontsize{9}{12}\selectfont
    \caption{Classification AUCs (\%) on BNCI2014008. The best metrics are marked in bold, and the second best by an underline.}
    \label{tab:14008_auc}
    \scalebox{0.66}{
    \begin{tabular}{c|c|w{c}{1.5cm}|w{c}{2.0cm}|*{8}{w{c}{0.88cm}}|w{c}{1.5cm}}
        \toprule
        Scenario & Tuning & Model Type & Approach & S0 & S1 & S2 & S3 & S4 & S5 & S6 & S7 & Avg. \\
        \midrule
    \multirow{36}{*}{\makecell{Cross-subject \\ (LOSO)}}%
    & \multirow{22}{*}{\makecell{Full \\ Fine-tuning}}%
    & \multirow{8}{*}{\makecell{Specialist \\ Models}}%
 & xDAWN+LDA & 74.40 & 73.89 & 82.98 & 73.59 & 73.49 & 75.63 & 76.78 & 83.04 & 76.72 \\
 &  &  & EEGNet & \textbf{76.93} & \textbf{76.03} & \textbf{84.89} & \underline{75.86} & \underline{79.61} & \textbf{79.52} & \underline{81.46} & \textbf{84.91} & \textbf{79.90}$_{\pm0.11}$ \\
 &  &  & ShallowConv & 73.45 & 73.71 & 81.35 & 73.90 & 76.49 & 78.36 & 78.21 & 84.30 & 77.47$_{\pm0.02}$ \\
 &  &  & LMDA & \underline{76.73} & 74.53 & \underline{83.30} & \textbf{75.88} & \textbf{79.79} & \underline{78.38} & \textbf{81.69} & \underline{84.50} & \underline{79.35}$_{\pm0.10}$ \\
 &  &  & CNN-T & 53.91 & 57.47 & 53.13 & 51.42 & 60.06 & 51.51 & 57.31 & 53.63 & 54.80$_{\pm0.35}$ \\
 &  &  & Deformer & 75.63 & 71.69 & 80.11 & 71.61 & 73.60 & 72.84 & 78.54 & 81.41 & 75.68$_{\pm0.48}$ \\
 &  &  & Conformer & 71.50 & 72.36 & 80.10 & 70.38 & 76.80 & 73.74 & 73.76 & 80.32 & 74.87$_{\pm0.15}$ \\
\bottomrule
    \end{tabular}
    }
    \vspace{-0em}
\end{table*}
\clearpage

\begin{table*}[!t]
    \centering
    \renewcommand{\arraystretch}{1.08}
    \vspace{-0em}
    \fontsize{9}{12}\selectfont
    \caption{Balanced classification accuracies (\%) on BNCI2014008. The best BCAs are marked in bold, and the second best by an underline.}
    \label{tab:14008_bca}
    \scalebox{0.66}{
    \begin{tabular}{c|c|w{c}{1.5cm}|w{c}{2.0cm}|*{8}{w{c}{0.88cm}}|w{c}{1.5cm}}
        \toprule
        Scenario & Tuning & Model Type & Approach & S0 & S1 & S2 & S3 & S4 & S5 & S6 & S7 & Avg. \\
        \midrule
    \multirow{36}{*}{\makecell{Cross-subject \\ (LOSO)}}%
    & \multirow{22}{*}{\makecell{Full \\ Fine-tuning}}%
    & \multirow{8}{*}{\makecell{Specialist \\ Models}}%
 & xDAWN+LDA & 55.06 & 58.16 & 62.73 & 63.89 & 53.66 & 55.24 & 63.33 & 57.50 & 58.69 \\
 &  &  & EEGNet & \textbf{69.66} & \textbf{69.00} & \textbf{76.86} & \underline{69.85} & \textbf{71.71} & \textbf{71.87} & \underline{73.18} & \textbf{76.22} & \textbf{72.29}$_{\pm 0.17}$ \\
 &  &  & ShallowConv & 66.29 & \underline{68.30} & 74.02 & 67.11 & 67.45 & \underline{70.63} & 70.27 & 75.25 & 69.92$_{\pm 0.05}$ \\
 &  &  & LMDA & 68.94 & 67.62 & \underline{75.95} & \textbf{70.29} & \underline{71.17} & 69.65 & \textbf{74.13} & \underline{76.21} & \underline{71.74}$_{\pm 0.12}$ \\
 &  &  & CNN-T & 50.16 & 53.36 & 51.38 & 50.58 & 54.02 & 50.65 & 54.78 & 50.54 & 51.93$_{\pm 0.16}$ \\
 &  &  & Deformer & 57.00 & 58.14 & 61.30 & 59.57 & 54.70 & 50.77 & 63.29 & 57.27 & 57.75$_{\pm 1.21}$ \\
 &  &  & Conformer & 62.48 & 66.20 & 72.50 & 65.30 & 69.10 & 67.80 & 67.15 & 72.43 & 67.87$_{\pm0.11}$ \\
\bottomrule
    \end{tabular}
    }
    \vspace{-0em}
\end{table*}
\clearpage



\begin{table*}[!t]
    \centering
    \renewcommand{\arraystretch}{1.09}
    \vspace{-0em}
    \fontsize{9}{12}\selectfont
    \caption{Classification AUCs (\%) on CHB-MIT. The best AUCs are marked in bold, and the second best by an underline.}
    \label{tab:chb_auc1}
    \scalebox{0.65}{
    \begin{tabular}{c|c|w{c}{1.25cm}|w{c}{1.9cm}|*{12}{w{c}{0.67cm}}}
        \toprule
        Scenario & Tuning & Model Type & Approach & S0 & S1 & S2 & S3 & S4 & S5 & S6 & S7 & S8 & S9 & S10 & S11 \\
        \midrule
        \multirow{36}{*}{\makecell{Cross-subject \\ (LOSO)}}%
        & \multirow{22}{*}{\makecell{Full \\ Fine-tuning}}%
        & \multirow{8}{*}{\makecell{Specialist \\ Models}}%
 & PSD+SVM & 87.68 & 87.33 & 90.45 & 56.32 & 96.28 & 64.13 & 91.22 & 95.00 & 98.39 & 96.49 & 97.08 & 63.53 \\
 &  &  & EEGNet & 99.50 & 98.04 & 95.99 & 95.61 & 91.61 & 78.02 & 91.87 & 75.30 & \textbf{98.44} & \underline{98.44} & 99.55 & 58.59 \\
 &  &  & ShallowConv & 99.90 & 97.39 & \underline{99.03} & 95.29 & 94.16 & 86.96 & \underline{99.07} & 94.44 & \underline{98.42} & 98.26 & 99.28 & 58.85 \\
 &  &  & LMDA & 99.88 & 96.44 & 93.43 & 90.55 & 95.63 & 80.83 & 94.44 & 86.41 & 97.54 & 97.76 & 97.48 & 64.77 \\
 &  &  & CNN-T & 98.86 & 90.07 & 97.24 & 95.99 & 95.70 & 85.79 & 95.67 & 85.60 & 97.95 & 96.05 & 98.83 & 57.31 \\
 &  &  & Deformer & \textbf{99.93} & \textbf{98.25} & 96.21 & 97.13 & 93.73 & 87.57 & 95.76 & 90.85 & 96.99 & 98.02 & 99.02 & 61.13 \\
 &  &  & Conformer & \underline{99.91} & \underline{98.15} & 98.53 & 95.30 & 86.09 & 85.77 & 97.42 & 96.56 & 91.95 & 97.13 & 99.00 & 55.21 \\
\bottomrule
    \end{tabular}
    }
    \vspace{-0em}
\end{table*}
\clearpage

\begin{table*}[!t]
    \centering
    \renewcommand{\arraystretch}{1.09}
    \vspace{-0em}
    \fontsize{9}{12}\selectfont
    \caption{Classification AUCs (\%) on CHB-MIT. The best AUCs are marked in bold, and the second best by an underline (continued).}
    \label{tab:chb_auc2}
    \scalebox{0.65}{
    \begin{tabular}{c|c|w{c}{1.25cm}|w{c}{1.9cm}|*{11}{w{c}{0.56cm}}|w{c}{1.6cm}}
        \toprule
        Scenario & Tuning & Model Type & Approach & S12 & S13 & S14 & S15 & S16 & S17 & S18 & S19 & S20 & S21 & S22 & Avg. \\
        \midrule
        \multirow{36}{*}{\makecell{Cross-subject \\ (LOSO)}}%
        & \multirow{22}{*}{\makecell{Full \\ Fine-tuning}}%
        & \multirow{8}{*}{\makecell{Specialist \\ Models}}%
 & PSD+SVM & 68.26 & \textbf{73.36} & 39.34 & 81.41 & 62.20 & 15.81 & 92.62 & 57.13 & 56.00 & 97.75 & 91.97 & 76.51 \\
 &  &  & EEGNet & 59.13 & 7.95 & 83.28 & 84.11 & 97.47 & \underline{93.76} & 92.25 & 92.38 & 83.81 & 91.65 & 98.38 & 85.44$_{\pm0.97}$ \\
 &  &  & ShallowConv & 63.10 & 8.07 & 79.66 & 85.00 & \textbf{99.49} & \textbf{95.08} & \textbf{100.00} & \textbf{95.95} & 95.70 & 99.74 & 99.33 & \textbf{88.79}$_{\pm0.12}$ \\
 &  &  & LMDA & 75.46 & 18.44 & 69.35 & 78.94 & 97.12 & 85.97 & 99.87 & 82.07 & 84.44 & 97.61 & \textbf{99.50} & 86.26$_{\pm0.01}$ \\
 &  &  & CNN-T & 79.56 & 56.84 & 47.28 & 82.91 & 90.73 & 87.12 & 99.47 & 84.34 & 95.06 & 96.58 & \underline{99.40} & 87.58$_{\pm0.21}$ \\
 &  &  & Deformer & 64.15 & 25.06 & \textbf{88.04} & 84.77 & \underline{99.14} & 82.23 & \underline{99.96} & 92.32 & 92.61 & 99.58 & 96.10 & \underline{88.63}$_{\pm0.53}$ \\
 &  &  & Conformer & 67.74 & 8.40 & 73.25 & 76.97 & 98.65 & 87.60 & 98.93 & 93.77 & 83.62 & 99.63 & 98.24 & 86.43$_{\pm0.39}$ \\
\bottomrule
    \end{tabular}
    }
    \vspace{-0em}
\end{table*}
\clearpage


\begin{table*}[!t]
    \centering
    \renewcommand{\arraystretch}{1.09}
    \vspace{-0em}
    \fontsize{9}{12}\selectfont
    \caption{Balanced classification accuracies (\%) on CHB-MIT. The best BCAs are marked in bold, and the second best by an underline.}
    \label{tab:chb_bca1}
    \scalebox{0.65}{
    \begin{tabular}{c|c|w{c}{1.25cm}|w{c}{1.9cm}|*{12}{w{c}{0.67cm}}}
        \toprule
        Scenario & Tuning & Model Type & Approach & S0 & S1 & S2 & S3 & S4 & S5 & S6 & S7 & S8 & S9 & S10 & S11 \\
        \midrule
    \multirow{36}{*}{\makecell{Cross-subject \\ (LOSO)}}%
    & \multirow{22}{*}{\makecell{Full \\ Fine-tuning}}%
    & \multirow{8}{*}{\makecell{Specialist \\ Models}}%
 & PSD+SVM & 83.80 & 79.44 & 74.27 & 64.11 & \textbf{91.05} & 49.18 & 87.97 & 84.67 & \textbf{96.18} & 89.68 & 93.17 & 51.31 \\
 &  &  & EEGNet & 96.15 & 92.11 & 93.94 & \underline{91.62} & 61.96 & 69.03 & 88.62 & 72.41 & 75.50 & \textbf{95.05} & 94.92 & 52.97 \\
 &  &  & ShallowConv & 97.66 & 93.91 & \textbf{95.79} & 89.89 & 61.29 & 68.82 & \textbf{93.27} & 82.61 & 81.86 & \underline{93.33} & 93.44 & 52.26 \\
 &  &  & LMDA & \textbf{98.31} & 93.15 & 91.70 & 87.03 & 67.41 & 72.49 & 84.68 & 76.37 & 65.53 & 89.04 & 88.50 & 56.30 \\
 &  &  & CNN-T & 92.69 & 77.10 & 84.21 & 87.60 & 81.60 & \underline{78.76} & 89.43 & 73.36 & 74.23 & 88.74 & 92.49 & 55.71 \\
 &  &  & Deformer & \underline{98.18} & 93.62 & 82.27 & \textbf{92.27} & 66.37 & 69.06 & 91.90 & 78.01 & 65.28 & 92.26 & 90.50 & 62.41 \\
 &  &  & Conformer & 95.97 & 92.37 & 94.42 & 90.89 & 64.39 & 65.20 & 90.40 & 86.45 & 65.36 & 88.16 & 84.17 & 58.91 \\
\bottomrule
    \end{tabular}
    }
    \vspace{-0em}
\end{table*}
\clearpage

\begin{table*}[!t]
    \centering
    \renewcommand{\arraystretch}{1.09}
    \vspace{-0em}
    \fontsize{9}{12}\selectfont
    \caption{Balanced classification accuracies (\%) on CHB-MIT. The best BCAs are marked in bold, and the second best by an underline (continued).}
    \label{tab:chb_bca2}
    \scalebox{0.65}{
    \begin{tabular}{c|c|w{c}{1.25cm}|w{c}{1.9cm}|*{11}{w{c}{0.56cm}}|w{c}{1.6cm}}
        \toprule
        Scenario & Tuning & Model Type & Approach & S12 & S13 & S14 & S15 & S16 & S17 & S18 & S19 & S20 & S21 & S22 & Avg. \\
        \midrule
\multirow{36}{*}{\makecell{Cross-subject \\ (LOSO)}}%
& \multirow{22}{*}{\makecell{Full \\ Fine-tuning}}%
& \multirow{8}{*}{\makecell{Specialist \\ Models}}%
 & PSD+SVM & 56.12 & 51.04 & 49.86 & 49.27 & 50.00 & 50.00 & 84.62 & 49.96 & 50.00 & 87.22 & 66.25 & 69.09$_{\pm0.04}$ \\
 &  &  & EEGNet & 54.37 & 30.62 & 54.02 & 75.54 & 94.87 & \textbf{86.18} & 82.05 & 75.56 & 57.77 & 90.78 & 93.24 & 77.36$_{\pm0.54}$ \\
 &  &  & ShallowConv & 54.55 & 38.57 & 69.35 & 75.29 & 91.68 & \underline{86.05} & 95.30 & 76.33 & 64.94 & 95.19 & 92.68 & \textbf{80.18}$_{\pm0.22}$ \\
 &  &  & LMDA & 61.56 & 39.29 & 58.09 & 74.02 & 88.82 & 79.03 & \underline{96.15} & 69.55 & 56.69 & 92.44 & \underline{95.52} & 77.47$_{\pm0.41}$ \\
 &  &  & CNN-T & \textbf{70.90} & \underline{55.64} & 51.79 & 57.07 & 64.55 & 73.09 & 93.48 & 59.63 & 68.34 & 89.22 & \textbf{96.10} & 76.34$_{\pm0.61}$ \\
 &  &  & Deformer & 55.98 & 41.81 & \underline{74.85} & 74.82 & \textbf{96.25} & 80.38 & \underline{96.15} & \underline{81.35} & \textbf{71.57} & \textbf{96.19} & 83.17 & \underline{79.77}$_{\pm0.14}$ \\
 &  &  & Conformer & 56.59 & 21.68 & 68.84 & 72.63 & \underline{95.31} & 80.66 & \textbf{98.29} & \textbf{83.68} & 64.63 & \underline{95.56} & 92.78 & 78.58$_{\pm0.56}$ \\
\bottomrule
    \end{tabular}
    }
    \vspace{-0em}
\end{table*}
\clearpage


\begin{table*}[!t]
    \centering
    \renewcommand{\arraystretch}{1.4}
    \vspace{-0em}
    \fontsize{9}{12}\selectfont
    \caption{Classification accuracies (\%) on TUAB. The best accuracies are marked in bold, and the second best by an underline.}
    \label{tab:tuab_acc_cross}
    \scalebox{0.73}{
    \begin{tabular}{c|c|w{c}{1.5cm}|w{c}{1.9cm}|*{10}{w{c}{0.52cm}}|w{c}{1.3cm}}
        \toprule
        Scenario & Tuning & Model Type & Approach & S0 & S1 & S2 & S3 & S4 & S5 & S6 & S7 & S8 & S9 & Avg. \\
        \midrule
    \multirow{36}{*}{\makecell{Cross-subject \\ (LOSO)}}%
    & \multirow{22}{*}{\makecell{Full \\ Fine-tuning}}%
    & \multirow{8}{*}{\makecell{Specialist \\ Models}}%
 & CSP+LDA & 67.01 & 65.16 & 67.06 & 65.41 & 65.19 & 63.65 & 68.40 & 65.69 & 66.18 & 66.58 & 66.03$_{\pm0.25}$ \\
 &  &  & EEGNet & 75.88 & 78.02 & 75.68 & 77.91 & 77.16 & 79.49 & 76.19 & 77.17 & 77.54 & 75.23 & 77.03$_{\pm0.67}$ \\
 &  &  & ShallowConv & 80.01 & 79.37 & 79.00 & 79.34 & 80.04 & 80.58 & 79.94 & 79.67 & 79.91 & 80.19 & 79.81$_{\pm0.12}$ \\
 &  &  & LMDA & 55.08 & 63.90 & 65.58 & 55.30 & 70.41 & 60.98 & 67.43 & 53.15 & 61.63 & 73.46 & 62.69$_{\pm1.52}$ \\
 &  &  & CNN-T & 75.73 & 77.69 & 71.97 & 68.19 & 73.36 & 75.70 & 73.68 & 68.20 & 75.16 & 72.29 & 73.20$_{\pm0.73}$ \\
 &  &  & Deformer & \underline{82.61} & \textbf{81.94} & 80.53 & 79.54 & 81.01 & \underline{82.45} & \underline{82.11} & \underline{82.14} & \underline{81.39} & 81.10 & \underline{81.48}$_{\pm0.21}$ \\
 &  &  & Conformer & 78.26 & 78.34 & 77.25 & 77.56 & 77.95 & 77.85 & 77.48 & 77.71 & 77.11 & 78.29 & 77.78$_{\pm0.04}$ \\
\bottomrule
    \end{tabular}
    }
    \vspace{-0em}
\end{table*}
\clearpage

\begin{table*}[!t]
    \centering
    \renewcommand{\arraystretch}{1.4}
    \vspace{-0em}
    \fontsize{9}{12}\selectfont
    \caption{Cohen's Kappa (\%) on TUAB. The best metrics are marked in bold, and the second best by an underline.}
    \label{tab:tuab_kappa_cross}
    \scalebox{0.73}{
    \begin{tabular}{c|c|w{c}{1.5cm}|w{c}{1.9cm}|*{10}{w{c}{0.52cm}}|w{c}{1.3cm}}
        \toprule
        Scenario & Tuning & Model Type & Approach & S0 & S1 & S2 & S3 & S4 & S5 & S6 & S7 & S8 & S9 & Avg. \\
        \midrule
    \multirow{36}{*}{\makecell{Cross-subject \\ (LOSO)}}%
    & \multirow{22}{*}{\makecell{Full \\ Fine-tuning}}%
    & \multirow{8}{*}{\makecell{Specialist \\ Models}}%
 & CSP+LDA & 33.95 & 29.97 & 34.34 & 31.04 & 30.14 & 27.16 & 36.63 & 30.71 & 32.16 & 32.80 & 31.89$_{\pm0.52}$ \\
 &  &  & EEGNet & 51.73 & 55.97 & 51.53 & 55.97 & 54.28 & 58.67 & 52.50 & 54.22 & 55.06 & 50.76 & 54.07$_{\pm1.29}$ \\
 &  &  & ShallowConv & 59.95 & 58.57 & 58.09 & 58.75 & 59.92 & 61.08 & 59.51 & 59.19 & 59.64 & 60.14 & 59.48$_{\pm0.24}$ \\
 &  &  & LMDA & 9.59 & 26.14 & 31.61 & 11.37 & 39.51 & 20.50 & 33.66 & 0.74 & 22.81 & 47.21 & 24.31$_{\pm3.77}$ \\
 &  &  & CNN-T & 51.32 & 54.80 & 44.72 & 37.22 & 46.31 & 51.00 & 46.44 & 34.39 & 50.03 & 44.05 & 46.03$_{\pm1.34}$ \\
 &  &  & Deformer & \underline{65.17} & \textbf{63.61} & 61.18 & 59.41 & 61.77 & \underline{64.83} & \underline{63.94} & \underline{63.86} & \underline{62.70} & 61.91 & \underline{62.84}$_{\pm0.41}$ \\
 &  &  & Conformer & 56.44 & 56.50 & 54.59 & 55.22 & 55.66 & 55.53 & 54.63 & 55.02 & 54.02 & 56.18 & 55.38$_{\pm0.09}$ \\
\bottomrule
    \end{tabular}
    }
    \vspace{-0em}
\end{table*}
\clearpage



\begin{table*}[!t]
    \centering
    \renewcommand{\arraystretch}{1.23}
    \vspace{-0em}
    \fontsize{9}{12}\selectfont
    \caption{Cohen's Kappa (\%) on Sleep-EDFx. The best metrics are marked in bold, and the second best by an underline.}
    \label{tab:sleep_kappa_cross}
    \scalebox{0.6}{
    \begin{tabular}{c|c|w{c}{1.5cm}|w{c}{1.9cm}|*{10}{w{c}{0.85cm}}|w{c}{1.6cm}}
        \toprule
        Scenario & Tuning & Model Type & Approach & S0 & S1 & S2 & S3 & S4 & S5 & S6 & S7 & S8 & S9 & Avg. \\
        \midrule
    \multirow{32}{*}{\makecell{Cross-subject \\ (LOSO)}}%
    & \multirow{20}{*}{\makecell{Full \\ Fine-tuning}}%
    & \multirow{8}{*}{\makecell{Specialist \\ Models}}%
 & PSD+LDA & 56.10 & 59.27 & 63.88 & 56.79 & 57.29 & 57.88 & 61.43 & 61.09 & 54.35 & 60.02 & 58.81$_{\pm0.18}$ \\
 &  &  & EEGNet & 63.53 & 64.88 & 67.99 & 64.09 & 57.62 & 59.65 & 66.20 & 66.81 & 63.21 & 69.66 & 64.36$_{\pm0.46}$ \\
 &  &  & ShallowConv & 68.04 & 73.86 & 73.79 & 70.82 & 70.45 & 74.03 & 69.73 & 73.05 & 69.78 & 74.67 & 71.82$_{\pm0.32}$ \\
 &  &  & LMDA & 68.32 & 71.47 & 73.13 & 67.84 & 67.83 & 71.82 & 69.69 & 72.17 & 65.08 & 71.87 & 69.92$_{\pm0.12}$ \\
 &  &  & CNN-T & 67.52 & 72.69 & 73.21 & 69.17 & 70.71 & 70.47 & 69.42 & 73.14 & 66.21 & 72.74 & 70.53$_{\pm0.29}$ \\
 &  &  & Deformer & \underline{72.88} & \underline{76.44} & \underline{78.41} & \underline{73.18} & \underline{72.55} & \underline{76.14} & \underline{74.63} & \underline{75.42} & 71.62 & \underline{77.20} & \underline{74.85}$_{\pm0.23}$ \\
 &  &  & Conformer & 59.78 & 68.40 & 66.12 & 59.30 & 58.51 & 63.49 & 60.56 & 66.83 & 54.80 & 67.53 & 62.53$_{\pm0.54}$ \\
\bottomrule
    \end{tabular}
    }
    \vspace{-0em}
\end{table*}

\begin{table*}[!t]
    \centering
    \renewcommand{\arraystretch}{1.23}
    \vspace{-0em}
    \fontsize{9}{12}\selectfont
    \caption{Balanced classification accuracies (\%) on Sleep-EDFx. The best BCAs are marked in bold, and the second best by an underline.}
    \label{tab:sleep_bca_cross}
    \scalebox{0.6}{
    \begin{tabular}{c|c|w{c}{1.5cm}|w{c}{1.9cm}|*{10}{w{c}{0.85cm}}|w{c}{1.6cm}}
        \toprule
        Scenario & Tuning & Model Type & Approach & S0 & S1 & S2 & S3 & S4 & S5 & S6 & S7 & S8 & S9 & Avg. \\
        \midrule
\multirow{32}{*}{Cross-subject}%
& \multirow{20}{*}{\makecell{Full \\ Fine-tuning}}%
& \multirow{8}{*}{\makecell{Specialist \\ Models}}%
 & PSD+LDA & 50.10 & 51.22 & 54.46 & 52.93 & 50.65 & 50.79 & 53.69 & 51.11 & 49.63 & 53.23 & 51.78$_{\pm0.22}$ \\
 &  &  & EEGNet & 72.61 & 73.33 & 76.61 & 73.54 & 71.52 & 72.23 & 73.94 & 73.97 & 73.22 & 76.48 & 73.75$_{\pm0.19}$ \\
 &  &  & ShallowConv & 71.21 & 75.28 & 76.94 & 74.56 & 75.21 & 76.35 & 73.75 & 73.93 & 73.90 & 77.52 & 74.86$_{\pm0.42}$ \\
 &  &  & LMDA & 71.89 & 74.39 & 77.64 & 73.91 & 73.72 & 75.17 & \underline{74.84} & 75.13 & 72.84 & 76.27 & 74.58$_{\pm0.34}$ \\
 &  &  & CNN-T & 73.65 & 75.88 & 76.88 & 74.88 & 76.47 & 76.70 & 73.84 & \underline{77.56} & 74.49 & 77.08 & 75.74$_{\pm0.48}$ \\
 &  &  & Deformer & \textbf{76.20} & \textbf{79.32} & \textbf{81.04} & \textbf{78.35} & \textbf{78.19} & \textbf{79.99} & \textbf{77.53} & \textbf{78.57} & \textbf{76.94} & \textbf{81.19} & \textbf{78.73}$_{\pm0.09}$ \\
 &  &  & Conformer & 65.74 & 69.70 & 71.25 & 69.89 & 67.08 & 70.07 & 68.00 & 68.04 & 63.71 & 70.53 & 68.40$_{\pm2.87}$ \\
\bottomrule
    \end{tabular}
    }
    \vspace{-0em}
\end{table*}
\clearpage


\begin{table*}[!t]
    \centering
    \renewcommand{\arraystretch}{1.38}
    \vspace{-0em}
    \fontsize{9}{12}\selectfont
    \caption{Cohen's Kappa (\%) on Sleep-EDFx. The best metrics are marked in bold, and the second best by an underline.}
    \label{tab:sleep_kappa1}
    \scalebox{0.56}{
    \begin{tabular}{w{c}{1.6cm}|w{c}{1.3cm}|w{c}{1.3cm}|w{c}{1.8cm}|*{20}{w{c}{0.45cm}}}
        \toprule
        Scenario & Tuning & Model Type & Approach & S0 & S1 & S2 & S3 & S4 & S5 & S6 & S7 & S8 & S9 & S10 & S11 & S12 & S13 & S14 & S15 & S16 & S17 & S18 & S19 \\
        \midrule
\multirow{32}{*}{\makecell{Within-subject \\ (Few-shot)}}%
& \multirow{20}{*}{\makecell{Full \\ Fine-tuning}}%
& \multirow{8}{*}{\makecell{Specialist \\ Models}}%
 & PSD+LDA & 76.05 & 76.34 & 71.08 & 72.37 & 70.51 & 71.57 & 78.11 & 75.49 & 69.06 & 73.75 & 74.44 & 54.00 & 64.18 & 62.49 & 81.67 & 70.96 & 60.14 & 60.24 & 75.96 & 64.20 \\
 &  &  & EEGNet & 68.07 & 66.92 & 65.91 & 65.31 & 64.62 & 58.55 & 70.20 & 66.64 & 60.08 & 64.74 & 67.14 & 49.53 & 54.38 & 56.74 & 67.67 & 66.85 & 53.43 & 55.28 & 70.70 & 51.05 \\
 &  &  & ShallowConv & 79.86 & 75.38 & 74.20 & 78.72 & 73.63 & 65.23 & \underline{79.29} & 75.15 & 70.52 & 72.05 & \underline{78.05} & 63.33 & 66.42 & 75.73 & 82.40 & 72.67 & 51.99 & 62.81 & 76.39 & \textbf{65.52} \\
 &  &  & LMDA & 68.53 & 60.54 & 64.10 & 71.80 & 56.80 & 60.37 & 64.88 & 64.24 & 65.13 & 57.15 & 56.75 & 51.41 & 55.35 & 32.78 & 60.03 & 57.98 & 33.87 & 41.19 & 72.18 & 47.23 \\
 &  &  & CNN-T & 78.99 & 79.04 & \textbf{83.43} & \underline{82.28} & \textbf{76.73} & 64.38 & 77.43 & 81.90 & 76.95 & \underline{75.15} & \textbf{79.48} & 65.79 & 57.00 & 60.84 & 72.67 & 74.04 & \textbf{75.59} & \underline{64.08} & 79.52 & 63.98 \\
 &  &  & Deformer & 74.48 & 76.89 & 76.21 & 74.37 & 71.64 & 75.37 & 77.22 & 76.09 & 69.98 & 69.73 & 75.19 & 48.38 & 65.95 & 63.33 & 70.70 & \underline{74.74} & 65.84 & 61.88 & 75.98 & 55.29 \\
 &  &  & Conformer & 80.60 & 68.84 & 80.33 & 81.97 & 73.85 & 69.94 & 73.32 & \underline{82.02} & 65.89 & 68.86 & 75.48 & \textbf{68.09} & 65.76 & 67.50 & \underline{86.49} & 74.70 & 68.47 & 49.82 & 73.78 & 61.65 \\
\bottomrule
    \end{tabular}
    }
    \vspace{-0em}
\end{table*}
\clearpage

\begin{table*}[!t]
    \centering
    \renewcommand{\arraystretch}{1.38}
    \vspace{-0em}
    \fontsize{9}{12}\selectfont
    \caption{Cohen's Kappa (\%) on Sleep-EDFx. The best metrics are marked in bold, and the second best by an underline (continued).}
    \label{tab:sleep_kappa2}
    \scalebox{0.56}{
    \begin{tabular}{w{c}{1.6cm}|w{c}{1.3cm}|w{c}{1.3cm}|w{c}{1.78cm}|*{20}{w{c}{0.46cm}}}
        \toprule
        Scenario & Tuning & Model Type & Approach & S40 & S41 & S42 & S43 & S44 & S45 & S46 & S47 & S48 & S49 & S50 & S51 & S52 & S53 & S54 & S55 & S56 & S57 & S58 & S59 \\
        \midrule
\multirow{32}{*}{\makecell{Within-subject \\ (Few-shot)}}%
& \multirow{20}{*}{\makecell{Full \\ Fine-tuning}}%
& \multirow{8}{*}{\makecell{Specialist \\ Models}}%
 & PSD+LDA & 60.81 & \underline{79.48} & 57.11 & \underline{62.76} & 68.15 & \textbf{72.18} & 34.91 & 70.09 & 65.06 & 35.52 & \underline{65.74} & 68.05 & 67.14 & 61.19 & 60.39 & 63.43 & 38.45 & \underline{64.77} & 55.56 & 56.93 \\
 &  &  & EEGNet & 50.38 & 64.13 & 58.96 & 55.70 & 63.84 & 64.56 & 29.35 & 58.49 & 52.60 & 27.76 & 53.26 & 47.94 & 63.90 & 56.26 & \underline{63.56} & 47.90 & 33.88 & 54.83 & 28.16 & 56.34 \\
 &  &  & ShallowConv & 62.44 & 76.83 & 67.35 & 61.78 & 72.67 & 68.42 & 36.94 & 61.12 & 63.81 & 25.67 & 53.44 & 56.91 & 63.39 & 62.93 & 63.05 & 59.87 & 50.78 & 62.06 & 35.74 & 61.59 \\
 &  &  & LMDA & 47.38 & 69.64 & 37.59 & 51.89 & 57.61 & 59.41 & 21.42 & 48.23 & 45.17 & 21.05 & 33.30 & 28.06 & 57.72 & 43.58 & 59.97 & 42.24 & 23.54 & 39.14 & 28.18 & 43.34 \\
 &  &  & CNN-T & 58.41 & 74.38 & 66.03 & 56.76 & \textbf{79.10} & 57.96 & 26.74 & 54.55 & \underline{68.41} & 24.70 & 59.89 & 63.16 & 61.25 & 61.30 & 53.36 & 60.27 & 49.76 & 45.95 & 35.16 & \textbf{65.55} \\
 &  &  & Deformer & 63.82 & 75.53 & 67.95 & 59.28 & 67.85 & 65.33 & 37.83 & 67.75 & 61.26 & 30.30 & 30.92 & \underline{69.33} & \underline{69.16} & 62.23 & 60.77 & \textbf{66.81} & 52.39 & 59.73 & 31.10 & 61.72 \\
 &  &  & Conformer & 56.19 & 69.18 & 67.74 & 48.95 & 62.50 & 64.98 & 40.67 & \textbf{77.18} & 56.99 & 28.16 & 60.73 & 56.52 & 59.94 & 61.42 & 52.65 & 61.10 & 45.71 & 50.14 & 47.21 & 62.99 \\
\bottomrule
    \end{tabular}
    }
    \vspace{-0em}
\end{table*}
\clearpage


\begin{table*}[!t]
    \centering
    \renewcommand{\arraystretch}{1.38}
    \vspace{-0em}
    \fontsize{9}{12}\selectfont
    \caption{Balanced classification accuracies (\%) on Sleep-EDFx. The best BCAs are marked in bold, and the second best by an underline.}
    \label{tab:sleep_bca1}
    \scalebox{0.56}{
    \begin{tabular}{w{c}{1.6cm}|w{c}{1.3cm}|w{c}{1.3cm}|w{c}{1.8cm}|*{20}{w{c}{0.45cm}}}
        \toprule
        Scenario & Tuning & Model Type & Approach & S0 & S1 & S2 & S3 & S4 & S5 & S6 & S7 & S8 & S9 & S10 & S11 & S12 & S13 & S14 & S15 & S16 & S17 & S18 & S19 \\
        \midrule
\multirow{32}{*}{\makecell{Within-subject \\ (Few-shot)}}%
& \multirow{20}{*}{\makecell{Full \\ Fine-tuning}}%
& \multirow{8}{*}{\makecell{Specialist \\ Models}}%
 & PSD+LDA & 71.43 & 73.06 & 67.93 & 64.96 & 68.87 & 62.69 & 68.73 & 64.23 & 66.30 & 66.07 & 54.66 & 51.15 & 67.60 & 62.80 & 73.51 & 70.79 & 54.38 & 59.56 & 63.48 & 63.97 \\
 &  &  & EEGNet & 58.14 & 54.34 & 53.86 & 50.42 & 54.41 & 54.70 & 56.53 & 58.14 & 55.69 & 53.49 & 46.63 & 42.77 & 49.06 & 43.13 & 55.60 & 61.21 & 48.94 & 46.11 & 58.51 & 47.10 \\
 &  &  & ShallowConv & 69.50 & 67.90 & 57.19 & 65.42 & 64.58 & 61.73 & 65.33 & 65.27 & 63.85 & 63.55 & 53.22 & 50.21 & 64.40 & 62.98 & 69.71 & 67.44 & 55.37 & 55.59 & 61.95 & 62.19 \\
 &  &  & LMDA & 59.61 & 53.53 & 52.33 & 58.97 & 46.58 & 52.21 & 56.92 & 56.80 & 62.22 & 52.15 & 46.31 & 44.57 & 55.39 & 35.66 & 54.66 & 59.04 & 43.11 & 42.31 & 60.06 & 47.75 \\
 &  &  & CNN-T & \textbf{74.44} & \underline{78.51} & \textbf{73.99} & 73.40 & \textbf{76.30} & 67.55 & 72.20 & 73.76 & \textbf{76.20} & 69.53 & 68.73 & \textbf{64.50} & 65.99 & 67.89 & 69.73 & \textbf{77.11} & \textbf{70.39} & \textbf{62.68} & \underline{72.12} & 65.82 \\
 &  &  & Deformer & 63.51 & 58.19 & 61.45 & 55.23 & 58.15 & 66.52 & 63.11 & 67.33 & 61.94 & 54.27 & 53.46 & 40.62 & 58.59 & 47.15 & 58.51 & 65.45 & 55.79 & 50.26 & 59.66 & 46.04 \\
 &  &  & Conformer & 72.04 & 70.56 & 70.92 & \textbf{75.80} & 75.44 & 66.47 & 65.62 & \underline{76.07} & 69.83 & 58.89 & 60.29 & 57.31 & 67.35 & 72.64 & \textbf{80.12} & 72.55 & 67.28 & 56.72 & 64.59 & \underline{72.69} \\
\bottomrule
    \end{tabular}
    }
    \vspace{-0em}
\end{table*}
\clearpage

\begin{table*}[!t]
    \centering
    \renewcommand{\arraystretch}{1.38}
    \vspace{-0em}
    \fontsize{9}{12}\selectfont
    \caption{Balanced classification accuracies (\%) on Sleep-EDFx. The best BCAs are marked in bold, and the second best by an underline (continued).}
    \label{tab:sleep_bca2}
    \scalebox{0.56}{
    \begin{tabular}{w{c}{1.6cm}|w{c}{1.3cm}|w{c}{1.3cm}|w{c}{1.78cm}|*{20}{w{c}{0.46cm}}}
        \toprule
        Scenario & Tuning & Model Type & Approach & S40 & S41 & S42 & S43 & S44 & S45 & S46 & S47 & S48 & S49 & S50 & S51 & S52 & S53 & S54 & S55 & S56 & S57 & S58 & S59 \\
        \midrule
\multirow{32}{*}{\makecell{Within-subject \\ (Few-shot)}}%
& \multirow{20}{*}{\makecell{Full \\ Fine-tuning}}%
& \multirow{8}{*}{\makecell{Specialist \\ Models}}%
 & PSD+LDA & 59.46 & 74.13 & 57.22 & 62.63 & 70.36 & 58.04 & 53.13 & 67.96 & 70.65 & 40.97 & 55.67 & 57.74 & 59.62 & 58.24 & 58.23 & 64.30 & 34.91 & 45.94 & \underline{59.78} & 55.52 \\
 &  &  & EEGNet & 47.10 & 53.53 & 44.73 & 46.44 & 61.93 & 44.32 & 42.13 & 54.34 & 45.10 & 35.09 & 38.44 & 35.91 & 49.83 & 44.05 & 44.62 & 43.49 & 34.12 & 43.43 & 39.93 & 51.22 \\
 &  &  & ShallowConv & 60.75 & 70.65 & 56.67 & 58.53 & 70.56 & 46.68 & 51.98 & 61.58 & 55.58 & 33.39 & 38.76 & 38.40 & 47.89 & 52.59 & 53.30 & 52.45 & 39.61 & 47.89 & 43.81 & 57.49 \\
 &  &  & LMDA & 49.21 & 62.67 & 42.43 & 50.75 & 60.69 & 43.64 & 42.73 & 48.26 & 40.86 & 28.87 & 30.79 & 29.19 & 44.80 & 39.68 & 47.44 & 39.42 & 29.73 & 36.64 & 39.23 & 48.87 \\
 &  &  & CNN-T & \textbf{71.05} & \underline{76.11} & \textbf{69.04} & \textbf{73.02} & \underline{79.44} & 44.92 & 51.49 & 65.71 & \textbf{77.53} & 45.01 & 51.08 & 63.56 & 57.63 & \textbf{75.15} & \underline{65.70} & 56.95 & 45.85 & 47.59 & 45.67 & \textbf{68.56} \\
 &  &  & Deformer & 56.30 & 69.29 & 52.96 & 53.22 & 62.77 & 43.87 & 54.71 & 58.60 & 48.95 & 34.51 & 29.67 & 39.49 & 51.37 & 46.50 & 50.29 & 52.21 & 44.49 & 44.13 & 39.48 & 53.39 \\
 &  &  & Conformer & 60.02 & 68.23 & 65.21 & 63.30 & 67.76 & 60.65 & 56.11 & \textbf{75.93} & 71.74 & \underline{47.85} & 54.51 & \textbf{65.00} & 56.96 & 70.55 & 60.03 & \underline{67.22} & \underline{58.71} & 60.61 & 45.84 & \underline{64.46} \\
\bottomrule
    \end{tabular}
    }
    \vspace{-0em}
\end{table*}
\clearpage


\begin{table*}[!t]
    \centering
    \renewcommand{\arraystretch}{1.25}
    \vspace{-0em}
    \fontsize{9}{12}\selectfont
    \caption{Classification accuracies (\%) on SEED. The best accuracies are marked in bold, and the second best by an underline.}
    \label{tab:seed_acc}
    \scalebox{0.56}{
    \begin{tabular}{c|c|w{c}{1.25cm}|w{c}{1.8cm}|*{15}{w{c}{0.56cm}}|w{c}{1.5cm}}
        \toprule
        Scenario & Tuning & Model Type & Approach & S0 & S1 & S2 & S3 & S4 & S5 & S6 & S7 & S8 & S9 & S10 & S11 & S12 & S13 & S14 & Avg. \\
        \midrule
    \multirow{36}{*}{\makecell{Cross-subject \\ (LOSO)}}%
    & \multirow{22}{*}{\makecell{Full \\ Fine-tuning}}%
    & \multirow{8}{*}{\makecell{Specialist \\ Models}}%
 & CSP+LDA & 33.68 & 46.91 & 45.02 & 58.93 & 37.07 & 58.66 & 38.66 & 59.63 & 41.07 & 57.13 & 62.67 & 48.00 & 59.13 & 52.68 & 34.47 & 48.91 \\
 &  &  & EEGNet & 32.40 & 39.06 & 47.20 & 54.10 & 43.64 & 53.91 & 41.42 & 65.06 & \underline{59.62} & 39.16 & 55.98 & 48.72 & 53.87 & 62.87 & 31.57 & 48.57$_{\pm 1.34}$ \\
 &  &  & ShallowConv & 38.14 & 48.57 & 36.32 & 55.76 & 52.64 & 57.65 & 50.71 & 69.46 & 54.11 & 48.23 & \textbf{66.54} & \textbf{59.75} & \underline{60.77} & \textbf{68.07} & 34.47 & \underline{53.41}$_{\pm 0.12}$ \\
 &  &  & LMDA & 33.02 & 41.07 & 50.85 & 59.70 & 41.44 & 60.80 & 38.19 & 67.78 & 50.91 & 53.78 & 51.10 & 55.17 & 50.36 & 58.90 & 38.79 & 50.12$_{\pm 0.43}$ \\
 &  &  & CNN-T & 31.58 & 33.36 & 44.49 & 56.87 & 37.94 & 32.16 & 38.35 & 60.03 & \textbf{61.22} & 51.66 & 52.70 & 44.89 & 43.86 & 45.98 & 33.34 & 44.56$_{\pm 1.40}$ \\
 &  &  & Deformer & 33.70 & \textbf{52.11} & 38.85 & 58.35 & 47.96 & 56.47 & 44.99 & 67.22 & 52.46 & 42.97 & \underline{66.27} & 49.66 & 59.12 & 66.39 & 29.31 & 51.05$_{\pm 0.97}$ \\
 &  &  & Conformer & 34.32 & 38.19 & 38.48 & 44.62 & 45.55 & 55.77 & 43.40 & 67.18 & 53.04 & 36.27 & 65.67 & 46.66 & \textbf{61.01} & 60.07 & 41.12 & 48.76$_{\pm 1.23}$ \\
\bottomrule
    \end{tabular}
    }
    \vspace{-0em}
\end{table*}
\clearpage

\begin{table*}[!t]
    \centering
    \renewcommand{\arraystretch}{1.25}
    \vspace{-0em}
    \fontsize{9}{12}\selectfont
    \caption{Cohen's Kappa (\%) on SEED. The best metrics are marked in bold, and the second best by an underline.}
    \label{tab:seed_kappa}
    \scalebox{0.56}{
    \begin{tabular}{c|c|w{c}{1.25cm}|w{c}{1.8cm}|*{15}{w{c}{0.56cm}}|w{c}{1.5cm}}
        \toprule
        Scenario & Tuning & Model Type & Approach & S0 & S1 & S2 & S3 & S4 & S5 & S6 & S7 & S8 & S9 & S10 & S11 & S12 & S13 & S14 & Avg. \\
        \midrule
    \multirow{36}{*}{\makecell{Cross-subject \\ (LOSO)}}%
    & \multirow{22}{*}{\makecell{Full \\ Fine-tuning}}%
    & \multirow{8}{*}{\makecell{Specialist \\ Models}}%
 & CSP+LDA & -1.19 & 20.93 & 17.19 & 38.53 & 4.22 & 37.92 & 8.86 & 39.11 & 10.36 & 35.85 & 44.10 & 21.48 & 38.18 & 29.26 & 0.00 & 22.99 \\
 &  &  & EEGNet & -0.79 & 9.48 & 20.03 & 31.48 & 15.43 & 31.17 & 12.77 & 47.38 & \underline{39.23} & 9.62 & 34.28 & 23.18 & 30.47 & 44.16 & -4.00 & 22.93$_{\pm 1.95}$ \\
 &  &  & ShallowConv & 6.58 & 23.44 & 2.94 & 33.66 & 28.73 & 36.16 & 26.52 & 54.04 & 30.76 & 22.86 & \textbf{49.85} & \textbf{39.62} & \underline{40.90} & \textbf{51.99} & 0.00 & \underline{29.87}$_{\pm 0.15}$ \\
 &  &  & LMDA & 0.03 & 12.50 & 25.79 & 39.56 & 12.22 & 40.93 & 8.23 & 51.66 & 25.92 & 30.99 & 27.16 & 33.05 & 25.64 & 38.40 & 8.76 & 25.39$_{\pm 0.62}$ \\
 &  &  & CNN-T & -4.14 & 1.21 & 16.22 & 35.20 & 7.31 & -2.98 & 8.46 & 39.94 & \textbf{41.69} & 27.67 & 29.11 & 17.63 & 16.44 & 18.86 & -1.03 & 16.77$_{\pm 2.09}$ \\
 &  &  & Deformer & -0.06 & \textbf{28.65} & 6.86 & 37.54 & 21.60 & 34.60 & 18.05 & 50.65 & 28.22 & 15.10 & \underline{49.46} & 24.57 & 38.29 & 49.53 & -6.78 & 26.42$_{\pm 1.49}$ \\
 &  &  & Conformer & 1.95 & 8.23 & 6.28 & 17.28 & 18.25 & 33.65 & 15.69 & 50.61 & 29.10 & 4.97 & 48.50 & 20.39 & \textbf{41.40} & 39.97 & 11.86 & 23.21$_{\pm 1.81}$ \\
\bottomrule
    \end{tabular}
    }
    \vspace{-0em}
\end{table*}
\clearpage


\begin{table*}[!t]
    \centering
    \renewcommand{\arraystretch}{1.1}
    \vspace{-0em}
    \fontsize{9}{12}\selectfont
    \caption{Classification accuracies (\%) on Nakanishi2015. The best accuracies are marked in bold, and the second best by an underline.}
    \label{tab:ssvep_acc}
    \scalebox{0.64}{
    \begin{tabular}{c|c|w{c}{1.5cm}|w{c}{2.0cm}|*{9}{w{c}{0.75cm}}|w{c}{1.7cm}}
        \toprule
        Scenario & Tuning & Model Type & Approach & S0 & S1 & S2 & S3 & S4 & S5 & S6 & S7 & S8 & Avg. \\
        \midrule
    \multirow{36}{*}{\makecell{Cross-subject \\ (LOSO)}}%
    & \multirow{22}{*}{\makecell{Full \\ Fine-tuning}}%
    & \multirow{8}{*}{\makecell{Specialist \\ Models}}%
 & TRCA & 91.67 & 72.78 & 85.00 & \textbf{100.00} & \underline{99.44} & \textbf{100.00} & \underline{99.44} & 98.89 & \underline{99.44} & 94.07 \\
 &  &  & EEGNet & \underline{92.78} & \underline{78.70} & \underline{92.96} & \textbf{100.00} & \textbf{99.63} & \textbf{100.00} & \underline{99.44} & \textbf{100.00} & \underline{99.44} & \underline{95.88}$_{\pm 0.18}$ \\
 &  &  & ShallowConv & 56.85 & 29.63 & 71.11 & 83.89 & 36.48 & 92.96 & 80.37 & 90.93 & 84.26 & 69.61$_{\pm 0.86}$ \\
 &  &  & LMDA & 62.59 & 42.59 & 81.48 & 98.89 & 87.59 & \underline{99.63} & 95.56 & \textbf{100.00} & 97.78 & 85.12$_{\pm 0.94}$ \\
 &  &  & CNN-T & 29.26 & 19.63 & 42.59 & 55.00 & 41.67 & 63.15 & 54.07 & 61.30 & 50.37 & 46.34$_{\pm 0.34}$ \\
 &  &  & Deformer & \textbf{93.33} & \textbf{86.85} & \textbf{94.81} & \textbf{100.00} & \textbf{99.63} & \textbf{100.00} & \textbf{100.00} & \textbf{100.00} & \textbf{100.00} & \textbf{97.18}$_{\pm 0.13}$ \\
 &  &  & Conformer & 82.41 & 65.56 & 86.48 & \textbf{100.00} & 92.96 & 99.44 & 96.48 & \underline{99.26} & \underline{99.44} & 91.34$_{\pm1.34}$ \\
\bottomrule
    \end{tabular}
    }
    \vspace{-0em}
\end{table*}
\clearpage

\begin{table*}[!t]
    \centering
    \renewcommand{\arraystretch}{1.1}
    \vspace{-0em}
    \fontsize{9}{12}\selectfont
    \caption{Cohen's Kappa (\%) on Nakanishi2015. The best metrics are marked in bold, and the second best by an underline.}
    \label{tab:ssvep_kappa}
    \scalebox{0.64}{
    \begin{tabular}{c|c|w{c}{1.5cm}|w{c}{2.0cm}|*{9}{w{c}{0.75cm}}|w{c}{1.7cm}}
        \toprule
        Scenario & Tuning & Model Type & Approach & S0 & S1 & S2 & S3 & S4 & S5 & S6 & S7 & S8 & Avg. \\
        \midrule
    \multirow{36}{*}{\makecell{Cross-subject \\ (LOSO)}}%
    & \multirow{22}{*}{\makecell{Full \\ Fine-tuning}}%
    & \multirow{8}{*}{\makecell{Specialist \\ Models}}%
 & TRCA & 90.91 & 70.30 & 83.64 & \textbf{100.00} & \underline{99.39} & \textbf{100.00} & \underline{99.39} & 98.79 & \underline{99.39} & 93.54 \\
 &  &  & EEGNet & \underline{92.12} & \underline{76.77} & \underline{92.32} & \textbf{100.00} & \textbf{99.60} & \textbf{100.00} & \underline{99.39} & \textbf{100.00} & \underline{99.39} & \underline{95.51}$_{\pm 0.19}$ \\
 &  &  & ShallowConv & 52.93 & 23.23 & 68.48 & 82.42 & 30.71 & 92.32 & 78.59 & 90.10 & 82.83 & 66.85$_{\pm 0.94}$ \\
 &  &  & LMDA & 59.19 & 37.37 & 79.80 & 98.79 & 86.46 & \underline{99.60} & 95.15 & \textbf{100.00} & 97.58 & 83.77$_{\pm 1.03}$ \\
 &  &  & CNN-T & 22.83 & 12.32 & 37.37 & 50.91 & 36.36 & 59.80 & 49.90 & 57.78 & 45.86 & 41.46$_{\pm 0.37}$ \\
 &  &  & Deformer & \textbf{92.73} & \textbf{85.66} & \textbf{94.34} & \textbf{100.00} & \textbf{99.60} & \textbf{100.00} & \textbf{100.00} & \textbf{100.00} & \textbf{100.00} & \textbf{96.92}$_{\pm 0.14}$ \\
 &  &  & Conformer & 80.81 & 62.42 & 85.25 & \textbf{100.00} & 92.32 & 99.39 & 96.16 & \underline{99.19} & \underline{99.39} & 90.55$_{\pm1.47}$ \\
\bottomrule
    \end{tabular}
    }
    \vspace{-0em}
\end{table*}
\clearpage



\begin{table*}[!t]
    \centering
    \renewcommand{\arraystretch}{1.3}
    \vspace{-0em}
    \fontsize{9}{12}\selectfont
    \caption{Classification accuracies (\%) on EEGMat. The best accuracies are marked in bold, and the second best by an underline.}
    \label{tab:eegmat_acc1}
    \scalebox{0.53}{
\begin{tabular}{w{c}{1.6cm}|w{c}{1.35cm}|w{c}{1.4cm}|w{c}{1.9cm}|*{19}{w{c}{0.54cm}}}
        \toprule
        Scenario & Tuning & Model Type & Approach & S0 & S1 & S2 & S3 & S4 & S5 & S6 & S7 & S8 & S9 & S10 & S11 & S12 & S13 & S14 & S15 & S16 & S17 & S18 \\
        \midrule
    \multirow{36}{*}{\makecell{Cross-subject \\ (LOSO)}}%
    & \multirow{22}{*}{\makecell{Full \\ Fine-tuning}}%
    & \multirow{8}{*}{\makecell{Specialist \\ Models}}%
 & CSP+LDA & \underline{83.33} & 50.00 & 50.00 & 50.00 & \textbf{100.00} & 53.33 & \textbf{93.33} & \underline{80.00} & 56.67 & 76.67 & 73.33 & 46.67 & 50.00 & 50.00 & 50.00 & 76.67 & 70.00 & 50.00 & 93.33 \\
 &  &  & EEGNet & 58.89 & 51.11 & 50.00 & \textbf{94.44} & 90.00 & 54.44 & 63.33 & 61.11 & 70.00 & 83.33 & 85.56 & 57.78 & 53.33 & 40.00 & 48.89 & 65.56 & 46.67 & 63.33 & 84.44 \\
 &  &  & ShallowConv & 68.89 & 58.89 & 61.11 & 75.56 & 86.67 & 57.78 & 86.67 & 66.67 & 65.56 & \underline{91.11} & 94.44 & 83.33 & \underline{56.67} & 47.78 & 53.33 & 63.33 & 74.44 & 70.00 & \textbf{97.78} \\
 &  &  & LMDA & 68.89 & 58.89 & 47.78 & \underline{91.11} & 82.22 & \textbf{81.11} & 63.33 & 55.56 & 71.11 & 85.56 & 75.56 & 58.89 & 38.89 & 61.11 & 55.56 & 70.00 & 50.00 & 75.56 & 75.56 \\
 &  &  & CNN-T & 68.89 & 67.78 & 53.33 & 57.78 & 74.44 & 73.33 & 70.00 & 62.22 & 81.11 & 74.44 & 84.44 & 73.33 & 36.67 & 42.22 & 48.89 & 62.22 & \underline{83.33} & 70.00 & 65.56 \\
 &  &  & Deformer & 77.78 & 57.78 & 54.44 & 82.22 & \underline{94.44} & 50.00 & 87.78 & 73.33 & 65.56 & 85.56 & 92.22 & 76.67 & 45.56 & 40.00 & 61.11 & 65.56 & 71.11 & \textbf{92.22} & 64.44 \\
 &  &  & Conformer & 78.89 & 64.44 & 55.56 & 71.11 & 88.89 & 55.56 & \underline{90.00} & 64.44 & 65.56 & 76.67 & \textbf{100.00} & \textbf{88.89} & 51.11 & 47.78 & 53.33 & 61.11 & 74.44 & 64.44 & 92.22 \\
\bottomrule
    \end{tabular}
    }
    \vspace{-0em}
\end{table*}
\clearpage

\begin{table*}[!t]
    \centering
    \renewcommand{\arraystretch}{1.3}
    \vspace{-0em}
    \fontsize{9}{12}\selectfont
    \caption{Classification accuracies (\%) on EEGMat. The best accuracies are marked in bold, and the second best by an underline.}
    \label{tab:eegmat_acc2}
    \scalebox{0.55}{
    \begin{tabular}{w{c}{1.6cm}|w{c}{1.25cm}|w{c}{1.25cm}|w{c}{1.9cm}|*{17}{w{c}{0.51cm}}|w{c}{1.5cm}}
        \toprule
        Scenario & Tuning & Model Type & Approach & S19 & S20 & S21 & S22 & S23 & S24 & S25 & S26 & S27 & S28 & S29 & S30 & S31 & S32 & S33 & S34 & S35 & Avg. \\
        \midrule
    \multirow{36}{*}{\makecell{Cross-subject \\ (LOSO)}}%
    & \multirow{22}{*}{\makecell{Full \\ Fine-tuning}}%
    & \multirow{8}{*}{\makecell{Specialist \\ Models}}%
 & CSP+LDA & 96.67 & \textbf{90.00} & 70.00 & 70.00 & 36.67 & 56.67 & 50.00 & \textbf{83.33} & 60.00 & \textbf{93.33} & 50.00 & 50.00 & 63.33 & 70.00 & 56.67 & 80.00 & \textbf{96.67} & 67.41 \\
 &  &  & EEGNet & 92.22 & 85.56 & 64.44 & 64.44 & 65.56 & 67.78 & 68.89 & 61.11 & 71.11 & 62.22 & \textbf{65.56} & 56.67 & 60.00 & 67.78 & 62.22 & 78.89 & \underline{81.11} & 66.60$_{\pm 0.63}$ \\
 &  &  & ShallowConv & 94.44 & 73.33 & 77.78 & 76.67 & 81.11 & \underline{84.44} & \textbf{86.67} & 70.00 & 60.00 & \underline{92.22} & 52.22 & 57.78 & 56.67 & 62.22 & 63.33 & 72.22 & 78.89 & 72.22$_{\pm 1.24}$ \\
 &  &  & LMDA & 82.22 & 65.56 & 64.44 & 91.11 & 71.11 & 50.00 & 73.33 & 71.11 & 77.78 & 62.22 & 60.00 & 73.33 & 43.33 & 51.11 & 70.00 & 87.78 & 67.78 & 67.47$_{\pm 1.21}$ \\
 &  &  & CNN-T & 90.00 & \underline{88.89} & 66.67 & 93.33 & 75.56 & 73.33 & 68.83 & 71.11 & 71.11 & 62.22 & 51.11 & 83.33 & \textbf{72.22} & \textbf{84.44} & \textbf{87.78} & 90.00 & 67.78 & 70.77$_{\pm 2.49}$ \\
 &  &  & Deformer & 97.78 & 75.56 & 82.22 & 77.78 & 83.33 & 83.33 & 82.22 & 70.00 & 58.89 & 83.33 & 50.00 & 62.22 & 62.22 & 68.89 & 56.67 & 74.44 & 75.56 & 71.73$_{\pm 0.37}$ \\
 &  &  & Conformer & 94.44 & 61.11 & 75.56 & 61.11 & \underline{85.53} & 76.67 & \underline{85.56} & 71.11 & 58.89 & 90.00 & 50.00 & 52.22 & 53.33 & 65.53 & 64.44 & 70.00 & 77.78 & 70.49$_{\pm 0.93}$ \\
\bottomrule
    \end{tabular}
    }
    \vspace{-0em}
\end{table*}
\clearpage


\begin{table*}[!t]
    \centering
    \renewcommand{\arraystretch}{1.3}
    \vspace{-0em}
    \fontsize{9}{12}\selectfont
    \caption{Cohen's Kappa (\%) on EEGMat. The best metrics are marked in bold, and the second best by an underline.}
    \label{tab:eegmat_kappa1}
    \scalebox{0.53}{
    %
    }
    \vspace{-0em}
\end{table*}
\clearpage

\begin{table*}[!t]
    \centering
    \renewcommand{\arraystretch}{1.3}
    \vspace{-0em}
    \fontsize{9}{12}\selectfont
    \caption{Cohen's Kappa (\%) on EEGMat. The best metrics are marked in bold, and the second best by an underline.}
    \label{tab:eegmat_kappa2}
    \scalebox{0.54}{
    %
    }
    \vspace{-0em}
\end{table*}
\clearpage


\begin{table*}[!t]
    \centering
    \renewcommand{\arraystretch}{1.2}
    \vspace{-0em}
    \fontsize{9}{12}\selectfont
    \caption{Classification Top-5 accuracies (\%) on Things-EEG2. The best accuracies are marked in bold, and the second best by an underline.}
    \label{tab:things2_top5}
    \scalebox{0.6}{
    %
    }
    \vspace{-0em}
\end{table*}
\clearpage

\begin{table*}[!t]
    \centering
    \renewcommand{\arraystretch}{1.25}
    \vspace{-0em}
    \fontsize{9}{12}\selectfont
    \caption{Classification 2-Way accuracies (\%) on Things-EEG2. The best accuracies are marked in bold, and the second best by an underline.}
    \label{tab:things2_2way}
    \scalebox{0.6}{
    %
    }
    \vspace{-0em}
\end{table*}
\clearpage



\begin{table*}[!t]
    \centering
    \renewcommand{\arraystretch}{1.12}
    \vspace{-0em}
    \fontsize{9}{12}\selectfont
    \caption{Root Mean Square Error (RMSE) in SEED-VIG. The best RMSEs are marked in bold, and the second best by an underline.}
    \label{tab:VIG_rmse}
    \scalebox{0.62}{
    %
    }
    \vspace{-0em}
\end{table*}
\clearpage

\begin{table*}[!t]
    \centering
    \renewcommand{\arraystretch}{1.12}
    \vspace{-0em}
    \fontsize{9}{12}\selectfont
    \caption{Root Mean Square Error (RMSE) in SEED-VIG. The best RMSEs are marked in bold, and the second best by an underline (continued).}
    \label{tab:VIG_rmse2}
    \scalebox{0.6}{
    %
    }
    \vspace{-0em}
\end{table*}
\clearpage


\begin{table*}[!t]
    \centering
    \renewcommand{\arraystretch}{1.12}
    \vspace{-0em}
    \fontsize{9}{12}\selectfont
    \caption{Correlation Coefficient (CC) in SEED-VIG. The best CCs are marked in bold, and the second best by an underline.}
    \label{tab:VIG_cc}
    \scalebox{0.62}{
    %
    }
    \vspace{-0em}
\end{table*}
\clearpage

\begin{table*}[!t]
    \centering
    \renewcommand{\arraystretch}{1.12}
    \vspace{-0em}
    \fontsize{9}{12}\selectfont
    \caption{Correlation Coefficient (CC) in SEED-VIG. The best CCs are marked in bold, and the second best by an underline (continued).}
    \label{tab:VIG_cc2}
    \scalebox{0.6}{
    %
    }
    \vspace{-0em}
\end{table*}
\clearpage

Table~\ref{tab:overall_rank} reports the overall ranking, average rank, and parameter size of the evaluated EEG decoding models across the 13 datasets under both the LOSO and within-subject few-shot scenarios. Table~\ref{tab:fm_rank} further groups the 12 evaluated EEG FMs by their pre-training objectives, and reports the corresponding average ranks and model sizes. We additionally summarized a leakage-aware ranking in Table~\ref{tab:overall_rank_woleak}.

\begin{table}[t] \centering
\renewcommand{\arraystretch}{1.4}
\fontsize{9}{11}\selectfont
\caption{Overall ranking of EEG FMs and specialist models across 13 datasets under LOSO and few-shot scenarios.}
\label{tab:overall_rank}
\scalebox{0.75}{
\begin{tabular}{c c c c}
\toprule
\textbf{Total Rank} & \textbf{Model} & \textbf{Avg. Rank} & \textbf{Model Size} \\
\midrule
1  & CBraMod (FM)          & 6.56  & 4.0M \\
2  & EEGNet                & 7.40  & 2K \\
3  & Conformer             & 7.40  & 0.16M \\
4  & ShallowConv           & 7.60  & 36K \\
5  & Deformer              & 7.72  & 0.8M \\
6  & LMDA                  & 7.76  & 3K \\
7  & Neuro-GPT (FM)        & 7.84  & 0.16M \\
8  & TSception             & 9.24  & 9K \\
9  & LaBraM (FM)           & 9.36  & 5.8M \\
10 & EEGMamba (FM)         & 9.36  & 3.3M \\
11 & Traditional ML        & 9.91  & -- \\
12 & BENDR (FM)            & 10.48 & 4.0M \\
13 & BIOT-6D (FM)          & 11.44 & 3.2M \\
14 & CNN-T                 & 11.96 & 2.8M \\
15 & EEGPT (FM)            & 12.08 & 25M \\
16 & BrainOmni-Tiny (FM)   & 12.33 & 8.4M \\
17 & BrainOmni-Base (FM)   & 12.76 & 33M \\
18 & LUNA-Base (FM)        & 14.44 & 7.0M \\
19 & SingLEM (FM)          & 15.12 & 3.3M \\
20 & TFM (FM)              & 16.20 & 1.9M \\
\bottomrule
\end{tabular}
}
\end{table}

\begin{table}[t]
\centering
\renewcommand{\arraystretch}{1.6}
\fontsize{9}{11}\selectfont
\caption{Overall ranking of evaluated EEG foundation models across 13 datasets under LOSO and few-shot scenarios, grouped by pre-training objective.}
\label{tab:fm_rank}
\scalebox{0.55}{
\begin{tabular}{c c c c c}
\toprule
\textbf{Overall Rank} & \textbf{Model} & \textbf{Pre-training Objective} & \textbf{Avg. Rank} & \textbf{Model Size} \\
\midrule
1  & CBraMod          & Raw Signals              & 6.56  & 4.0M  \\
7  & Neuro-GPT        & Embedded Tokens          & 7.84  & 0.16M \\
9  & LaBraM           & EEG Codebook Index       & 9.36  & 5.8M  \\
10 & EEGMamba         & Raw Signals              & 9.36  & 3.3M  \\
12 & BENDR            & Embedded Tokens          & 10.48 & 4.0M  \\
13 & BIOT-6D          & Embedded Tokens          & 11.44 & 3.2M  \\
15 & EEGPT            & Raw Signals              & 12.08 & 25M   \\
16 & BrainOmni-Tiny   & Codebook Index           & 12.33 & 8.4M  \\
17 & BrainOmni-Base   & Codebook Index           & 12.76 & 33M   \\
18 & LUNA-Base        & Embedded Tokens          & 14.44 & 7.0M  \\
19 & SingLEM          & Raw Signals + Frequency  & 15.12 & 3.3M  \\
20 & TFM              & Codebook Index          & 16.20 & 1.9M  \\
\bottomrule
\end{tabular}
}
\end{table}

\begin{table}[t]
\centering
\renewcommand{\arraystretch}{1.4}
\fontsize{9}{11}\selectfont
\caption{Leakage-aware overall ranking of EEG FMs and specialist models across 13 datasets under LOSO and few-shot scenarios. For each model, rankings on downstream datasets that overlap with its pretraining data are excluded before computing the average rank.}
\label{tab:overall_rank_woleak}
\scalebox{0.75}{
\begin{tabular}{c c c c}
\toprule
\textbf{Total Rank} & \textbf{Model} & \textbf{Avg. Rank} & \textbf{Model Size} \\
\midrule
1  & CBraMod (FM)          & 6.63  & 4.0M \\
2  & EEGNet                & 7.08  & 2K \\
3  & Conformer             & 7.08  & 0.16M \\
4  & LMDA                  & 7.44  & 3K \\
5  & ShallowConv           & 7.48  & 36K \\
6  & Deformer              & 7.64  & 0.8M \\
7  & Neuro-GPT (FM)        & 7.75  & 0.16M \\
8  & TSception             & 8.96  & 9K \\
9  & LaBraM (FM)           & 9.17  & 5.8M \\
10 & Traditional ML        & 9.43  & -- \\
11 & EEGMamba (FM)         & 9.58  & 3.3M \\
12 & BENDR (FM)            & 10.38 & 4.0M \\
13 & BIOT-6D (FM)          & 11.41 & 3.2M \\
14 & EEGPT (FM)            & 11.57 & 25M \\
15 & CNN-T                 & 11.60 & 2.8M \\
16 & BrainOmni-Tiny (FM)   & 11.90 & 8.4M \\
17 & BrainOmni-Base (FM)   & 12.52 & 33M \\
18 & SingLEM (FM)          & 14.60 & 3.3M \\
19 & LUNA-Base (FM)        & 14.83 & 7.0M \\
20 & TFM (FM)              & 16.00 & 1.9M \\
\bottomrule
\end{tabular}
}
\end{table}

\subsubsection{Computational Efficiency}

Tables~\ref{tab:compute_cost_flatten} and~\ref{tab:compute_cost_pooling} compare the fine-tuning computational cost of all evaluated models under the \textit{flatten} and \textit{pooling} feature aggregation strategies. Each cell reports \textit{Full~/~Linear}, corresponding to full-parameter fine-tuning and head-only fine-tuning, respectively; specialist models have no linear-probing counterpart and are marked as ``--''. Here, \# Trainable Params denotes the parameters actually updated during optimization, Feature Dim is the dimensionality of the aggregated feature fed into the classification head, and FLOPs~(G) measures the floating-point operations of a single forward pass per sample. All measurements were obtained with PyTorch~2.6.0 (CUDA~12.4) on a single NVIDIA~A100-SXM4-80GB GPU with a fixed batch size of $32$.

\begin{table*}[t]
\centering
\renewcommand{\arraystretch}{1.45}
\setlength{\tabcolsep}{1.2pt}
\caption{Fine-tuning computational cost under the flatten feature aggregation strategy. Values are reported as \textit{Full / Linear}, where \textit{Full} denotes full-parameter fine-tuning and \textit{Linear} denotes head-only fine-tuning. Specialist models do not have a linear-probing counterpart, and unavailable entries are denoted by ``--''.}
\label{tab:compute_cost_flatten}
\resizebox{\textwidth}{!}{
\begin{tabular}{c|c|c|c|c c c c c c c c c c}
\toprule
\textbf{Channel} & \textbf{Duration} & \textbf{Category} & \textbf{Model} & \makecell{\textbf{Time}\\\textbf{Pts}} & \makecell{\# \textbf{Backbone}\\\textbf{Params}} & \makecell{\# \textbf{MLP}\\\textbf{Params}} & \makecell{\# \textbf{Trainable}\\\textbf{Params}} & \makecell{\textbf{Feature}\\\textbf{Dim}} & \makecell{\textbf{FLOPs}\\\textbf{(G)}} & \makecell{\textbf{Peak Mem.}\\\textbf{(MB)}} & \makecell{\textbf{FT Time}\\\textbf{(ms / batch)}} & \makecell{\textbf{Latency}\\\textbf{(ms / sample)}} & \makecell{\textbf{Throughput}\\\textbf{(sample / s)}} \\
\midrule
\multirow{36}{*}{10} & \multirow{18}{*}{1s} & \multirow{7}{*}{\makecell{Specialist\\Models}} & EEGNet & 250 / -- & 1.3K / -- & 452 / -- & 1.7K / -- & -- / -- & 0.09 / -- & 25.0 / -- & 2.2 / -- & 0.016 / -- & 61,566 / -- \\
 &  &  & ShallowConv & 250 / -- & 17.2K / -- & 1.8K / -- & 18.9K / -- & -- / -- & 0.38 / -- & 59.1 / -- & 1.7 / -- & 0.011 / -- & 94,679 / -- \\
 &  &  & LMDA & 250 / -- & 2.5K / -- & 256 / -- & 2.8K / -- & -- / -- & 0.26 / -- & 67.1 / -- & 3.0 / -- & 0.027 / -- & 36,585 / -- \\
 &  &  & CNN-T & 250 / -- & 2.75M / -- & 1.0K / -- & 2.76M / -- & -- / -- & 1.50 / -- & 88.2 / -- & 25.8 / -- & 0.297 / -- & 3,370 / -- \\
 &  &  & Deformer & 250 / -- & 352.9K / -- & 2.8K / -- & 355.7K / -- & -- / -- & 1.81 / -- & 153.1 / -- & 13.2 / -- & 0.120 / -- & 8,299 / -- \\
 &  &  & Conformer & 250 / -- & 137.1K / -- & 244 / -- & 137.4K / -- & -- / -- & 0.46 / -- & 60.9 / -- & 17.7 / -- & 0.140 / -- & 7,119 / -- \\
 &  &  & TSception & 128 / -- & 5.5K / -- & 644 / -- & 6.1K / -- & -- / -- & 0.09 / -- & 29.2 / -- & 3.7 / -- & 0.032 / -- & 30,801 / -- \\
\cmidrule(lr){3-14}
 &  & \multirow{11}{*}{\makecell{Foundation\\Models}} & BENDR & 256 / 256 & 3.97M / 3.97M & 6.1K / 6.1K & 3.98M / 6.4K & 1,536 / 1,536 & 0.0004 / 0.0004 & 113.3 / 68.4 & 4.5 / 3.1 & 0.037 / 0.036 & 27,335 / 27,631 \\
 &  &  & BIOT-6D & 200 / 200 & 3.19M / 3.19M & 18.4K / 18.4K & 3.21M / 18.6K & 4,608 / 4,608 & 3.73 / 3.73 & 102.8 / 64.0 & 29.5 / 24.9 & 0.272 / 0.266 & 3,676 / 3,756 \\
 &  &  & LaBraM & 200 / 200 & 5.82M / 5.82M & 8.8K / 8.8K & 5.83M / 8.8K & 2,200 / 2,200 & 4.10 / 4.10 & 159.4 / 42.7 & 33.7 / 11.4 & 0.280 / 0.281 & 3,565 / 3,557 \\
 &  &  & Neuro-GPT & 250 / 250 & 156.6K / 156.6K & 1.8K / 1.8K & 158.3K / 2.0K & 440 / 440 & 0.91 / 0.91 & 106.1 / 93.6 & 18.6 / 14.0 & 0.153 / 0.149 & 6,545 / 6,701 \\
 &  &  & EEGPT & 256 / 256 & 25.29M / 25.29M & 32.8K / 32.8K & 25.32M / 32.9K & 8,192 / 8,192 & 90.30 / 90.30 & 793.4 / 428.9 & 24.2 / 19.7 & 0.240 / 0.240 & 4,174 / 4,158 \\
 &  &  & CBraMod & 200 / 200 & 4.88M / 4.88M & 8.0K / 8.0K & 4.89M / 8.0K & 2,000 / 2,000 & 3.15 / 3.15 & 135.7 / 38.5 & 53.2 / 16.6 & 0.287 / 0.288 & 3,479 / 3,474 \\
 &  &  & TFM & 200 / 200 & 1.89M / 1.89M & 4.4K / 4.4K & 1.90M / 1.17M & 1,088 / 1,088 & 2.92 / 2.92 & 97.7 / 92.2 & 23.1 / 11.2 & 0.296 / 0.295 & 3,383 / 3,388 \\
 &  &  & BrainOmni-Tiny & 256 / 256 & 13.48M / 13.48M & 131.1K / 131.1K & 13.61M / 131.1K & 32,768 / 32,768 & 70.38 / 70.38 & 1126.8 / 201.4 & 50.7 / 19.4 & 0.571 / 0.572 & 1,752 / 1,748 \\
 &  &  & BrainOmni-Base & 256 / 256 & 37.76M / 37.76M & 262.1K / 262.1K & 38.02M / 262.1K & 65,536 / 65,536 & 264.80 / 264.80 & 2394.1 / 297.7 & 79.0 / 30.7 & 0.882 / 0.883 & 1,133 / 1,133 \\
 &  &  & SingLEM & 128 / 128 & 3.27M / 3.27M & 644 / 644 & 3.27M / 644 & 160 / 160 & 4.60 / 4.60 & 226.0 / 54.3 & 27.1 / 7.6 & 0.208 / 0.210 & 4,810 / 4,767 \\
 &  &  & LUNA-Base & 256 / 256 & 7.36M / 7.36M & 1.0K / 1.0K & 7.36M / 1.0K & 256 / 256 & 3.35 / 3.35 & 178.2 / 58.6 & 42.2 / 16.0 & 0.462 / 0.451 & 2,163 / 2,219 \\
\cmidrule(lr){2-14}
 & \multirow{18}{*}{4s} & \multirow{7}{*}{\makecell{Specialist\\Models}} & EEGNet & 1,000 / -- & 1.3K / -- & 2.0K / -- & 3.3K / -- & -- / -- & 0.35 / -- & 52.5 / -- & 2.8 / -- & 0.017 / -- & 59,844 / -- \\
 &  &  & ShallowConv & 1,000 / -- & 17.2K / -- & 9.8K / -- & 26.9K / -- & -- / -- & 1.62 / -- & 176.7 / -- & 2.6 / -- & 0.012 / -- & 86,717 / -- \\
 &  &  & LMDA & 1,000 / -- & 2.5K / -- & 1.3K / -- & 3.9K / -- & -- / -- & 1.35 / -- & 267.8 / -- & 7.8 / -- & 0.047 / -- & 21,250 / -- \\
 &  &  & CNN-T & 1,000 / -- & 2.75M / -- & 1.0K / -- & 2.76M / -- & -- / -- & 4.19 / -- & 153.4 / -- & 25.9 / -- & 0.298 / -- & 3,358 / -- \\
 &  &  & Deformer & 1,000 / -- & 751.1K / -- & 9.0K / -- & 760.0K / -- & -- / -- & 5.64 / -- & 292.8 / -- & 14.9 / -- & 0.122 / -- & 8,175 / -- \\
 &  &  & Conformer & 1,000 / -- & 137.1K / -- & 244 / -- & 137.4K / -- & -- / -- & 2.19 / -- & 183.7 / -- & 17.9 / -- & 0.150 / -- & 6,671 / -- \\
 &  &  & TSception & 512 / -- & 5.5K / -- & 644 / -- & 6.1K / -- & -- / -- & 0.50 / -- & 80.9 / -- & 3.9 / -- & 0.037 / -- & 27,203 / -- \\
\cmidrule(lr){3-14}
 &  & \multirow{11}{*}{\makecell{Foundation\\Models}} & BENDR & 1,024 / 1,024 & 3.97M / 3.97M & 22.5K / 22.5K & 3.99M / 22.7K & 5,632 / 5,632 & 0.0014 / 0.0014 & 224.0 / 169.0 & 4.4 / 3.0 & 0.037 / 0.036 & 27,016 / 27,644 \\
 &  &  & BIOT-6D & 800 / 800 & 3.19M / 3.19M & 129.0K / 129.0K & 3.32M / 129.2K & 32,256 / 32,256 & 26.13 / 26.13 & 373.2 / 244.4 & 31.1 / 25.9 & 0.275 / 0.275 & 3,632 / 3,630 \\
 &  &  & LaBraM & 800 / 800 & 5.82M / 5.82M & 32.8K / 32.8K & 5.86M / 32.8K & 8,200 / 8,200 & 15.67 / 15.67 & 393.6 / 55.6 & 34.5 / 11.1 & 0.275 / 0.278 & 3,639 / 3,591 \\
 &  &  & Neuro-GPT & 1,000 / 1,000 & 156.6K / 156.6K & 9.8K / 9.8K & 166.3K / 10.0K & 2,440 / 2,440 & 4.14 / 4.14 & 358.6 / 244.9 & 23.9 / 21.2 & 0.150 / 0.146 & 6,671 / 6,831 \\
 &  &  & EEGPT & 1,024 / 1,024 & 25.29M / 25.29M & 131.1K / 131.1K & 25.42M / 131.2K & 32,768 / 32,768 & 361.18 / 361.18 & 2212.0 / 1371.1 & 79.4 / 54.2 & 0.769 / 0.771 & 1,300 / 1,296 \\
 &  &  & CBraMod & 800 / 800 & 4.88M / 4.88M & 32.0K / 32.0K & 4.92M / 32.0K & 8,000 / 8,000 & 12.60 / 12.60 & 337.8 / 49.6 & 55.4 / 17.0 & 0.295 / 0.289 & 3,395 / 3,460 \\
 &  &  & TFM & 800 / 800 & 1.89M / 1.89M & 28.9K / 28.9K & 1.92M / 1.20M & 7,232 / 7,232 & 20.38 / 20.38 & 492.9 / 488.6 & 24.0 / 11.5 & 0.305 / 0.302 & 3,274 / 3,308 \\
 &  &  & BrainOmni-Tiny & 1,024 / 1,024 & 13.48M / 13.48M & 393.2K / 393.2K & 13.87M / 393.2K & 98,304 / 98,304 & 210.44 / 210.44 & 3094.5 / 452.1 & 74.9 / 32.7 & 0.923 / 0.917 & 1,084 / 1,091 \\
 &  &  & BrainOmni-Base & 1,024 / 1,024 & 37.76M / 37.76M & 786.4K / 786.4K & 38.54M / 786.4K & 196,608 / 196,608 & 793.67 / 793.67 & 6335.2 / 548.8 & 204.0 / 70.0 & 2.136 / 2.137 & 468 / 468 \\
 &  &  & SingLEM & 512 / 512 & 3.27M / 3.27M & 2.6K / 2.6K & 3.27M / 2.6K & 640 / 640 & 18.38 / 18.38 & 744.7 / 130.3 & 34.0 / 11.2 & 0.339 / 0.339 & 2,949 / 2,950 \\
 &  &  & LUNA-Base & 1,024 / 1,024 & 7.36M / 7.36M & 1.0K / 1.0K & 7.36M / 1.0K & 256 / 256 & 12.43 / 12.43 & 360.8 / 77.1 & 42.6 / 16.7 & 0.459 / 0.458 & 2,178 / 2,182 \\
\midrule
\multirow{36}{*}{32} & \multirow{18}{*}{1s} & \multirow{7}{*}{\makecell{Specialist\\Models}} & EEGNet & 250 / -- & 1.6K / -- & 452 / -- & 2.1K / -- & -- / -- & 0.27 / -- & 42.8 / -- & 2.6 / -- & 0.016 / -- & 61,418 / -- \\
 &  &  & ShallowConv & 250 / -- & 52.4K / -- & 1.8K / -- & 54.1K / -- & -- / -- & 1.20 / -- & 149.4 / -- & 2.5 / -- & 0.011 / -- & 94,754 / -- \\
 &  &  & LMDA & 250 / -- & 2.9K / -- & 256 / -- & 3.2K / -- & -- / -- & 0.84 / -- & 175.1 / -- & 5.9 / -- & 0.033 / -- & 30,011 / -- \\
 &  &  & CNN-T & 250 / -- & 2.77M / -- & 1.0K / -- & 2.77M / -- & -- / -- & 2.07 / -- & 99.3 / -- & 32.5 / -- & 0.501 / -- & 1,994 / -- \\
 &  &  & Deformer & 250 / -- & 443.0K / -- & 2.8K / -- & 445.8K / -- & -- / -- & 3.50 / -- & 251.5 / -- & 14.0 / -- & 0.125 / -- & 8,008 / -- \\
 &  &  & Conformer & 250 / -- & 172.3K / -- & 244 / -- & 172.6K / -- & -- / -- & 1.29 / -- & 151.2 / -- & 17.7 / -- & 0.150 / -- & 6,645 / -- \\
 &  &  & TSception & 128 / -- & 12.0K / -- & 644 / -- & 12.6K / -- & -- / -- & 0.27 / -- & 55.7 / -- & 3.8 / -- & 0.036 / -- & 27,850 / -- \\
\cmidrule(lr){3-14}
 &  & \multirow{11}{*}{\makecell{Foundation\\Models}} & BENDR & 256 / 256 & 3.97M / 3.97M & 6.1K / 6.1K & 3.98M / 6.8K & 1,536 / 1,536 & 0.0004 / 0.0004 & 114.9 / 69.1 & 4.5 / 3.0 & 0.037 / 0.036 & 27,123 / 27,639 \\
 &  &  & BIOT-6D & 200 / 200 & 3.19M / 3.19M & 18.4K / 18.4K & 3.21M / 19.0K & 4,608 / 4,608 & 3.73 / 3.73 & 103.3 / 64.5 & 29.4 / 24.8 & 0.268 / 0.268 & 3,731 / 3,734 \\
 &  &  & LaBraM & 200 / 200 & 5.82M / 5.82M & 26.4K / 26.4K & 5.85M / 26.4K & 6,600 / 6,600 & 12.53 / 12.53 & 324.9 / 50.8 & 34.5 / 11.1 & 0.277 / 0.279 & 3,610 / 3,584 \\
 &  &  & Neuro-GPT & 250 / 250 & 157.0K / 157.0K & 1.8K / 1.8K & 158.8K / 2.5K & 440 / 440 & 0.91 / 0.91 & 106.8 / 94.3 & 18.1 / 14.0 & 0.153 / 0.146 & 6,555 / 6,848 \\
 &  &  & EEGPT & 256 / 256 & 25.29M / 25.29M & 32.8K / 32.8K & 25.32M / 33.8K & 8,192 / 8,192 & 232.24 / 232.24 & 1544.1 / 922.2 & 51.7 / 34.6 & 0.498 / 0.498 & 2,009 / 2,010 \\
 &  &  & CBraMod & 200 / 200 & 4.88M / 4.88M & 25.6K / 25.6K & 4.91M / 25.6K & 6,400 / 6,400 & 10.18 / 10.18 & 292.6 / 46.2 & 52.7 / 16.4 & 0.288 / 0.287 & 3,467 / 3,485 \\
 &  &  & TFM & 200 / 200 & 1.89M / 1.89M & 4.4K / 4.4K & 1.90M / 1.17M & 1,088 / 1,088 & 2.92 / 2.92 & 98.3 / 92.7 & 23.3 / 11.2 & 0.298 / 0.294 & 3,354 / 3,396 \\
 &  &  & BrainOmni-Tiny & 256 / 256 & 13.48M / 13.48M & 131.1K / 131.1K & 13.61M / 131.1K & 32,768 / 32,768 & 72.68 / 72.68 & 1127.7 / 469.4 & 54.1 / 23.2 & 0.648 / 0.670 & 1,542 / 1,492 \\
 &  &  & BrainOmni-Base & 256 / 256 & 37.76M / 37.76M & 262.1K / 262.1K & 38.02M / 262.1K & 65,536 / 65,536 & 267.09 / 267.09 & 2394.8 / 564.9 & 81.2 / 37.0 & 0.989 / 0.990 & 1,011 / 1,010 \\
 &  &  & SingLEM & 128 / 128 & 3.27M / 3.27M & 2.1K / 2.1K & 3.27M / 2.1K & 512 / 512 & 14.71 / 14.71 & 600.7 / 110.0 & 28.6 / 9.2 & 0.277 / 0.276 & 3,616 / 3,617 \\
 &  &  & LUNA-Base & 256 / 256 & 7.36M / 7.36M & 1.0K / 1.0K & 7.36M / 1.0K & 256 / 256 & 3.71 / 3.71 & 206.4 / 69.7 & 42.9 / 16.6 & 0.458 / 0.449 & 2,182 / 2,229 \\
\cmidrule(lr){2-14}
 & \multirow{18}{*}{4s} & \multirow{7}{*}{\makecell{Specialist\\Models}} & EEGNet & 1,000 / -- & 1.6K / -- & 2.0K / -- & 3.6K / -- & -- / -- & 1.09 / -- & 122.9 / -- & 6.1 / -- & 0.024 / -- & 42,424 / -- \\
 &  &  & ShallowConv & 1,000 / -- & 52.4K / -- & 9.8K / -- & 62.1K / -- & -- / -- & 5.20 / -- & 512.2 / -- & 4.5 / -- & 0.031 / -- & 32,458 / -- \\
 &  &  & LMDA & 1,000 / -- & 2.9K / -- & 1.3K / -- & 4.3K / -- & -- / -- & 4.30 / -- & 776.6 / -- & 25.4 / -- & 0.118 / -- & 8,503 / -- \\
 &  &  & CNN-T & 1,000 / -- & 2.77M / -- & 1.0K / -- & 2.77M / -- & -- / -- & 6.34 / -- & 225.0 / -- & 33.0 / -- & 0.510 / -- & 1,962 / -- \\
 &  &  & Deformer & 1,000 / -- & 841.2K / -- & 9.0K / -- & 850.1K / -- & -- / -- & 12.40 / -- & 835.6 / -- & 16.6 / -- & 0.128 / -- & 7,815 / -- \\
 &  &  & Conformer & 1,000 / -- & 172.3K / -- & 244 / -- & 172.6K / -- & -- / -- & 5.77 / -- & 514.0 / -- & 19.5 / -- & 0.146 / -- & 6,859 / -- \\
 &  &  & TSception & 512 / -- & 12.0K / -- & 644 / -- & 12.6K / -- & -- / -- & 1.55 / -- & 223.9 / -- & 5.0 / -- & 0.034 / -- & 29,742 / -- \\
\cmidrule(lr){3-14}
 &  & \multirow{11}{*}{\makecell{Foundation\\Models}} & BENDR & 1,024 / 1,024 & 3.97M / 3.97M & 22.5K / 22.5K & 4.00M / 23.2K & 5,632 / 5,632 & 0.0014 / 0.0014 & 226.5 / 172.4 & 4.4 / 3.1 & 0.037 / 0.037 & 27,298 / 27,364 \\
 &  &  & BIOT-6D & 800 / 800 & 3.19M / 3.19M & 129.0K / 129.0K & 3.32M / 129.6K & 32,256 / 32,256 & 26.13 / 26.13 & 375.3 / 245.4 & 31.0 / 26.0 & 0.278 / 0.279 & 3,603 / 3,583 \\
 &  &  & LaBraM & 800 / 800 & 5.82M / 5.82M & 103.2K / 103.2K & 5.93M / 103.2K & 25,800 / 25,800 & 52.81 / 52.81 & 1183.1 / 111.8 & 42.9 / 15.0 & 0.364 / 0.364 & 2,750 / 2,750 \\
 &  &  & Neuro-GPT & 1,000 / 1,000 & 157.0K / 157.0K & 9.8K / 9.8K & 166.8K / 10.5K & 2,440 / 2,440 & 4.14 / 4.14 & 361.4 / 247.8 & 24.9 / 19.7 & 0.148 / 0.153 & 6,761 / 6,520 \\
 &  &  & EEGPT & 1,024 / 1,024 & 25.29M / 25.29M & 131.1K / 131.1K & 25.42M / 132.1K & 32,768 / 32,768 & 928.95 / 928.95 & 5195.3 / 3339.4 & 188.1 / 126.5 & 1.765 / 1.765 & 566 / 567 \\
 &  &  & CBraMod & 800 / 800 & 4.88M / 4.88M & 102.4K / 102.4K & 4.99M / 102.4K & 25,600 / 25,600 & 40.77 / 40.77 & 932.8 / 79.9 & 56.0 / 17.6 & 0.295 / 0.293 & 3,385 / 3,414 \\
 &  &  & TFM & 800 / 800 & 1.89M / 1.89M & 28.9K / 28.9K & 1.92M / 1.20M & 7,232 / 7,232 & 20.38 / 20.38 & 495.9 / 490.4 & 24.0 / 11.5 & 0.302 / 0.302 & 3,307 / 3,306 \\
 &  &  & BrainOmni-Tiny & 1,024 / 1,024 & 13.48M / 13.48M & 393.2K / 393.2K & 13.87M / 393.2K & 98,304 / 98,304 & 215.75 / 215.75 & 3097.8 / 1255.8 & 83.9 / 41.2 & 1.225 / 1.225 & 816 / 816 \\
 &  &  & BrainOmni-Base & 1,024 / 1,024 & 37.76M / 37.76M & 786.4K / 786.4K & 38.54M / 786.4K & 196,608 / 196,608 & 798.99 / 798.99 & 6336.6 / 1353.6 & 213.7 / 79.8 & 2.445 / 2.447 & 409 / 409 \\
 &  &  & SingLEM & 512 / 512 & 3.27M / 3.27M & 8.2K / 8.2K & 3.28M / 8.2K & 2,048 / 2,048 & 58.83 / 58.83 & 2227.0 / 353.3 & 96.7 / 32.3 & 0.992 / 0.993 & 1,008 / 1,007 \\
 &  &  & LUNA-Base & 1,024 / 1,024 & 7.36M / 7.36M & 1.0K / 1.0K & 7.36M / 1.0K & 256 / 256 & 13.70 / 13.70 & 461.8 / 118.1 & 42.6 / 16.9 & 0.460 / 0.476 & 2,172 / 2,099 \\
\bottomrule
\end{tabular}
}
\end{table*}

\begin{table*}[t]
\centering
\renewcommand{\arraystretch}{1.35}
\setlength{\tabcolsep}{1.2pt}
\caption{Fine-tuning computational cost under the pooling feature aggregation strategy. Values are reported as \textit{Full / Linear}, where \textit{Full} denotes full-parameter fine-tuning and \textit{Linear} denotes head-only fine-tuning. Specialist models do not have a linear-probing counterpart, and unavailable entries are denoted by ``--''.}
\label{tab:compute_cost_pooling}
\resizebox{\textwidth}{!}{
\begin{tabular}{c|c|c|c|c c c c c c c c c c}
\toprule
\textbf{Channel} & \textbf{Duration} & \textbf{Category} & \textbf{Model} & \makecell{\textbf{Time}\\\textbf{Pts}} & \makecell{\# \textbf{Backbone}\\\textbf{Params}} & \makecell{\# \textbf{MLP}\\\textbf{Params}} & \makecell{\# \textbf{Trainable}\\\textbf{Params}} & \makecell{\textbf{Feature}\\\textbf{Dim}} & \makecell{\textbf{FLOPs}\\\textbf{(G)}} & \makecell{\textbf{Peak Mem.}\\\textbf{(MB)}} & \makecell{\textbf{FT Time}\\\textbf{(ms / batch)}} & \makecell{\textbf{Latency}\\\textbf{(ms / sample)}} & \makecell{\textbf{Throughput}\\\textbf{(sample / s)}} \\
\midrule
\multirow{36}{*}{10} & \multirow{18}{*}{1s} & \multirow{7}{*}{\makecell{Specialist\\Models}} & EEGNet & 250 / -- & 1.3K / -- & 452 / -- & 1.7K / -- & -- / -- & 0.09 / -- & 25.0 / -- & 2.2 / -- & 0.016 / -- & 61,566 / -- \\
 &  &  & ShallowConv & 250 / -- & 17.2K / -- & 1.8K / -- & 18.9K / -- & -- / -- & 0.38 / -- & 59.1 / -- & 1.7 / -- & 0.011 / -- & 94,679 / -- \\
 &  &  & LMDA & 250 / -- & 2.5K / -- & 256 / -- & 2.8K / -- & -- / -- & 0.26 / -- & 67.1 / -- & 3.0 / -- & 0.027 / -- & 36,585 / -- \\
 &  &  & CNN-T & 250 / -- & 2.75M / -- & 1.0K / -- & 2.76M / -- & -- / -- & 1.50 / -- & 88.2 / -- & 25.8 / -- & 0.297 / -- & 3,370 / -- \\
 &  &  & Deformer & 250 / -- & 352.9K / -- & 2.8K / -- & 355.7K / -- & -- / -- & 1.81 / -- & 153.1 / -- & 13.2 / -- & 0.120 / -- & 8,299 / -- \\
 &  &  & Conformer & 250 / -- & 137.1K / -- & 244 / -- & 137.4K / -- & -- / -- & 0.46 / -- & 60.9 / -- & 17.7 / -- & 0.140 / -- & 7,119 / -- \\
 &  &  & TSception & 128 / -- & 5.5K / -- & 644 / -- & 6.1K / -- & -- / -- & 0.09 / -- & 29.2 / -- & 3.7 / -- & 0.032 / -- & 30,801 / -- \\
\cmidrule(lr){3-14}
 &  & \multirow{11}{*}{\makecell{Foundation\\Models}} & BENDR & 256 / 256 & 3.97M / 3.97M & 2.1K / 2.1K & 3.97M / 2.3K & 512 / 512 & 0.0001 / 0.0001 & 114.2 / 68.4 & 4.6 / 3.1 & 0.037 / 0.037 & 26,909 / 27,120 \\
 &  &  & BIOT-6D & 200 / 200 & 3.19M / 3.19M & 1.0K / 1.0K & 3.19M / 1.2K & 256 / 256 & 3.73 / 3.73 & 102.5 / 63.7 & 29.5 / 25.4 & 0.265 / 0.273 & 3,773 / 3,659 \\
 &  &  & LaBraM & 200 / 200 & 5.82M / 5.82M & 804 / 804 & 5.83M / 804 & 200 / 200 & 4.10 / 4.10 & 159.2 / 42.6 & 34.6 / 11.5 & 0.287 / 0.287 & 3,490 / 3,488 \\
 &  &  & Neuro-GPT & 250 / 250 & 156.6K / 156.6K & 164 / 164 & 156.7K / 406 & 40 / 40 & 0.91 / 0.91 & 106.1 / 93.6 & 18.4 / 14.3 & 0.155 / 0.152 & 6,463 / 6,559 \\
 &  &  & EEGPT & 256 / 256 & 25.29M / 25.29M & 2.1K / 2.1K & 25.29M / 2.2K & 512 / 512 & 90.29 / 90.29 & 792.9 / 428.5 & 26.6 / 16.3 & 0.240 / 0.240 & 4,175 / 4,166 \\
 &  &  & CBraMod & 200 / 200 & 4.88M / 4.88M & 804 / 804 & 4.88M / 804 & 200 / 200 & 3.15 / 3.15 & 135.6 / 38.4 & 52.7 / 16.5 & 0.288 / 0.289 & 3,477 / 3,463 \\
 &  &  & TFM & 200 / 200 & 1.89M / 1.89M & 260 / 260 & 1.89M / 1.17M & 64 / 64 & 2.92 / 2.92 & 97.7 / 92.1 & 23.2 / 11.4 & 0.296 / 0.301 & 3,379 / 3,323 \\
 &  &  & BrainOmni-Tiny & 256 / 256 & 13.48M / 13.48M & 1.0K / 1.0K & 13.48M / 1.0K & 256 / 256 & 70.38 / 70.38 & 1124.8 / 199.9 & 51.1 / 19.5 & 0.575 / 0.573 & 1,739 / 1,747 \\
 &  &  & BrainOmni-Base & 256 / 256 & 37.76M / 37.76M & 2.1K / 2.1K & 37.76M / 2.1K & 512 / 512 & 264.78 / 264.78 & 2390.1 / 294.7 & 79.2 / 32.9 & 0.885 / 0.883 & 1,130 / 1,133 \\
 &  &  & SingLEM & 128 / 128 & 3.27M / 3.27M & 68 / 68 & 3.27M / 68 & 16 / 16 & 4.60 / 4.60 & 226.0 / 54.3 & 27.0 / 7.5 & 0.211 / 0.206 & 4,745 / 4,850 \\
 &  &  & LUNA-Base & 256 / 256 & 7.36M / 7.36M & 1.0K / 1.0K & 7.36M / 1.0K & 256 / 256 & 3.35 / 3.35 & 178.2 / 58.6 & 43.1 / 16.6 & 0.469 / 0.465 & 2,132 / 2,149 \\
\cmidrule(lr){2-14}
 & \multirow{18}{*}{4s} & \multirow{7}{*}{\makecell{Specialist\\Models}} & EEGNet & 1,000 / -- & 1.3K / -- & 2.0K / -- & 3.3K / -- & -- / -- & 0.35 / -- & 52.5 / -- & 2.8 / -- & 0.017 / -- & 59,844 / -- \\
 &  &  & ShallowConv & 1,000 / -- & 17.2K / -- & 9.8K / -- & 26.9K / -- & -- / -- & 1.62 / -- & 176.7 / -- & 2.6 / -- & 0.012 / -- & 86,717 / -- \\
 &  &  & LMDA & 1,000 / -- & 2.5K / -- & 1.3K / -- & 3.9K / -- & -- / -- & 1.35 / -- & 267.8 / -- & 7.8 / -- & 0.047 / -- & 21,250 / -- \\
 &  &  & CNN-T & 1,000 / -- & 2.75M / -- & 1.0K / -- & 2.76M / -- & -- / -- & 4.19 / -- & 153.4 / -- & 25.9 / -- & 0.298 / -- & 3,358 / -- \\
 &  &  & Deformer & 1,000 / -- & 751.1K / -- & 9.0K / -- & 760.0K / -- & -- / -- & 5.64 / -- & 292.8 / -- & 14.9 / -- & 0.122 / -- & 8,175 / -- \\
 &  &  & Conformer & 1,000 / -- & 137.1K / -- & 244 / -- & 137.4K / -- & -- / -- & 2.19 / -- & 183.7 / -- & 17.9 / -- & 0.150 / -- & 6,671 / -- \\
 &  &  & TSception & 512 / -- & 5.5K / -- & 644 / -- & 6.1K / -- & -- / -- & 0.50 / -- & 80.9 / -- & 3.9 / -- & 0.037 / -- & 27,203 / -- \\
\cmidrule(lr){3-14}
 &  & \multirow{11}{*}{\makecell{Foundation\\Models}} & BENDR & 1,024 / 1,024 & 3.97M / 3.97M & 2.1K / 2.1K & 3.97M / 2.3K & 512 / 512 & 0.0001 / 0.0001 & 223.7 / 168.7 & 4.5 / 3.1 & 0.037 / 0.037 & 27,030 / 26,894 \\
 &  &  & BIOT-6D & 800 / 800 & 3.19M / 3.19M & 1.0K / 1.0K & 3.19M / 1.2K & 256 / 256 & 26.12 / 26.12 & 371.3 / 242.1 & 30.8 / 26.0 & 0.274 / 0.277 & 3,654 / 3,613 \\
 &  &  & LaBraM & 800 / 800 & 5.82M / 5.82M & 804 / 804 & 5.83M / 804 & 200 / 200 & 15.67 / 15.67 & 393.1 / 55.2 & 35.4 / 11.3 & 0.284 / 0.284 & 3,522 / 3,517 \\
 &  &  & Neuro-GPT & 1,000 / 1,000 & 156.6K / 156.6K & 164 / 164 & 156.7K / 406 & 40 / 40 & 4.14 / 4.14 & 358.4 / 244.7 & 23.8 / 19.1 & 0.155 / 0.145 & 6,445 / 6,889 \\
 &  &  & EEGPT & 1,024 / 1,024 & 25.29M / 25.29M & 2.1K / 2.1K & 25.29M / 2.2K & 512 / 512 & 361.17 / 361.17 & 2210.1 / 1369.1 & 79.3 / 54.1 & 0.769 / 0.770 & 1,300 / 1,299 \\
 &  &  & CBraMod & 800 / 800 & 4.88M / 4.88M & 804 / 804 & 4.88M / 804 & 200 / 200 & 12.60 / 12.60 & 337.3 / 49.2 & 56.3 / 17.1 & 0.299 / 0.290 & 3,347 / 3,448 \\
 &  &  & TFM & 800 / 800 & 1.89M / 1.89M & 260 / 260 & 1.89M / 1.17M & 64 / 64 & 20.38 / 20.38 & 492.6 / 488.3 & 24.1 / 11.8 & 0.304 / 0.310 & 3,294 / 3,222 \\
 &  &  & BrainOmni-Tiny & 1,024 / 1,024 & 13.48M / 13.48M & 1.0K / 1.0K & 13.48M / 1.0K & 256 / 256 & 210.41 / 210.41 & 3087.8 / 444.8 & 73.8 / 32.6 & 0.913 / 0.913 & 1,096 / 1,096 \\
 &  &  & BrainOmni-Base & 1,024 / 1,024 & 37.76M / 37.76M & 2.1K / 2.1K & 37.76M / 2.1K & 512 / 512 & 793.62 / 793.62 & 6323.0 / 539.9 & 203.7 / 69.6 & 2.135 / 2.138 & 468 / 468 \\
 &  &  & SingLEM & 512 / 512 & 3.27M / 3.27M & 68 / 68 & 3.27M / 68 & 16 / 16 & 18.38 / 18.38 & 744.6 / 130.2 & 34.0 / 11.1 & 0.338 / 0.338 & 2,956 / 2,960 \\
 &  &  & LUNA-Base & 1,024 / 1,024 & 7.36M / 7.36M & 1.0K / 1.0K & 7.36M / 1.0K & 256 / 256 & 12.43 / 12.43 & 360.8 / 77.1 & 43.2 / 16.8 & 0.468 / 0.466 & 2,136 / 2,145 \\
\midrule
\multirow{36}{*}{32} & \multirow{18}{*}{1s} & \multirow{7}{*}{\makecell{Specialist\\Models}} & EEGNet & 250 / -- & 1.6K / -- & 452 / -- & 2.1K / -- & -- / -- & 0.27 / -- & 42.8 / -- & 2.6 / -- & 0.016 / -- & 61,418 / -- \\
 &  &  & ShallowConv & 250 / -- & 52.4K / -- & 1.8K / -- & 54.1K / -- & -- / -- & 1.20 / -- & 149.4 / -- & 2.5 / -- & 0.011 / -- & 94,754 / -- \\
 &  &  & LMDA & 250 / -- & 2.9K / -- & 256 / -- & 3.2K / -- & -- / -- & 0.84 / -- & 175.1 / -- & 5.9 / -- & 0.033 / -- & 30,011 / -- \\
 &  &  & CNN-T & 250 / -- & 2.77M / -- & 1.0K / -- & 2.77M / -- & -- / -- & 2.07 / -- & 99.3 / -- & 32.5 / -- & 0.501 / -- & 1,994 / -- \\
 &  &  & Deformer & 250 / -- & 443.0K / -- & 2.8K / -- & 445.8K / -- & -- / -- & 3.50 / -- & 251.5 / -- & 14.0 / -- & 0.125 / -- & 8,008 / -- \\
 &  &  & Conformer & 250 / -- & 172.3K / -- & 244 / -- & 172.6K / -- & -- / -- & 1.29 / -- & 151.2 / -- & 17.7 / -- & 0.150 / -- & 6,645 / -- \\
 &  &  & TSception & 128 / -- & 12.0K / -- & 644 / -- & 12.6K / -- & -- / -- & 0.27 / -- & 55.7 / -- & 3.8 / -- & 0.036 / -- & 27,850 / -- \\
\cmidrule(lr){3-14}
 &  & \multirow{11}{*}{\makecell{Foundation\\Models}} & BENDR & 256 / 256 & 3.97M / 3.97M & 2.1K / 2.1K & 3.97M / 2.7K & 512 / 512 & 0.0001 / 0.0001 & 114.8 / 69.1 & 4.6 / 3.1 & 0.037 / 0.037 & 26,753 / 26,927 \\
 &  &  & BIOT-6D & 200 / 200 & 3.19M / 3.19M & 1.0K / 1.0K & 3.19M / 1.6K & 256 / 256 & 3.73 / 3.73 & 103.1 / 64.2 & 29.3 / 24.8 & 0.264 / 0.268 & 3,795 / 3,725 \\
 &  &  & LaBraM & 200 / 200 & 5.82M / 5.82M & 804 / 804 & 5.83M / 804 & 200 / 200 & 12.53 / 12.53 & 324.5 / 50.5 & 35.0 / 12.0 & 0.286 / 0.298 & 3,495 / 3,359 \\
 &  &  & Neuro-GPT & 250 / 250 & 157.0K / 157.0K & 164 / 164 & 157.2K / 890 & 40 / 40 & 0.91 / 0.91 & 106.8 / 94.2 & 18.4 / 14.4 & 0.151 / 0.149 & 6,608 / 6,698 \\
 &  &  & EEGPT & 256 / 256 & 25.29M / 25.29M & 2.1K / 2.1K & 25.29M / 3.1K & 512 / 512 & 232.24 / 232.24 & 1543.6 / 921.7 & 51.7 / 34.4 & 0.497 / 0.498 & 2,011 / 2,010 \\
 &  &  & CBraMod & 200 / 200 & 4.88M / 4.88M & 804 / 804 & 4.88M / 804 & 200 / 200 & 10.18 / 10.18 & 292.2 / 45.9 & 53.8 / 16.3 & 0.293 / 0.286 & 3,417 / 3,496 \\
 &  &  & TFM & 200 / 200 & 1.89M / 1.89M & 260 / 260 & 1.89M / 1.17M & 64 / 64 & 2.92 / 2.92 & 98.2 / 92.7 & 24.2 / 11.5 & 0.308 / 0.304 & 3,249 / 3,287 \\
 &  &  & BrainOmni-Tiny & 256 / 256 & 13.48M / 13.48M & 1.0K / 1.0K & 13.48M / 1.0K & 256 / 256 & 72.67 / 72.67 & 1125.8 / 467.9 & 54.1 / 22.3 & 0.656 / 0.646 & 1,523 / 1,548 \\
 &  &  & BrainOmni-Base & 256 / 256 & 37.76M / 37.76M & 2.1K / 2.1K & 37.76M / 2.1K & 512 / 512 & 267.07 / 267.07 & 2390.8 / 561.9 & 81.2 / 36.0 & 0.989 / 0.990 & 1,011 / 1,010 \\
 &  &  & SingLEM & 128 / 128 & 3.27M / 3.27M & 68 / 68 & 3.27M / 68 & 16 / 16 & 14.71 / 14.71 & 600.6 / 110.0 & 28.8 / 9.2 & 0.277 / 0.277 & 3,616 / 3,616 \\
 &  &  & LUNA-Base & 256 / 256 & 7.36M / 7.36M & 1.0K / 1.0K & 7.36M / 1.0K & 256 / 256 & 3.71 / 3.71 & 206.4 / 69.7 & 43.0 / 16.8 & 0.466 / 0.467 & 2,144 / 2,143 \\
\cmidrule(lr){2-14}
 & \multirow{18}{*}{4s} & \multirow{7}{*}{\makecell{Specialist\\Models}} & EEGNet & 1,000 / -- & 1.6K / -- & 2.0K / -- & 3.6K / -- & -- / -- & 1.09 / -- & 122.9 / -- & 6.1 / -- & 0.024 / -- & 42,424 / -- \\
 &  &  & ShallowConv & 1,000 / -- & 52.4K / -- & 9.8K / -- & 62.1K / -- & -- / -- & 5.20 / -- & 512.2 / -- & 4.5 / -- & 0.031 / -- & 32,458 / -- \\
 &  &  & LMDA & 1,000 / -- & 2.9K / -- & 1.3K / -- & 4.3K / -- & -- / -- & 4.30 / -- & 776.6 / -- & 25.4 / -- & 0.118 / -- & 8,503 / -- \\
 &  &  & CNN-T & 1,000 / -- & 2.77M / -- & 1.0K / -- & 2.77M / -- & -- / -- & 6.34 / -- & 225.0 / -- & 33.0 / -- & 0.510 / -- & 1,962 / -- \\
 &  &  & Deformer & 1,000 / -- & 841.2K / -- & 9.0K / -- & 850.1K / -- & -- / -- & 12.40 / -- & 835.6 / -- & 16.6 / -- & 0.128 / -- & 7,815 / -- \\
 &  &  & Conformer & 1,000 / -- & 172.3K / -- & 244 / -- & 172.6K / -- & -- / -- & 5.77 / -- & 514.0 / -- & 19.5 / -- & 0.146 / -- & 6,859 / -- \\
 &  &  & TSception & 512 / -- & 12.0K / -- & 644 / -- & 12.6K / -- & -- / -- & 1.55 / -- & 223.9 / -- & 5.0 / -- & 0.034 / -- & 29,742 / -- \\
\cmidrule(lr){3-14}
 &  & \multirow{11}{*}{\makecell{Foundation\\Models}} & BENDR & 1,024 / 1,024 & 3.97M / 3.97M & 2.1K / 2.1K & 3.97M / 2.7K & 512 / 512 & 0.0001 / 0.0001 & 226.2 / 172.1 & 4.1 / 3.1 & 0.037 / 0.037 & 26,823 / 26,775 \\
 &  &  & BIOT-6D & 800 / 800 & 3.19M / 3.19M & 1.0K / 1.0K & 3.19M / 1.6K & 256 / 256 & 26.12 / 26.12 & 373.4 / 243.4 & 31.2 / 25.8 & 0.279 / 0.274 & 3,590 / 3,647 \\
 &  &  & LaBraM & 800 / 800 & 5.82M / 5.82M & 804 / 804 & 5.83M / 804 & 200 / 200 & 52.80 / 52.80 & 1181.5 / 110.6 & 41.2 / 12.5 & 0.364 / 0.365 & 2,745 / 2,737 \\
 &  &  & Neuro-GPT & 1,000 / 1,000 & 157.0K / 157.0K & 164 / 164 & 157.2K / 890 & 40 / 40 & 4.14 / 4.14 & 361.3 / 247.6 & 24.9 / 20.4 & 0.152 / 0.145 & 6,561 / 6,900 \\
 &  &  & EEGPT & 1,024 / 1,024 & 25.29M / 25.29M & 2.1K / 2.1K & 25.29M / 3.1K & 512 / 512 & 928.95 / 928.95 & 5193.3 / 3337.4 & 188.2 / 126.3 & 1.764 / 1.764 & 567 / 567 \\
 &  &  & CBraMod & 800 / 800 & 4.88M / 4.88M & 804 / 804 & 4.88M / 804 & 200 / 200 & 40.76 / 40.76 & 931.2 / 78.7 & 56.4 / 17.0 & 0.295 / 0.291 & 3,385 / 3,439 \\
 &  &  & TFM & 800 / 800 & 1.89M / 1.89M & 260 / 260 & 1.89M / 1.17M & 64 / 64 & 20.38 / 20.38 & 495.6 / 490.0 & 24.9 / 12.4 & 0.317 / 0.322 & 3,156 / 3,106 \\
 &  &  & BrainOmni-Tiny & 1,024 / 1,024 & 13.48M / 13.48M & 1.0K / 1.0K & 13.48M / 1.0K & 256 / 256 & 215.72 / 215.72 & 3090.8 / 1250.9 & 83.9 / 41.9 & 1.225 / 1.224 & 816 / 817 \\
 &  &  & BrainOmni-Base & 1,024 / 1,024 & 37.76M / 37.76M & 2.1K / 2.1K & 37.76M / 2.1K & 512 / 512 & 798.93 / 798.93 & 6324.6 / 1344.6 & 213.8 / 79.6 & 2.447 / 2.447 & 409 / 409 \\
 &  &  & SingLEM & 512 / 512 & 3.27M / 3.27M & 68 / 68 & 3.27M / 68 & 16 / 16 & 58.83 / 58.83 & 2226.9 / 353.2 & 96.7 / 32.3 & 0.991 / 0.993 & 1,009 / 1,007 \\
 &  &  & LUNA-Base & 1,024 / 1,024 & 7.36M / 7.36M & 1.0K / 1.0K & 7.36M / 1.0K & 256 / 256 & 13.70 / 13.70 & 461.8 / 118.1 & 43.3 / 16.8 & 0.470 / 0.469 & 2,129 / 2,132 \\
\bottomrule
\end{tabular}
}
\end{table*}

\subsubsection{Parameter-Efficient Fine-tuning}

To assess the feasibility of parameter-efficient fine-tuning (PEFT) for EEG FMs, we conducted a preliminary experiment using Low-Rank Adaptation (LoRA)~\cite{hu2022lora} on two representative FMs, CBraMod and Neuro-GPT, evaluated on the BNCI2014001 and BNCI2014009 datasets under both the LOSO and within-subject few-shot scenarios. The results are reported in Table~\ref{tab:lora}.

\begin{table*}[!t]
    \centering
    \renewcommand{\arraystretch}{1.4}
    \fontsize{9}{12}\selectfont
    \caption{Balanced classification accuracies (\%) of Full Fine-tuning, Linear Probing, and LoRA fine-tuning on EEG FMs on BNCI2014001 and BNCI2014009.}
    \label{tab:lora}
    \scalebox{0.58}{
    \begin{tabular}{w{c}{2.0cm}|w{c}{1.9cm}|w{c}{1.7cm}|w{c}{1.6cm}|*{10}{w{c}{0.95cm}}|w{c}{1.6cm}}
        \toprule
        Scenario & Dataset & Tuning & Model & S0 & S1 & S2 & S3 & S4 & S5 & S6 & S7 & S8 & S9 & Avg. \\
        \midrule
\multirow{12}{*}{\makecell{Cross-subject \\ (LOSO)}} & \multirow{6}{*}{BNCI2014001} & \multirow{2}{*}{\makecell{Full \\ Fine-tuning}} & CBraMod & 54.51 & 46.99 & 63.19 & 47.92 & 43.29 & 44.91 & 54.75 & 60.19 & 61.57 & -- & 53.03$_{\pm 0.22}$ \\
 &  &  & Neuro-GPT & 59.72 & 32.18 & 62.62 & 34.49 & 31.71 & 37.27 & 39.81 & 65.62 & 59.26 & -- & 46.97$_{\pm 0.71}$ \\
\cmidrule(lr){3-15}
 &  & \multirow{2}{*}{\makecell{Linear \\ Probing}} & CBraMod & 49.31 & 31.37 & 55.67 & 34.61 & 29.28 & 29.17 & 32.75 & 56.60 & 54.28 & -- & 41.45$_{\pm 0.50}$ \\
 &  &  & Neuro-GPT & 54.63 & 38.43 & 62.85 & 47.45 & 32.87 & 33.33 & 39.81 & 64.12 & 60.65 & -- & 48.24$_{\pm 1.04}$ \\
\cmidrule(lr){3-15}
 &  & \multirow{2}{*}{LoRA} & CBraMod & 46.30 & 27.08 & 52.66 & 31.02 & 25.69 & 33.56 & 28.24 & 59.03 & 55.44 & -- & 39.89$_{\pm 0.38}$ \\
 &  &  & Neuro-GPT & 53.36 & 33.80 & 66.44 & 39.24 & 37.50 & 37.27 & 41.55 & 63.54 & 58.45 & -- & 47.90$_{\pm 1.44}$ \\
\cmidrule(lr){2-15}
 & \multirow{6}{*}{BNCI2014009} & \multirow{2}{*}{\makecell{Full \\ Fine-tuning}} & CBraMod & 66.42 & 74.76 & 67.81 & 81.98 & 85.76 & 69.83 & 72.22 & 77.88 & 91.67 & 84.65 & 77.30$_{\pm 0.28}$ \\
 &  &  & Neuro-GPT & 72.05 & 80.52 & 66.42 & 72.33 & 83.89 & 69.62 & 72.40 & 73.85 & 86.74 & 81.91 & 75.97$_{\pm 0.53}$ \\
\cmidrule(lr){3-15}
 &  & \multirow{2}{*}{\makecell{Linear \\ Probing}} & CBraMod & 57.15 & 55.52 & 58.09 & 56.98 & 65.87 & 55.21 & 54.93 & 53.40 & 67.53 & 61.63 & 58.63$_{\pm 0.54}$ \\
 &  &  & Neuro-GPT & 55.45 & 54.48 & 56.42 & 59.17 & 67.05 & 54.06 & 51.91 & 54.41 & 65.83 & 68.23 & 58.70$_{\pm 0.26}$ \\
\cmidrule(lr){3-15}
 &  & \multirow{2}{*}{LoRA} & CBraMod & 53.16 & 50.28 & 54.90 & 60.69 & 67.95 & 51.22 & 53.75 & 62.15 & 65.45 & 56.74 & 57.63$_{\pm 0.31}$ \\
 &  &  & Neuro-GPT & 68.19 & 78.99 & 66.81 & 71.88 & 82.78 & 70.24 & 72.26 & 71.35 & 83.96 & 83.19 & 74.97$_{\pm 0.14}$ \\
        \midrule
\multirow{12}{*}{\makecell{Within-subject \\ (Few-shot)}} & \multirow{6}{*}{BNCI2014001} & \multirow{2}{*}{\makecell{Full \\ Fine-tuning}} & CBraMod & 56.37 & 46.57 & 69.61 & 38.40 & 39.38 & 30.07 & 56.54 & 61.93 & 54.25 & -- & 50.34$_{\pm 1.18}$ \\
 &  &  & Neuro-GPT & 44.77 & 40.20 & 62.42 & 37.91 & 35.29 & 39.54 & 38.40 & 64.54 & 54.41 & -- & 46.39$_{\pm 1.92}$ \\
\cmidrule(lr){3-15}
 &  & \multirow{2}{*}{\makecell{Linear \\ Probing}} & CBraMod & 34.97 & 37.42 & 43.95 & 35.78 & 27.94 & 32.52 & 34.15 & 60.62 & 59.31 & -- & 40.74$_{\pm 1.06}$ \\
 &  &  & Neuro-GPT & 55.23 & 41.01 & 60.29 & 42.81 & 36.11 & 41.18 & 46.08 & 71.24 & 54.41 & -- & 49.82$_{\pm 1.55}$ \\
\cmidrule(lr){3-15}
 &  & \multirow{2}{*}{LoRA} & CBraMod & 35.95 & 44.61 & 67.65 & 40.52 & 30.39 & 33.66 & 41.50 & 66.18 & 72.55 & -- & 48.11$_{\pm 0.33}$ \\
 &  &  & Neuro-GPT & 48.04 & 39.05 & 48.86 & 40.52 & 32.84 & 35.29 & 39.54 & 56.70 & 51.96 & -- & 43.65$_{\pm 1.22}$ \\
\cmidrule(lr){2-15}
 & \multirow{6}{*}{BNCI2014009} & \multirow{2}{*}{\makecell{Full \\ Fine-tuning}} & CBraMod & 57.18 & 53.56 & 53.63 & 55.73 & 60.83 & 54.67 & 53.08 & 55.10 & 64.30 & 60.47 & 56.85$_{\pm 1.03}$ \\
 &  &  & Neuro-GPT & 52.39 & 55.36 & 50.00 & 54.17 & 62.09 & 60.17 & 51.19 & 51.76 & 52.26 & 60.83 & 55.02$_{\pm 0.97}$ \\
\cmidrule(lr){3-15}
 &  & \multirow{2}{*}{\makecell{Linear \\ Probing}} & CBraMod & 50.00 & 50.22 & 49.96 & 50.00 & 51.10 & 49.92 & 49.65 & 49.81 & 50.15 & 50.95 & 50.18$_{\pm 0.15}$ \\
 &  &  & Neuro-GPT & 50.00 & 50.00 & 49.96 & 50.00 & 50.34 & 50.46 & 49.96 & 50.38 & 51.34 & 54.02 & 50.65$_{\pm 0.31}$ \\
\cmidrule(lr){3-15}
 &  & \multirow{2}{*}{LoRA} & CBraMod & 49.88 & 53.88 & 49.56 & 54.12 & 56.03 & 52.69 & 52.69 & 50.11 & 55.75 & 52.89 & 52.76$_{\pm 0.42}$ \\
 &  &  & Neuro-GPT & 56.42 & 66.40 & 54.04 & 63.59 & 69.44 & 58.50 & 55.62 & 63.77 & 64.81 & 68.53 & 62.11$_{\pm 0.88}$ \\
        \bottomrule
    \end{tabular}
    }
\end{table*}

\subsubsection{Statistical Significance Analysis} \label{appendix:statistic}

For each EEG FM, we conducted a scenario-level paired Wilcoxon signed-rank test between full fine-tuning and linear probing over dataset--scenario pairs. Each valid pair corresponds to the full and linear results of the same model on the same dataset--scenario. A positive $\Delta$ indicates that full fine-tuning outperforms linear probing; for SEED-VIG, where a lower RMSE is better, the sign was reversed accordingly. ``Full wins'', ``Linear wins'', and ``Ties'' denote the numbers of dataset--scenario pairs favoring full fine-tuning, linear probing, and neither strategy, respectively. The rank-biserial correlation $r$ is reported as the effect size. The results are summarized in Table~\ref{tab:full_linear_wilcoxon}.

\begin{table*}[!t]
\centering
\renewcommand{\arraystretch}{1.4}
\setlength{\tabcolsep}{4pt}
\fontsize{8}{10}\selectfont
\caption{Scenario-level paired Wilcoxon signed-rank tests comparing full fine-tuning and linear probing for EEG FMs.}
\label{tab:full_linear_wilcoxon}
\scalebox{0.96}{
\begin{tabular}{c|c c c c c c c c c}
\toprule
\textbf{Model} &
\makecell{\textbf{N valid}\\\textbf{pairs}} &
\makecell{\textbf{Full}\\\textbf{wins}} &
\makecell{\textbf{Linear}\\\textbf{wins}} &
\textbf{Ties} &
\makecell{\textbf{Median}\\$\boldsymbol{\Delta}$} &
\makecell{\textbf{Wilcoxon}\\$\boldsymbol{W}$} &
\textbf{$p$-value} &
\textbf{Sig.} &
\makecell{\textbf{Rank-biserial}\\$\boldsymbol{r}$} \\
\midrule
BENDR            & 25 & 23 & 2  & 0 & 12.2200 & 4.0000   & $4.17\times10^{-7}$ & **** & 0.9754 \\
BIOT-6D          & 25 & 21 & 4  & 0 & 2.2200  & 34.0000  & $2.17\times10^{-4}$ & ***  & 0.7908 \\
LaBraM           & 25 & 19 & 6  & 0 & 2.0300  & 64.0000  & 0.0067              & **   & 0.6062 \\
Neuro-GPT        & 25 & 18 & 7  & 0 & 2.8200  & 52.5000  & 0.0031              & **   & 0.6769 \\
EEGPT            & 25 & 7  & 18 & 0 & -1.3600 & 80.0000  & 0.0255               & *    & -0.5077 \\
CBraMod          & 25 & 23 & 2  & 0 & 6.6700  & 7.0000   & $1.13\times10^{-6}$ & **** & 0.9569 \\
TFM              & 25 & 24 & 0  & 1 & 1.6500  & 0.0000   & $1.82\times10^{-5}$ & **** & 1.0000 \\
BrainOmni-Tiny   & 21 & 15 & 6  & 0 & 1.8000  & 64.0000  & 0.0760               & ns   & 0.4459 \\
BrainOmni-Base   & 21 & 16 & 5  & 0 & 1.1700  & 27.5000  & 0.0022              & **   & 0.7619 \\
EEGMamba         & 25 & 19 & 6  & 0 & 4.1100  & 34.0000  & $2.17\times10^{-4}$ & ***  & 0.7908 \\
SingLEM          & 25 & 15 & 10 & 0 & 0.2400  & 117.0000 & 0.2304               & ns   & 0.2800 \\
LUNA-Base        & 25 & 17 & 8  & 0 & 0.5400  & 71.0000  & 0.0125               & *    & 0.5631 \\
\bottomrule
\end{tabular}
}
\vspace{1mm}  
\parbox{0.95\textwidth}{
    \scriptsize
    \textbf{Note:} ****: $p < 0.0001$; ***: $p < 0.001$; **: $p < 0.01$; *: $p < 0.05$; ns: not significant.
}
\end{table*}

\begin{figure*}[!t]
\centering
\includegraphics[width=\linewidth,clip]{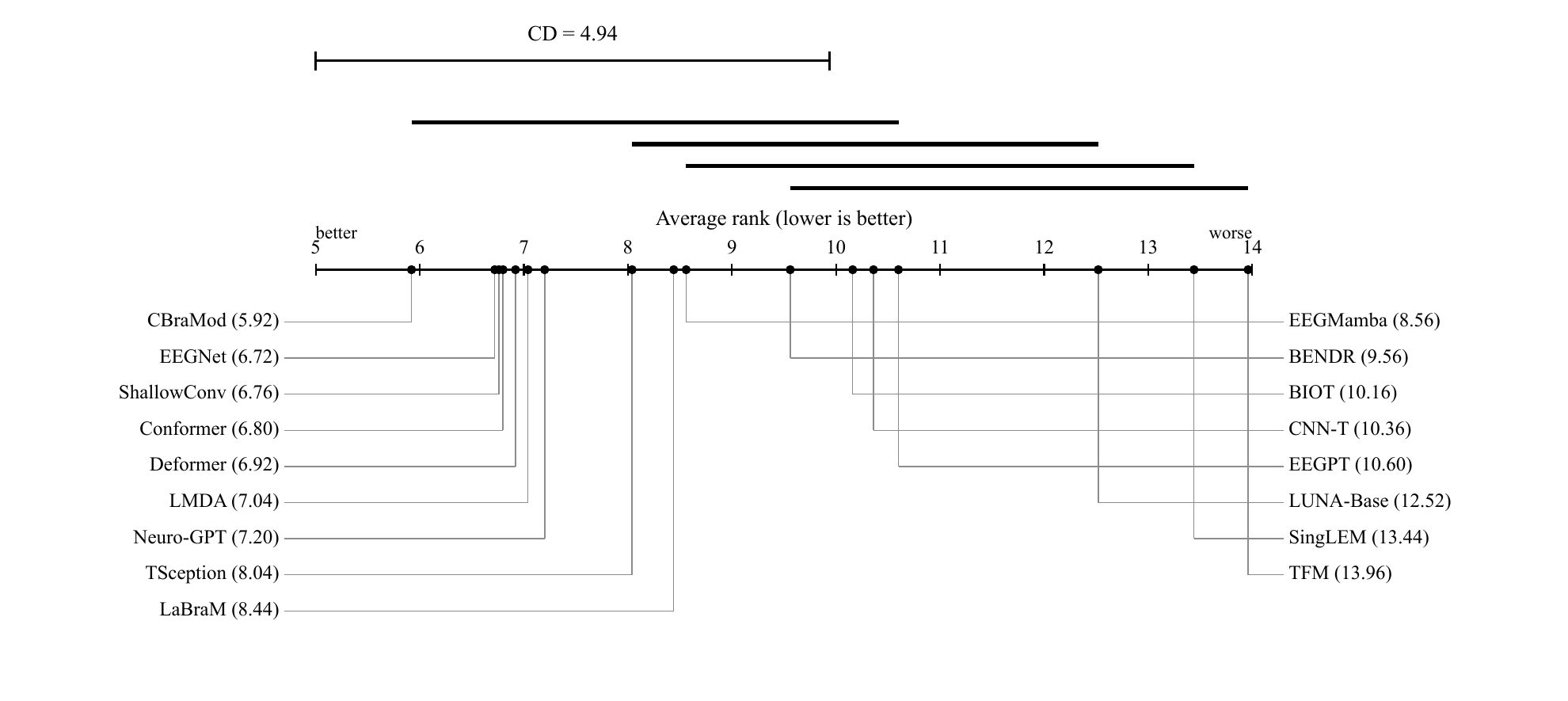
}
\caption{Critical difference (CD) diagram of the average ranks of all evaluated models across the 25 dataset--scenario combinations, based on the Nemenyi post-hoc test. A lower rank indicates better performance, and models connected by a horizontal bar are not significantly different at the 0.05 level.}
\label{fig:cd_diagram}
\end{figure*}

In addition, we assessed the statistical reliability of the overall model rankings using the Friedman test followed by the Nemenyi post-hoc test. The 25 dataset--scenario combinations were used as evaluation blocks. To satisfy the complete-block assumption of the Friedman/Nemenyi procedure, models with incomplete results, including Traditional ML, BrainOmni-Tiny, and BrainOmni-Base, were excluded, and ranks were recomputed within each scenario for the remaining 17 models. The Friedman test rejected the null hypothesis that all models have equivalent ranks ($\chi_F^2=98.19$, $p=7.56\times10^{-14}$), and the Iman-Davenport corrected statistic was also significant ($F_F=7.81$, $p=4.34\times10^{-16}$). The Nemenyi post-hoc test yielded a critical difference of $4.94$ at $\alpha=0.05$. The resulting critical difference diagram is shown in Fig.~\ref{fig:cd_diagram}, where models connected by a horizontal bar are not significantly different at the 0.05 level.

\subsubsection{Comparison of different fine-tuning ratios} \label{appendix:bench_13579}

\begin{figure*}[t]
\centering
\includegraphics[width=1.0\linewidth,clip]{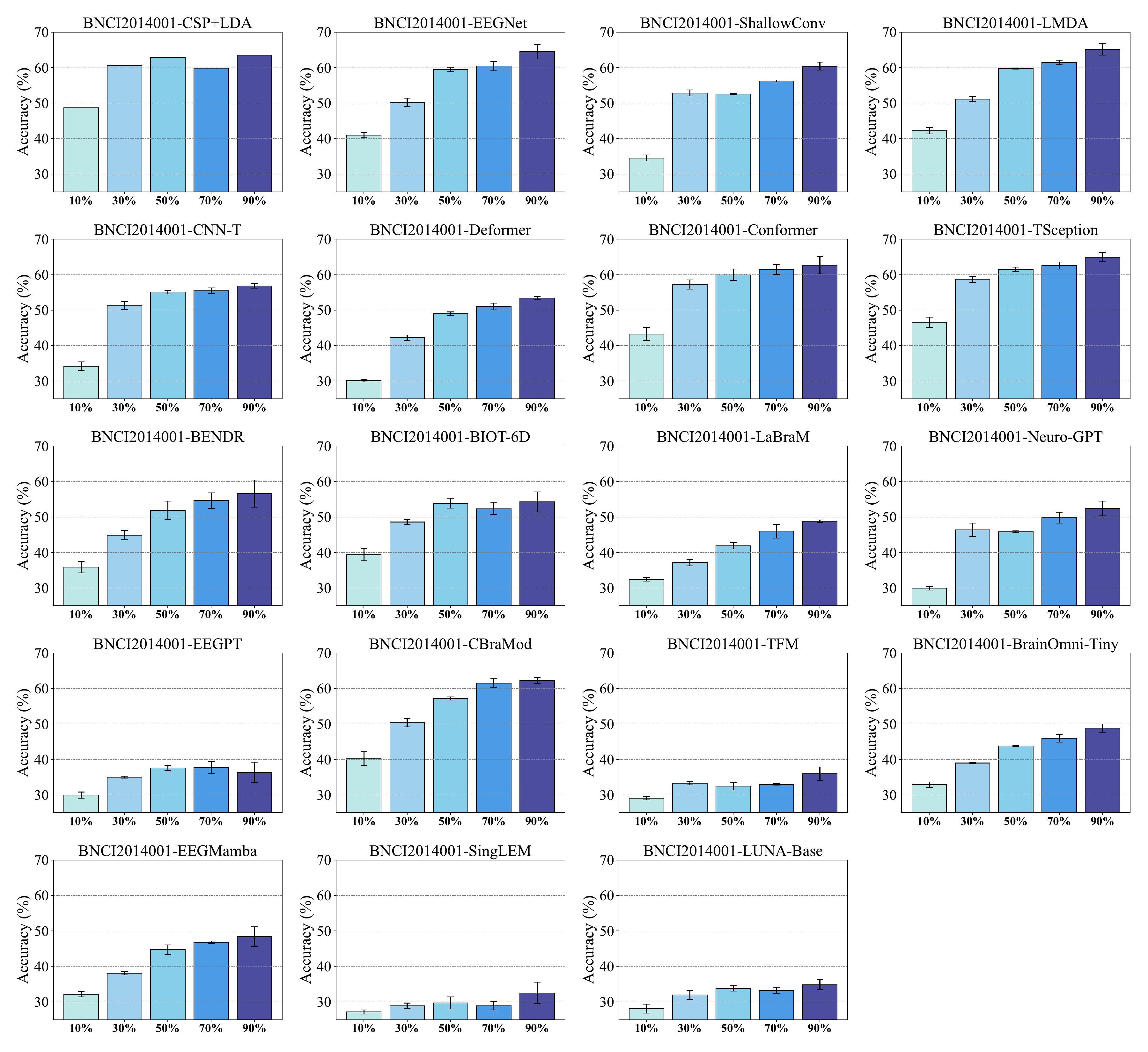}
\caption{Accuracy comparison on the BNCI2014001 dataset across different fine-tuning ratios.}
\label{fig:fint_ratio_14001_2}
\end{figure*}

\begin{figure*}[t]
\centering
\includegraphics[width=1.0\linewidth,clip]{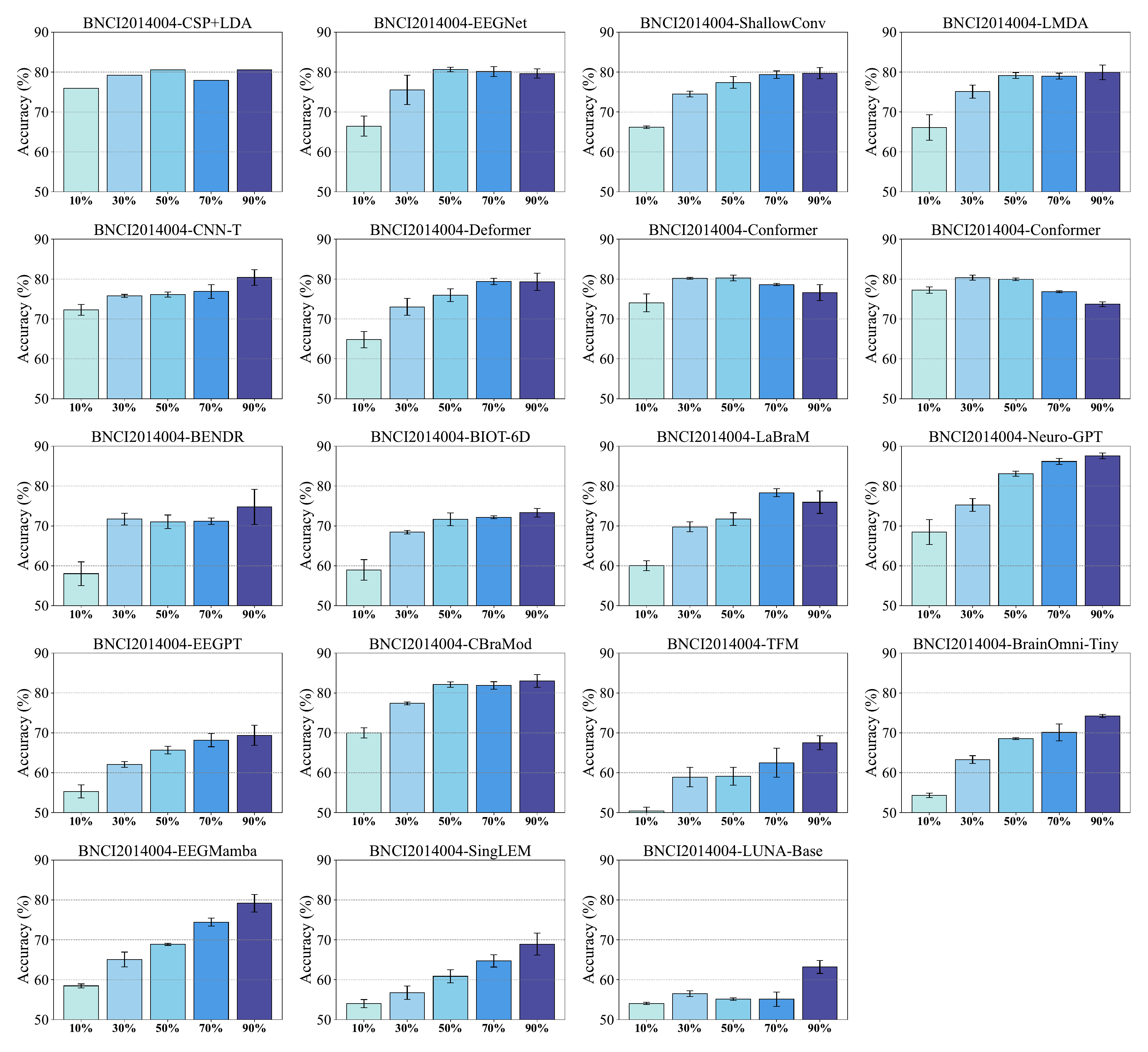}
\caption{Accuracy comparison on the BNCI2014004 dataset across different fine-tuning ratios.}
\label{fig:fint_ratio_14004}
\end{figure*}

\begin{figure*}[t]
\centering
\includegraphics[width=1.0\linewidth,clip]{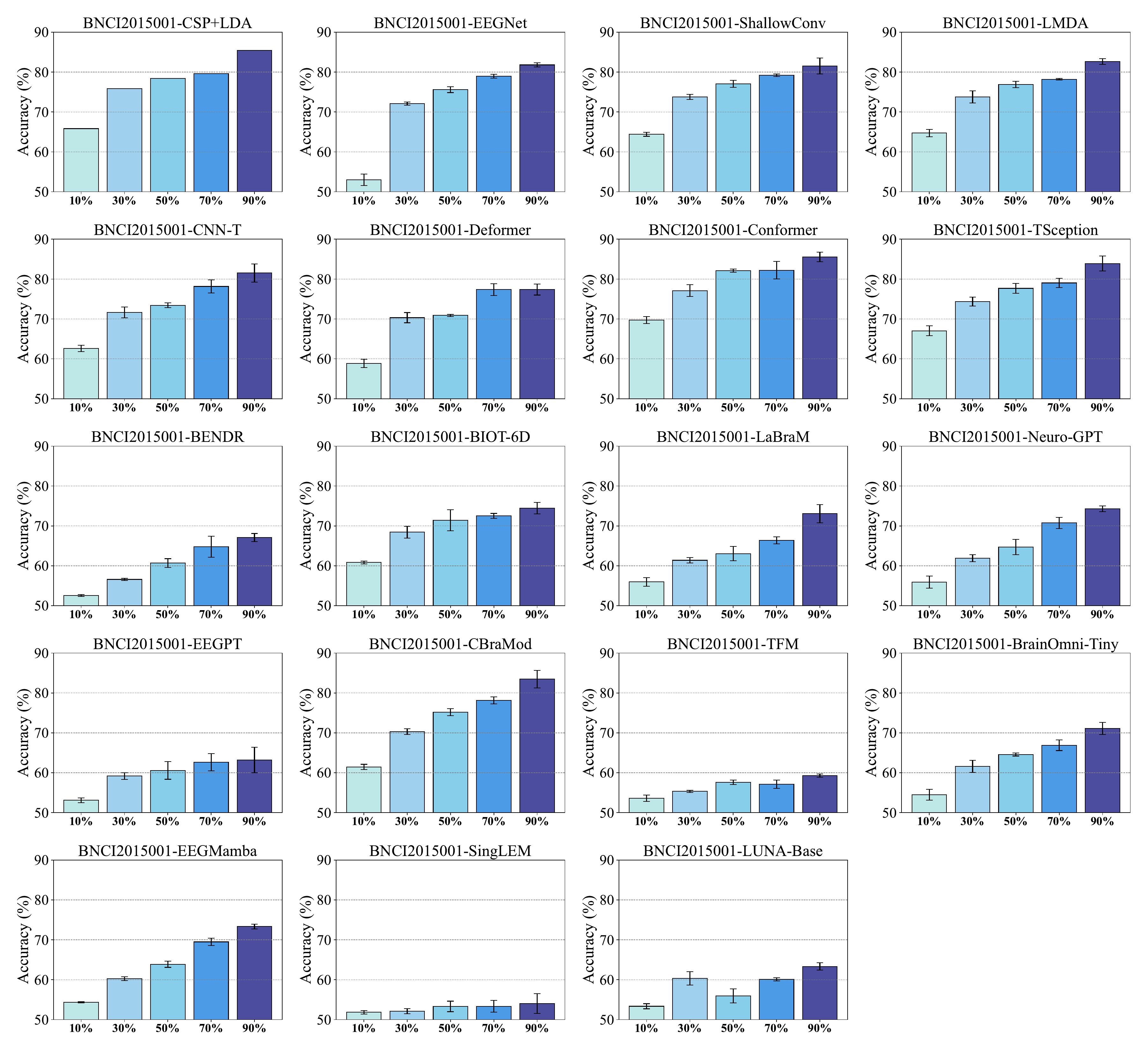}
\caption{Accuracy comparison on the BNCI2015001 dataset across different fine-tuning ratios.}
\label{fig:fint_ratio_15001}
\end{figure*}

\begin{figure*}[t]
\centering
\includegraphics[width=1.0\linewidth,clip]{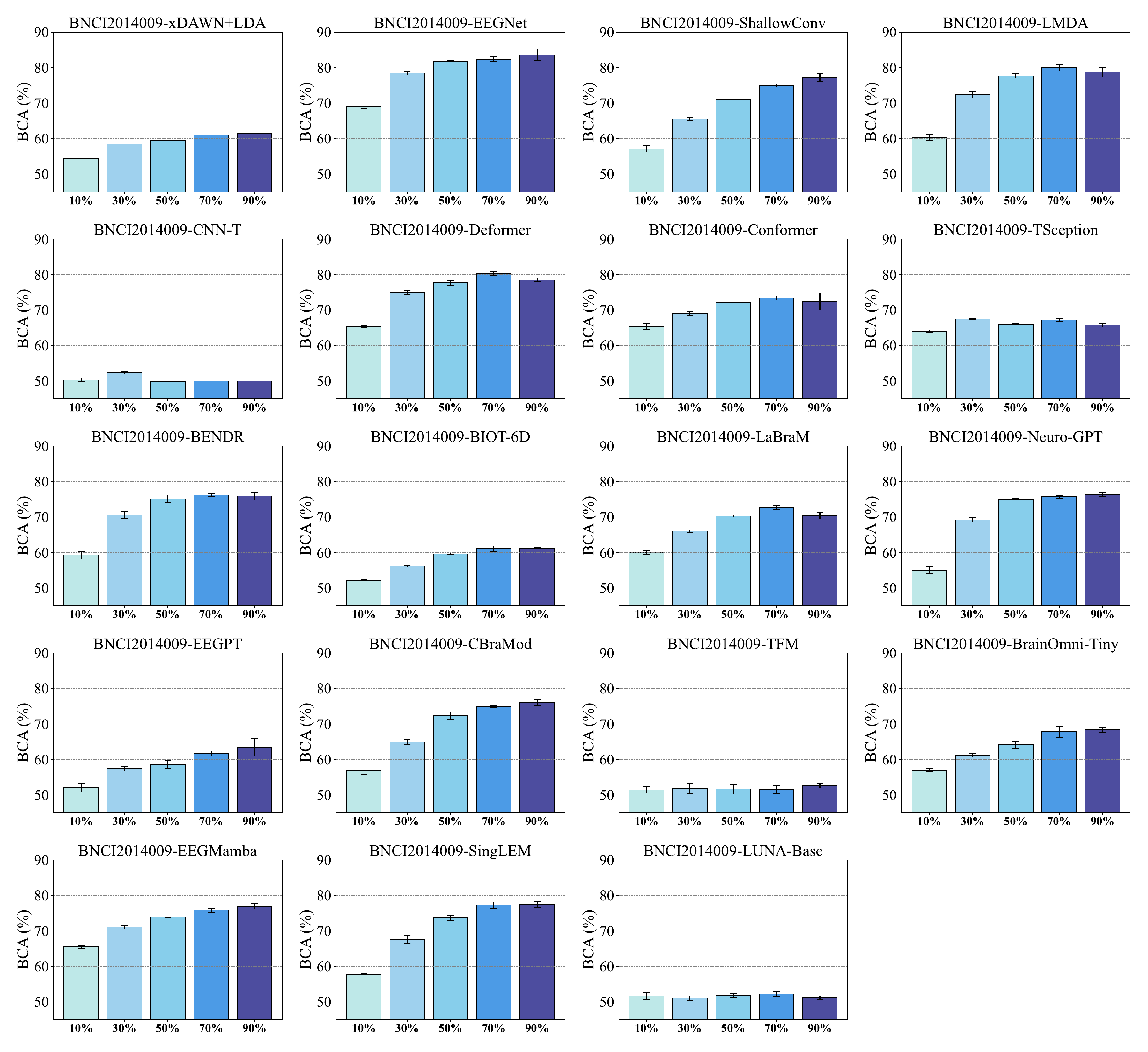}
\caption{Balanced classification accuracy comparison on the BNCI2014009 dataset across different fine-tuning ratios.}
\label{fig:fint_ratio_14009}
\end{figure*}

\begin{figure*}[t]
\centering
\includegraphics[width=1.0\linewidth,clip]{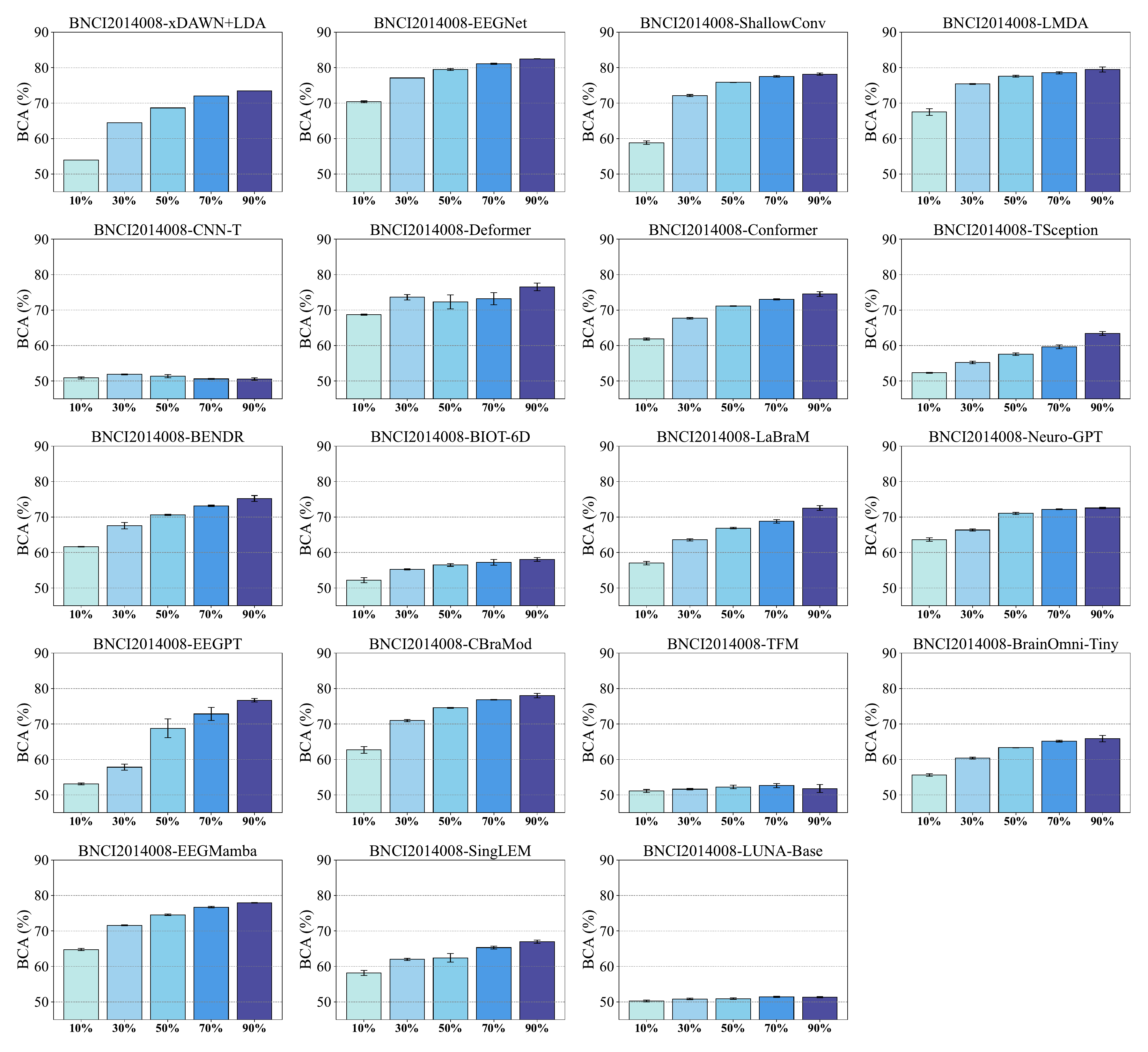}
\caption{Balanced classification accuracy comparison on the BNCI2014008 dataset across different fine-tuning ratios.}
\label{fig:fint_ratio_14008}
\end{figure*}

\begin{figure*}[t] \centering
\includegraphics[width=1.0\linewidth,clip]{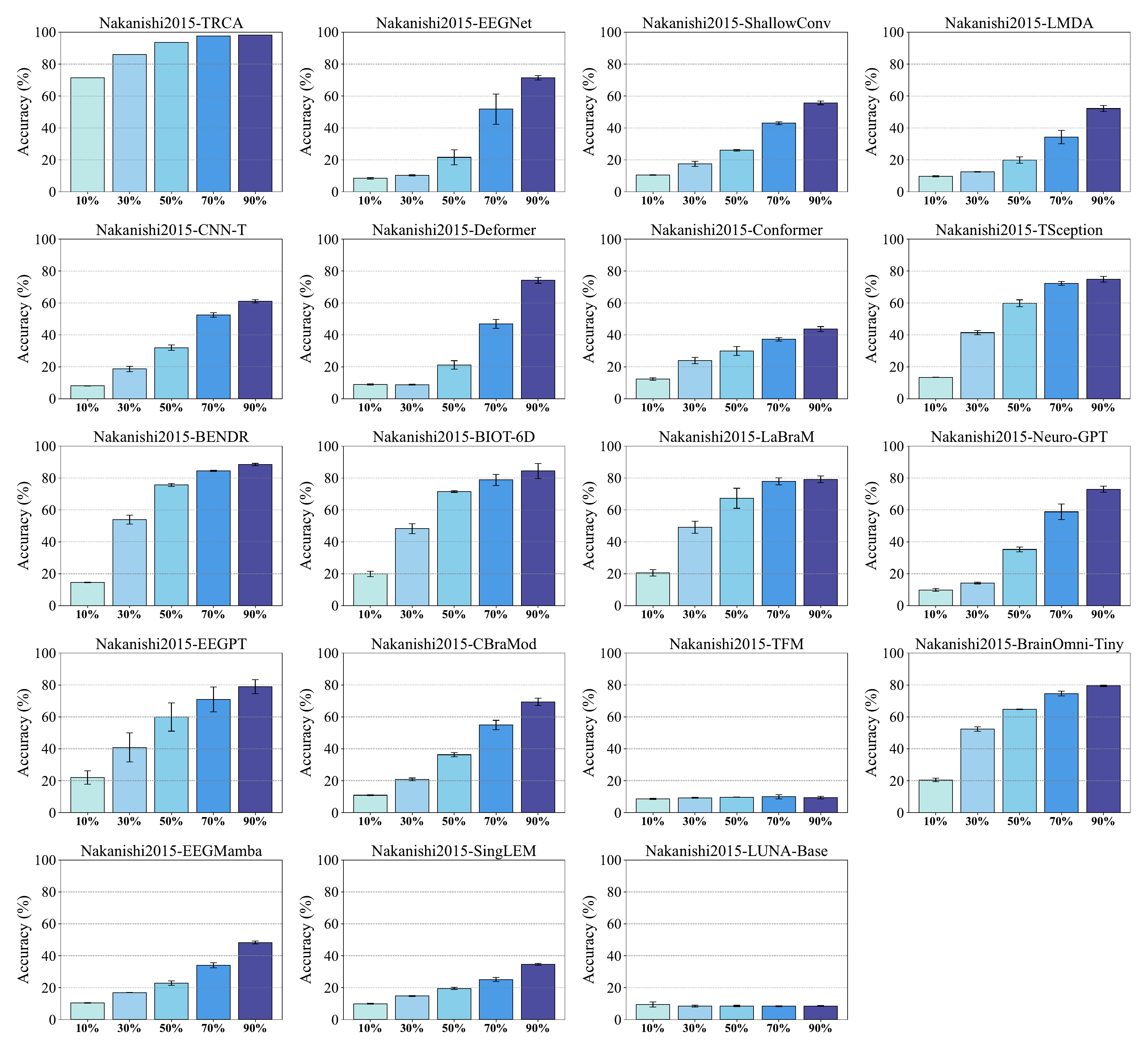}
\caption{Accuracy comparison on the Nakanishi2015 dataset across different fine-tuning ratios.}
\label{fig:fint_ratio_ssvep2}
\end{figure*}

We conducted an analysis on different fine-tuning data ratios, varying from 10\% to 90\% in increments of 20\%, across all specialist and FMs. The results are presented in Figs.~\ref{fig:fint_ratio_14001_2} - ~\ref{fig:fint_ratio_ssvep2}.

Most models exhibited consistent performance improvements as the fine-tuning data ratio increased from 10\% to 90\%, which aligns with intuitive expectations. Notably, the relative ranking among models remained largely stable across different fine-tuning ratios. For instance, TRCA consistently achieved the highest accuracy on the Nakanishi2015 dataset regardless of the data ratio, while EEGNet maintained competitive performance across all settings on the BNCI2014008 dataset. This observation suggests that model superiority is relatively independent of fine-tuning data availability, and that a well-performing model under low-data conditions tends to preserve its advantage as more data becomes available. However, minimal calibration or even calibration-free adaptation remains a critical requirement for practical BCI deployment. Developing models capable of rapid adaptation to downstream tasks with limited calibration data continues to be an important and open challenge.

\subsection{Representation Visualization}

\begin{figure*}[t]
\centering
\subfigure[]{\includegraphics[width=0.19\linewidth,clip]{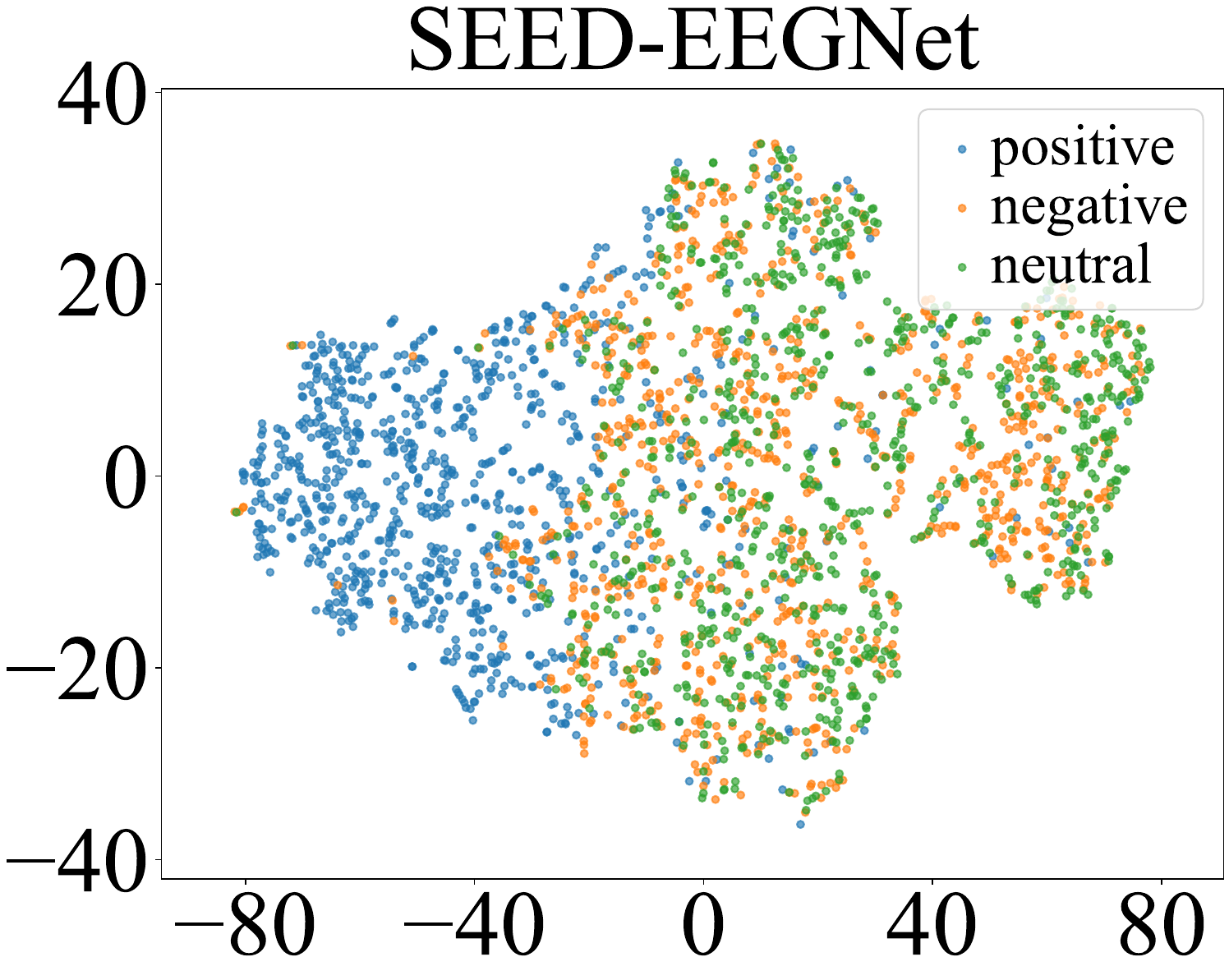}}
\subfigure[]{\includegraphics[width=0.19\linewidth,clip]{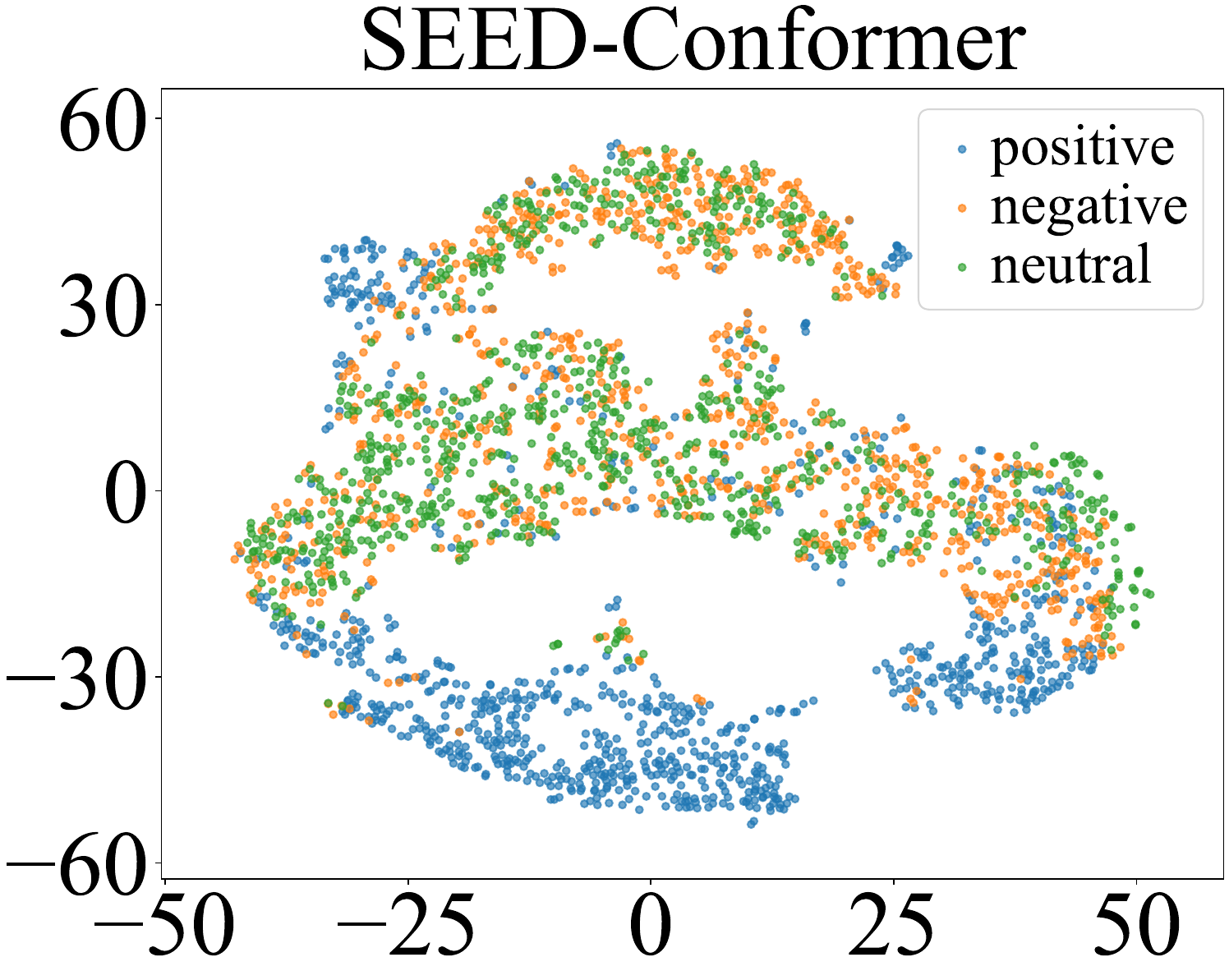}}
\subfigure[]{\includegraphics[width=0.19\linewidth,clip]{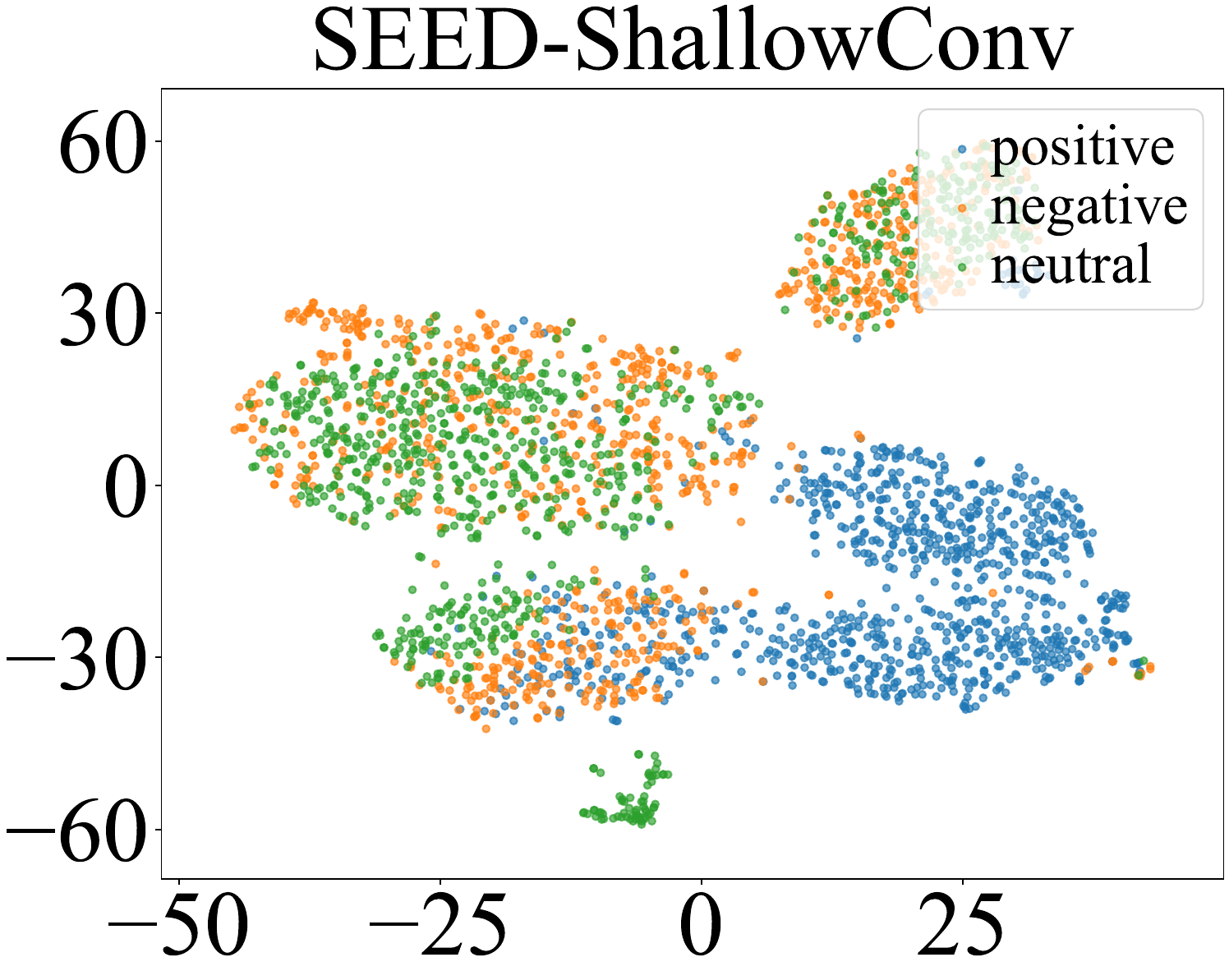}}
\subfigure[]{\includegraphics[width=0.19\linewidth,clip]{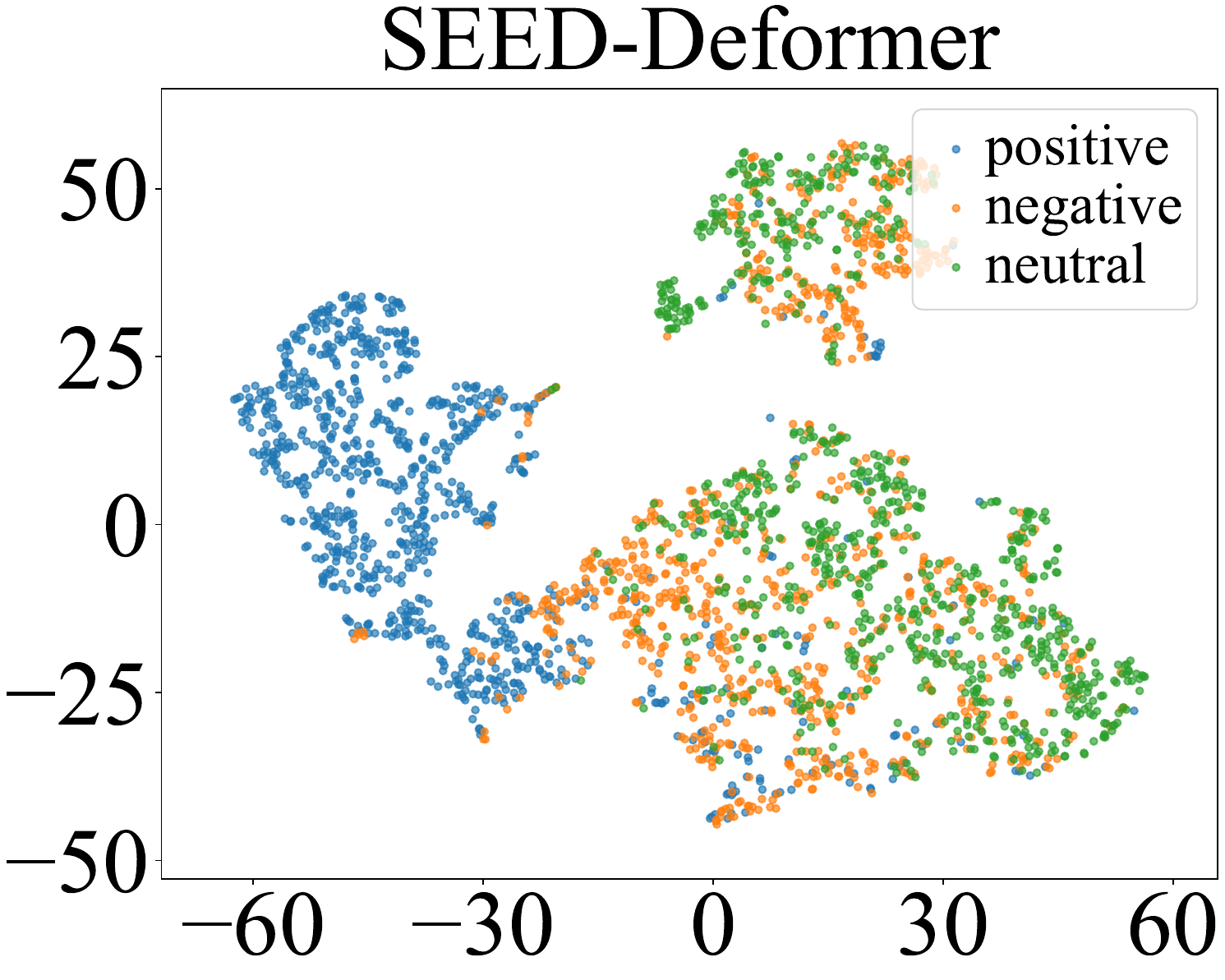}}
\subfigure[]{\includegraphics[width=0.19\linewidth,clip]{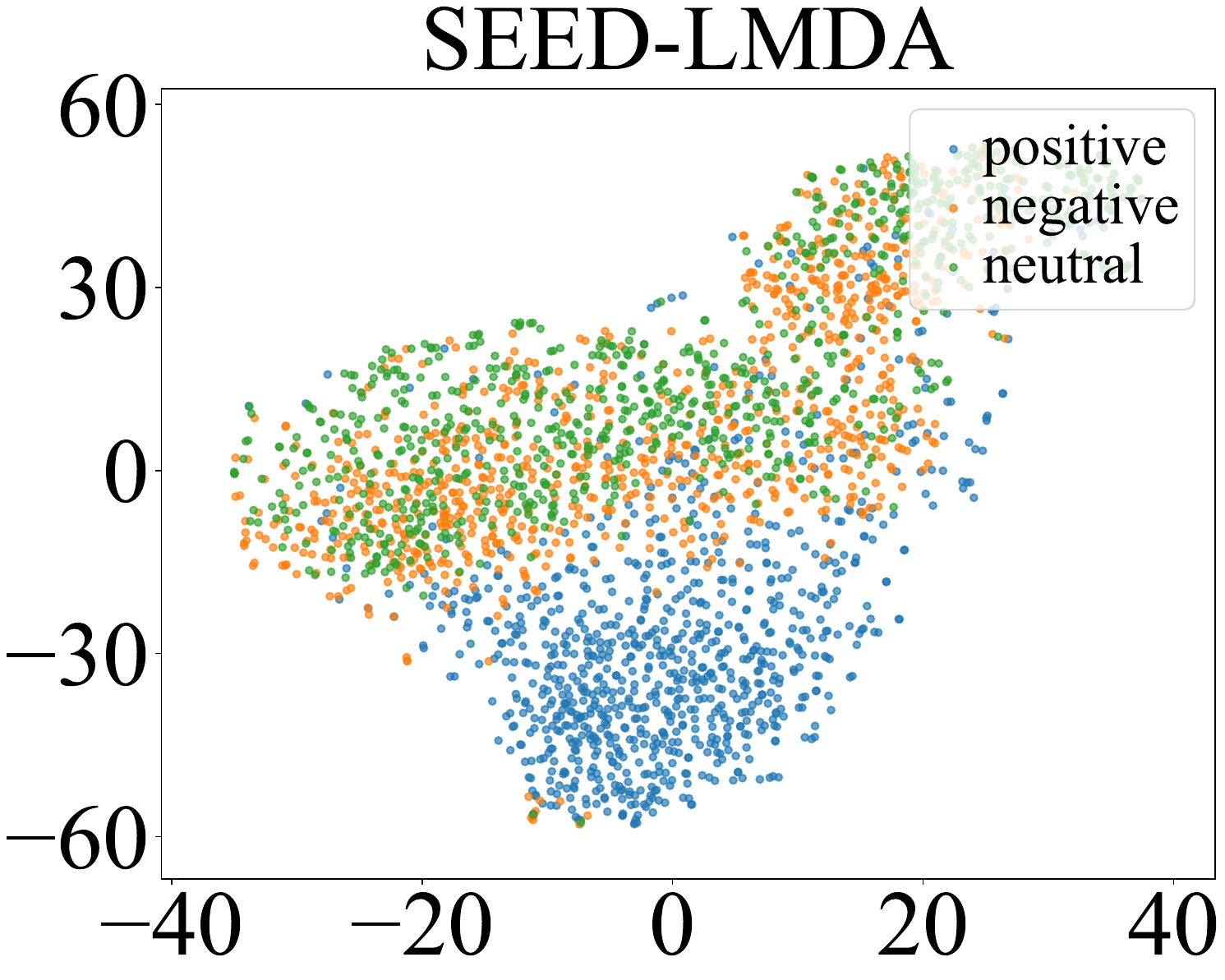}}\\
\subfigure[]{\includegraphics[width=0.19\linewidth,clip]{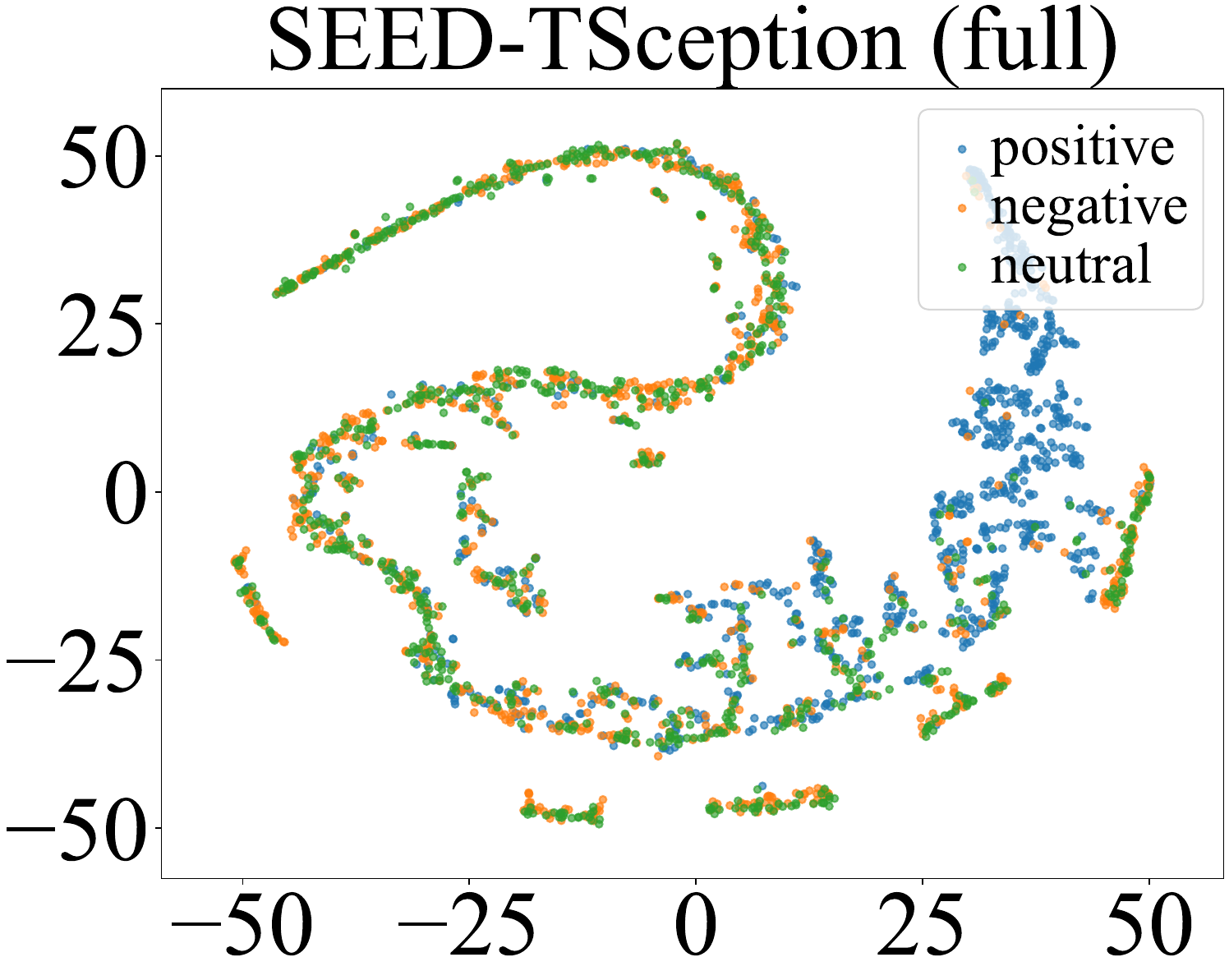}}
\subfigure[]{\includegraphics[width=0.19\linewidth,clip]{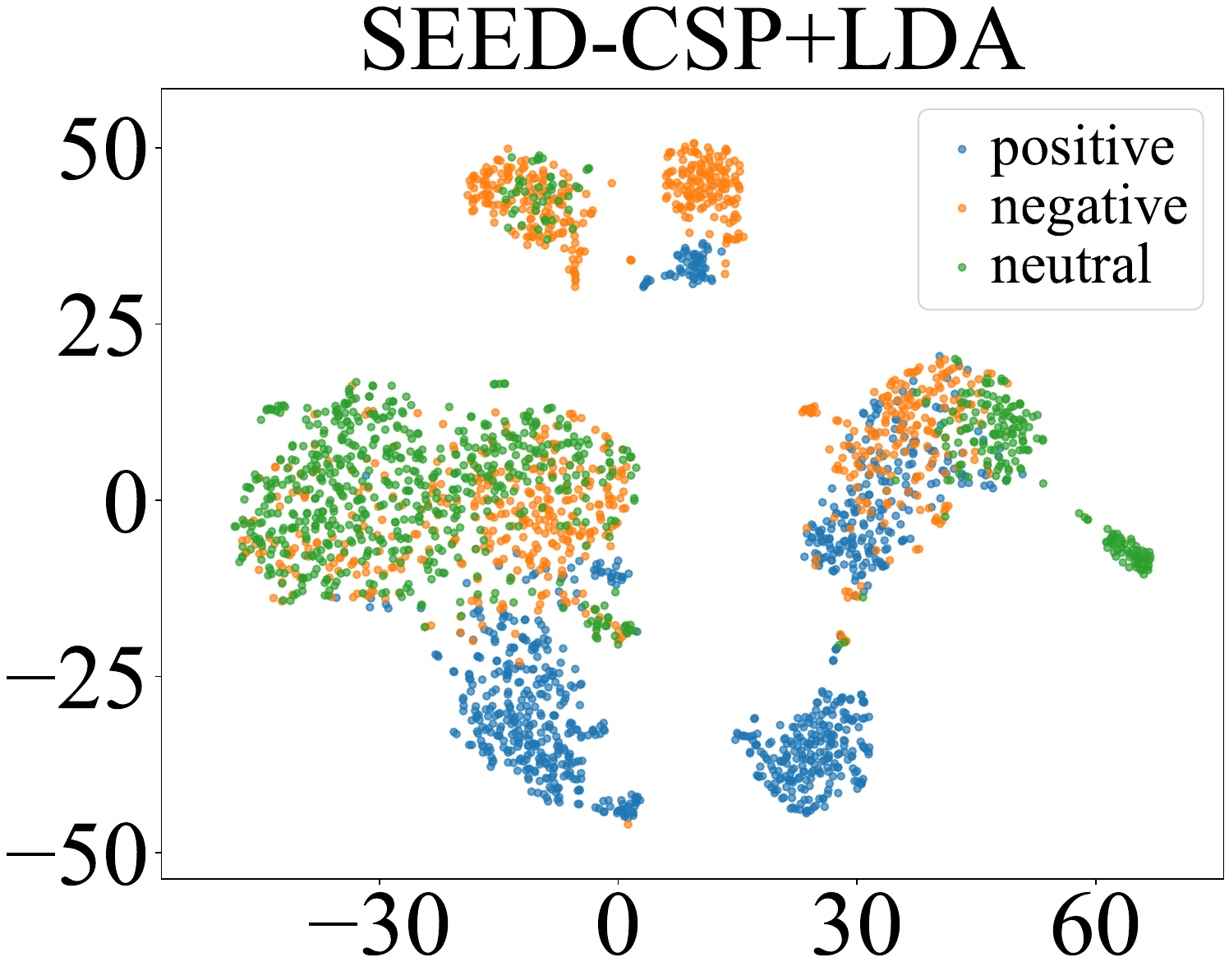}}
\subfigure[]{\includegraphics[width=0.19\linewidth,clip]{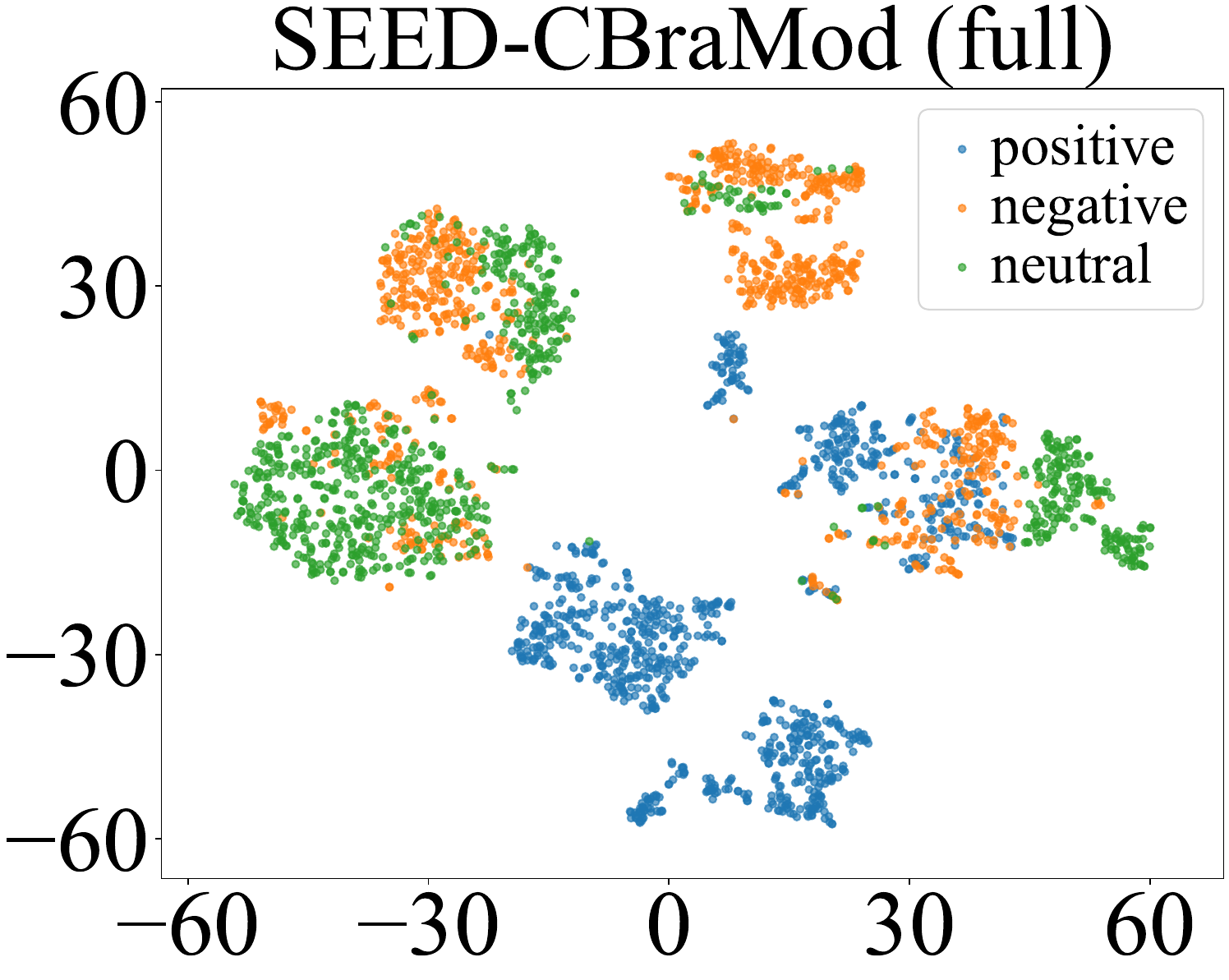}}
\subfigure[]{\includegraphics[width=0.19\linewidth,clip]{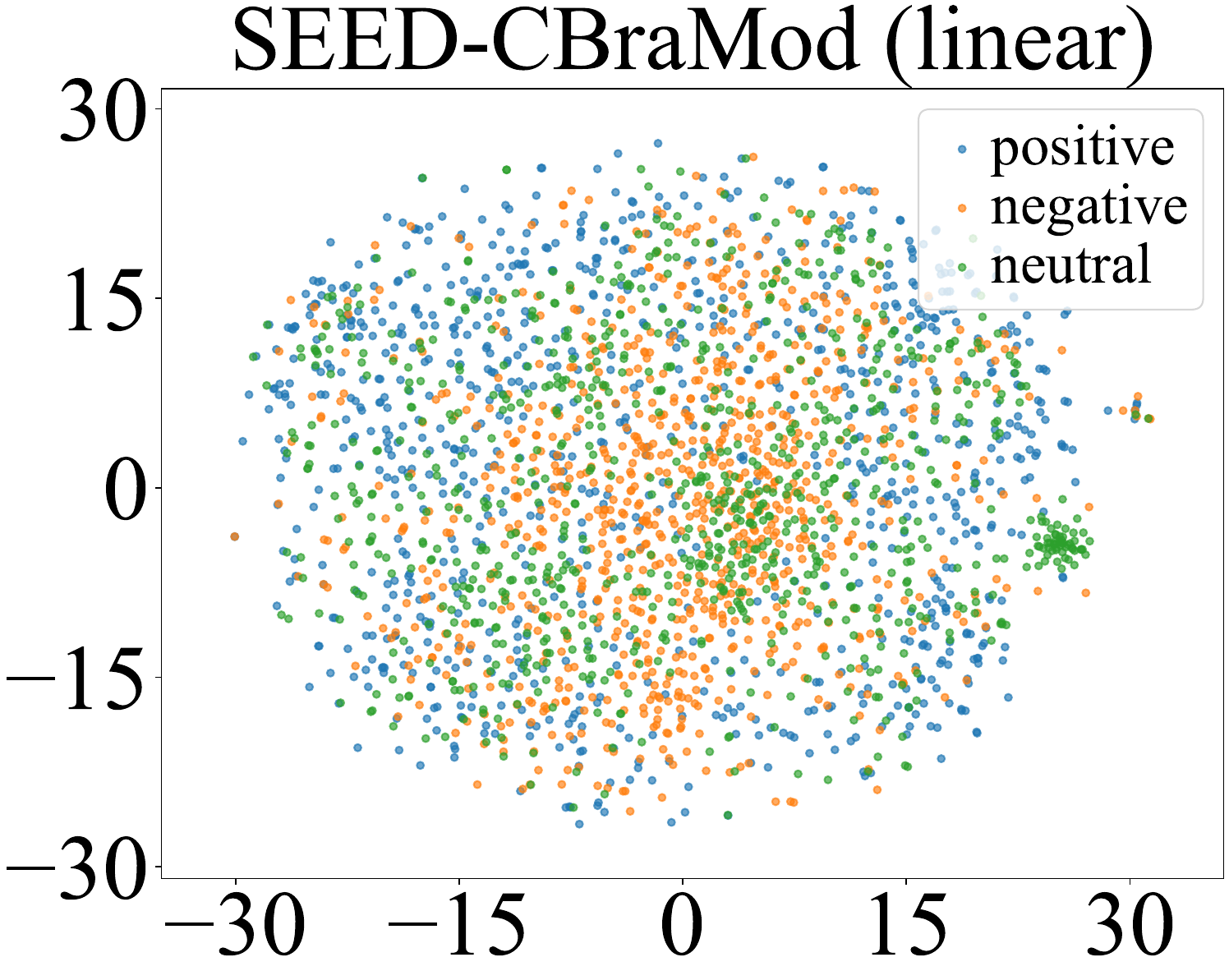}}
\subfigure[]{\includegraphics[width=0.19\linewidth,clip]{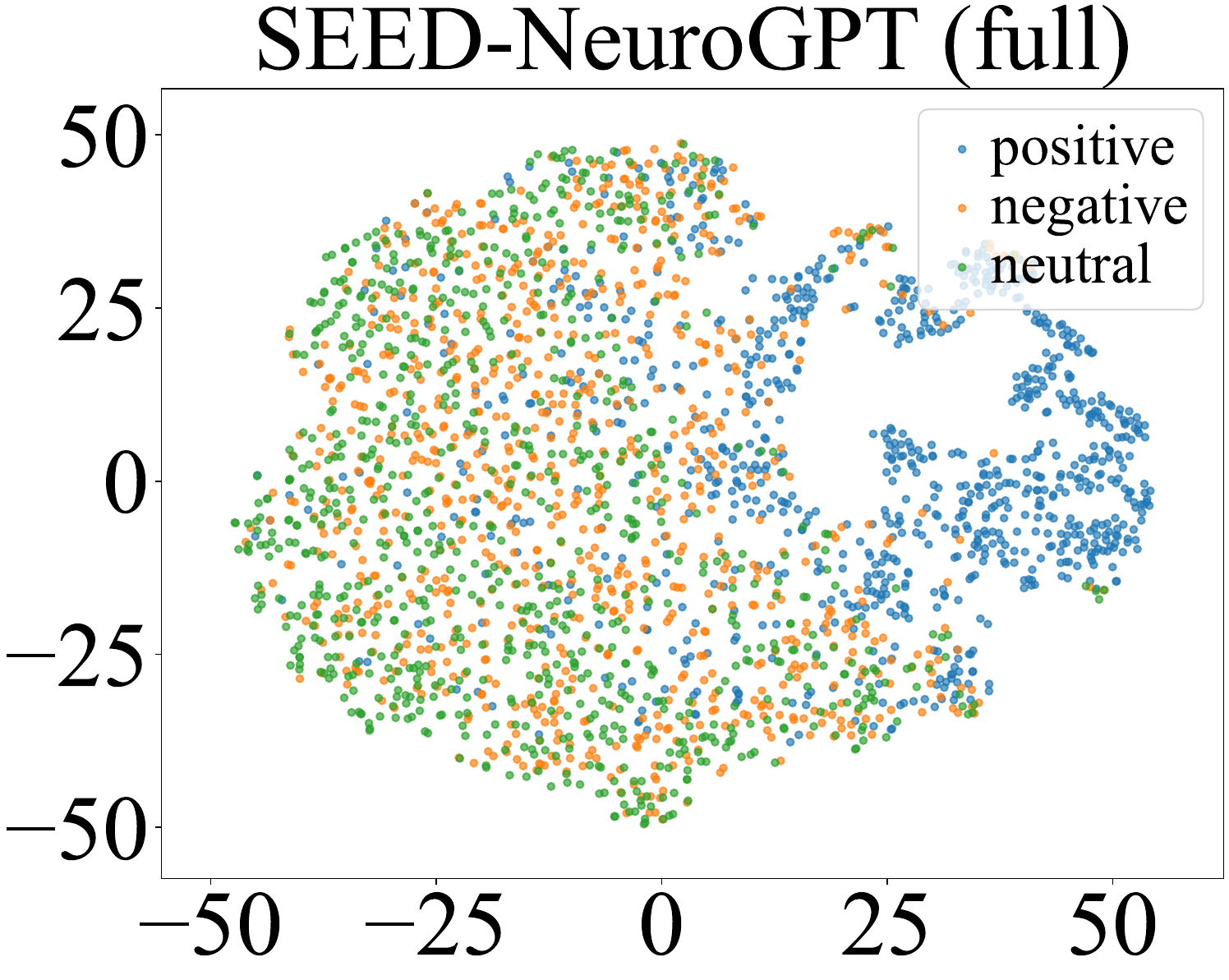}} \\
\subfigure[]{\includegraphics[width=0.19\linewidth,clip]{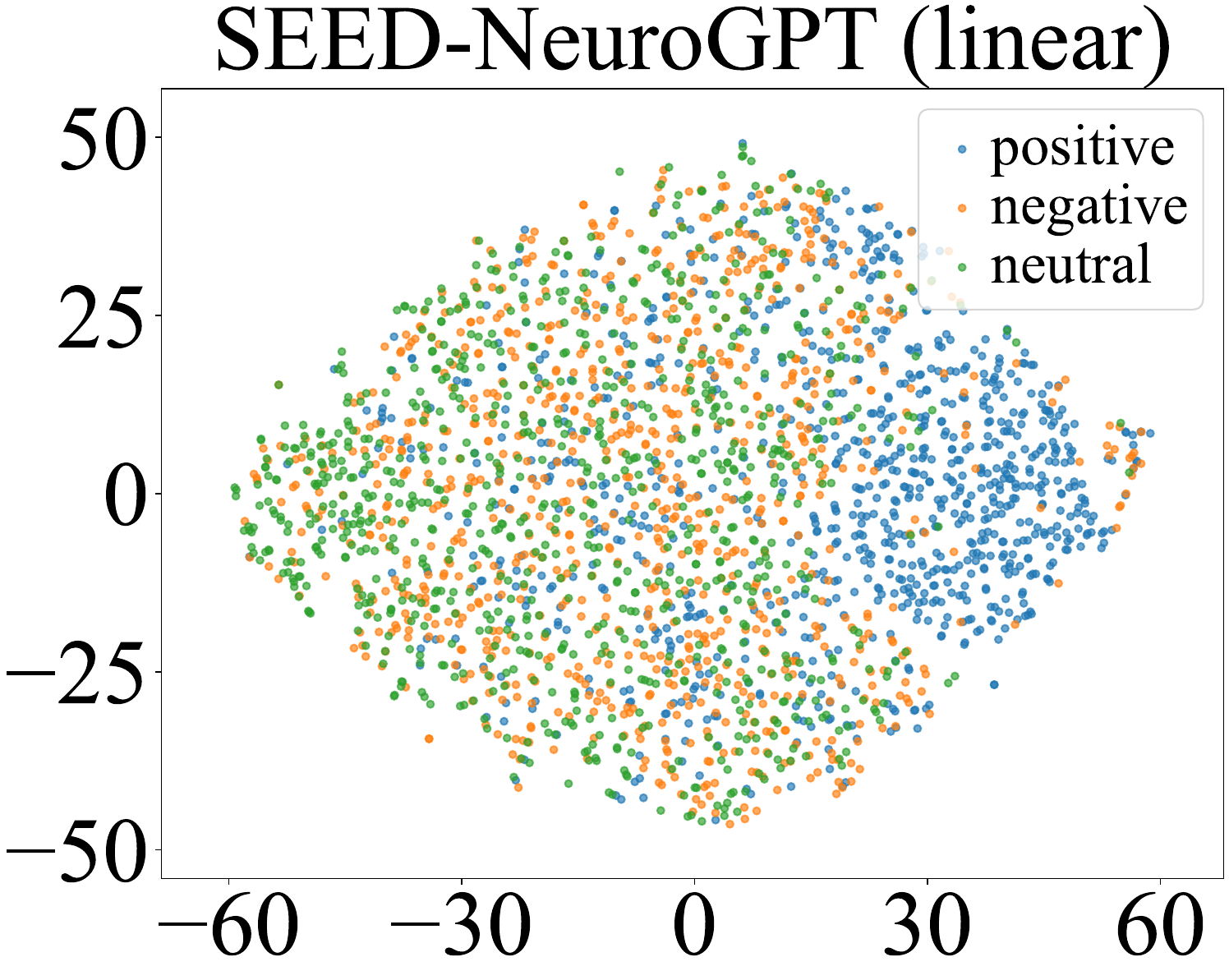}}
\subfigure[]{\includegraphics[width=0.19\linewidth,clip]{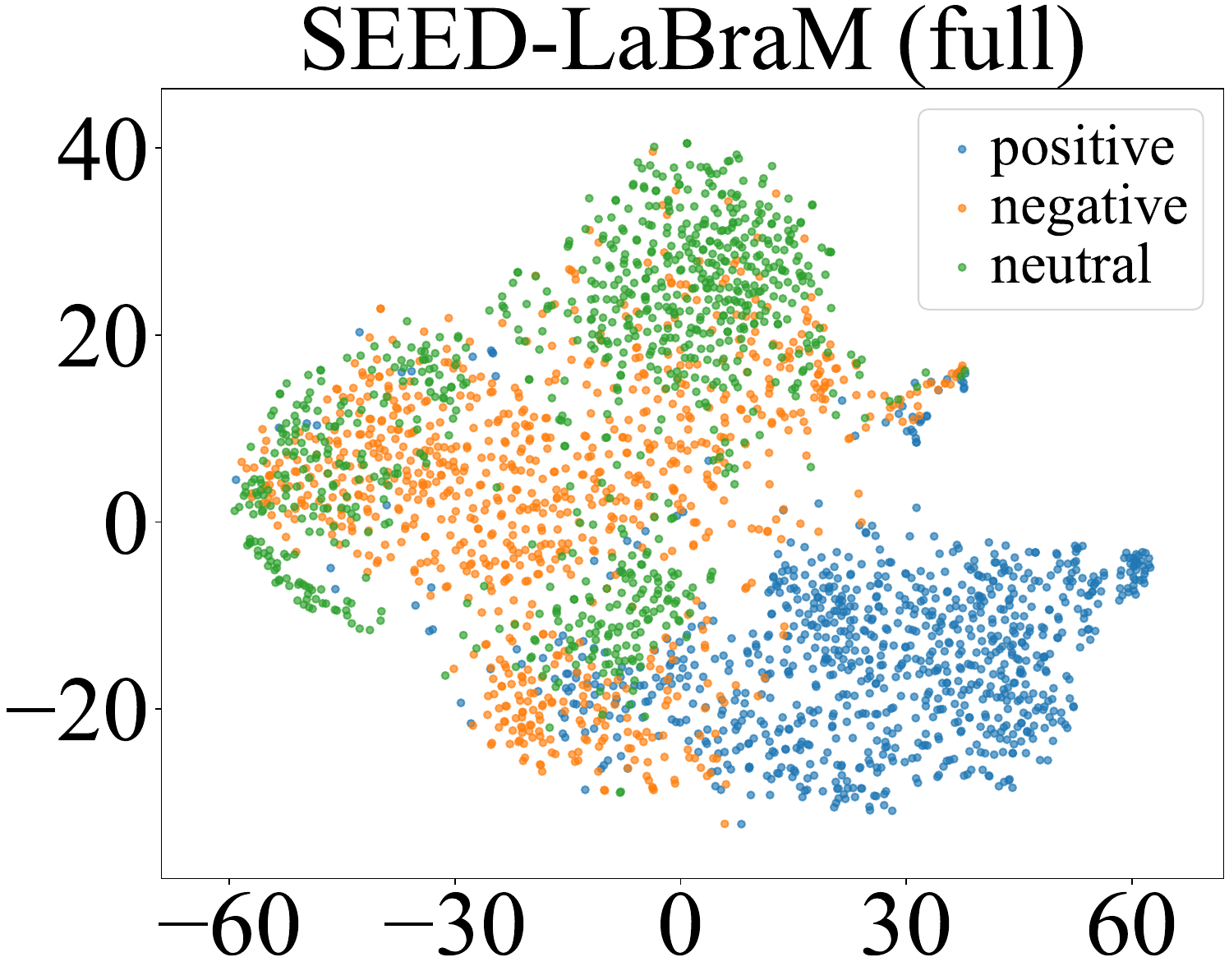}}
\subfigure[]{\includegraphics[width=0.19\linewidth,clip]{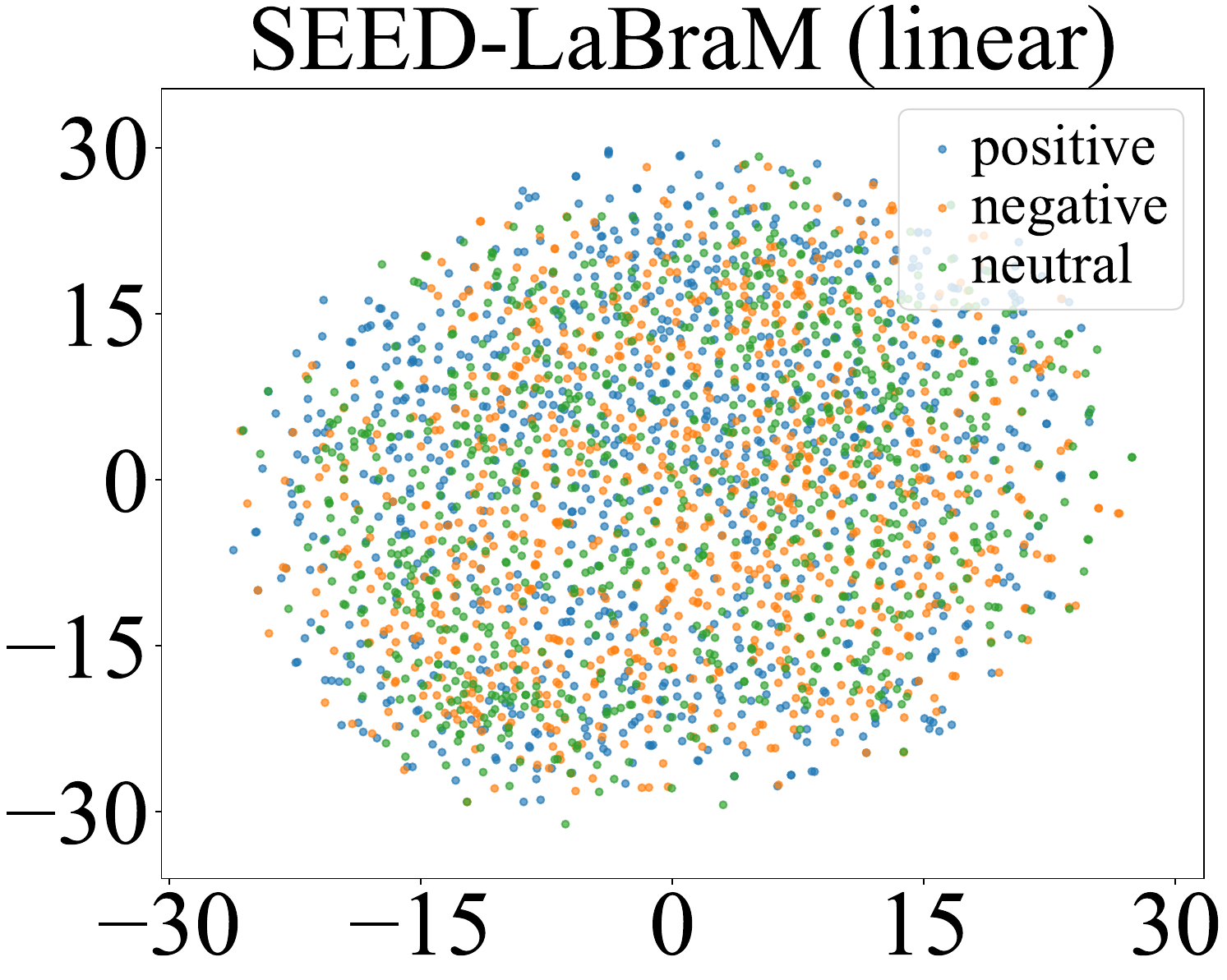}}
\subfigure[]{\includegraphics[width=0.19\linewidth,clip]{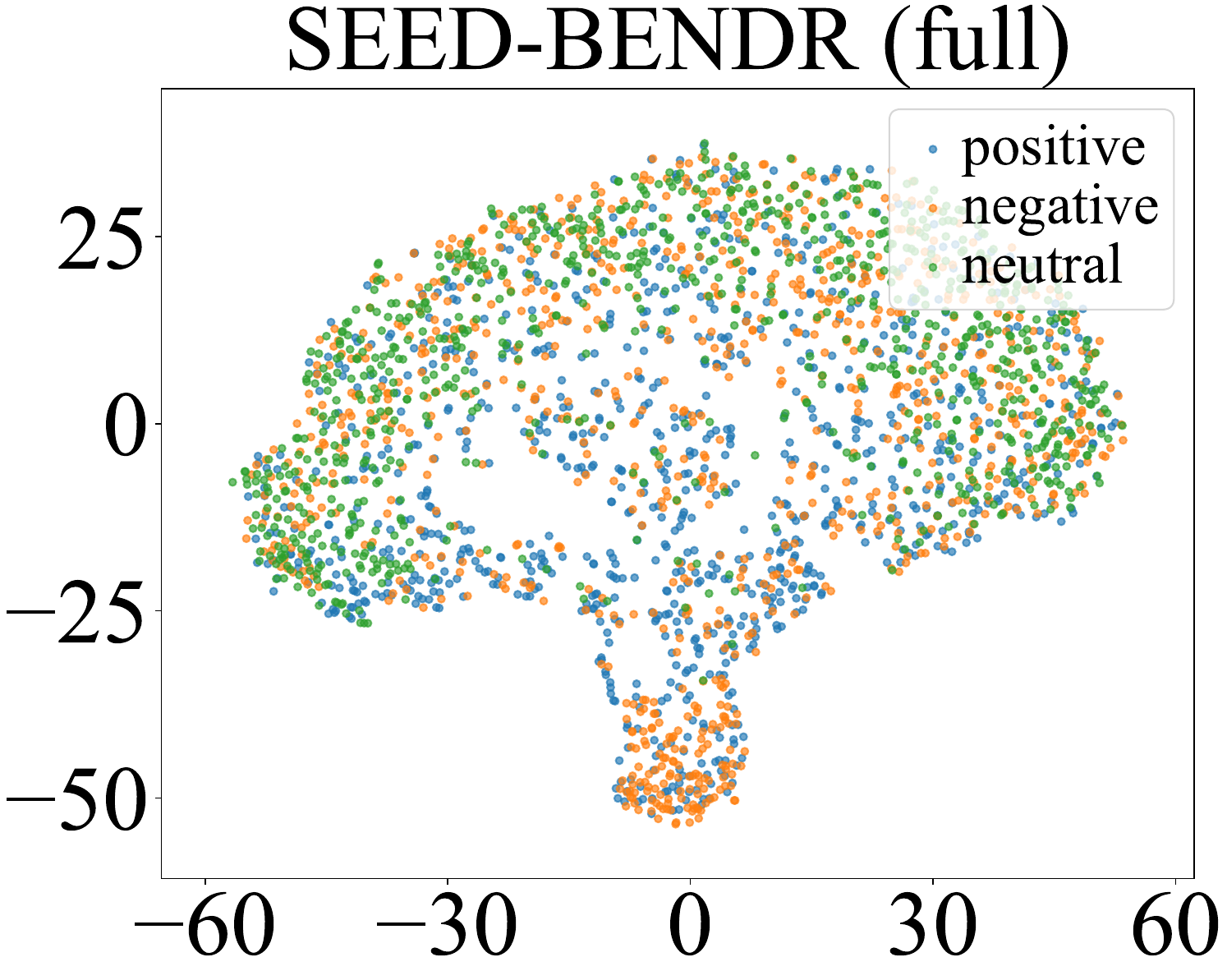}}
\subfigure[]{\includegraphics[width=0.19\linewidth,clip]{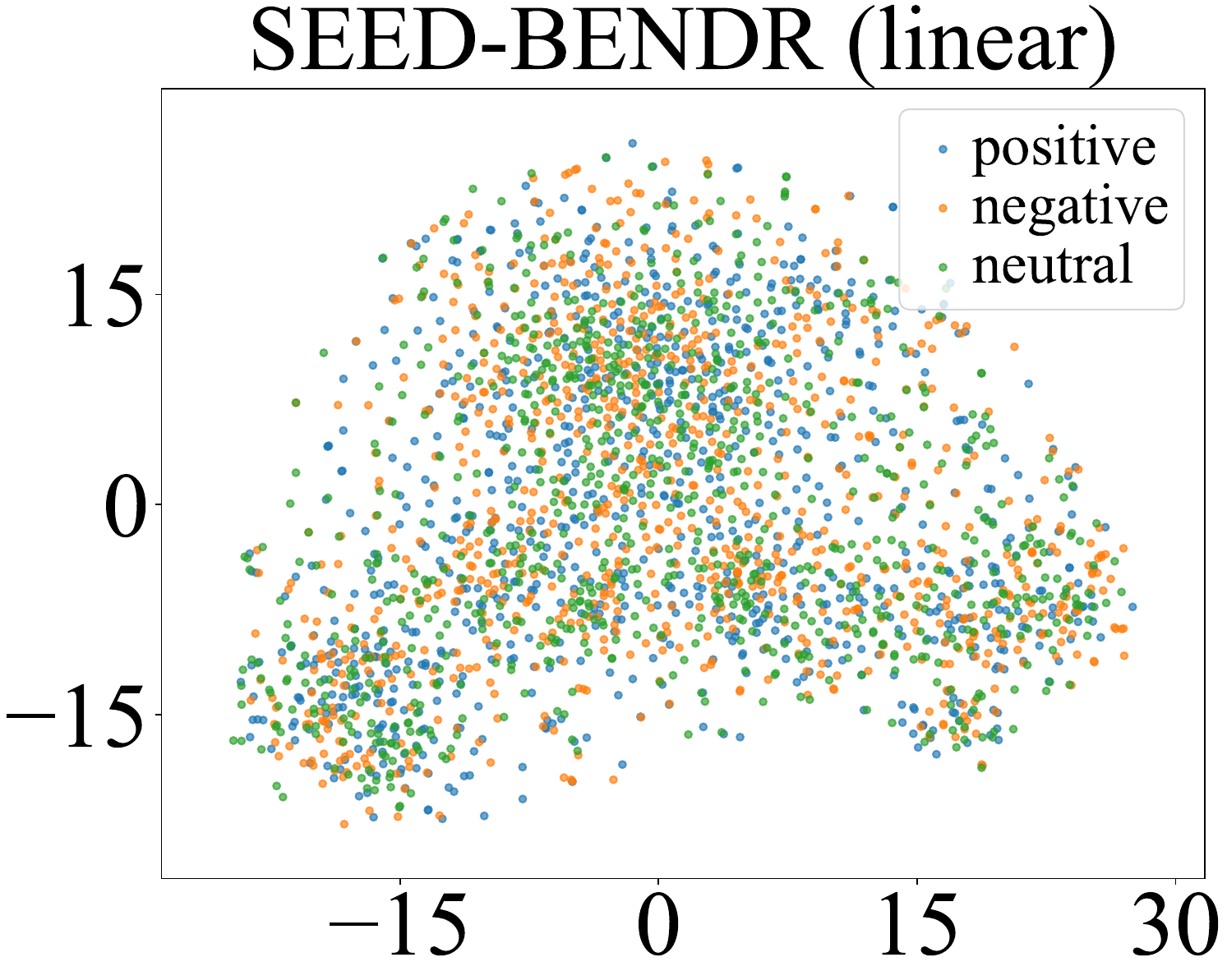}}\\
\subfigure[]{\includegraphics[width=0.19\linewidth,clip]{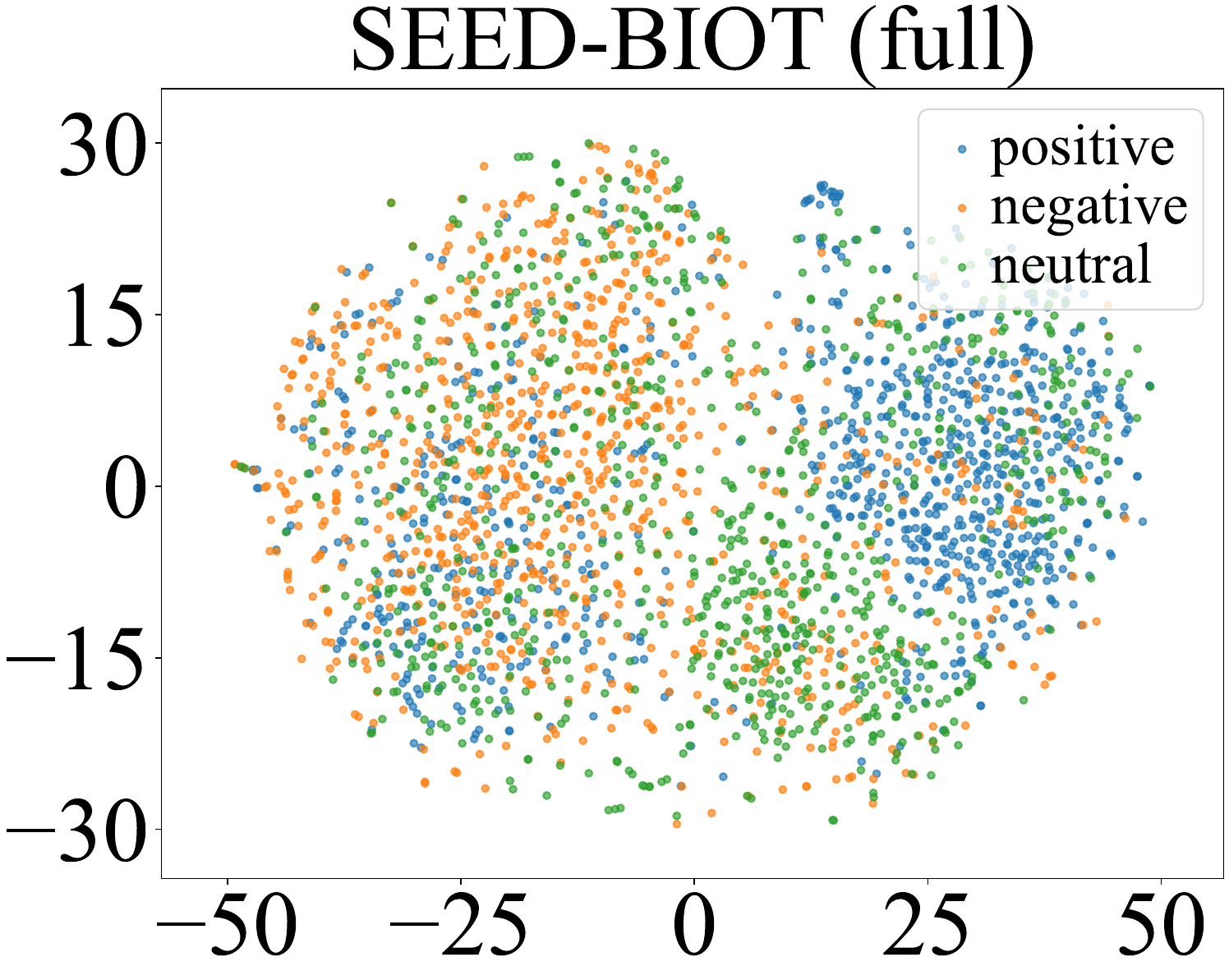}}
\subfigure[]{\includegraphics[width=0.19\linewidth,clip]{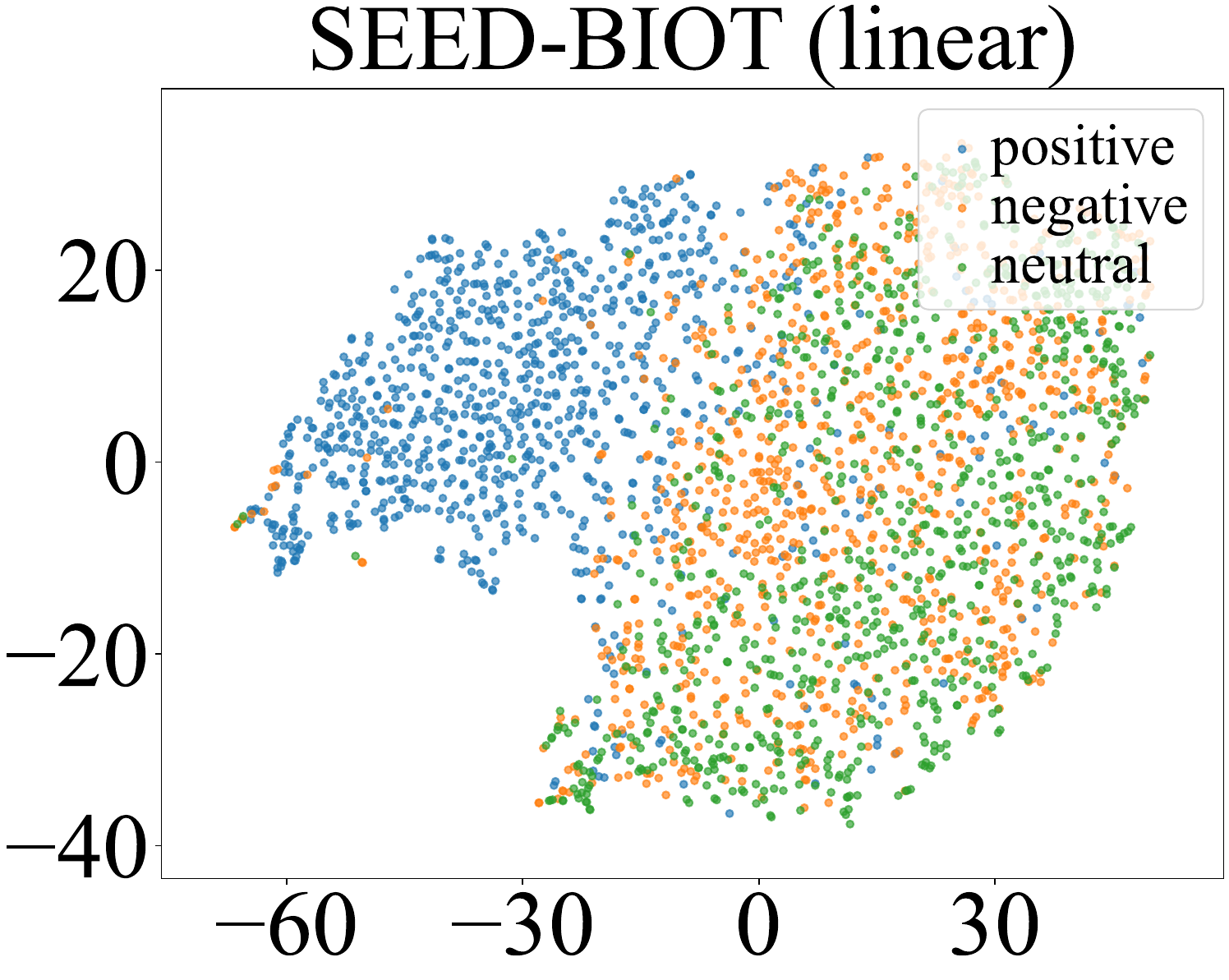}}
\subfigure[]{\includegraphics[width=0.19\linewidth,clip]{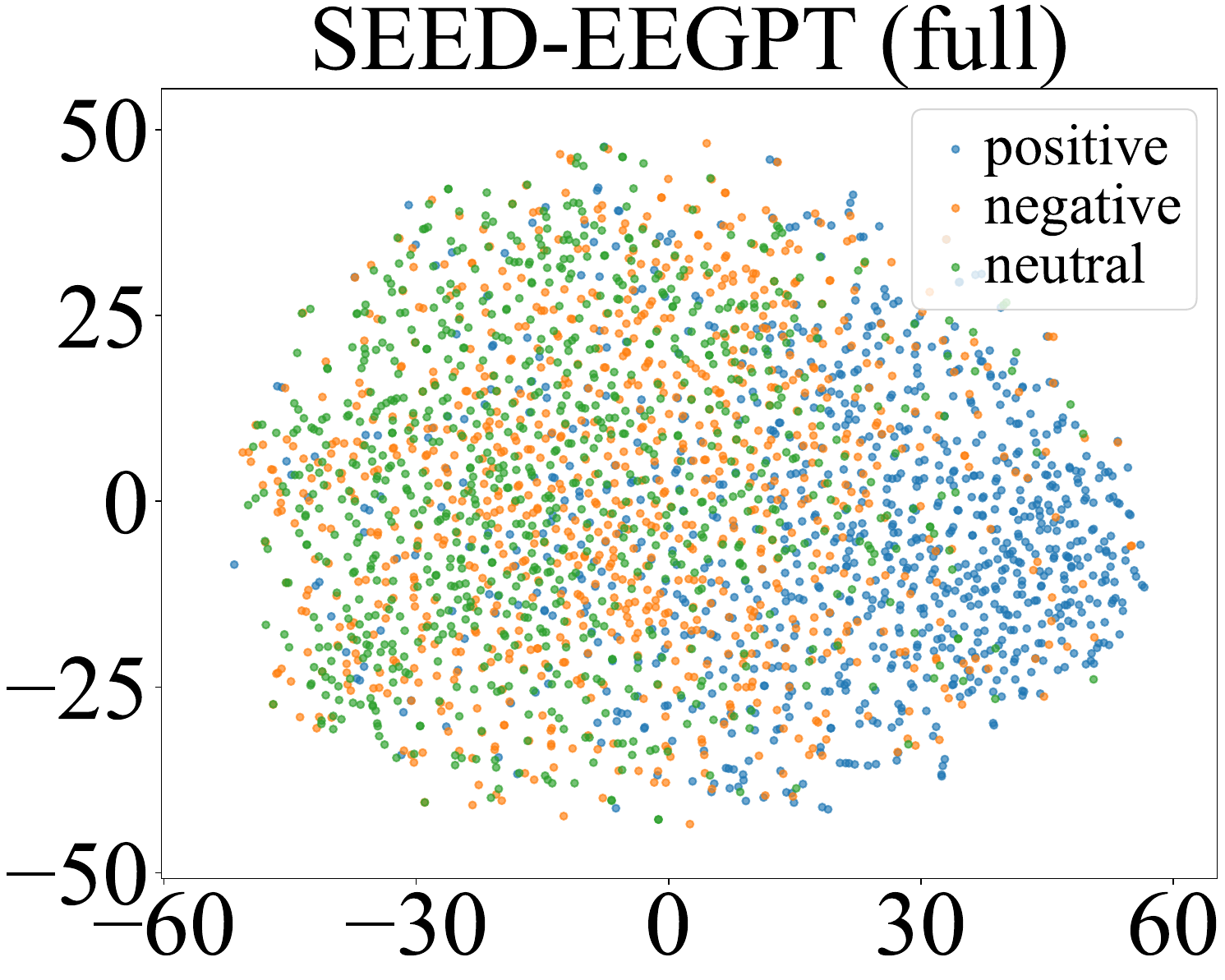}}
\subfigure[]{\includegraphics[width=0.19\linewidth,clip]{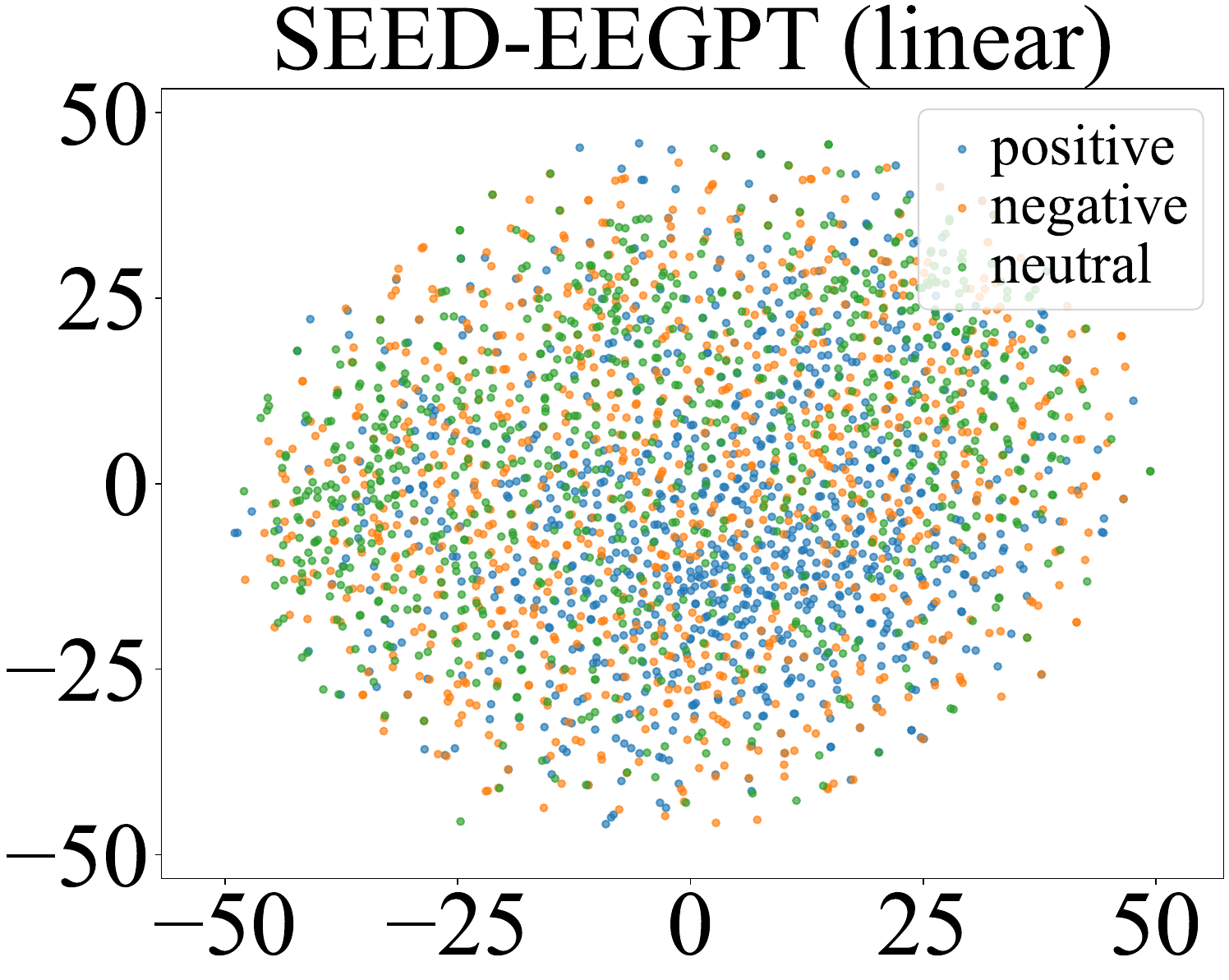}}
\subfigure[]{\includegraphics[width=0.19\linewidth,clip]{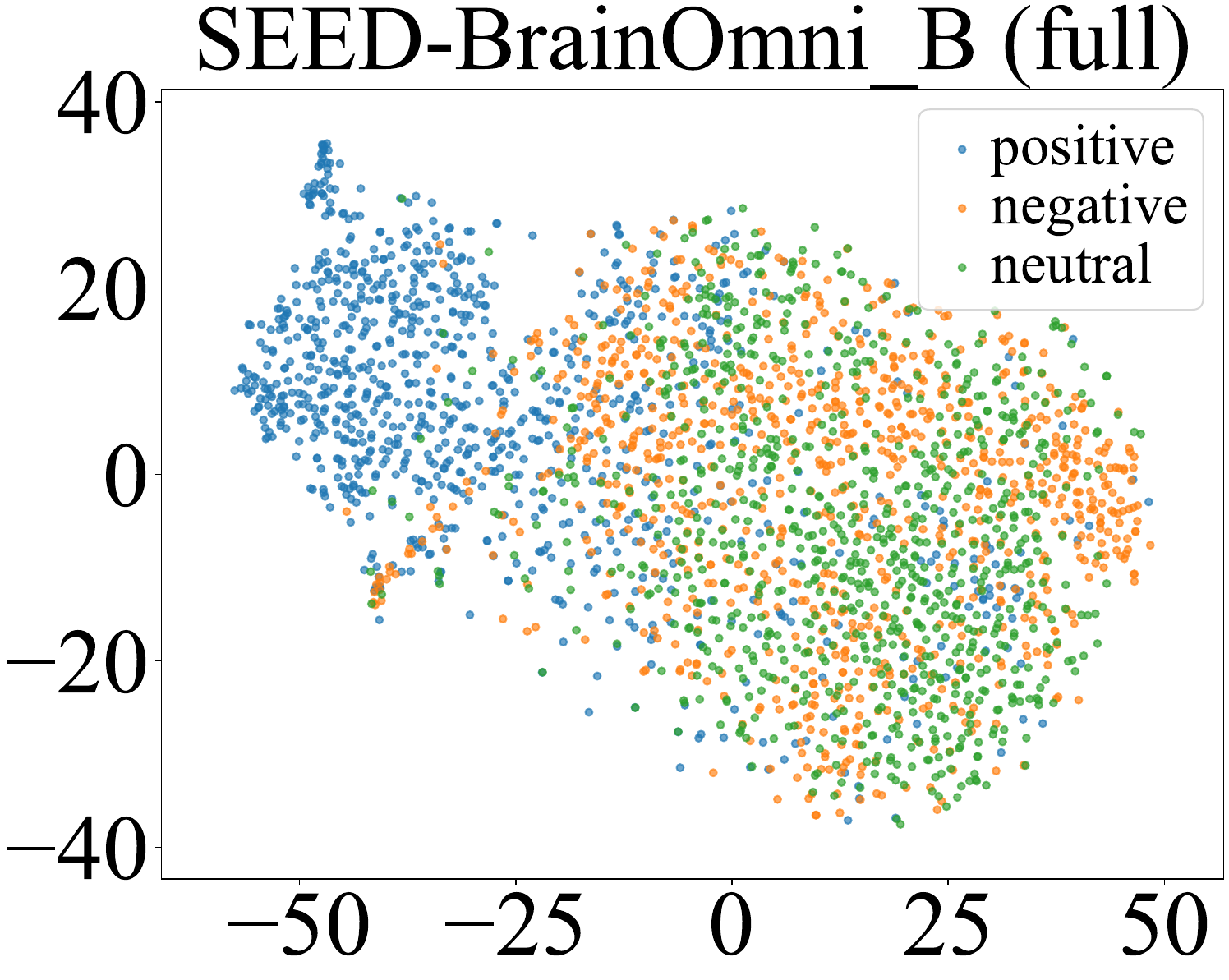}} \\
\subfigure[]{\includegraphics[width=0.19\linewidth,clip]{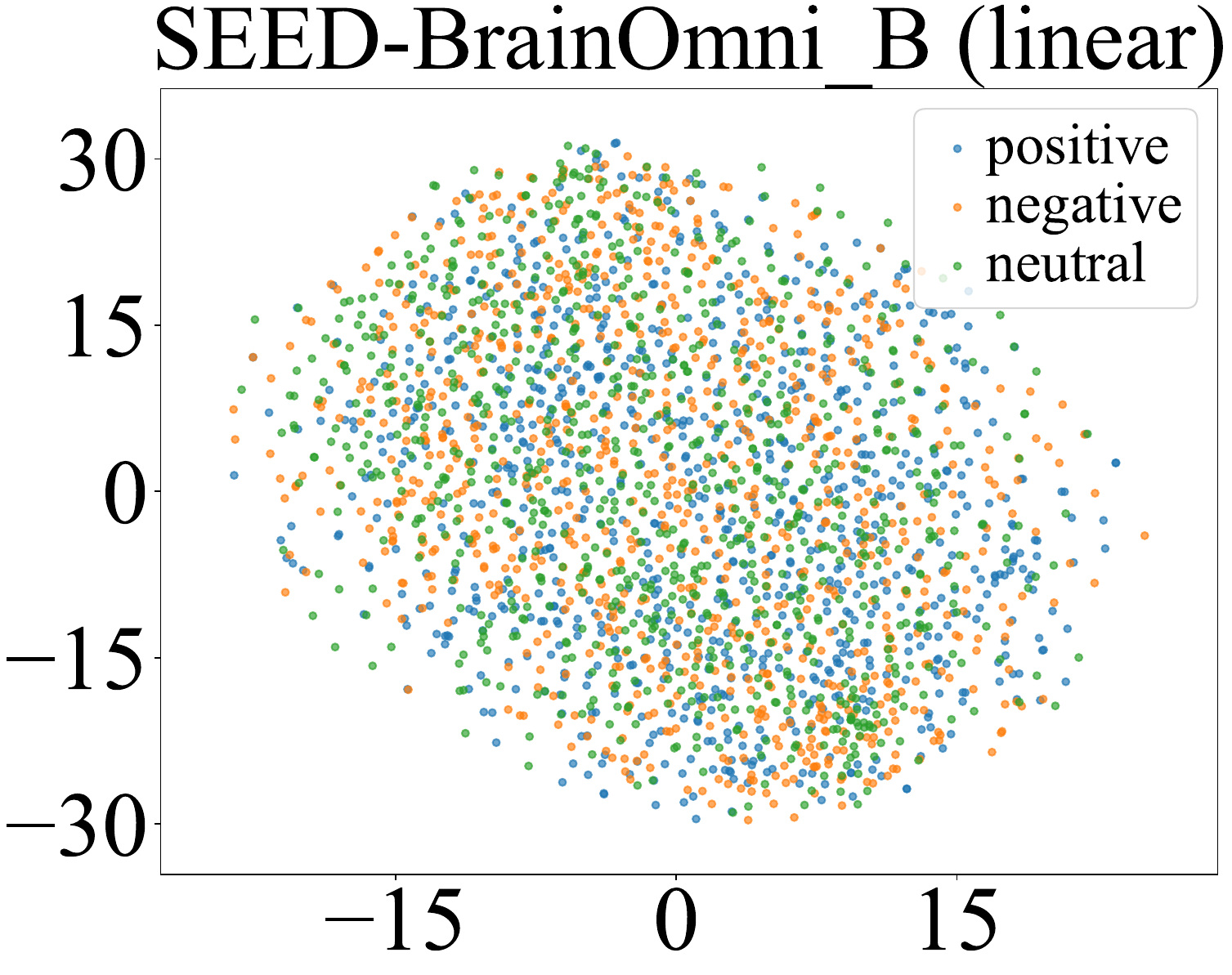}}
\subfigure[]{\includegraphics[width=0.19\linewidth,clip]{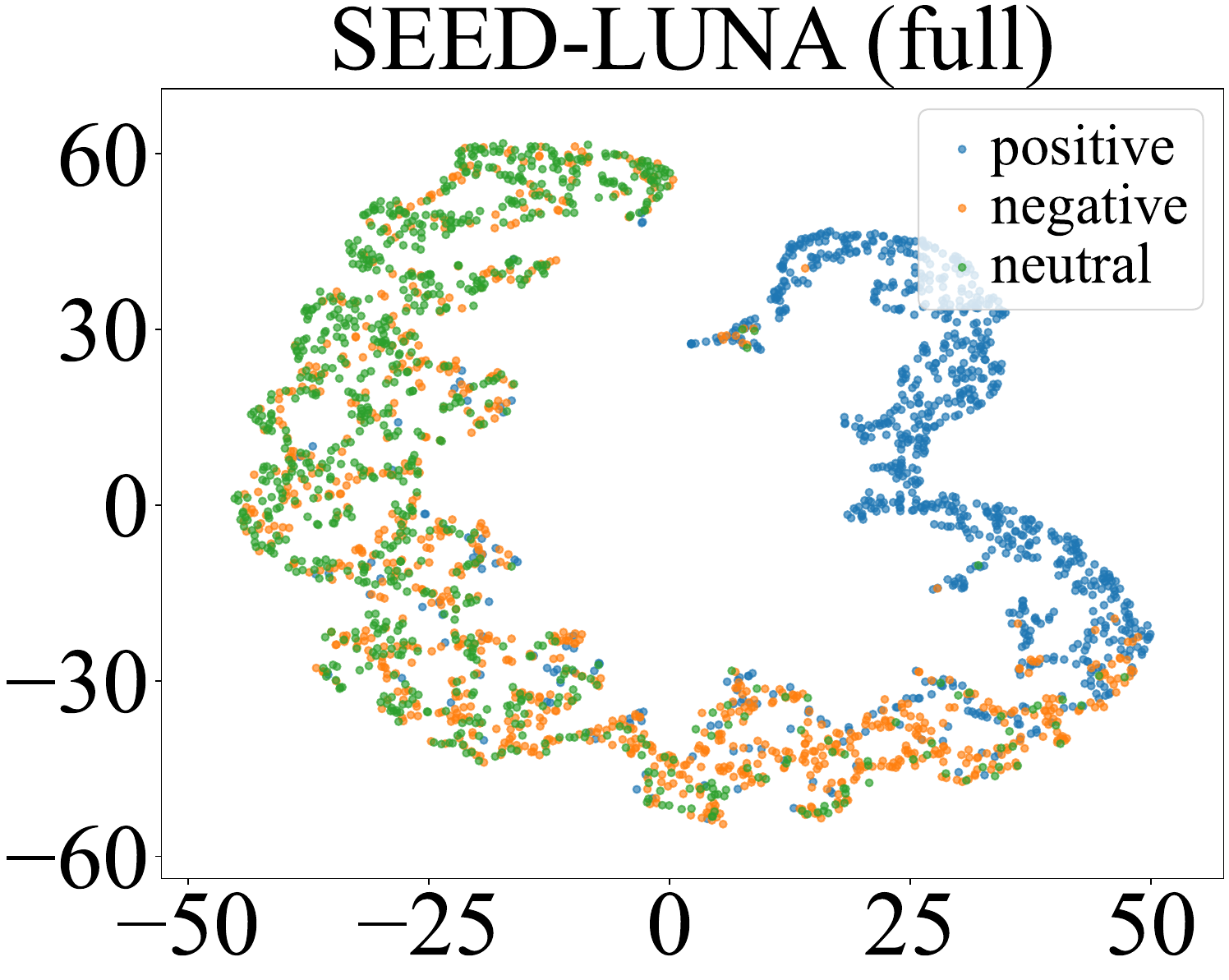}}
\subfigure[]{\includegraphics[width=0.19\linewidth,clip]{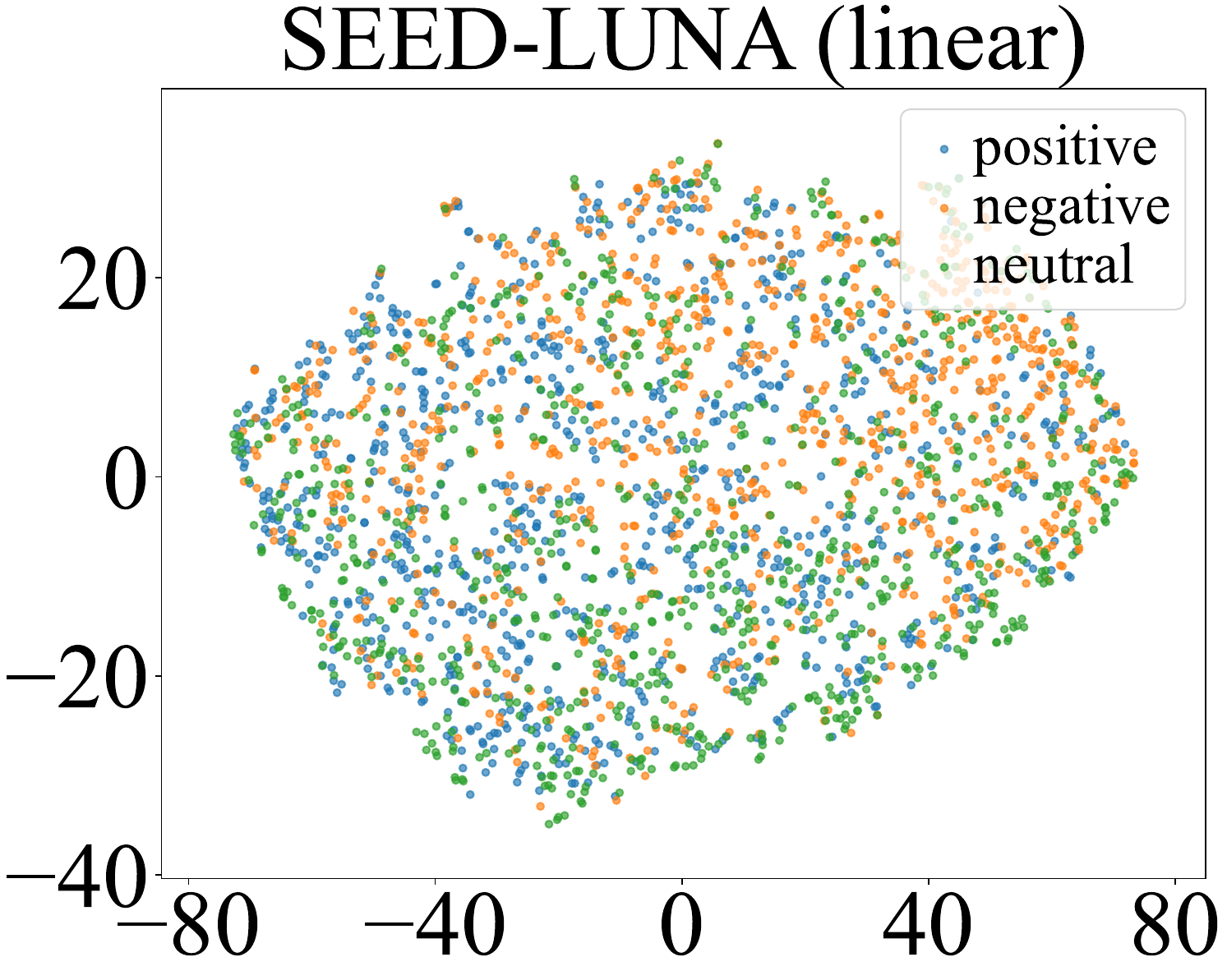}}
\subfigure[]{\includegraphics[width=0.19\linewidth,clip]{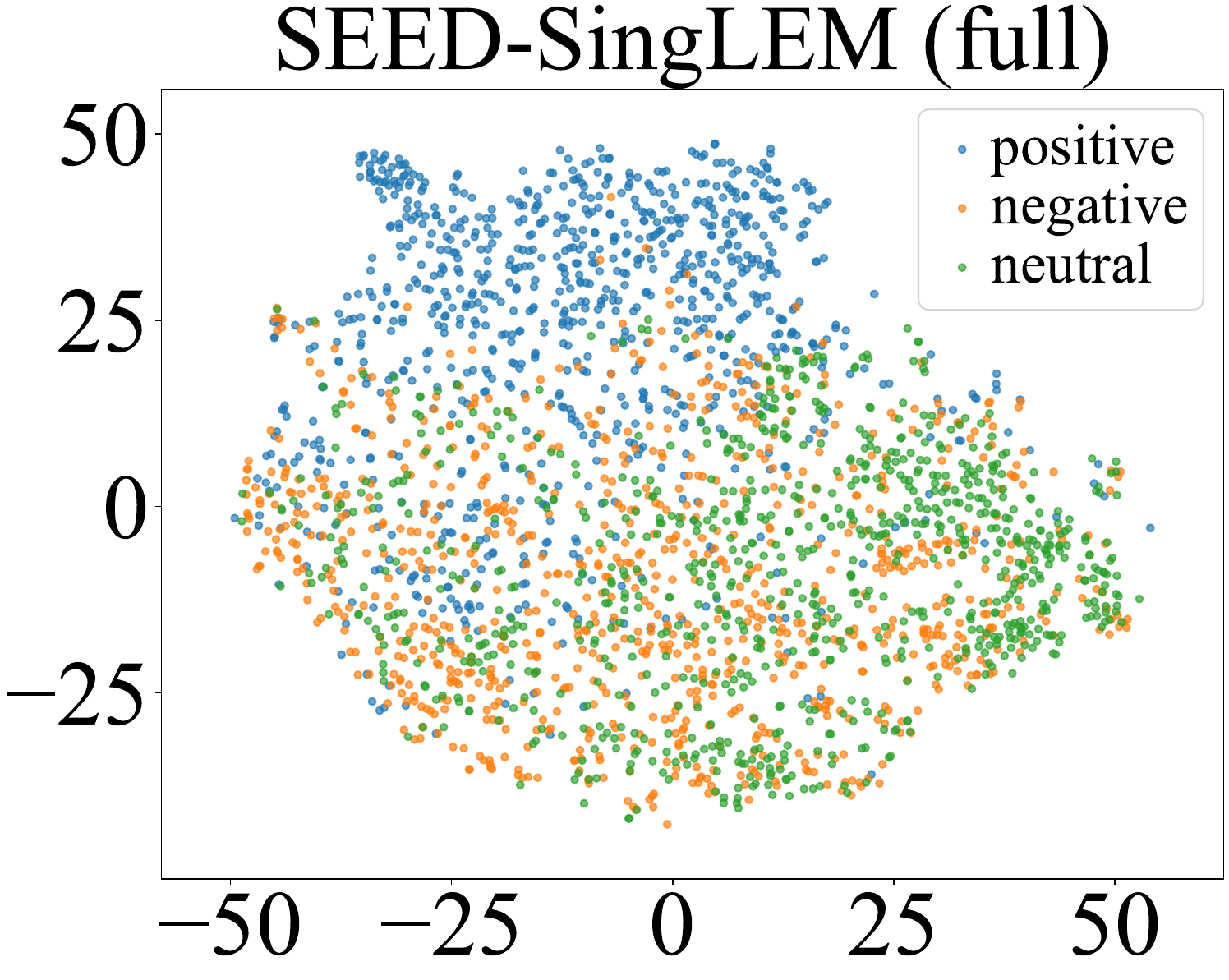}}
\subfigure[]{\includegraphics[width=0.19\linewidth,clip]{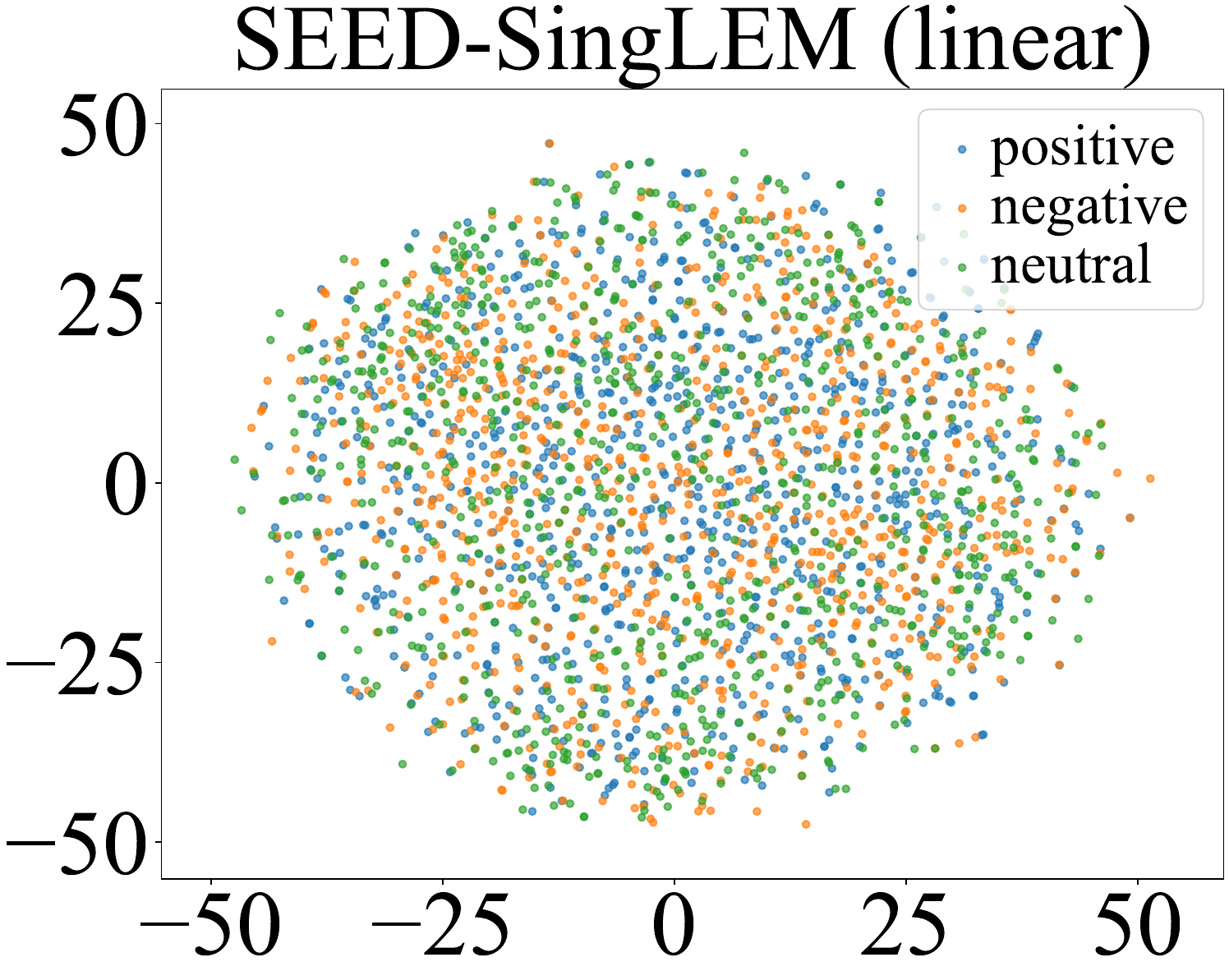}} \\
\caption{$t$-SNE visualization of the SEED datasets.} \label{fig:tsne_rep1_appd}
\end{figure*}

\begin{figure*}[t]
\centering
\subfigure[]{\includegraphics[width=0.19\linewidth,clip]{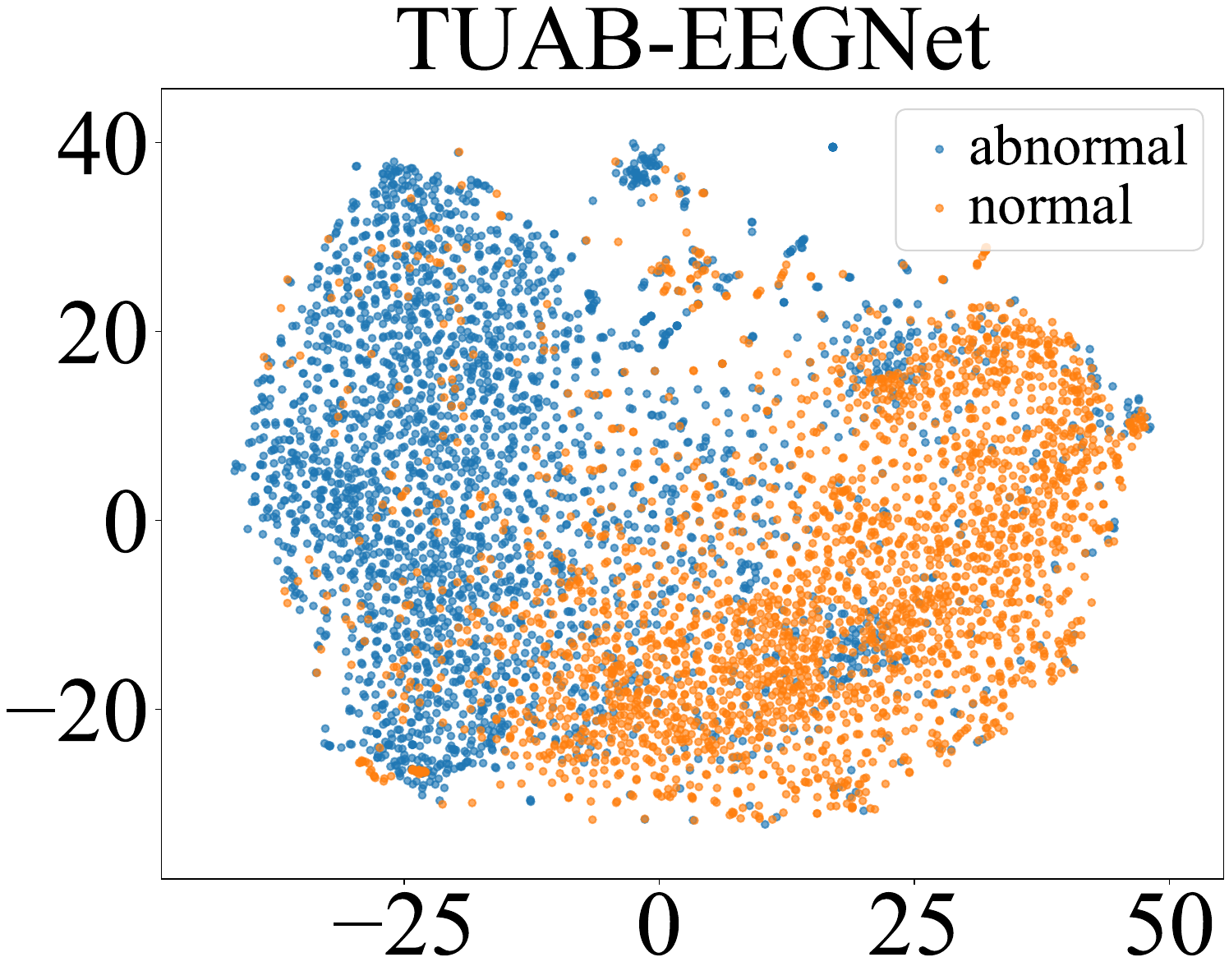}}
\subfigure[]{\includegraphics[width=0.19\linewidth,clip]{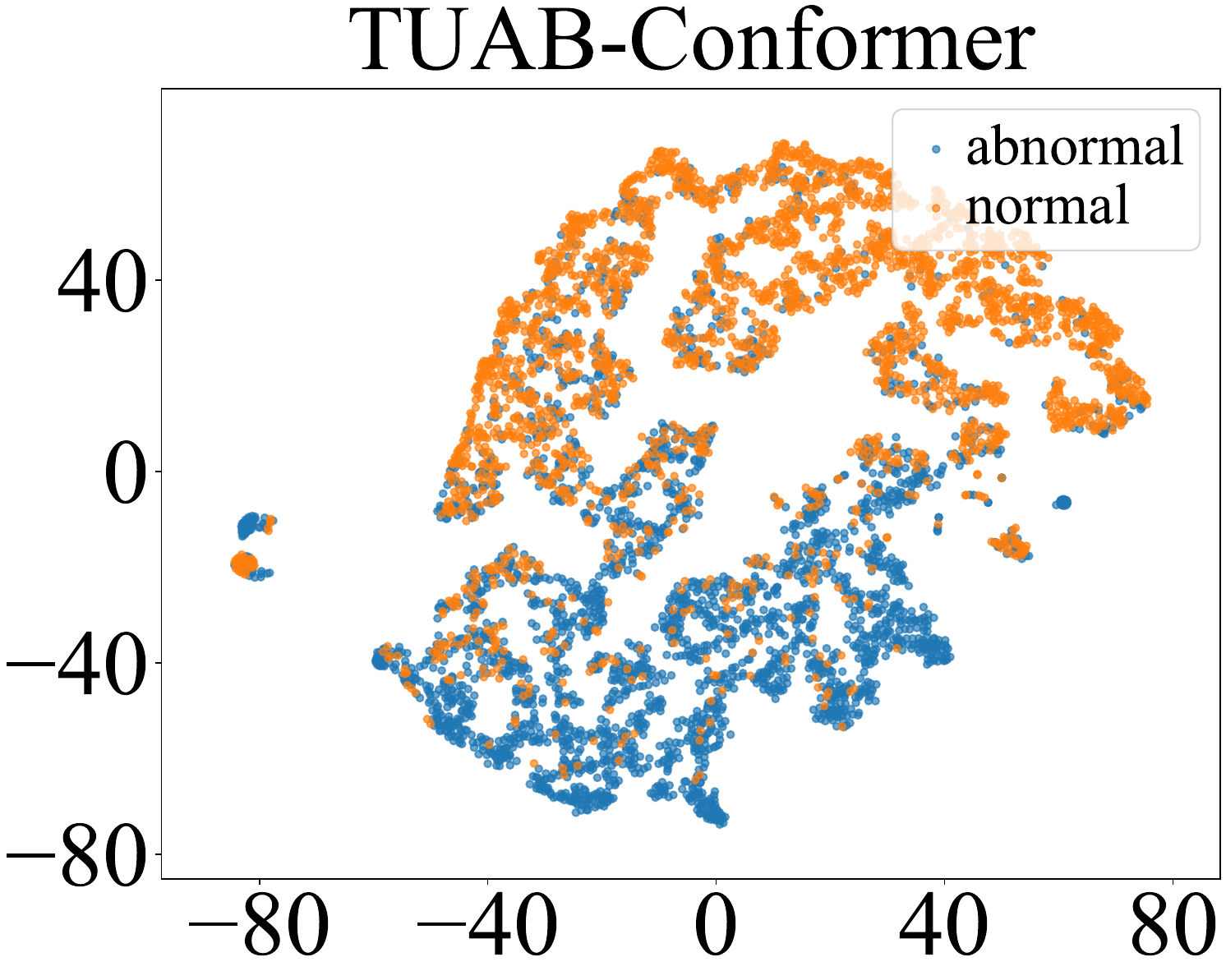}}
\subfigure[]{\includegraphics[width=0.19\linewidth,clip]{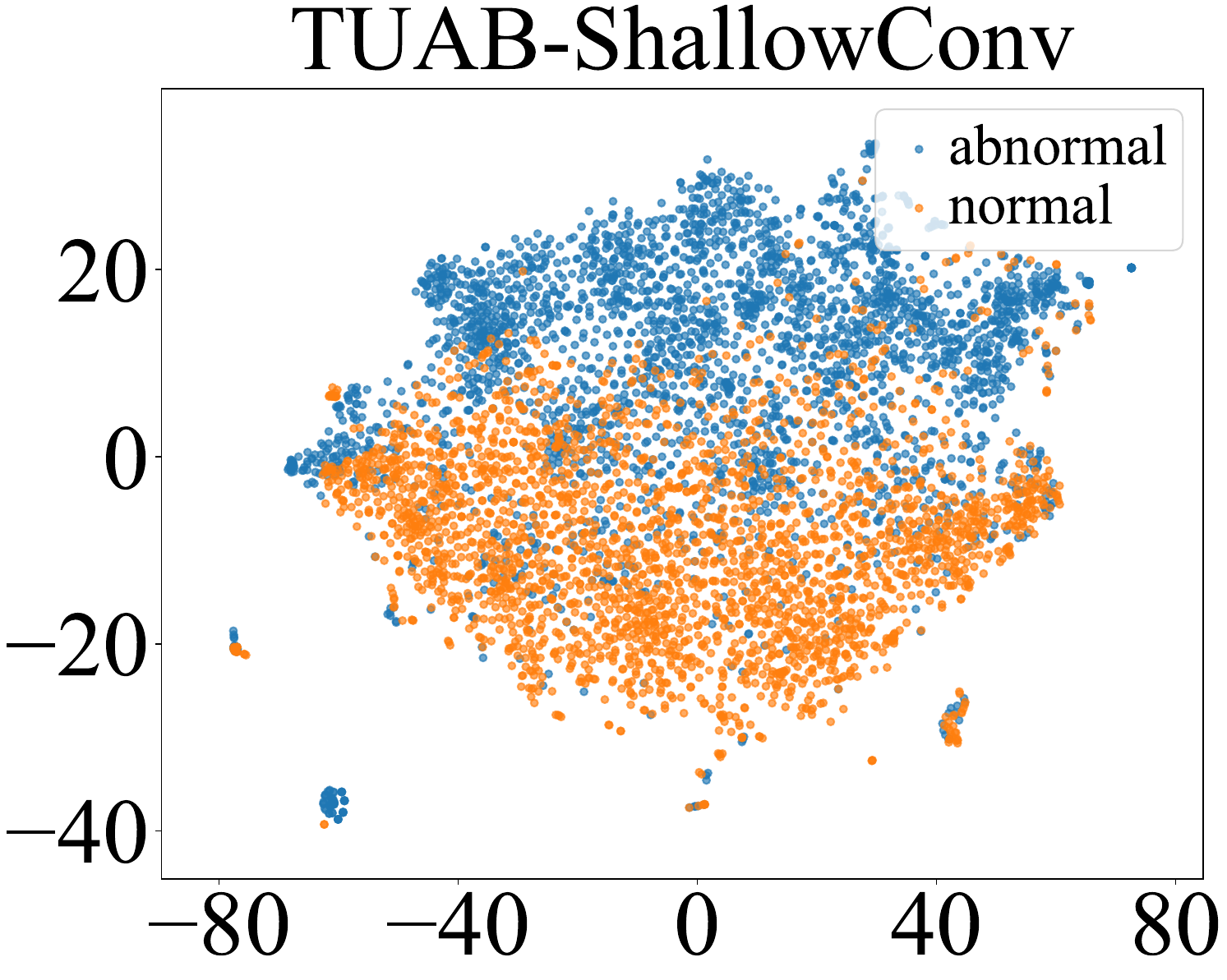}}
\subfigure[]{\includegraphics[width=0.19\linewidth,clip]{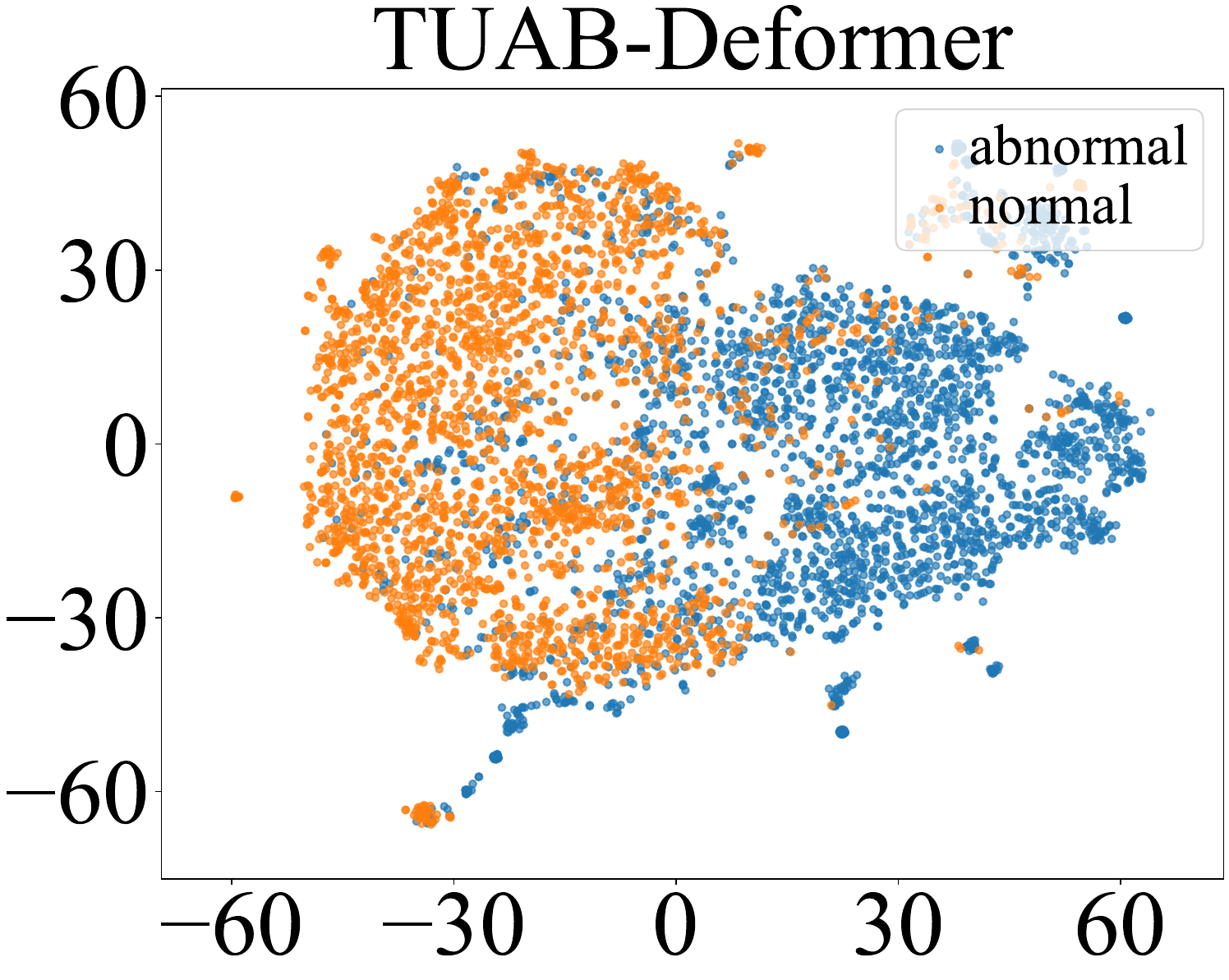}}
\subfigure[]{\includegraphics[width=0.19\linewidth,clip]{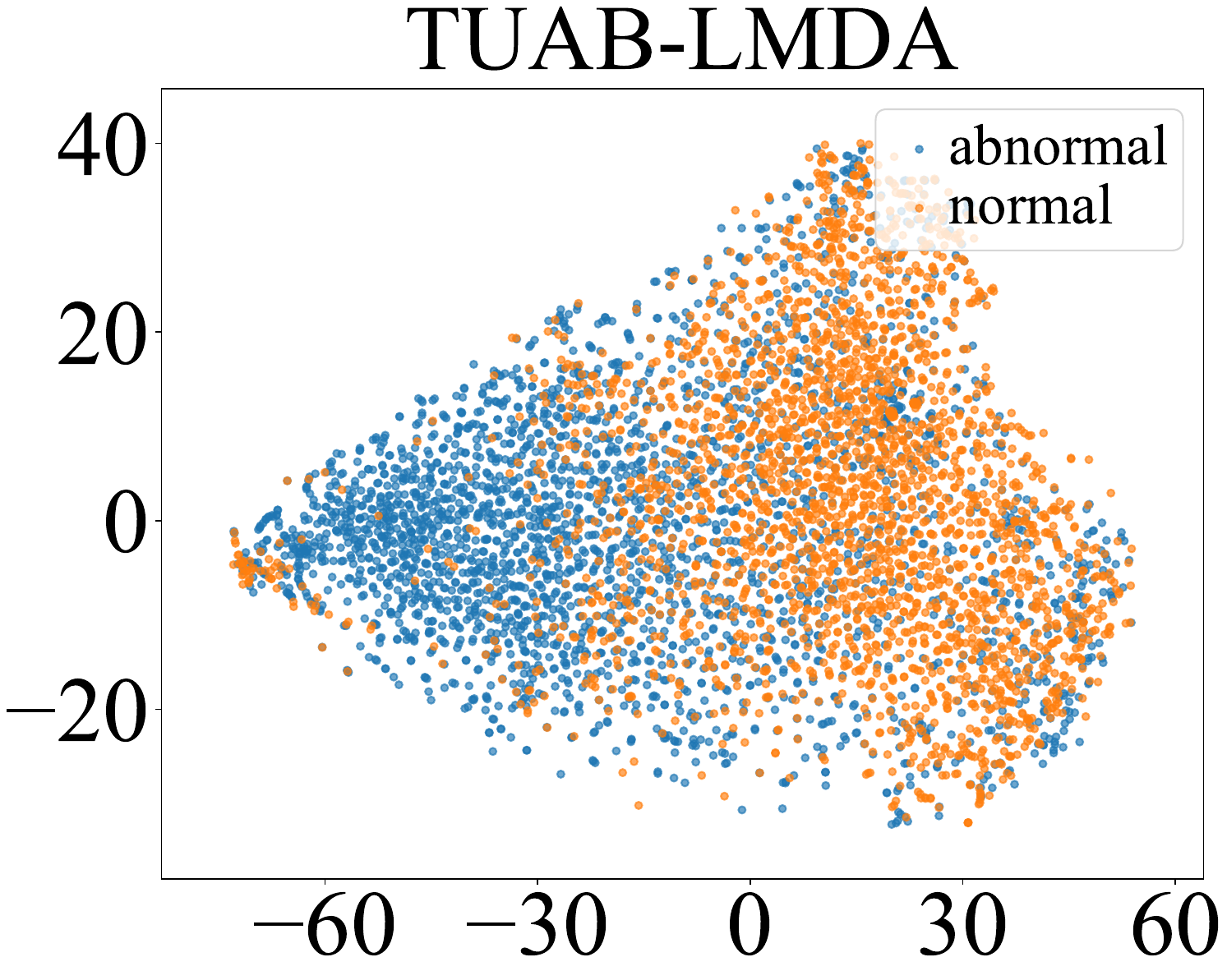}}\\
\subfigure[]{\includegraphics[width=0.19\linewidth,clip]{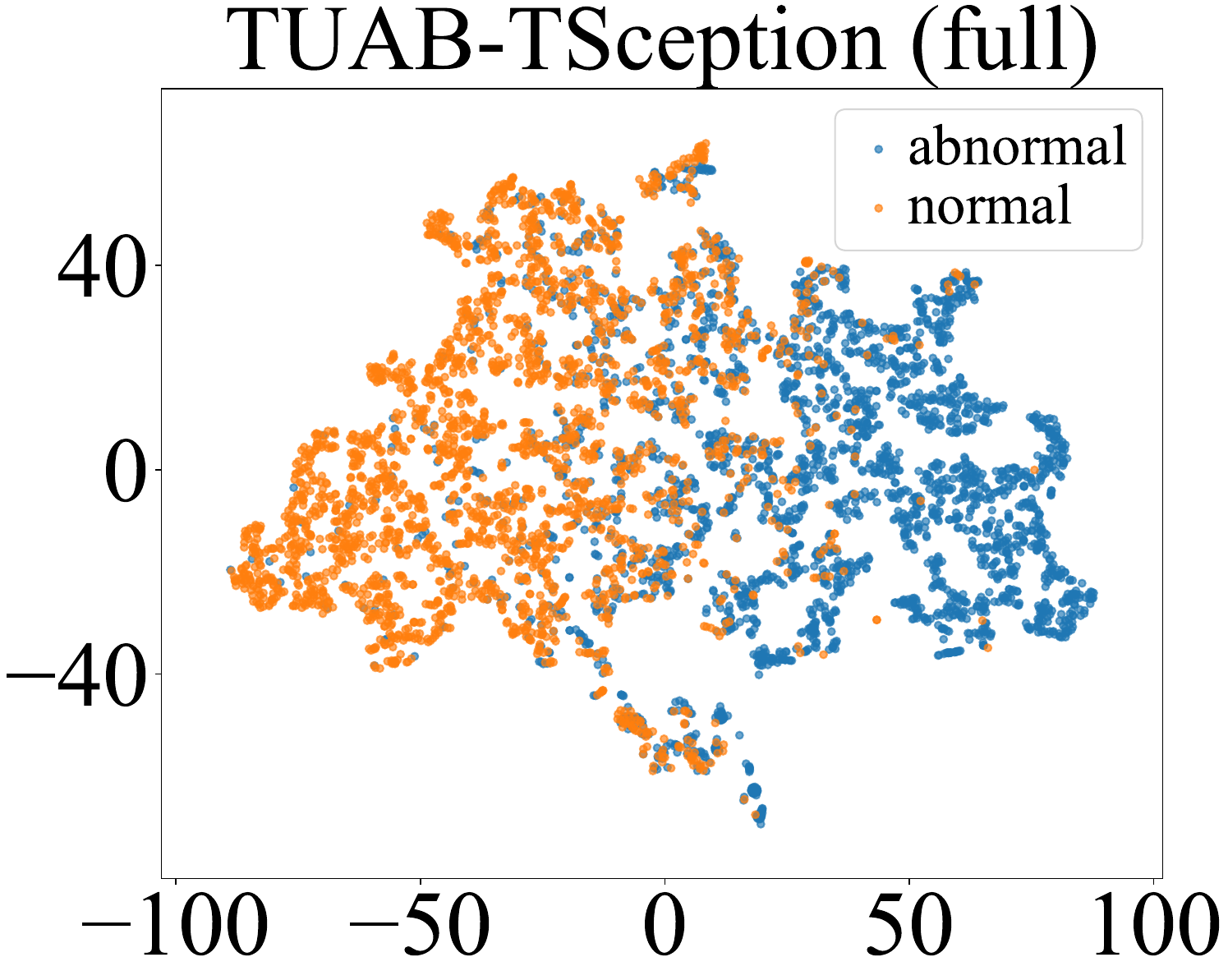}}
\subfigure[]{\includegraphics[width=0.19\linewidth,clip]{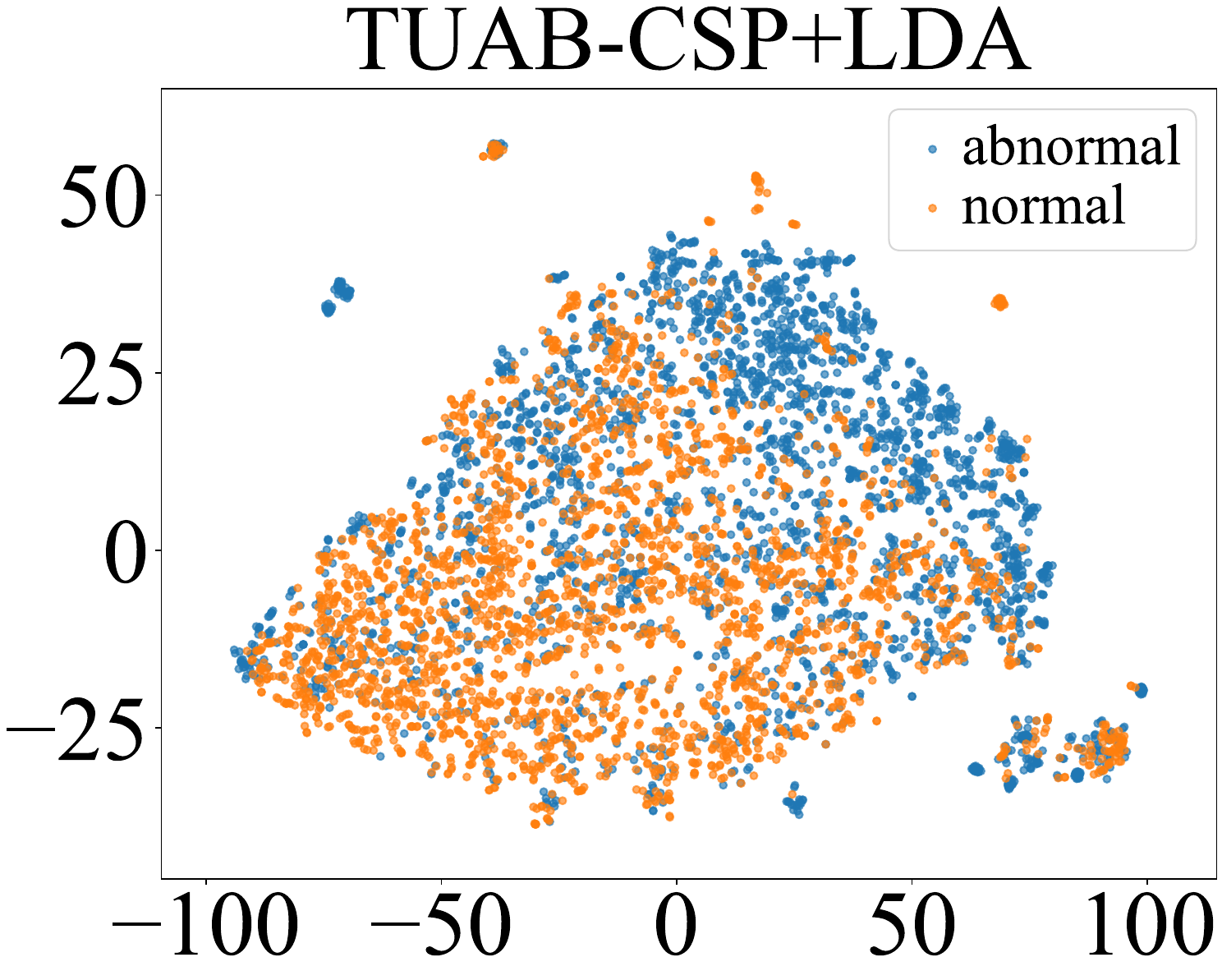}}
\subfigure[]{\includegraphics[width=0.19\linewidth,clip]{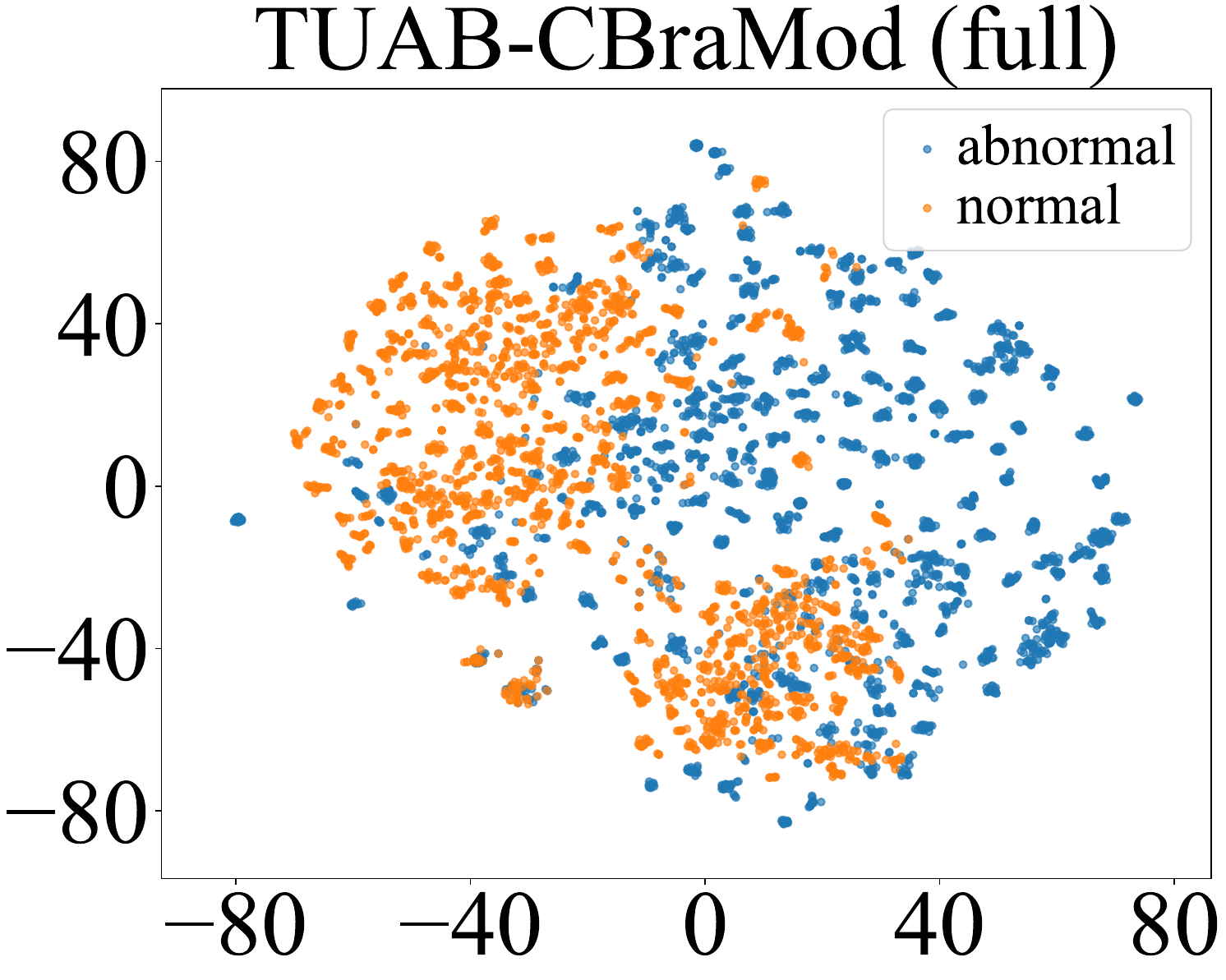}}
\subfigure[]{\includegraphics[width=0.19\linewidth,clip]{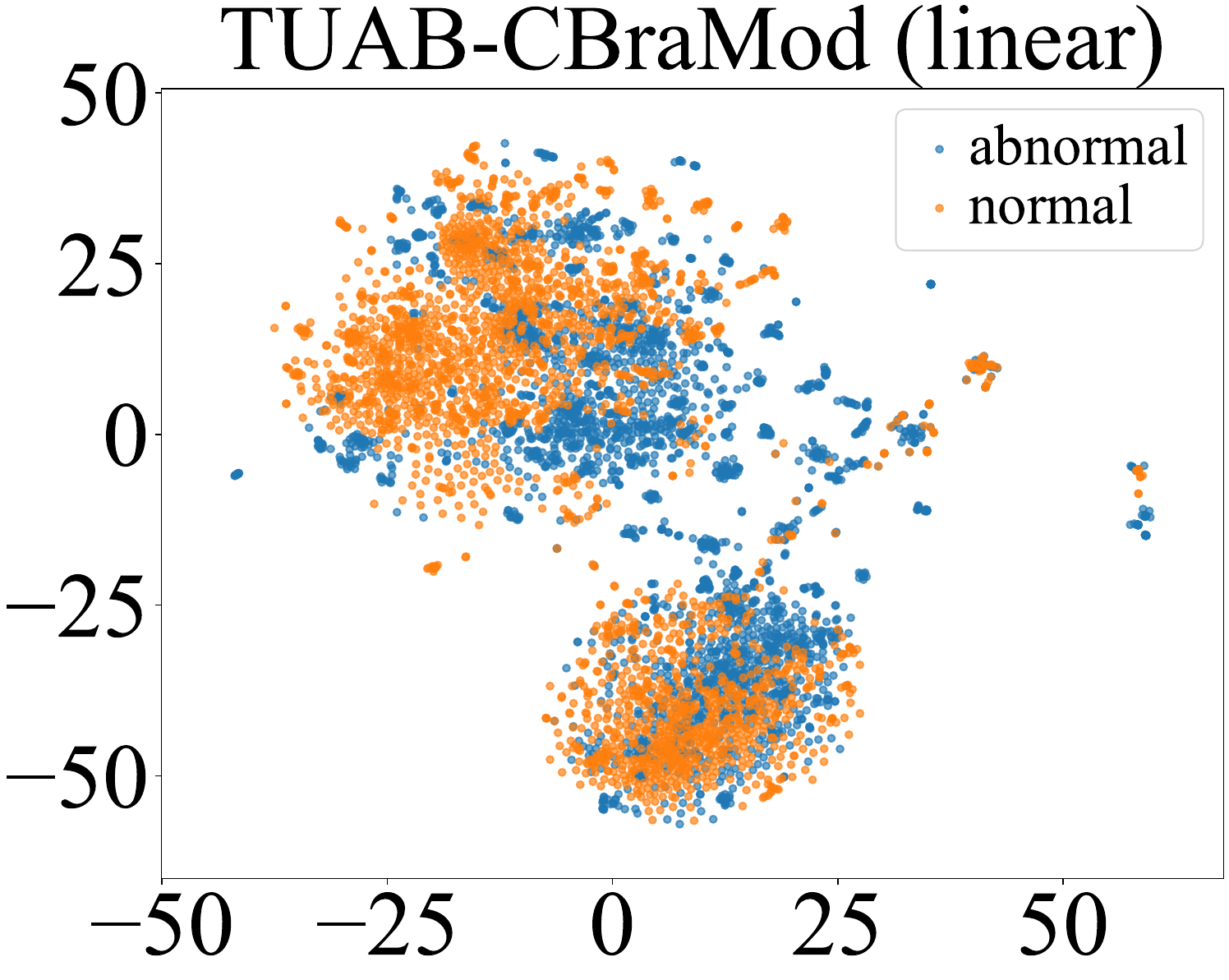}}
\subfigure[]{\includegraphics[width=0.19\linewidth,clip]{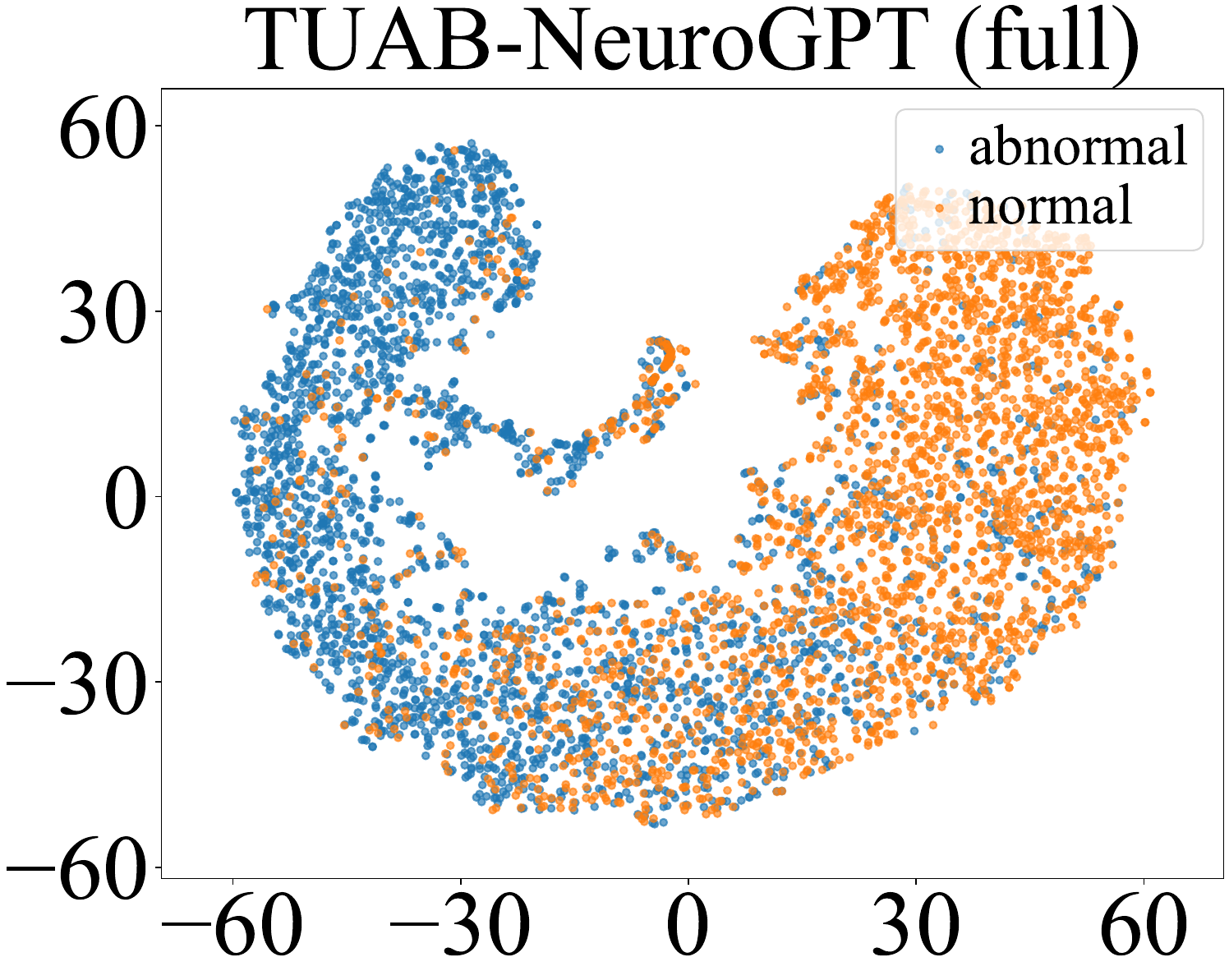}} \\
\subfigure[]{\includegraphics[width=0.19\linewidth,clip]{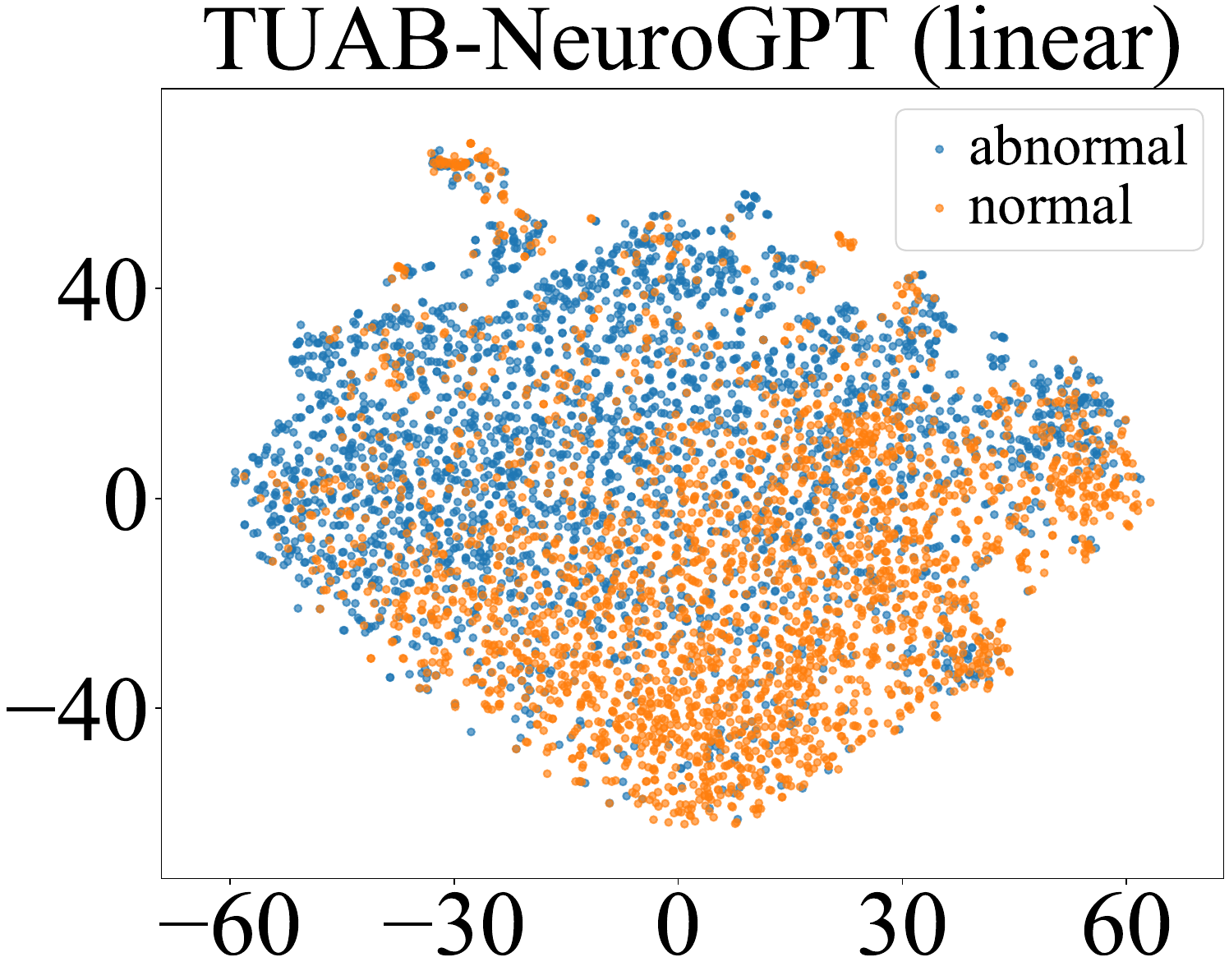}}
\subfigure[]{\includegraphics[width=0.19\linewidth,clip]{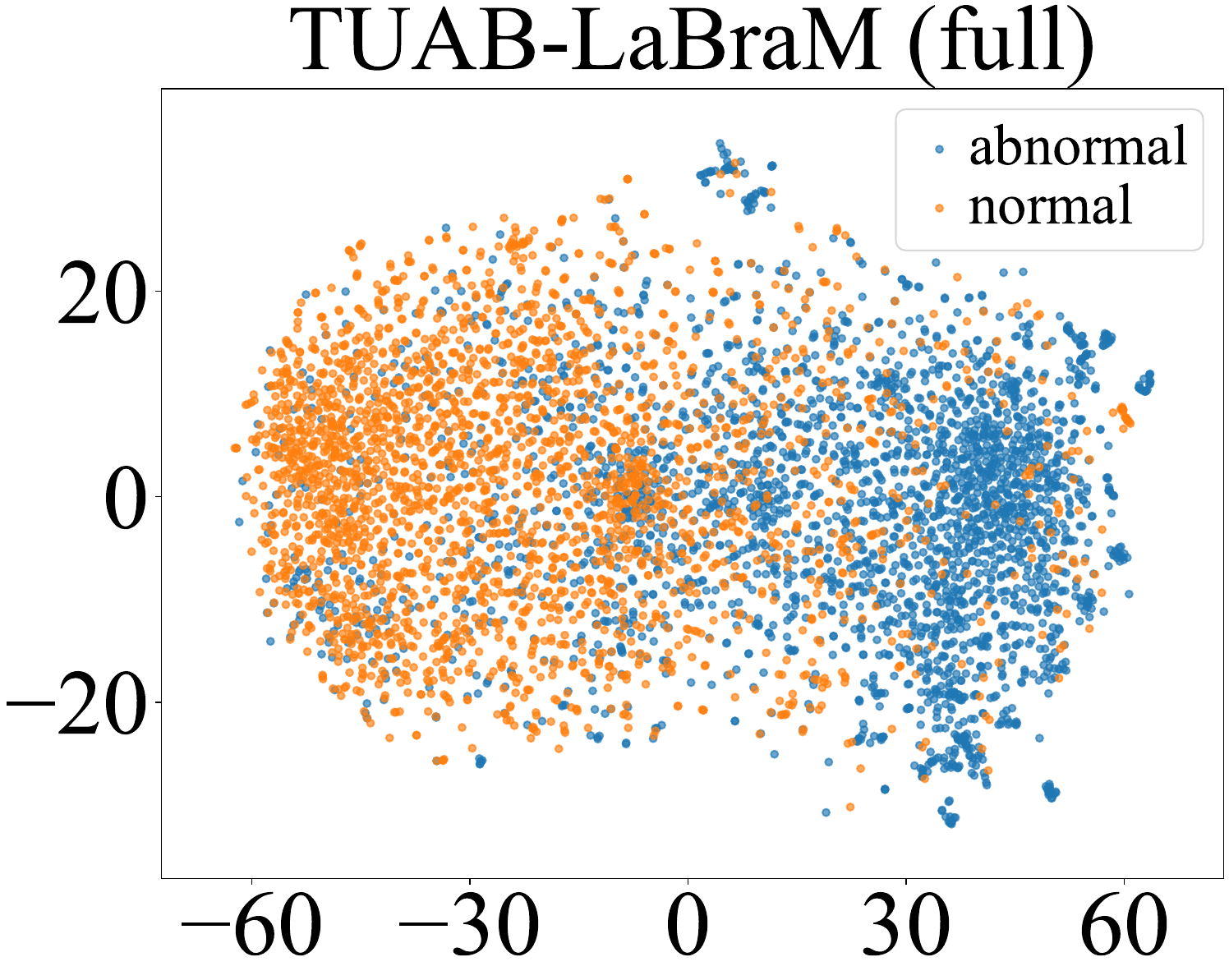}}
\subfigure[]{\includegraphics[width=0.19\linewidth,clip]{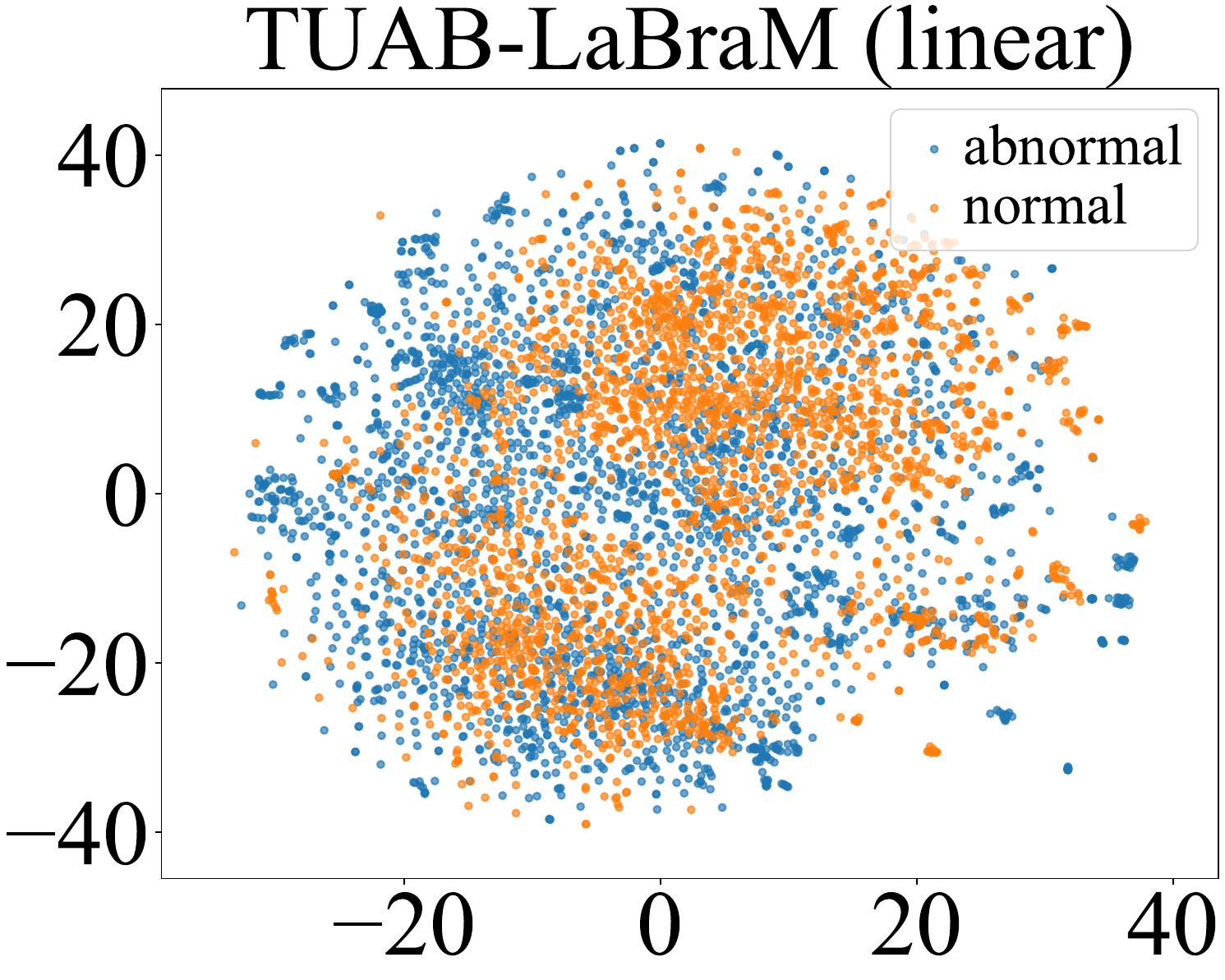}}
\subfigure[]{\includegraphics[width=0.19\linewidth,clip]{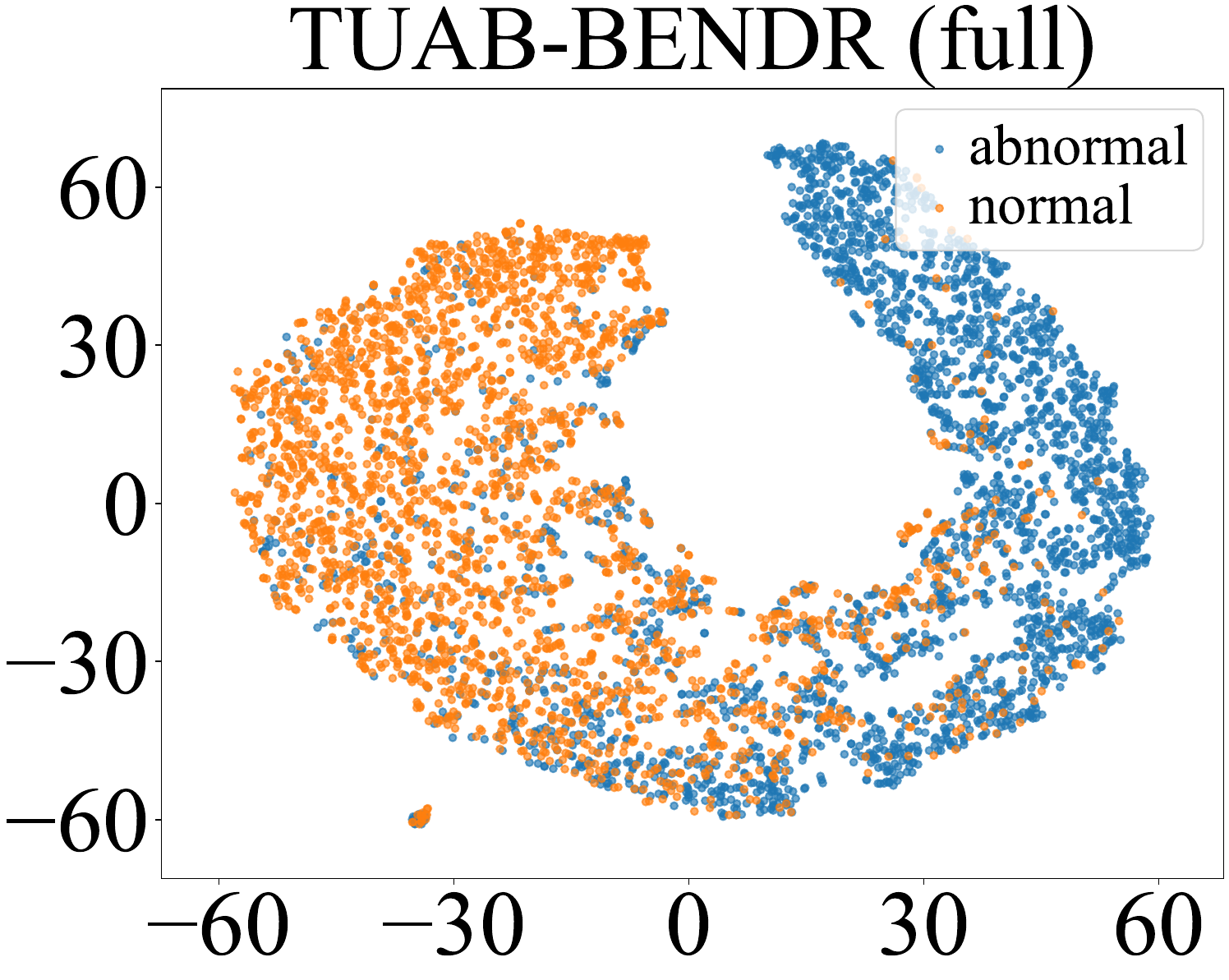}}
\subfigure[]{\includegraphics[width=0.19\linewidth,clip]{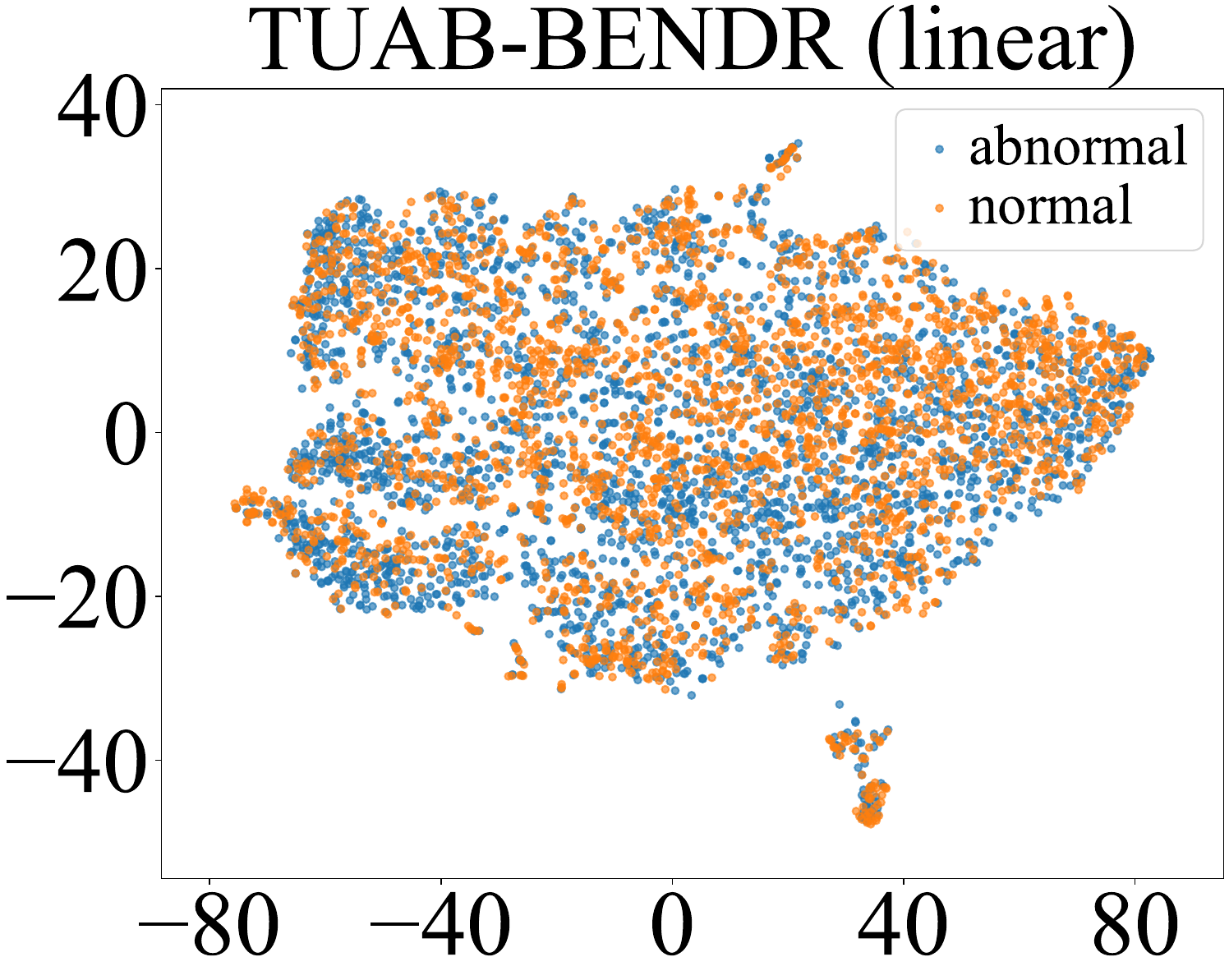}}\\
\subfigure[]{\includegraphics[width=0.19\linewidth,clip]{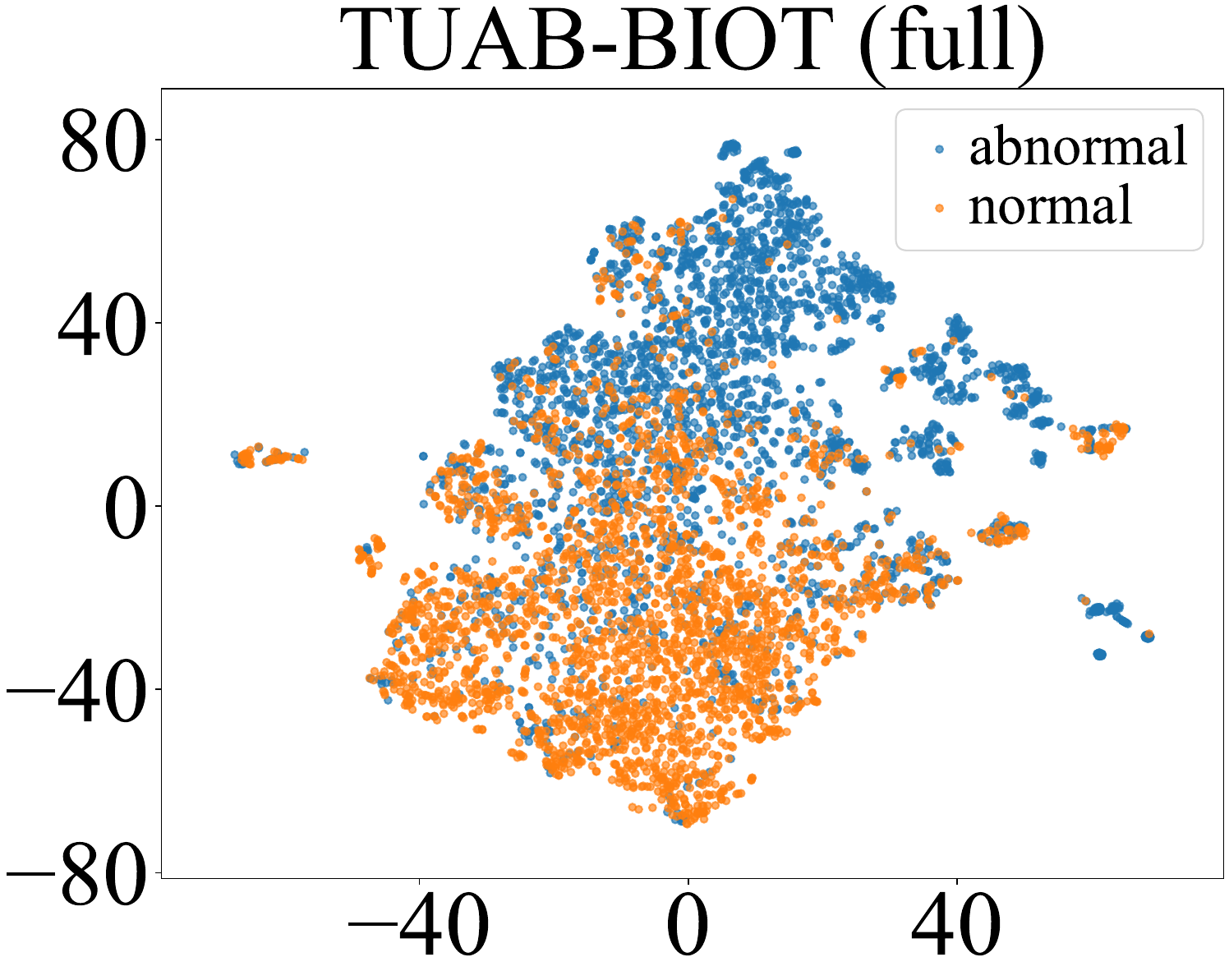}}
\subfigure[]{\includegraphics[width=0.19\linewidth,clip]{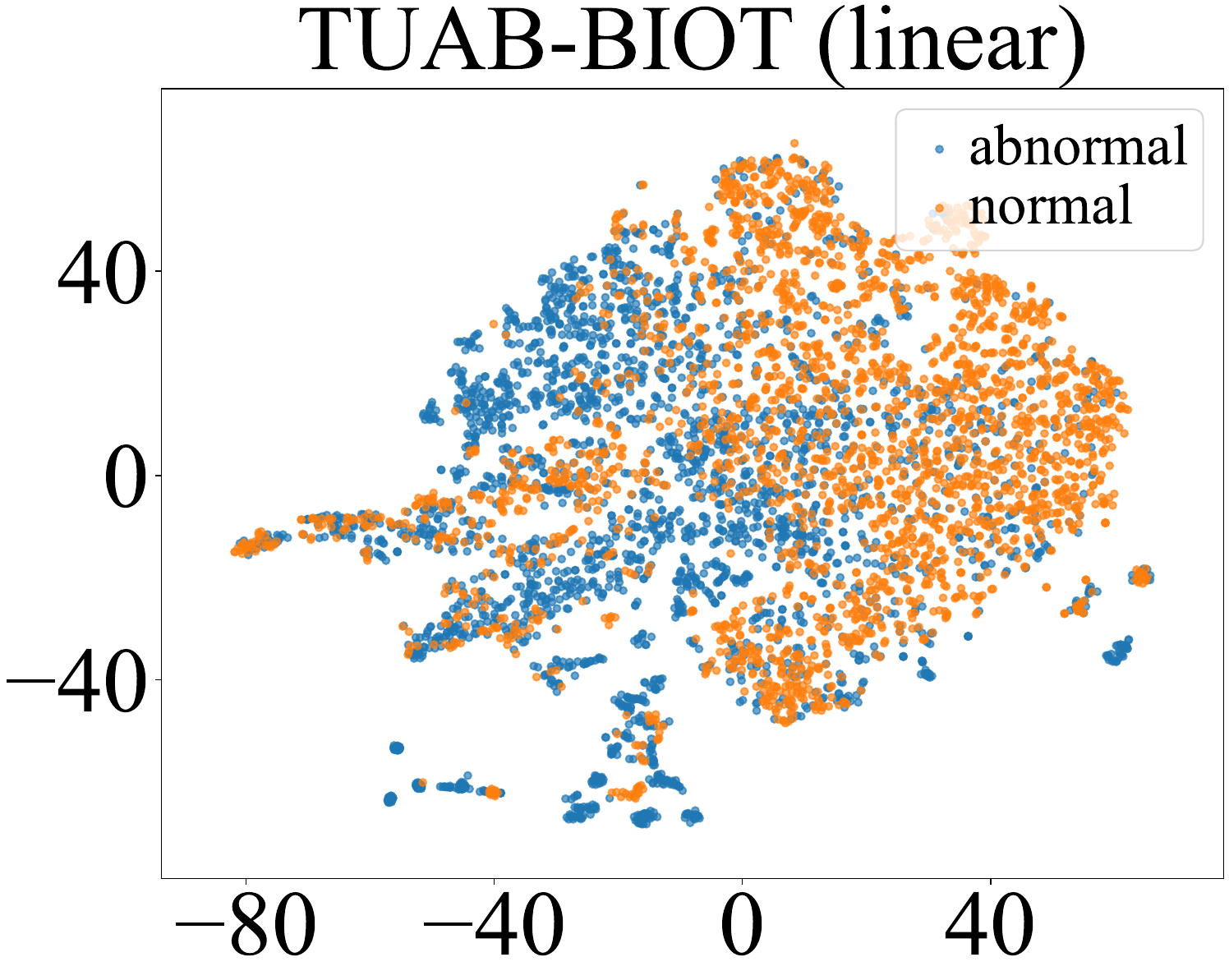}}
\subfigure[]{\includegraphics[width=0.19\linewidth,clip]{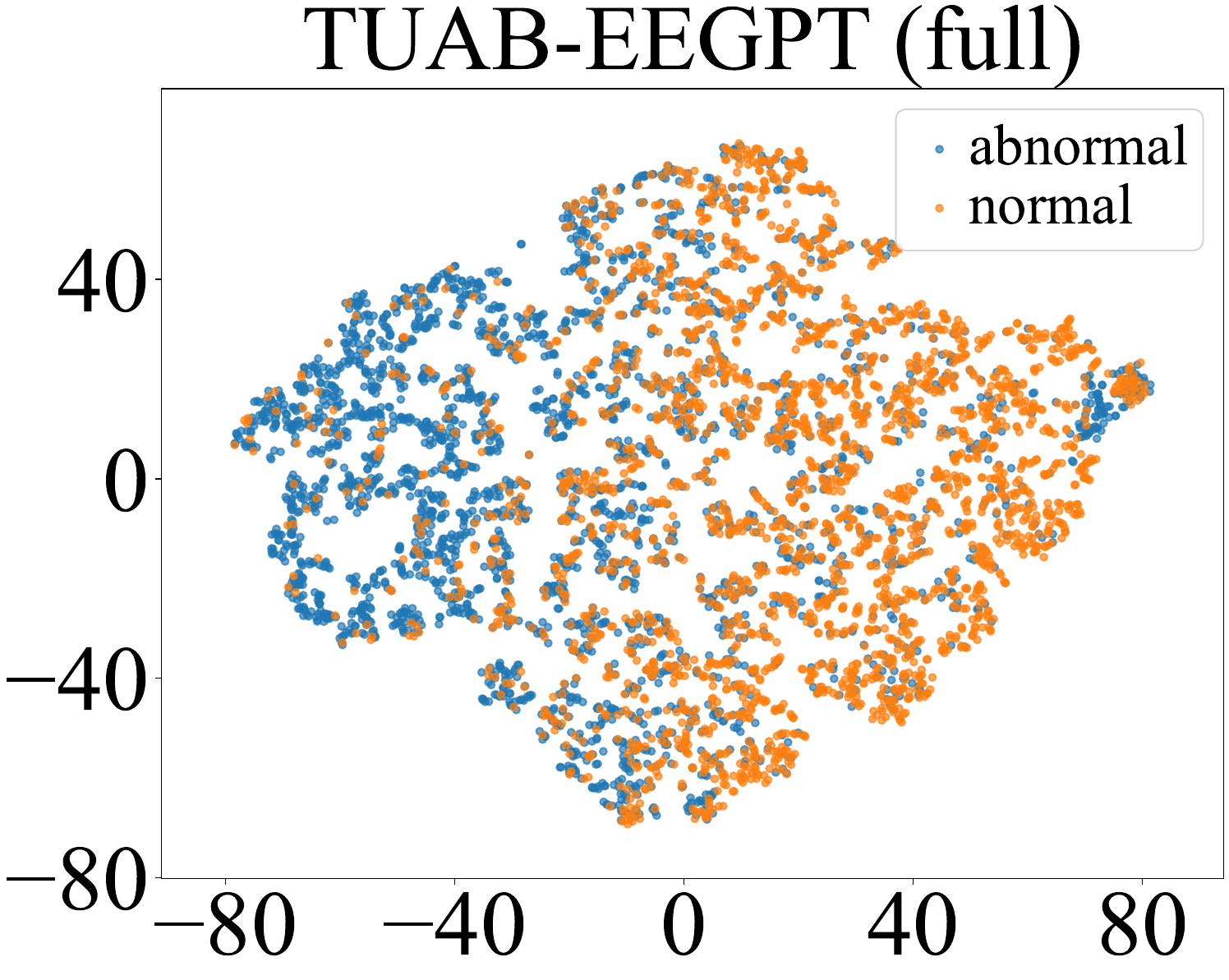}}
\subfigure[]{\includegraphics[width=0.19\linewidth,clip]{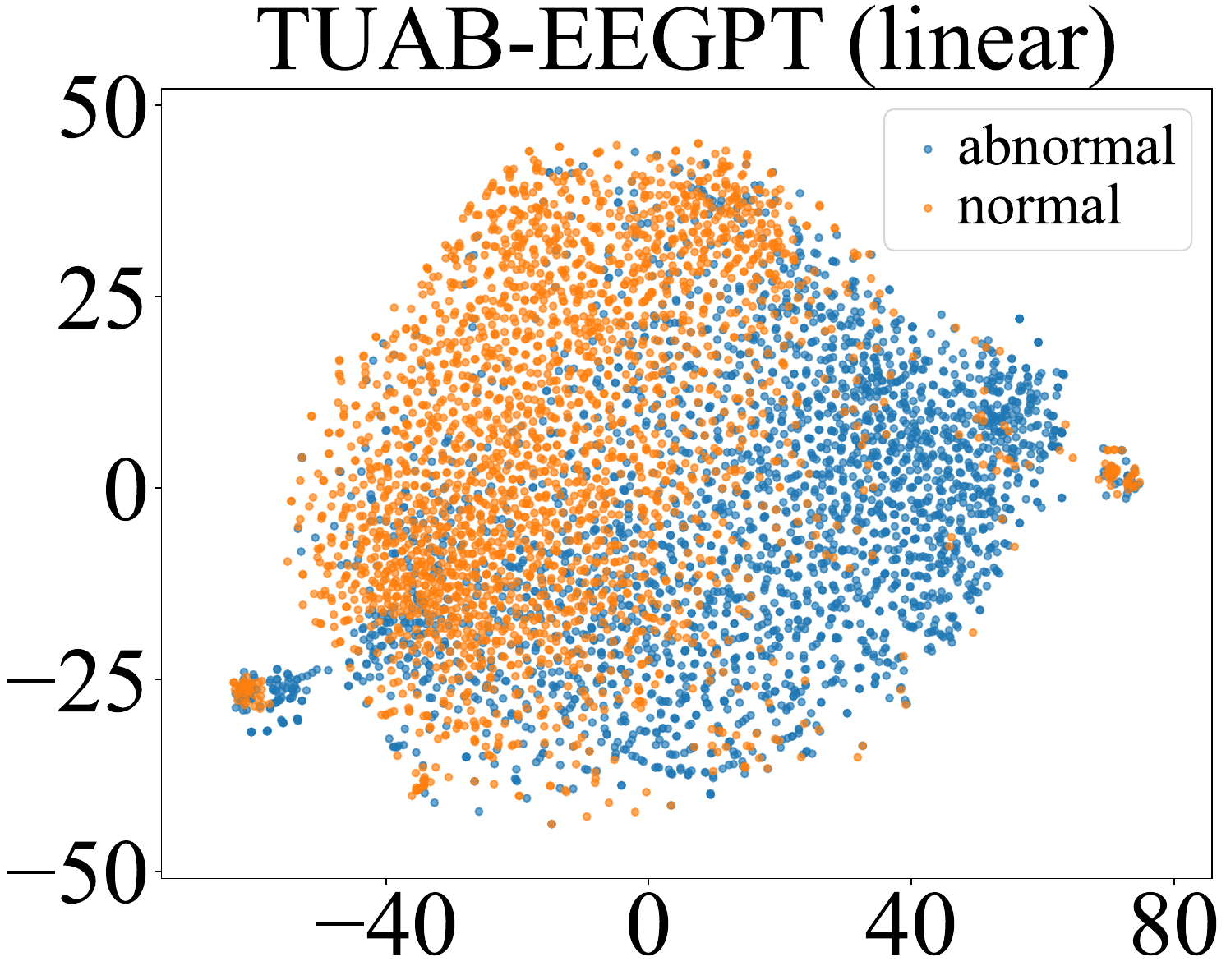}}
\subfigure[]{\includegraphics[width=0.19\linewidth,clip]{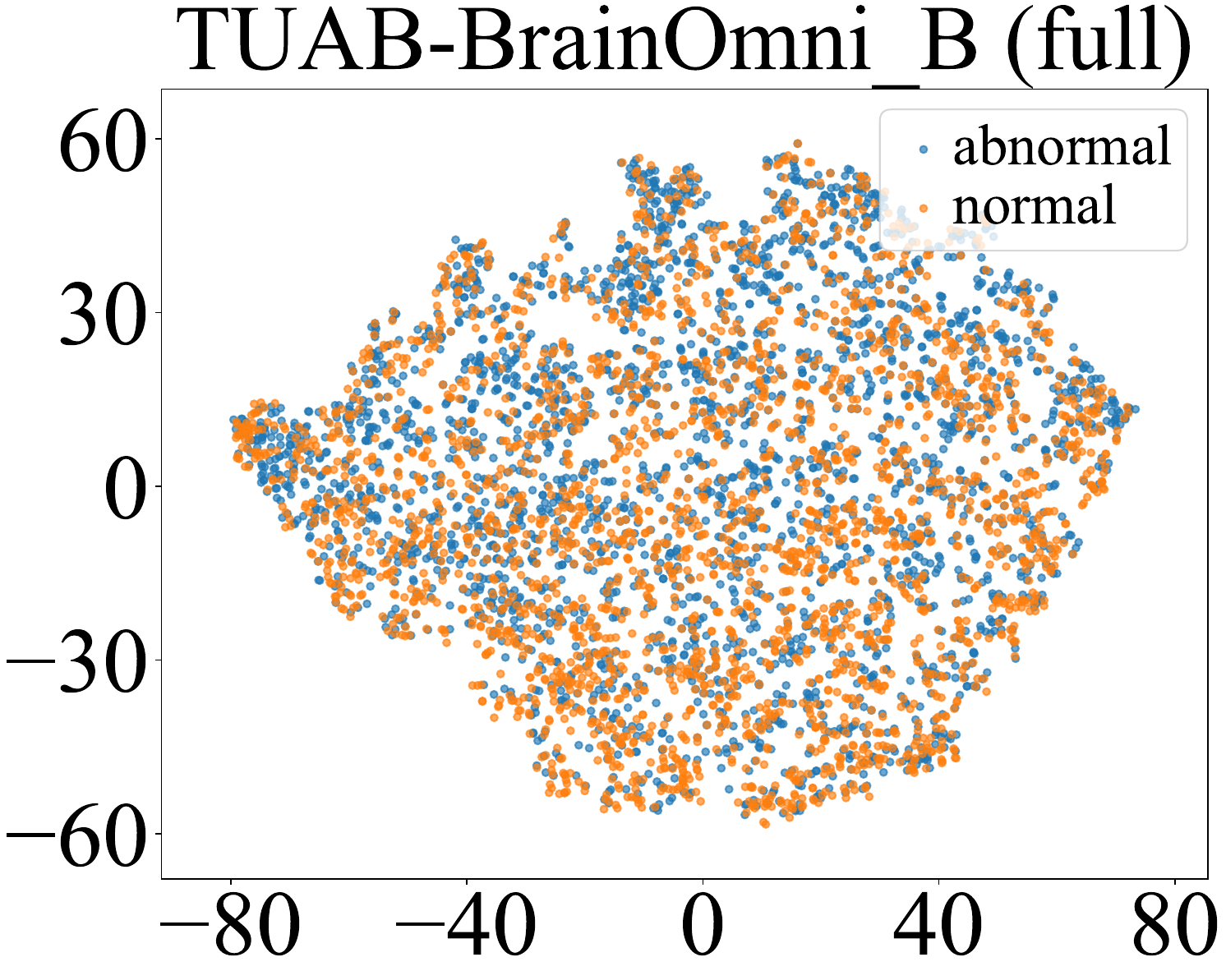}} \\
\subfigure[]{\includegraphics[width=0.19\linewidth,clip]{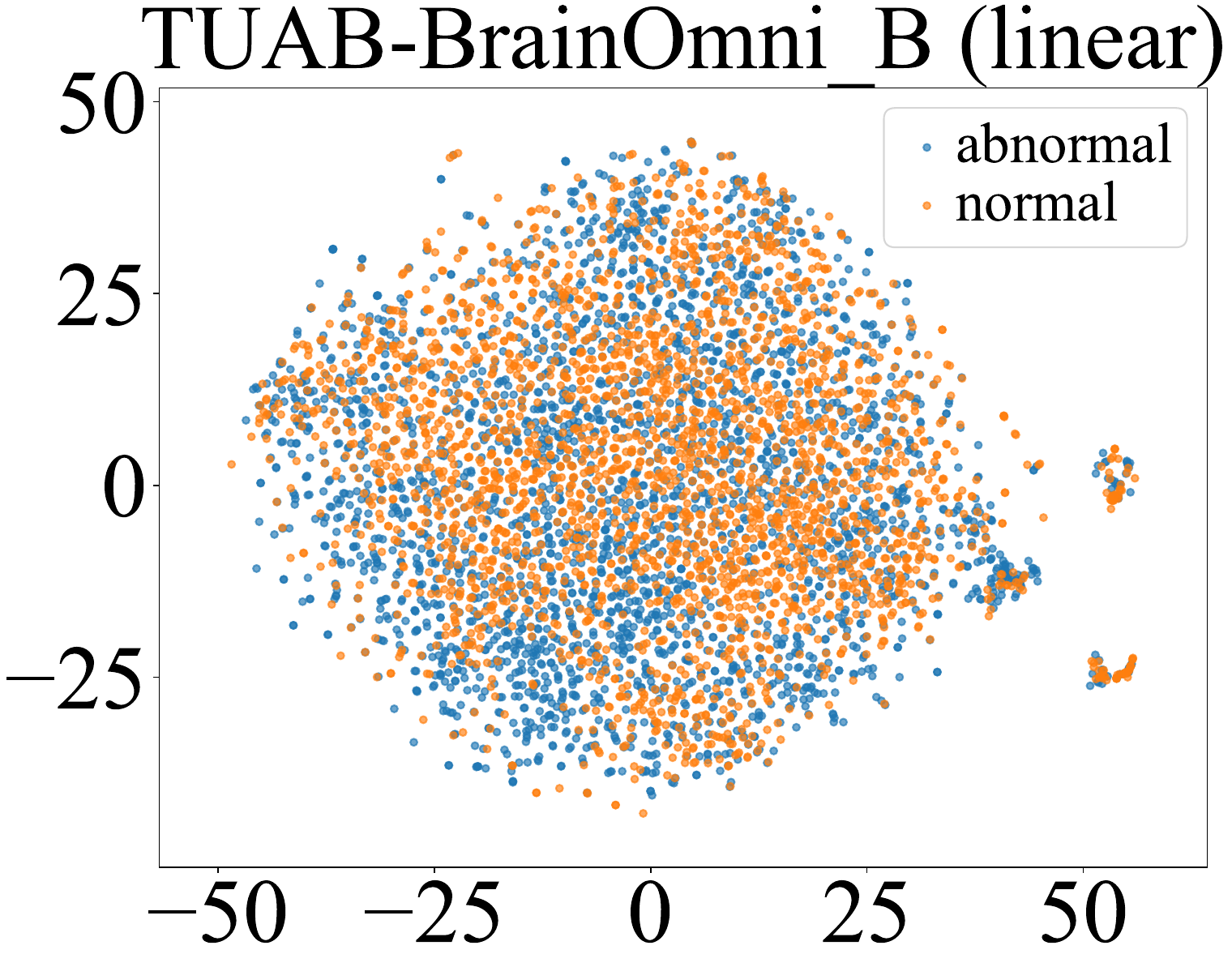}}
\subfigure[]{\includegraphics[width=0.19\linewidth,clip]{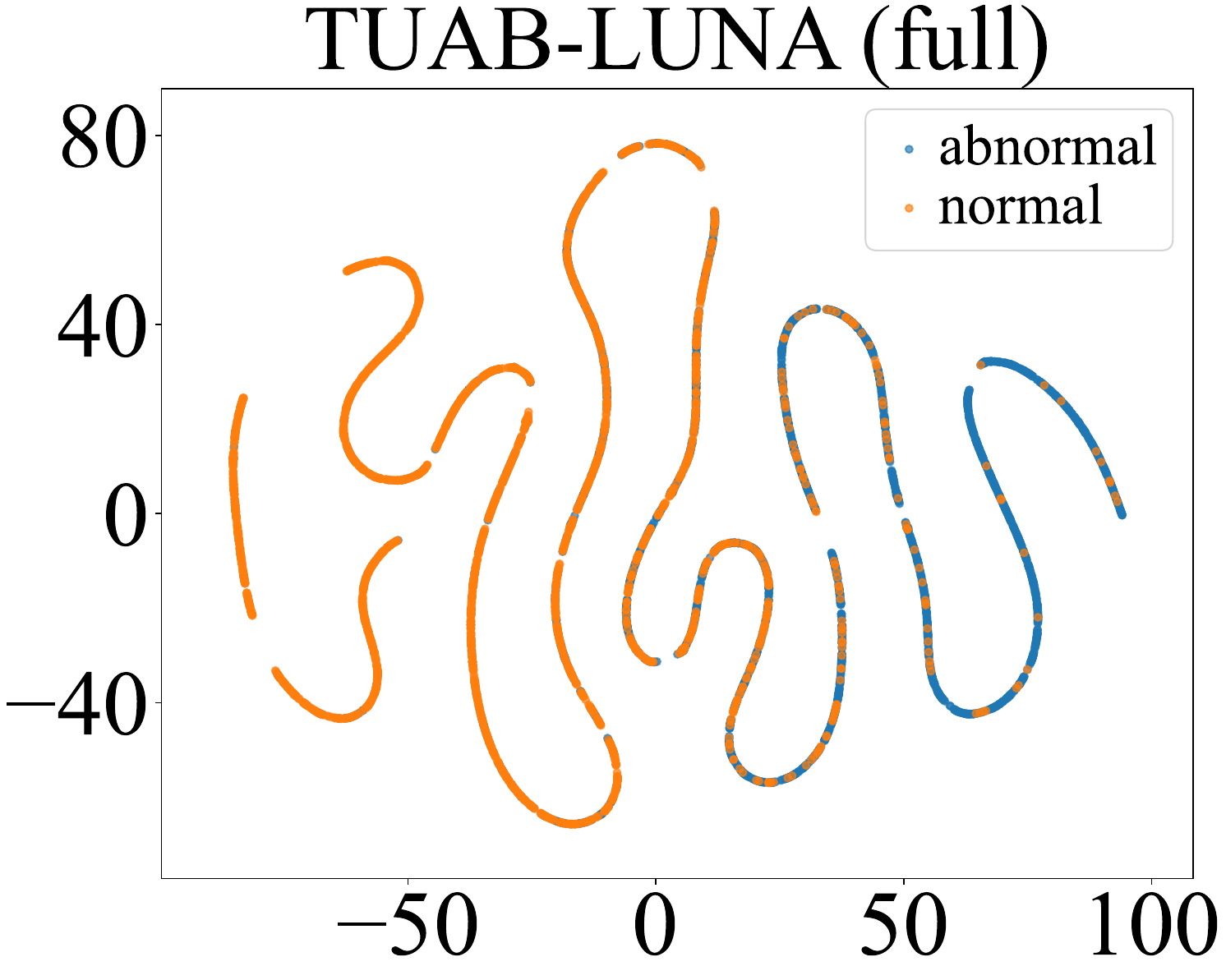}}
\subfigure[]{\includegraphics[width=0.19\linewidth,clip]{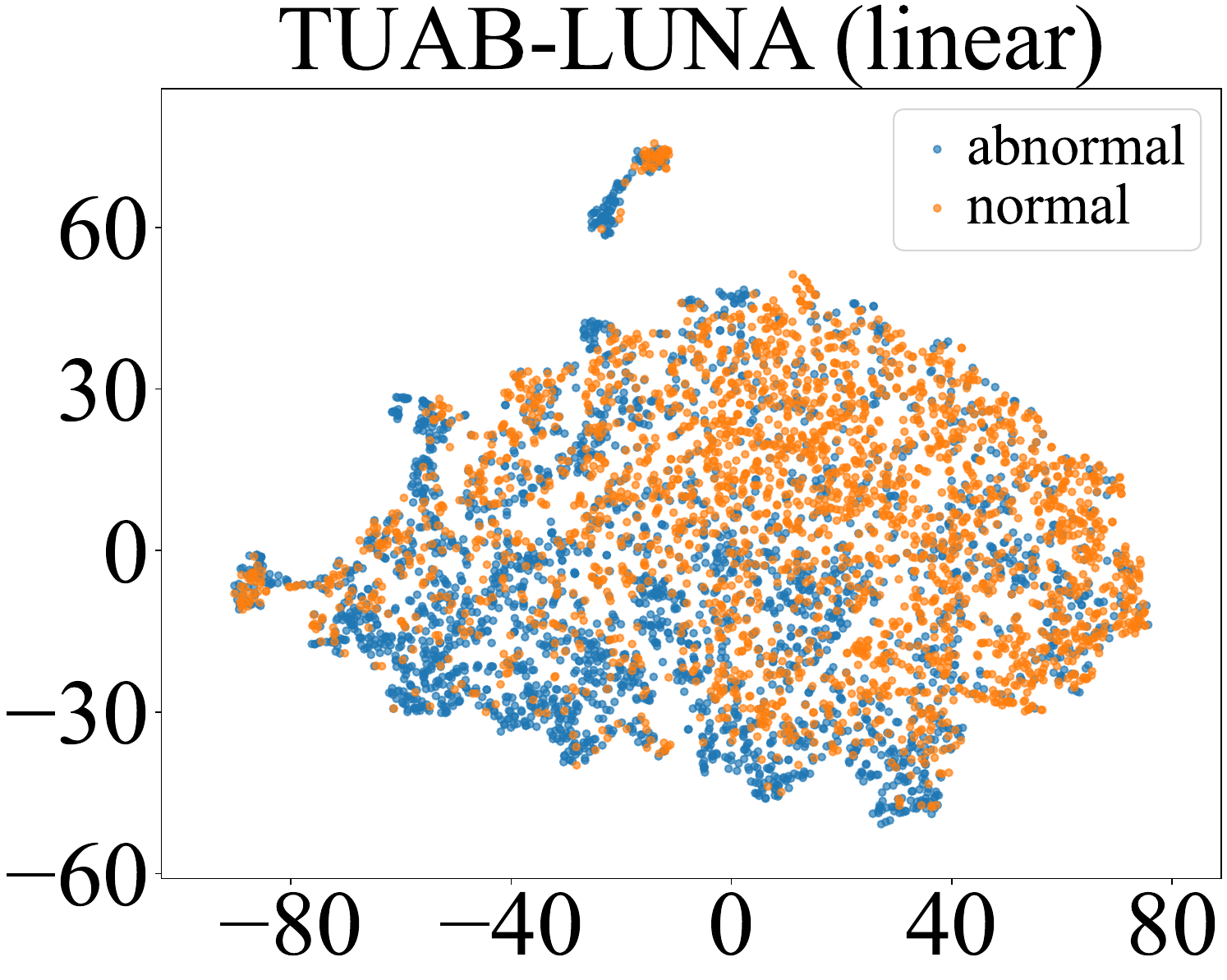}}
\subfigure[]{\includegraphics[width=0.19\linewidth,clip]{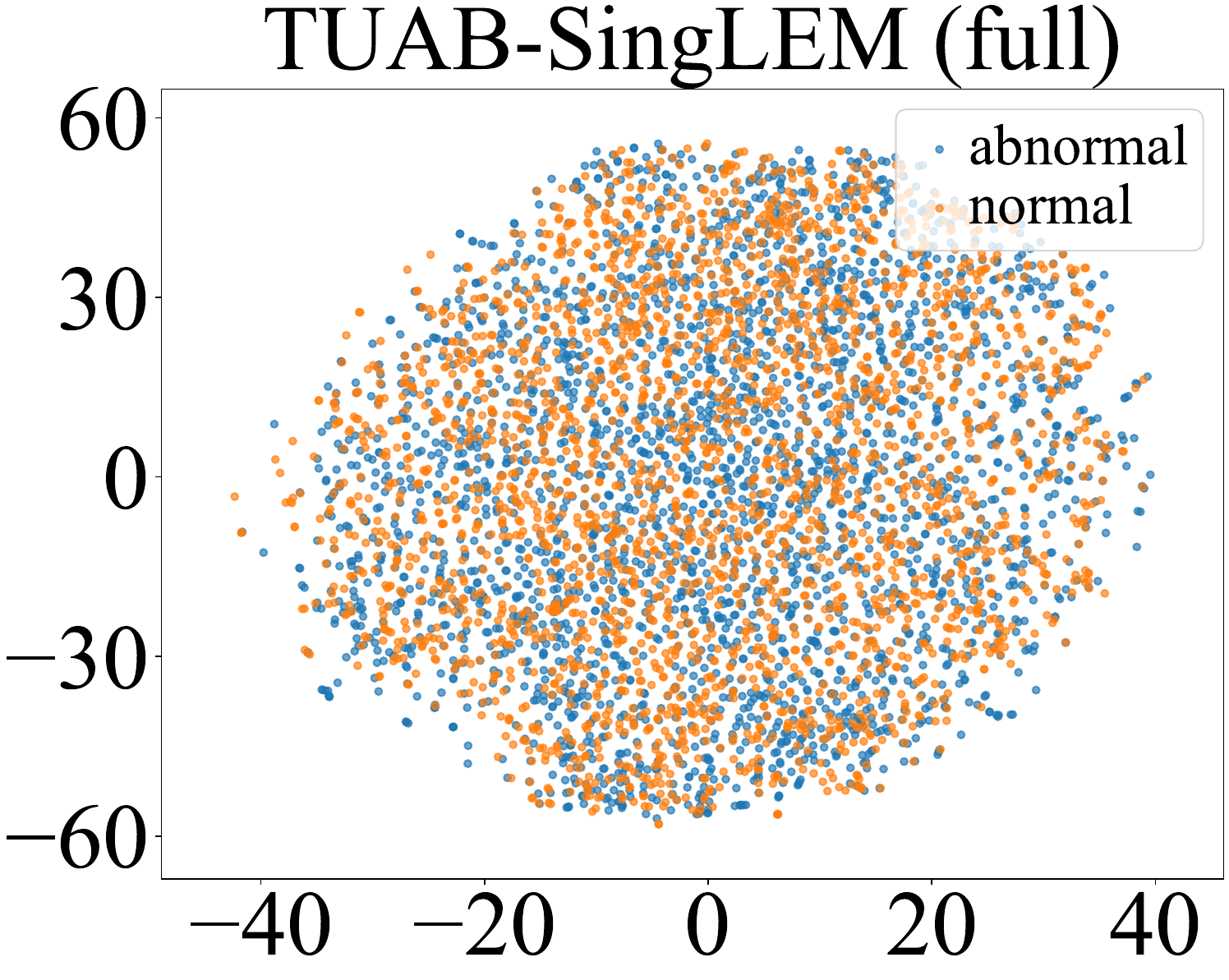}}
\subfigure[]{\includegraphics[width=0.19\linewidth,clip]{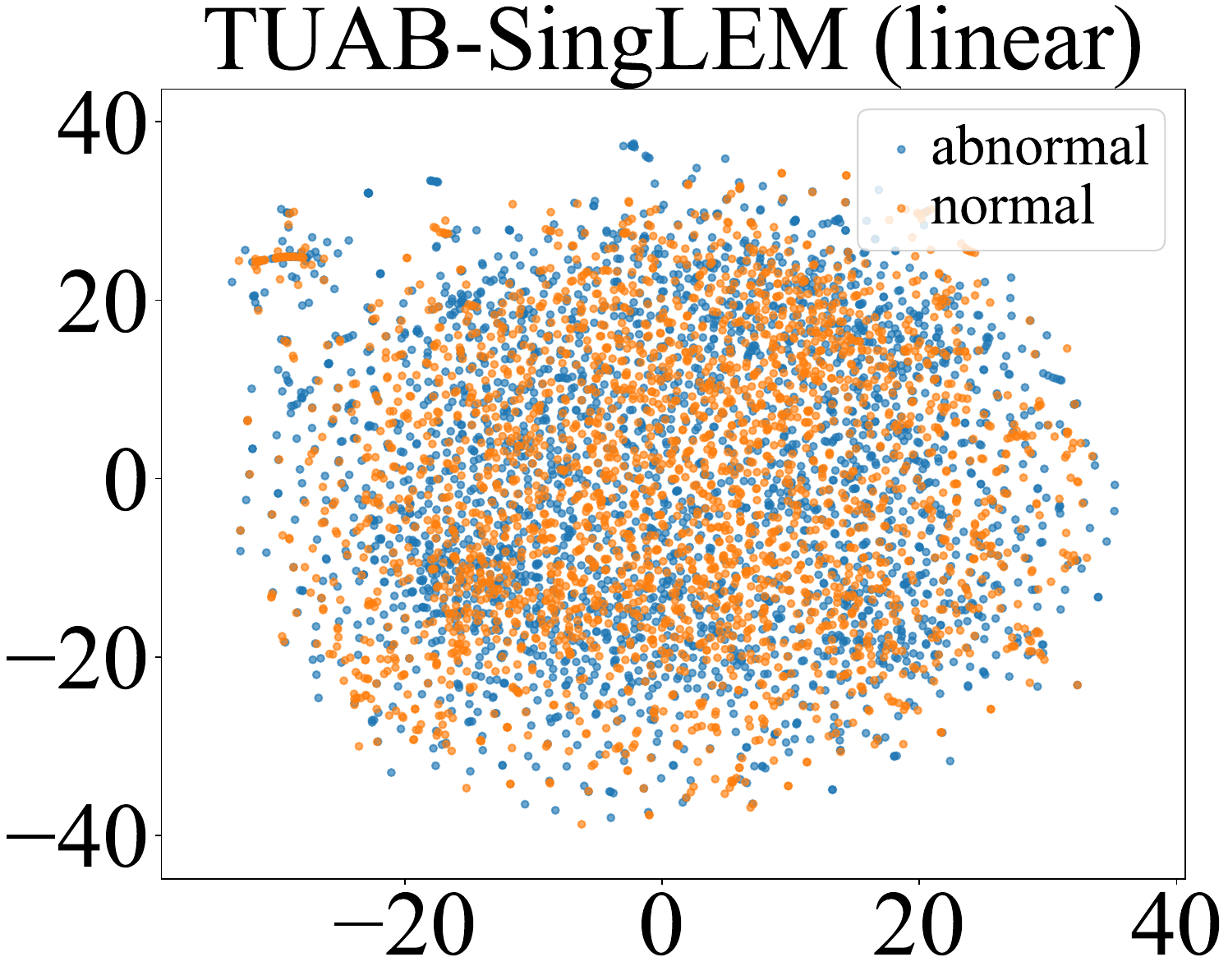}} \\
\caption{$t$-SNE visualization of the TUAB datasets.} \label{fig:tsne_rep2_appd}
\end{figure*}

\begin{figure*}[t]
\centering
\subfigure[]{\includegraphics[width=0.19\linewidth,clip]{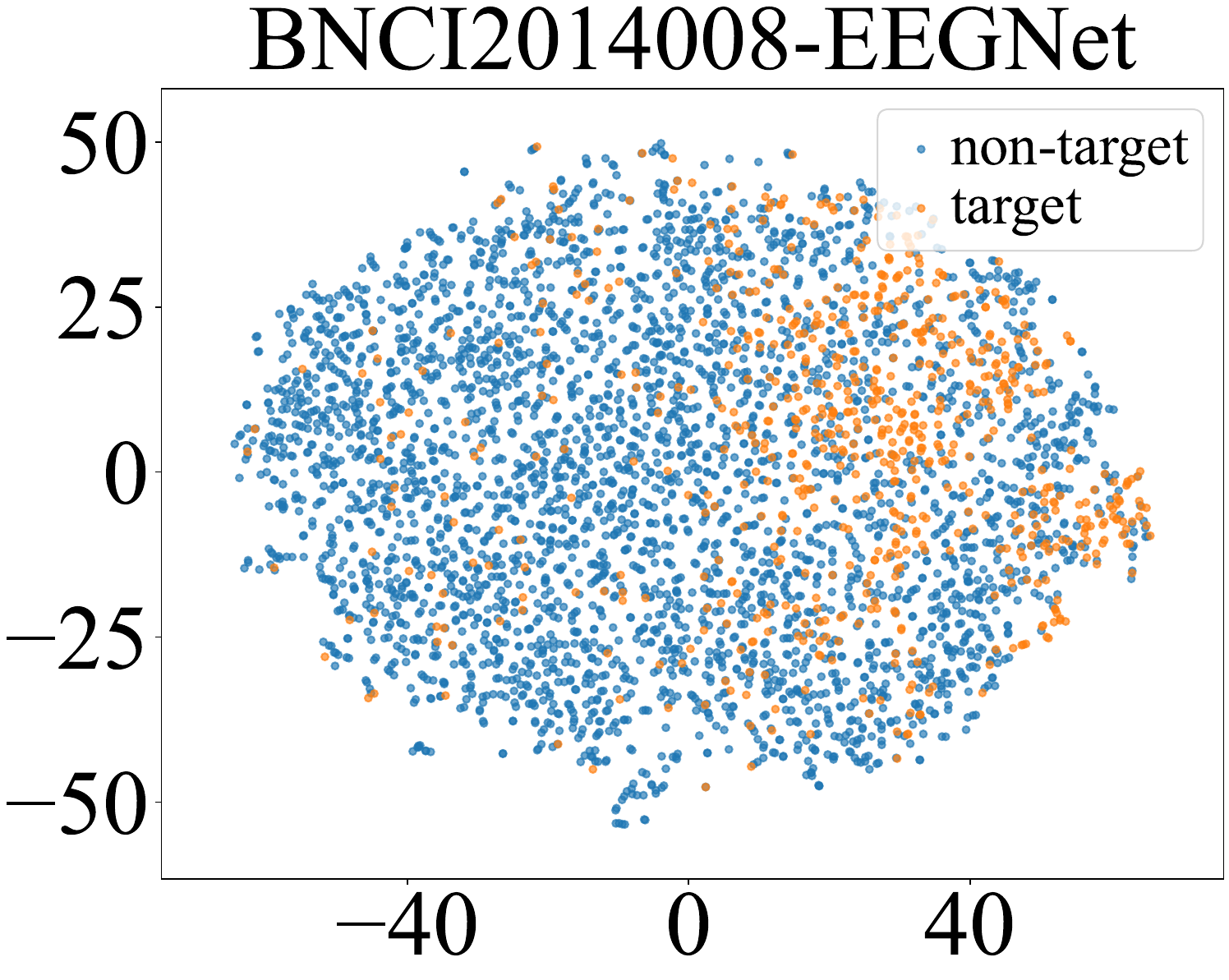}}
\subfigure[]{\includegraphics[width=0.19\linewidth,clip]{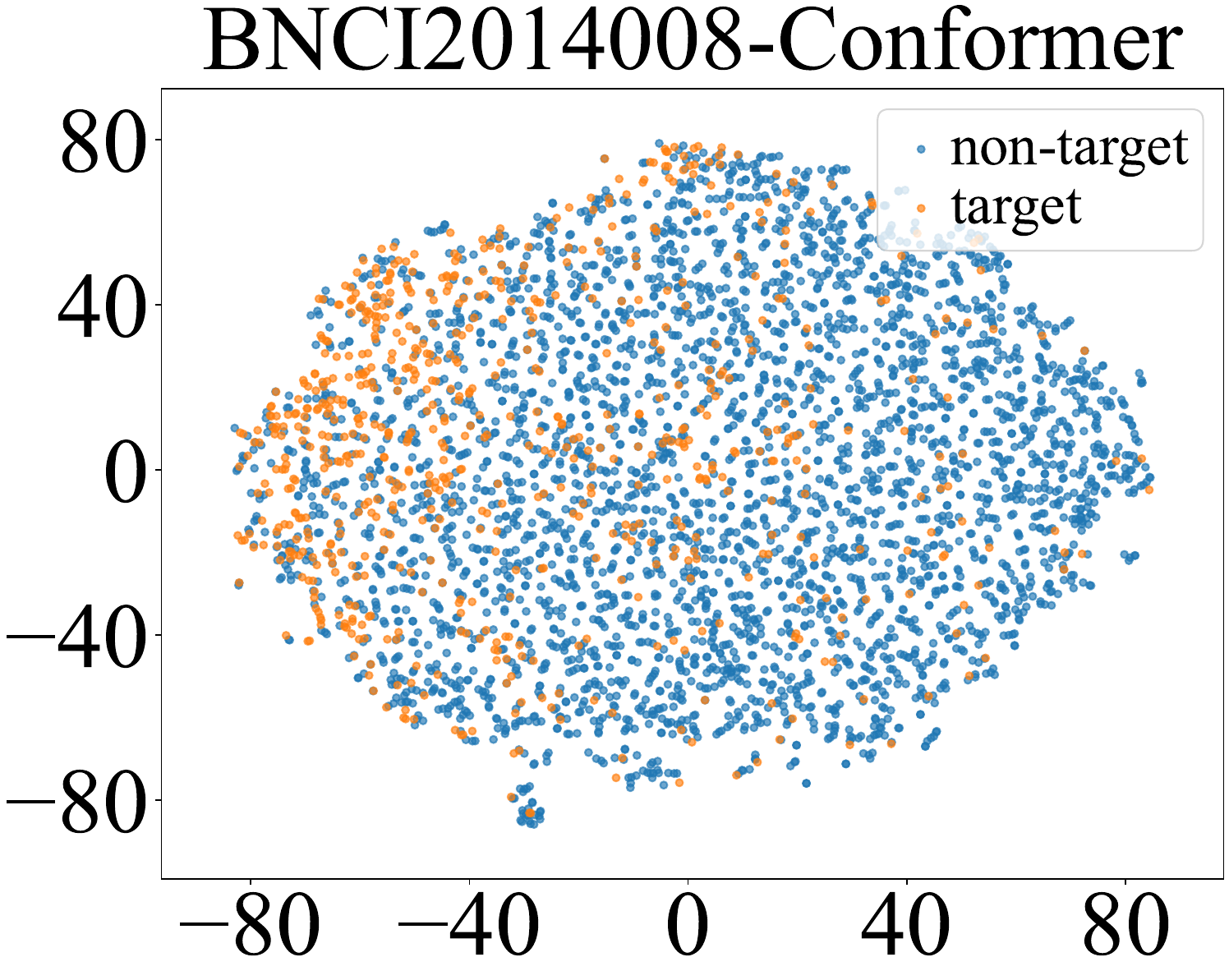}}
\subfigure[]{\includegraphics[width=0.19\linewidth,clip]{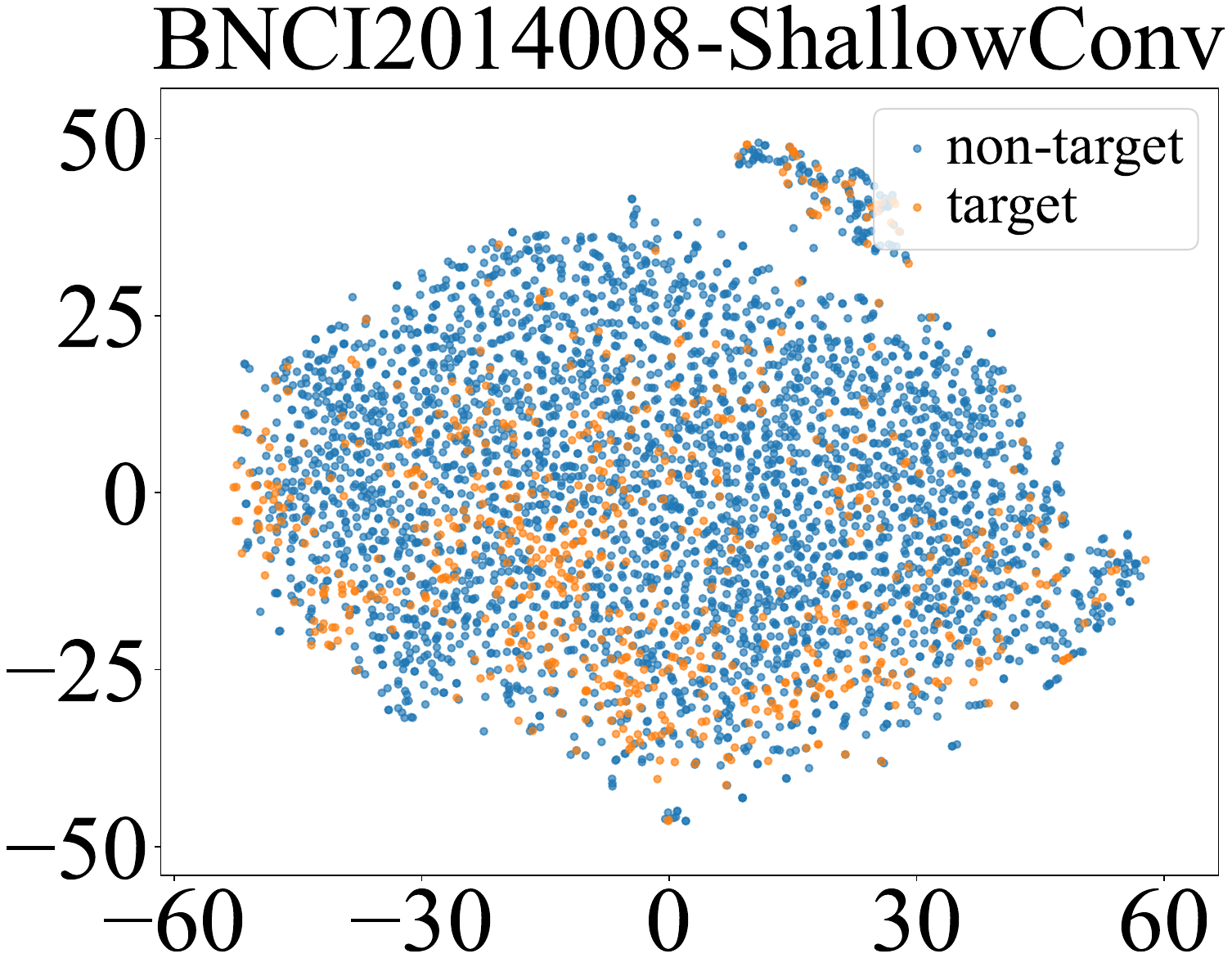}}
\subfigure[]{\includegraphics[width=0.19\linewidth,clip]{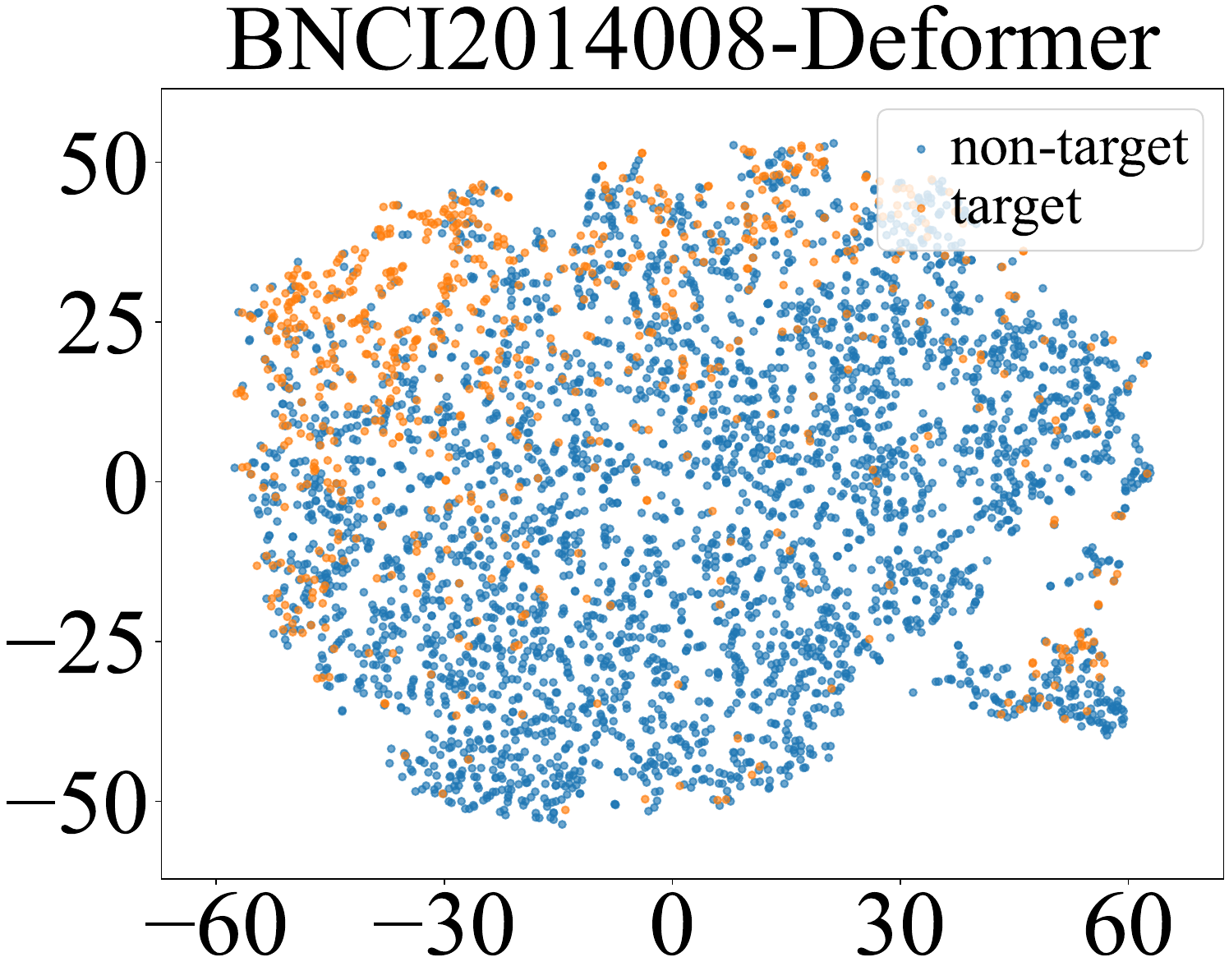}}
\subfigure[]{\includegraphics[width=0.19\linewidth,clip]{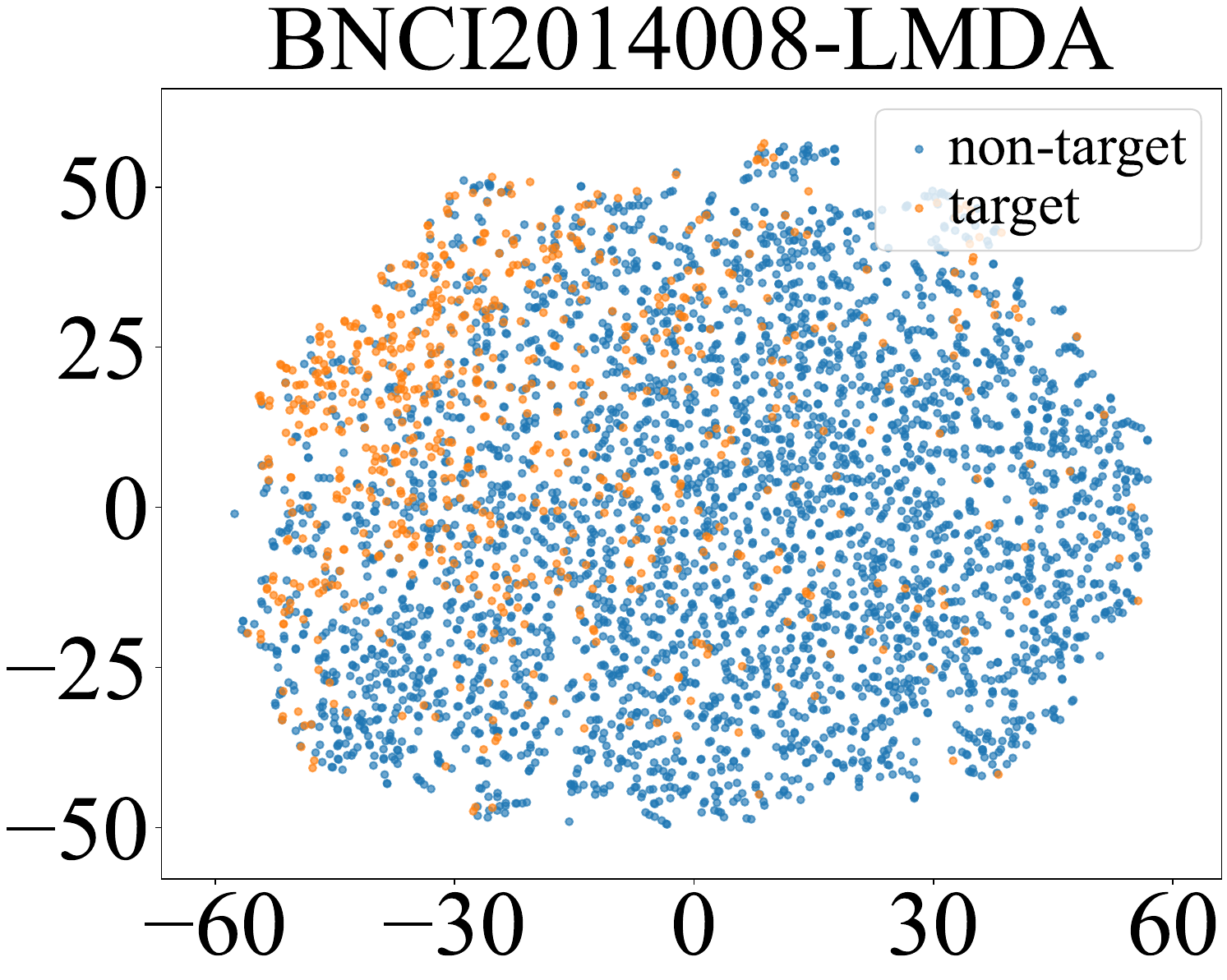}}\\
\subfigure[]{\includegraphics[width=0.19\linewidth,clip]{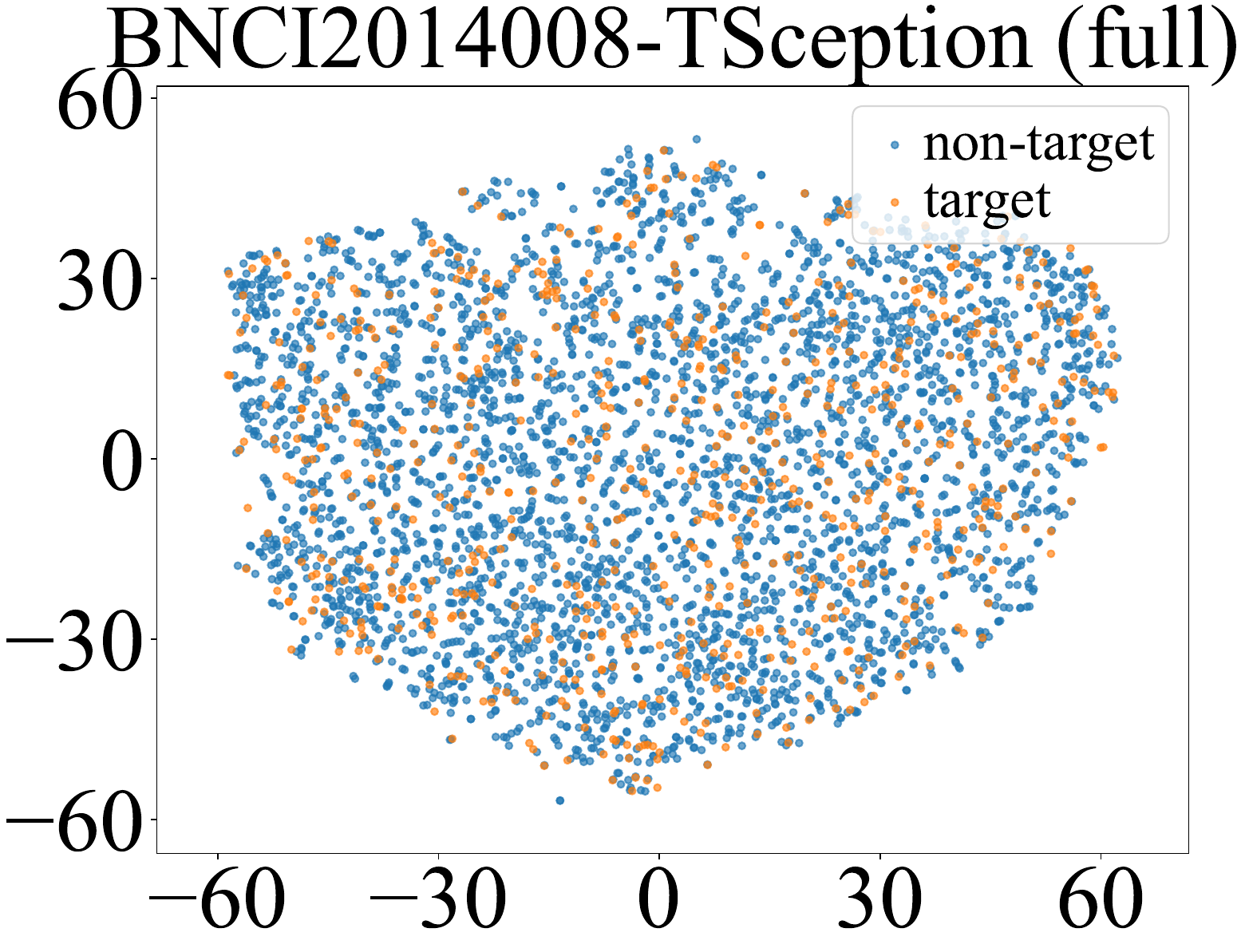}}
\subfigure[]{\includegraphics[width=0.19\linewidth,clip]{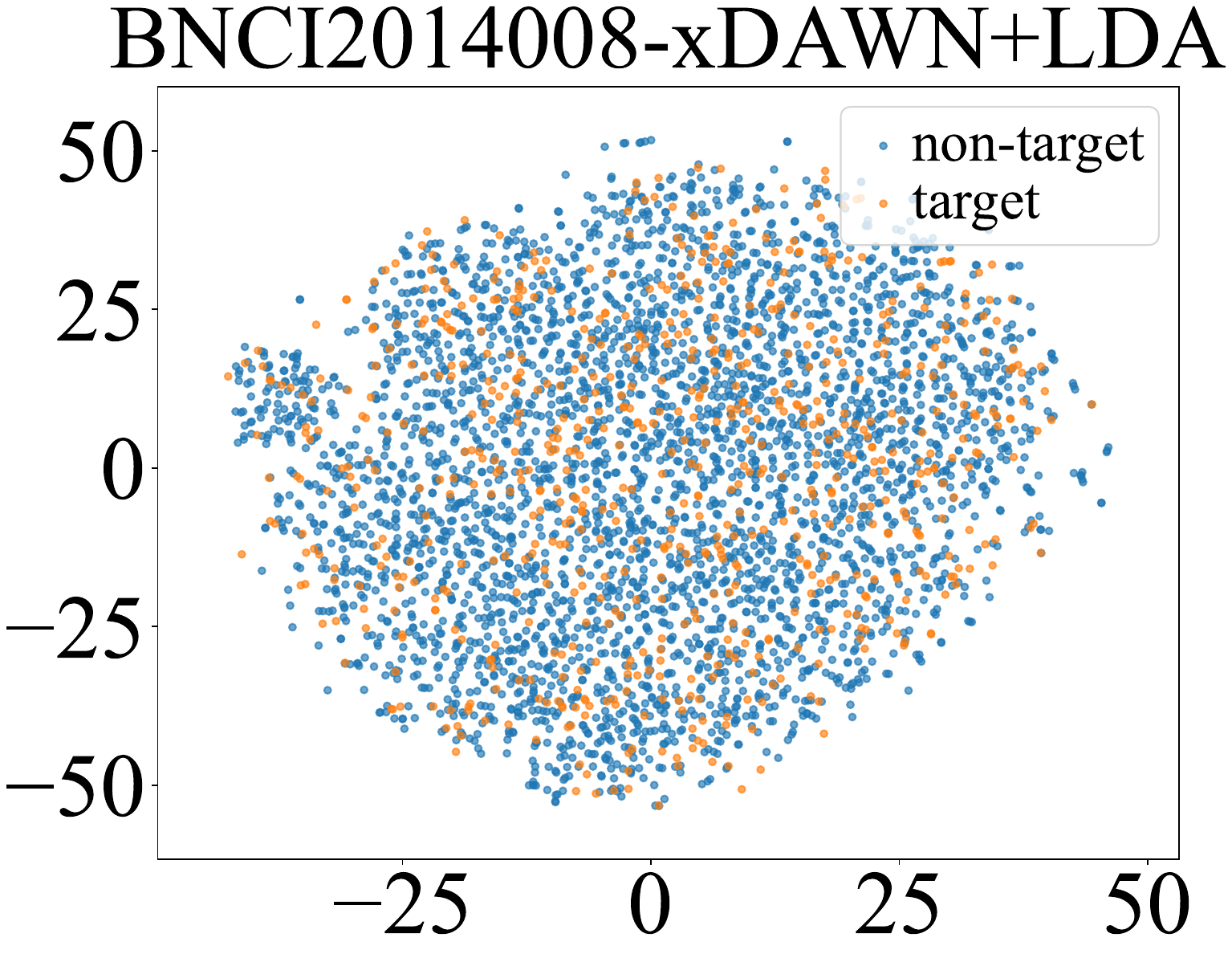}}
\subfigure[]{\includegraphics[width=0.19\linewidth,clip]{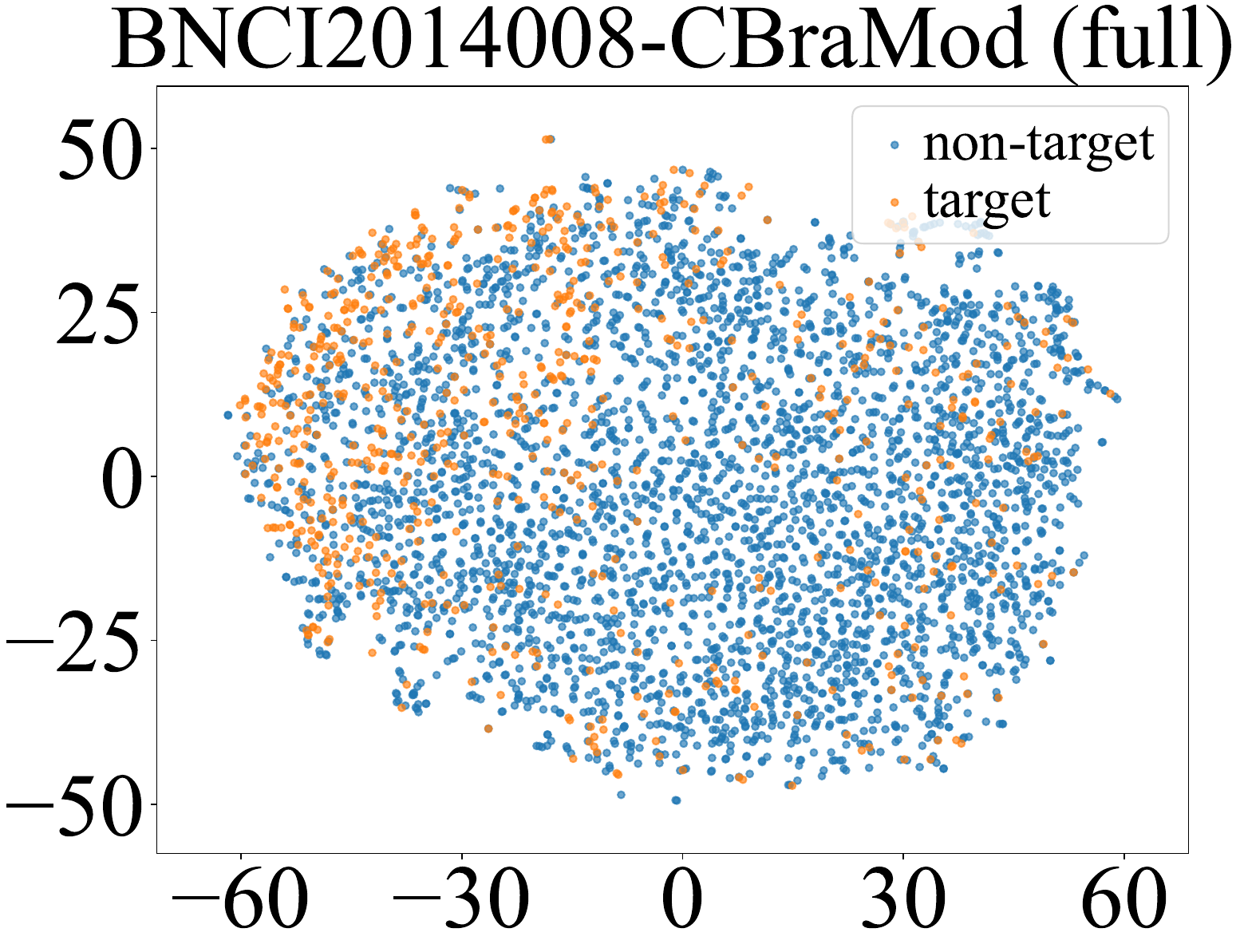}}
\subfigure[]{\includegraphics[width=0.19\linewidth,clip]{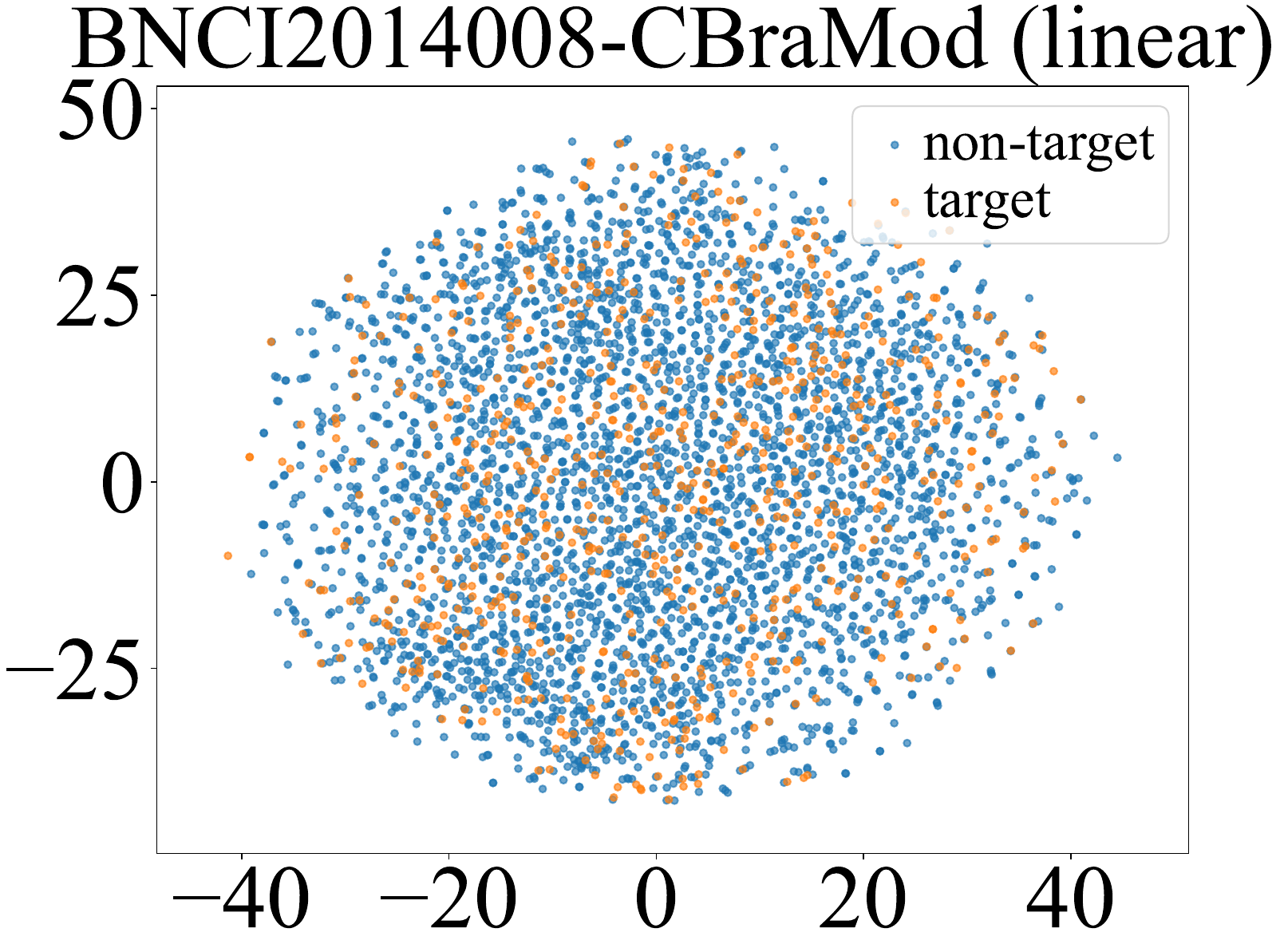}}
\subfigure[]{\includegraphics[width=0.19\linewidth,clip]{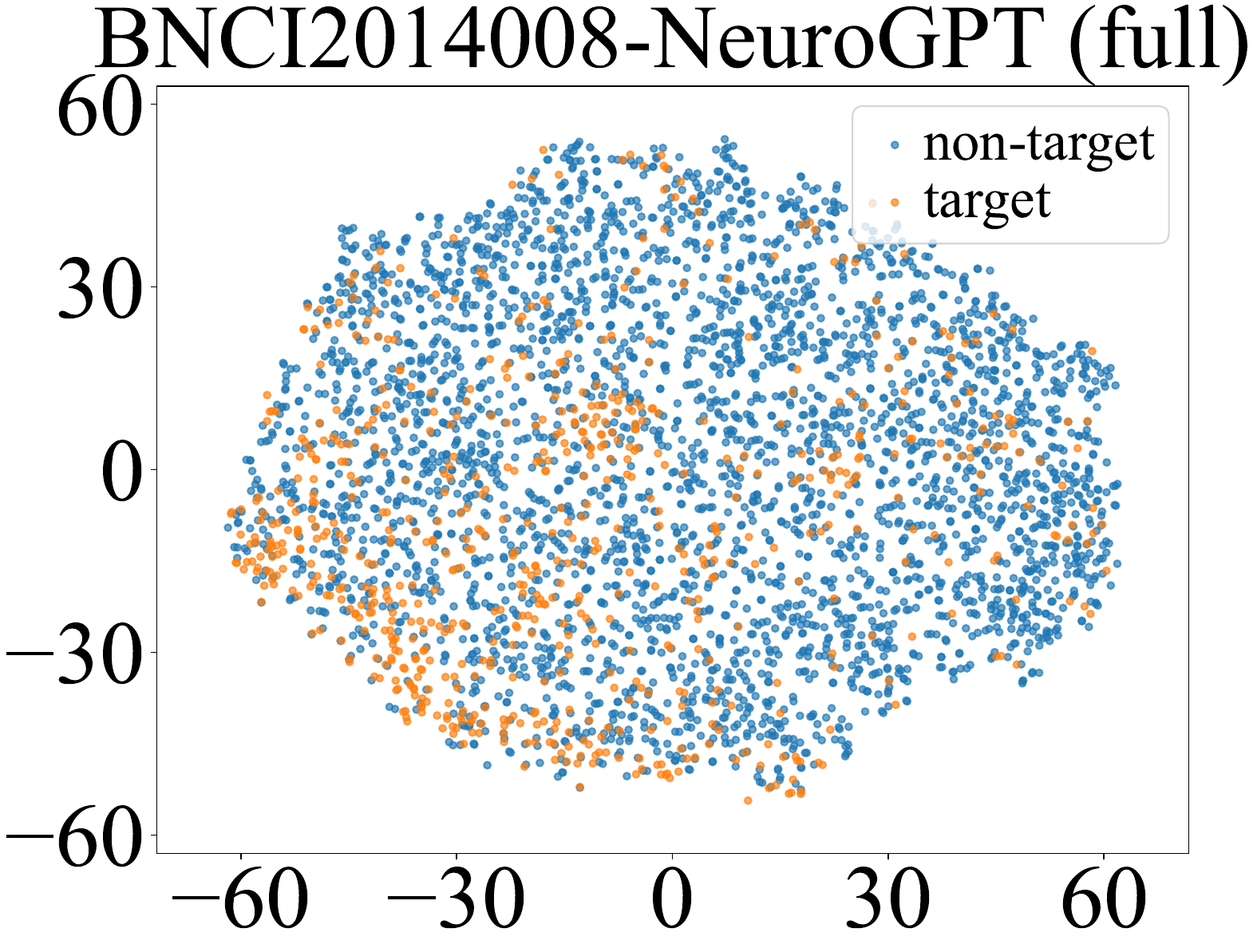}} \\
\subfigure[]{\includegraphics[width=0.19\linewidth,clip]{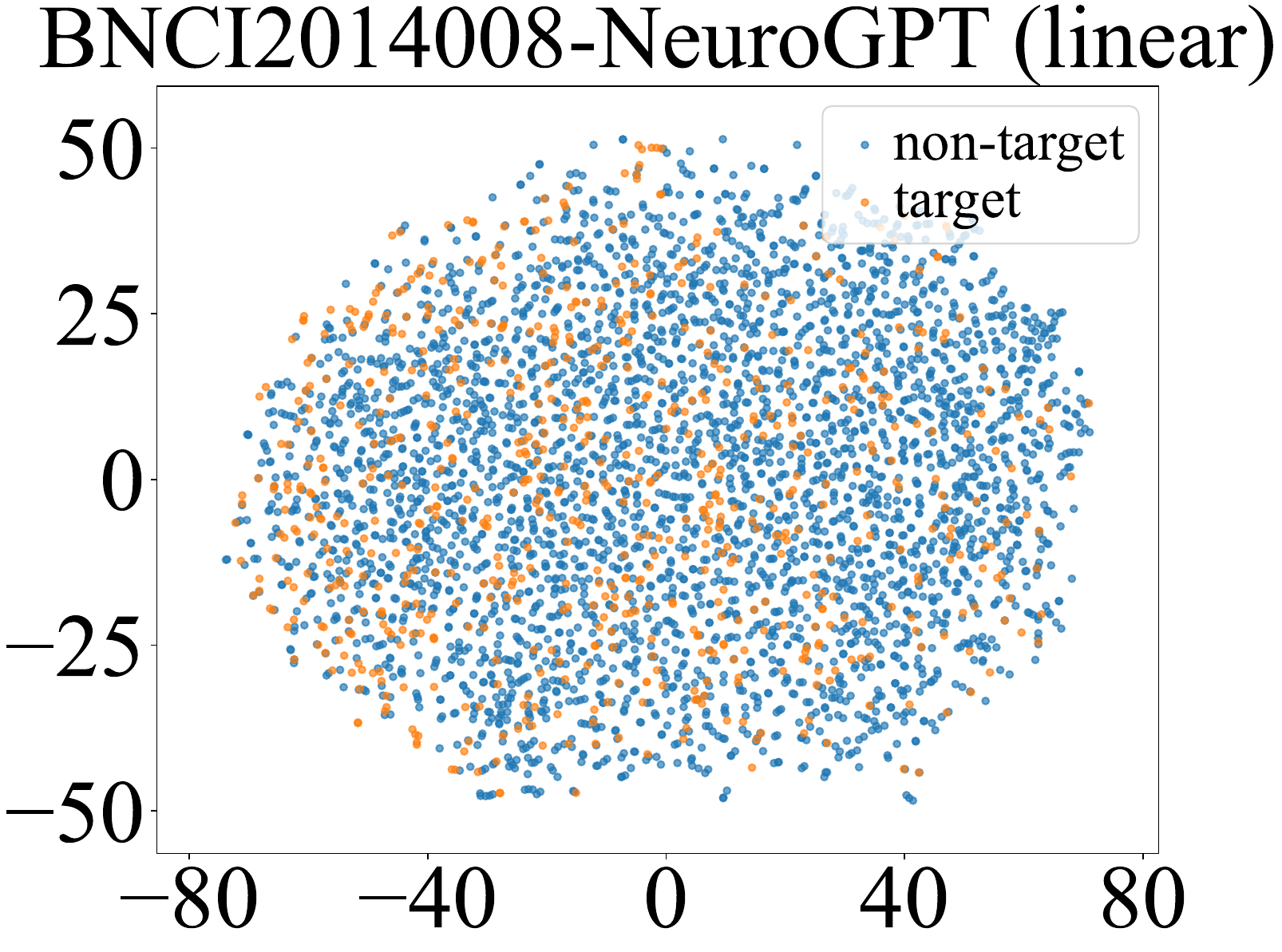}}
\subfigure[]{\includegraphics[width=0.19\linewidth,clip]{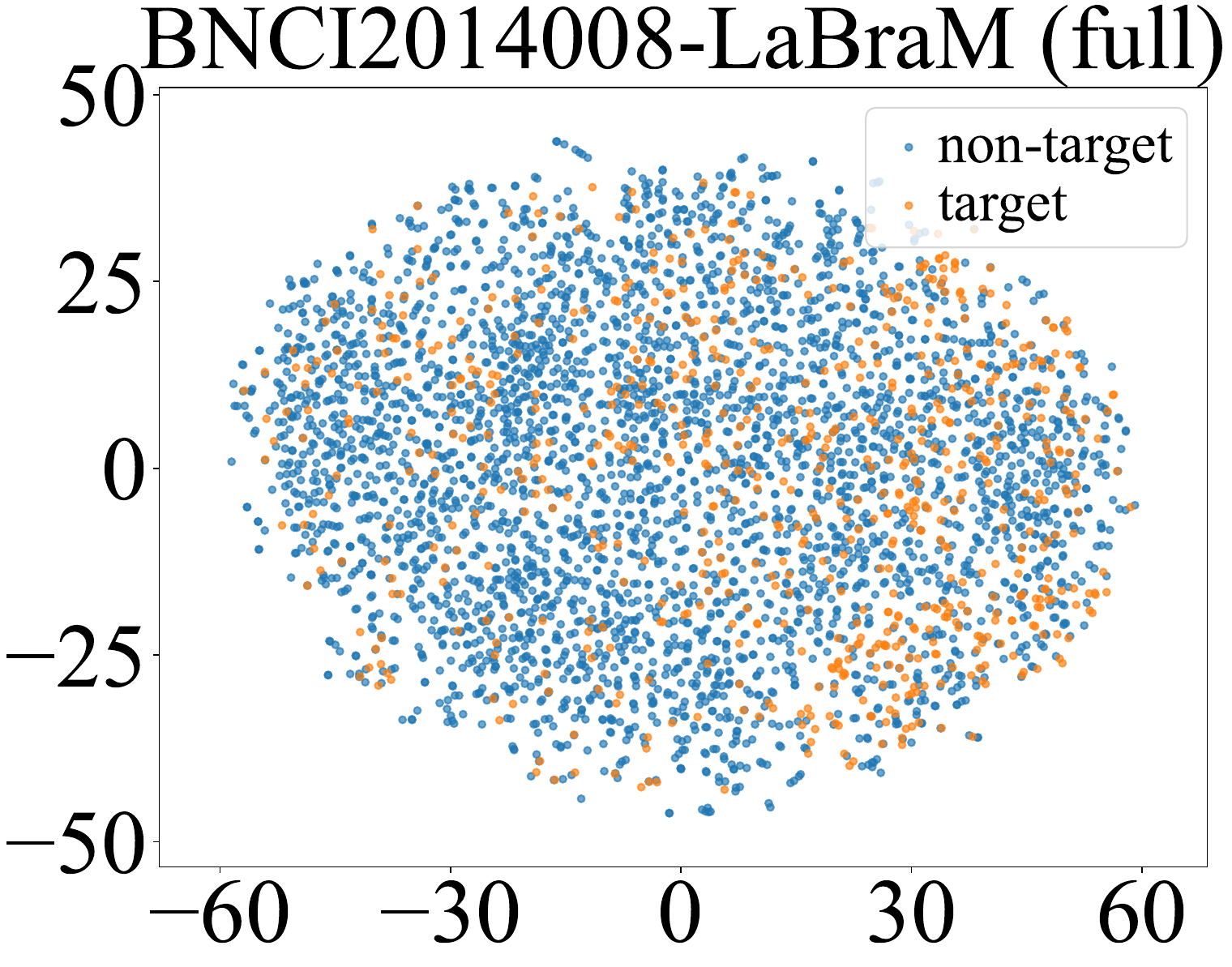}}
\subfigure[]{\includegraphics[width=0.19\linewidth,clip]{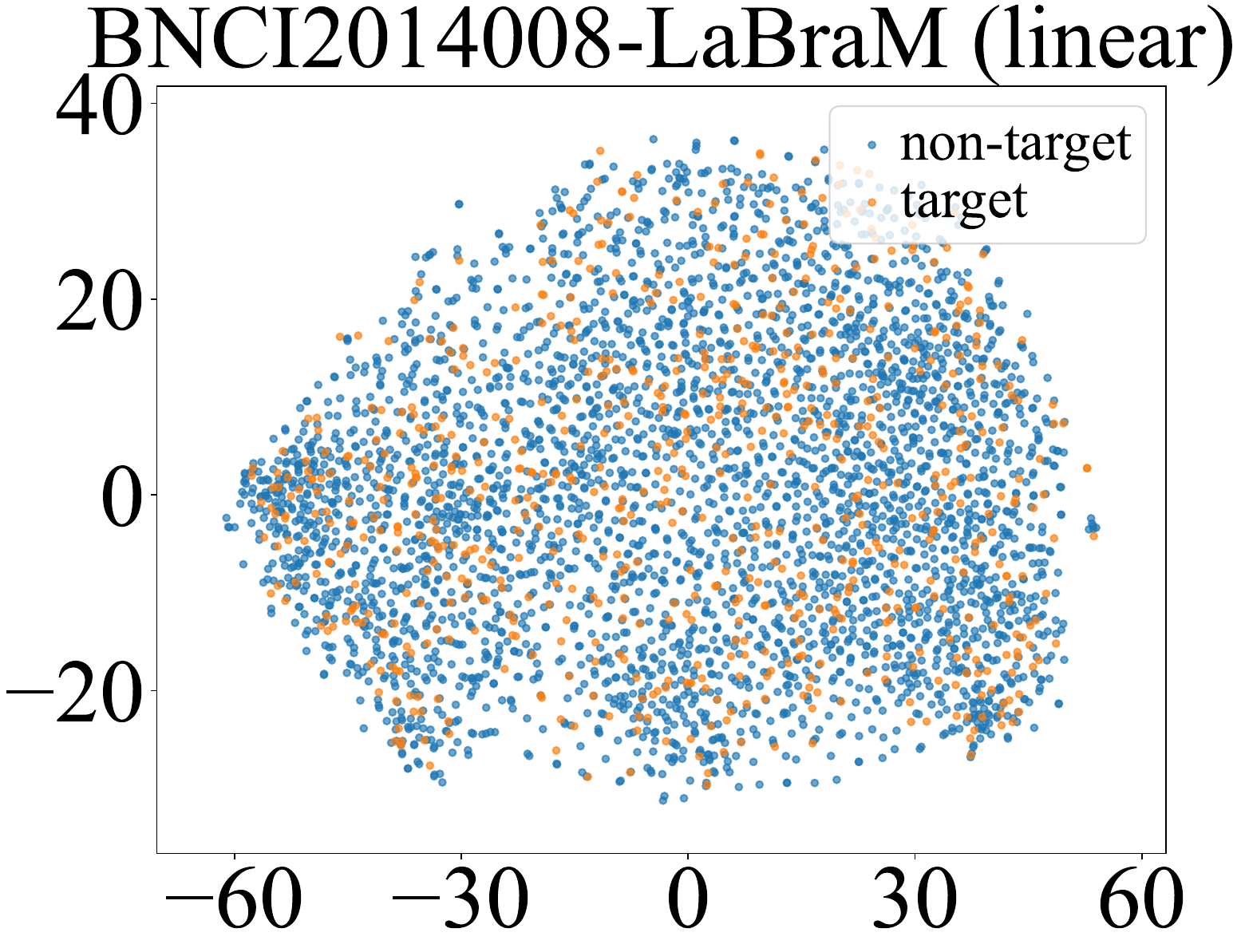}}
\subfigure[]{\includegraphics[width=0.19\linewidth,clip]{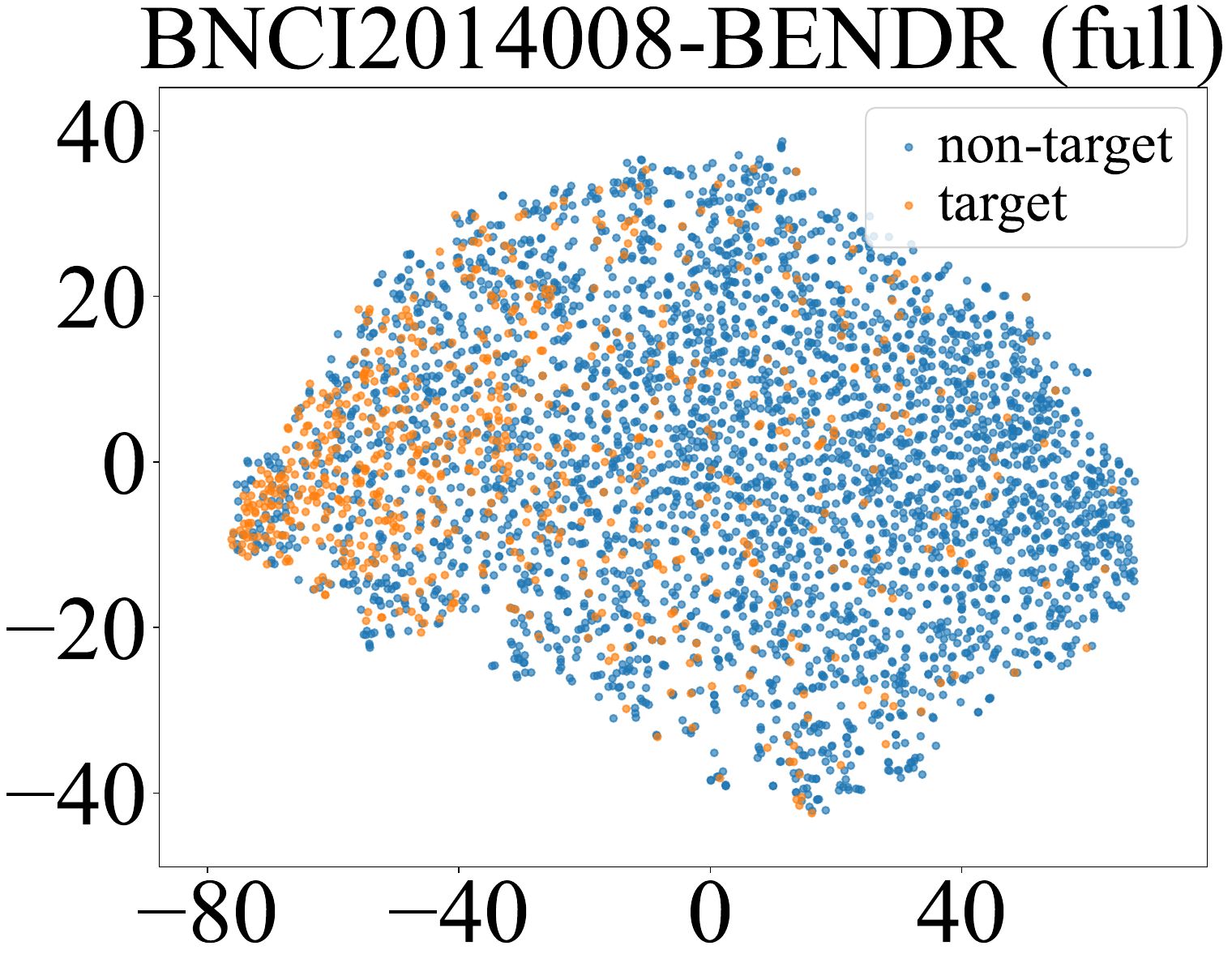}}
\subfigure[]{\includegraphics[width=0.19\linewidth,clip]{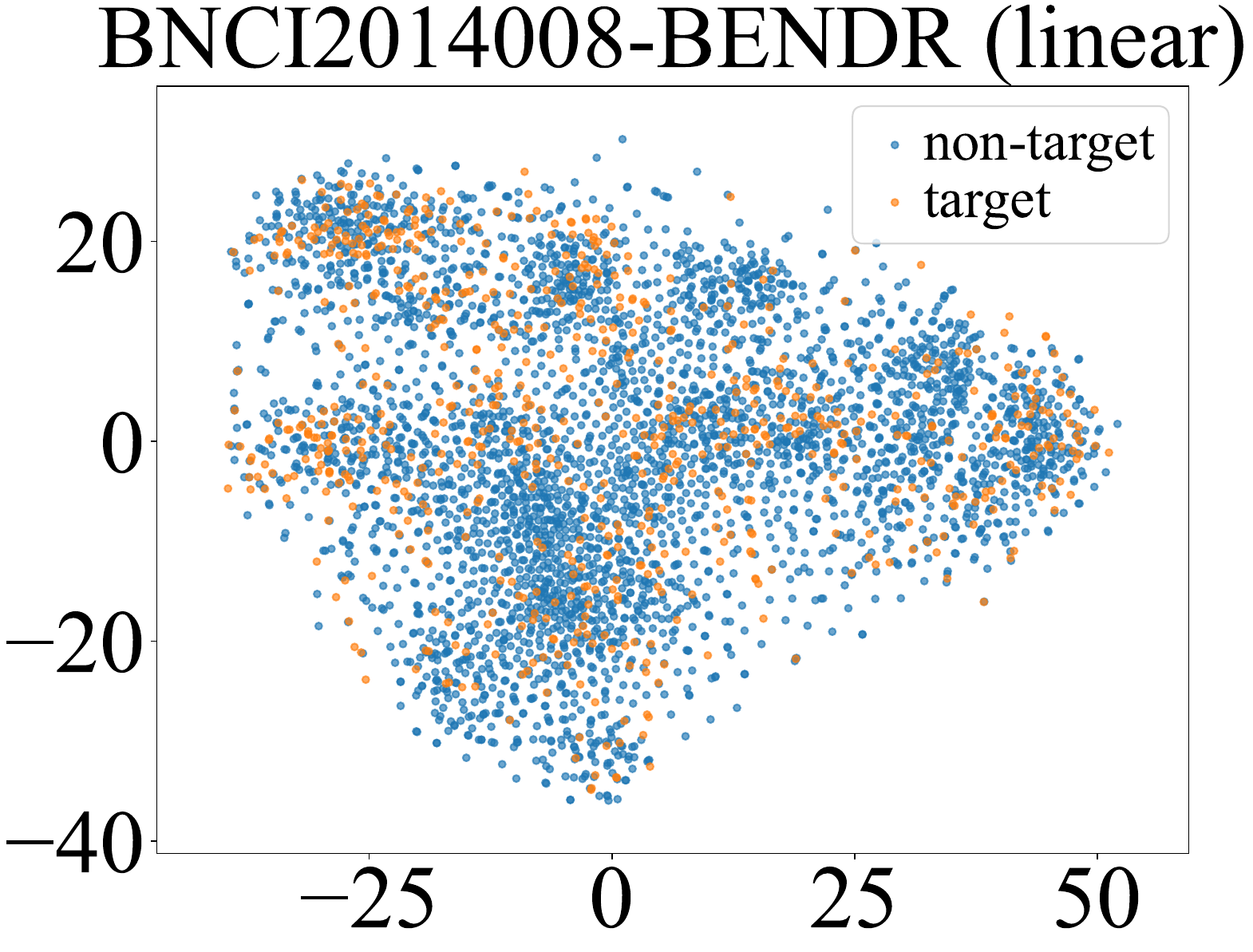}}\\
\subfigure[]{\includegraphics[width=0.19\linewidth,clip]{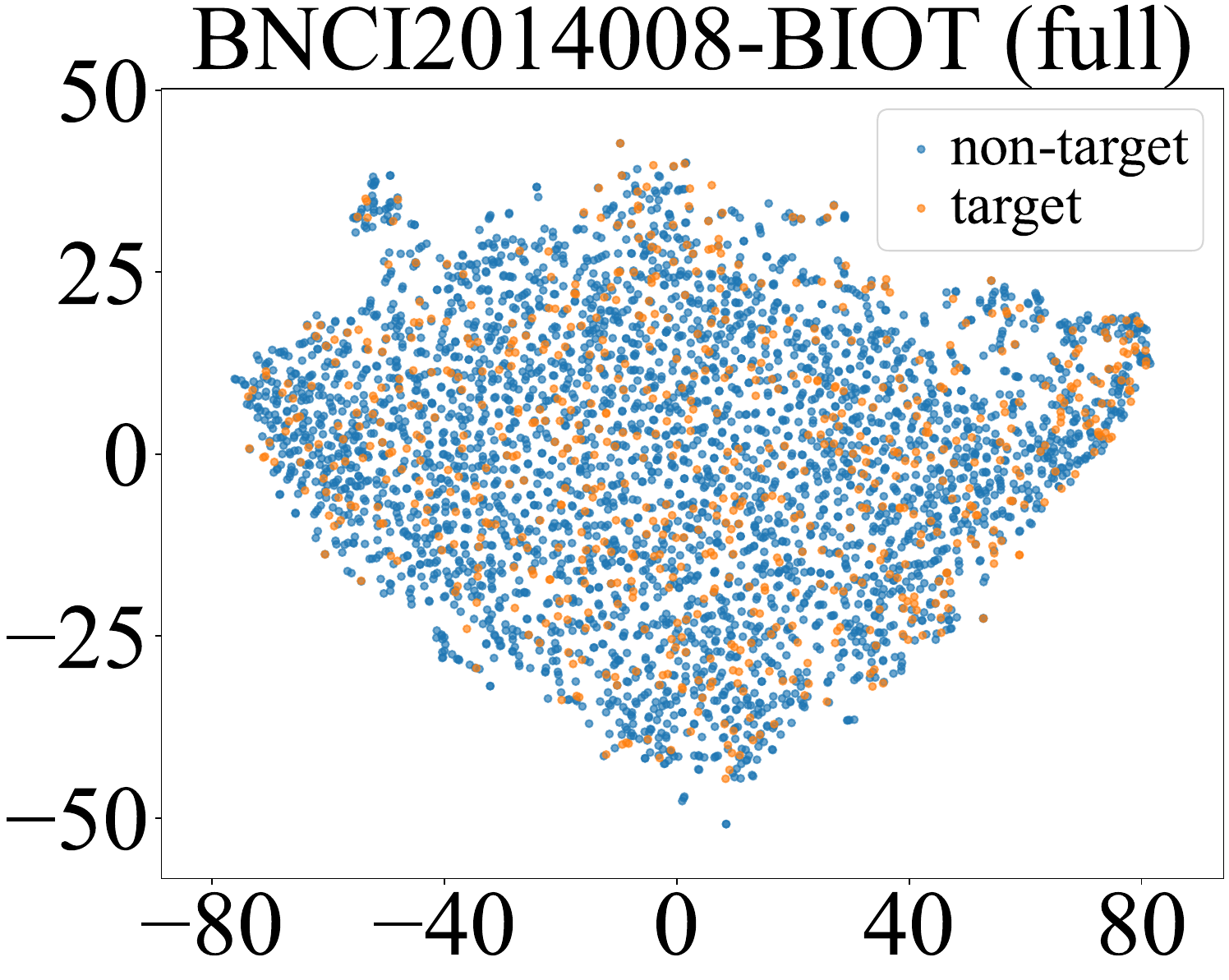}}
\subfigure[]{\includegraphics[width=0.19\linewidth,clip]{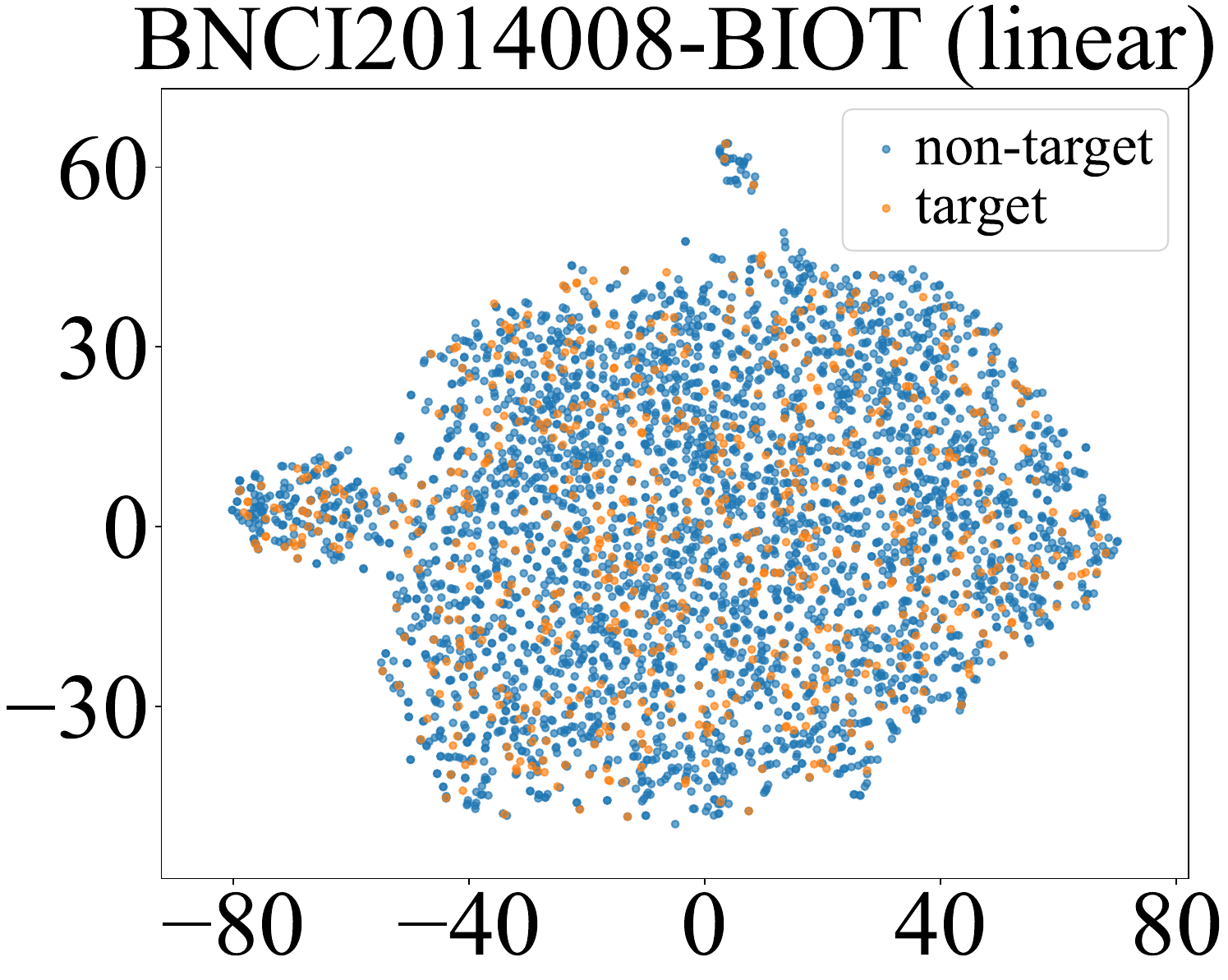}}
\subfigure[]{\includegraphics[width=0.19\linewidth,clip]{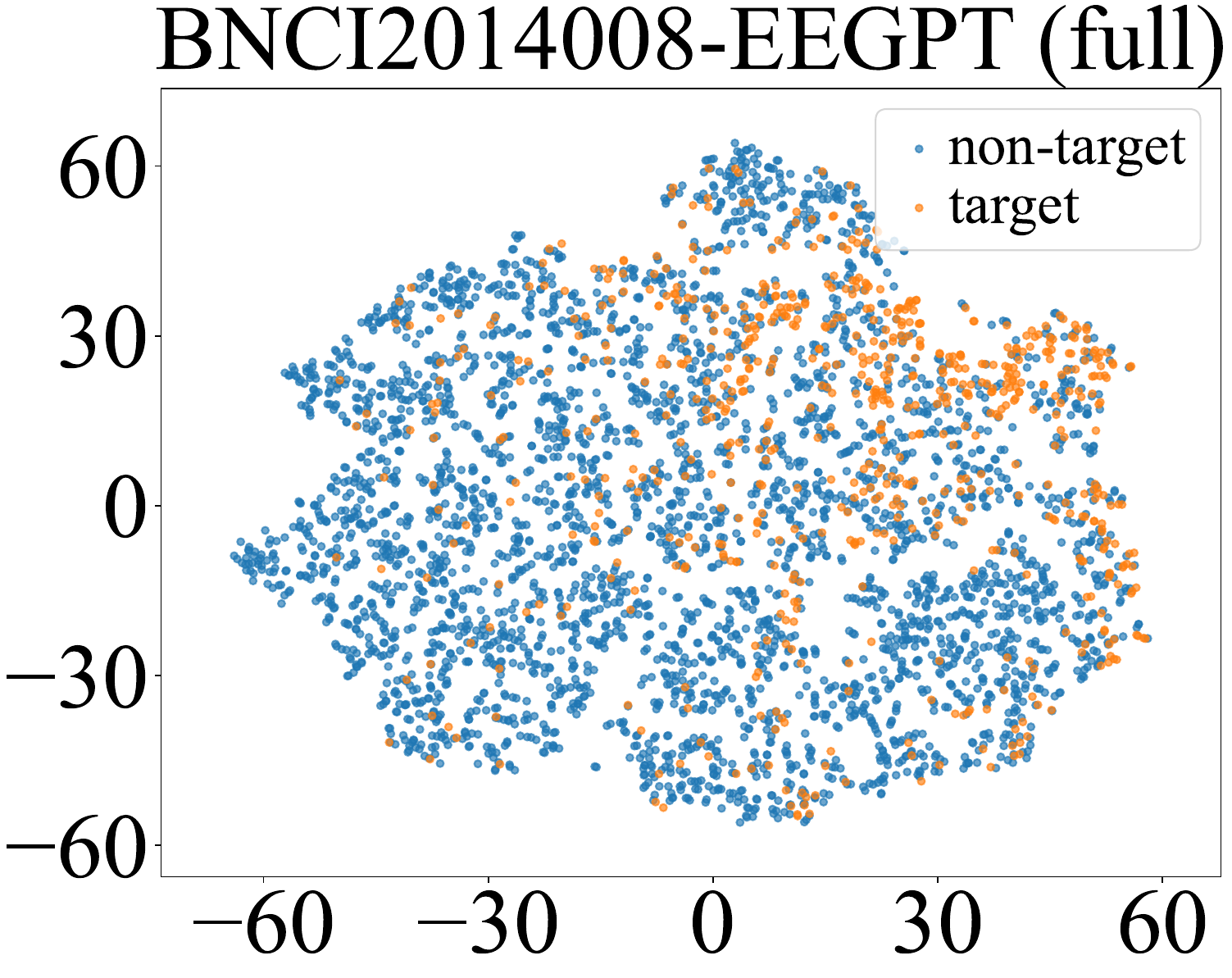}}
\subfigure[]{\includegraphics[width=0.19\linewidth,clip]{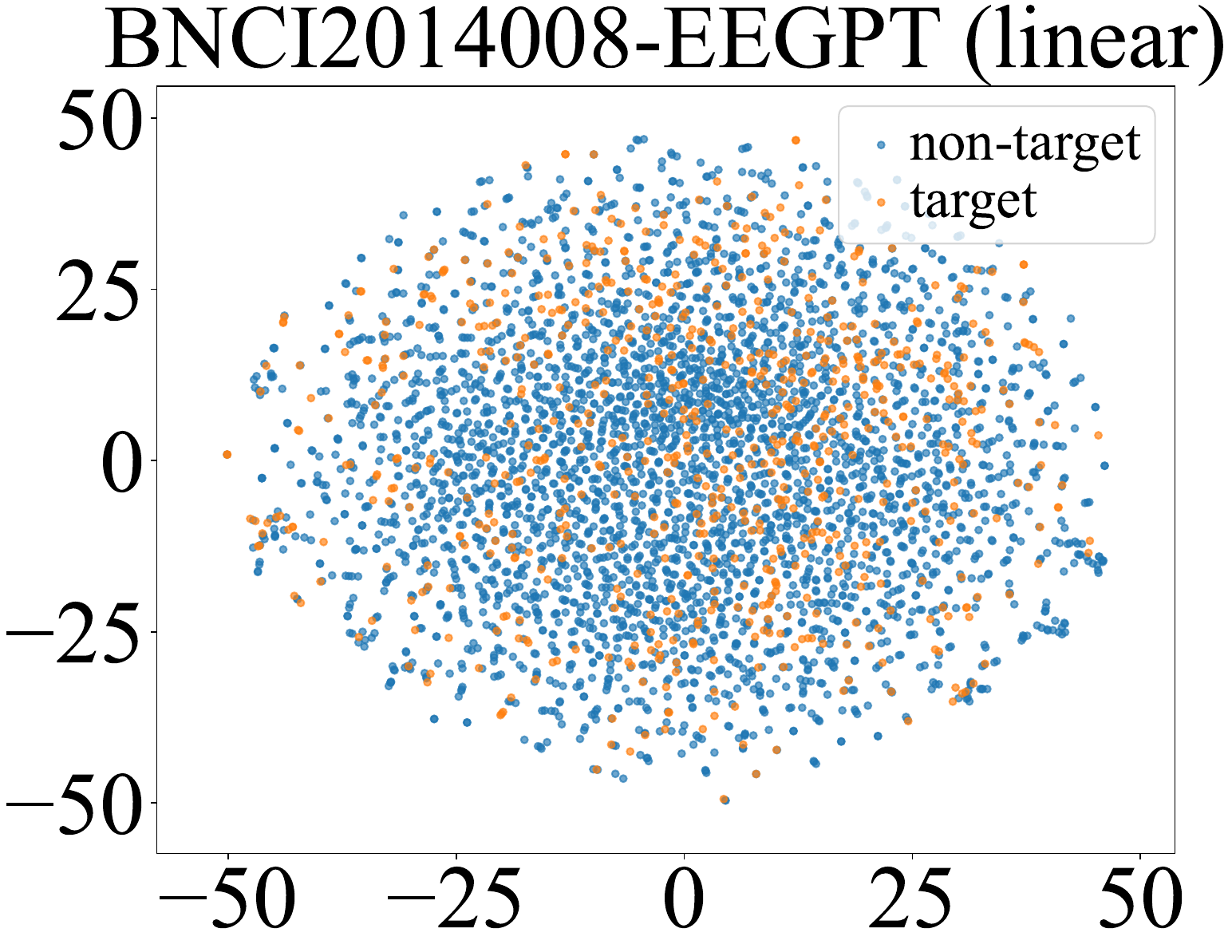}}
\subfigure[]{\includegraphics[width=0.19\linewidth,clip]{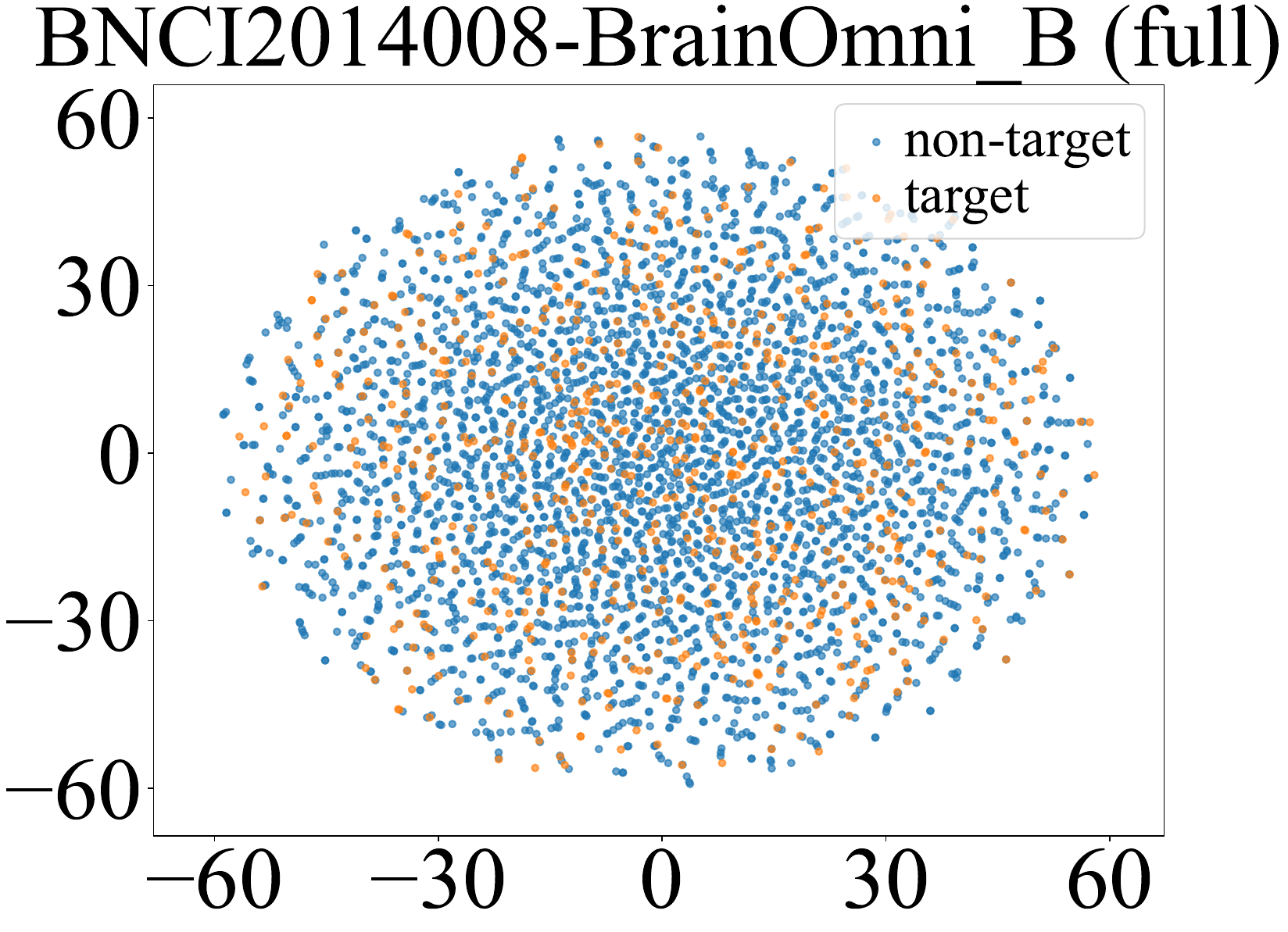}} \\
\subfigure[]{\includegraphics[width=0.19\linewidth,clip]{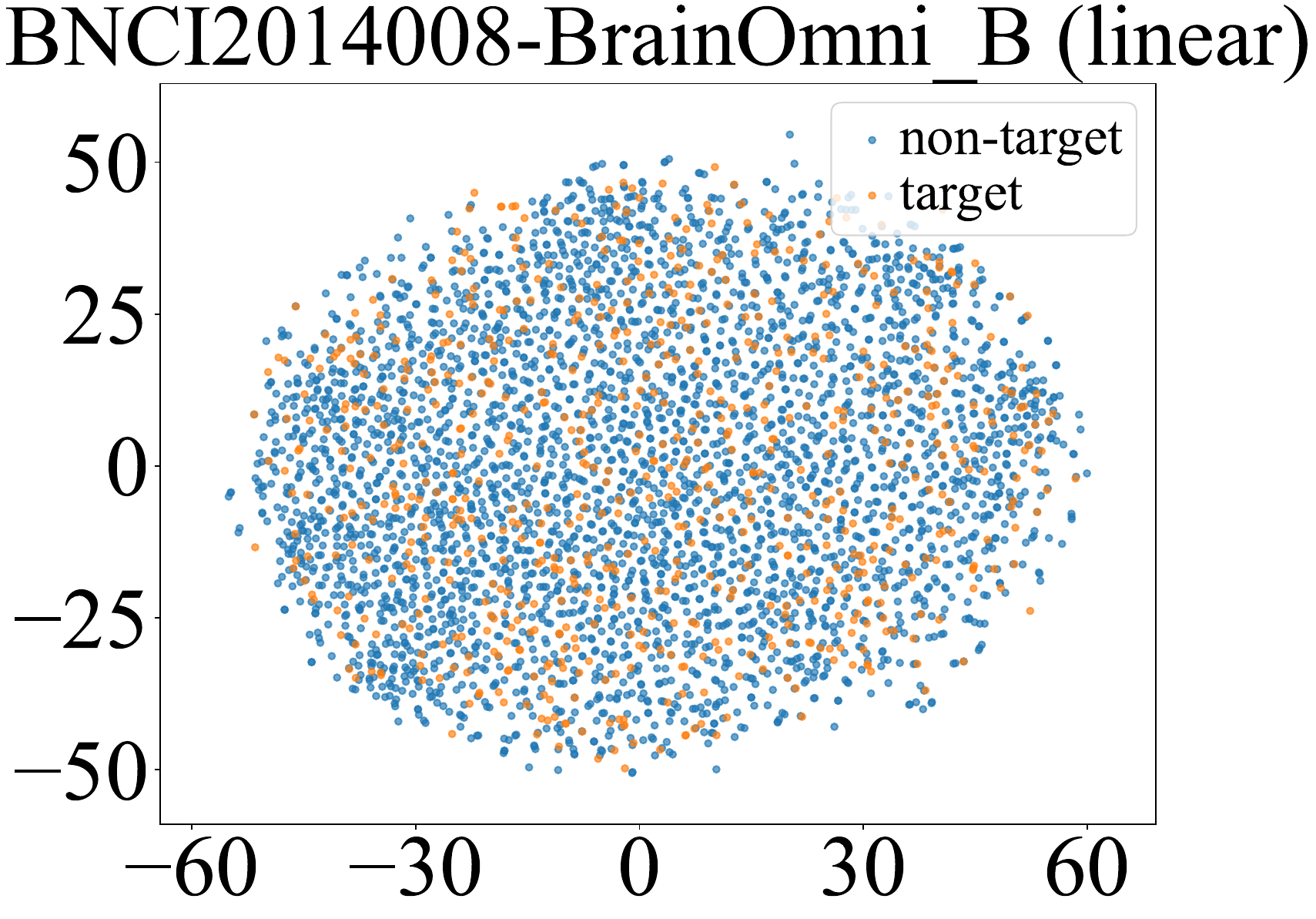}}
\subfigure[]{\includegraphics[width=0.19\linewidth,clip]{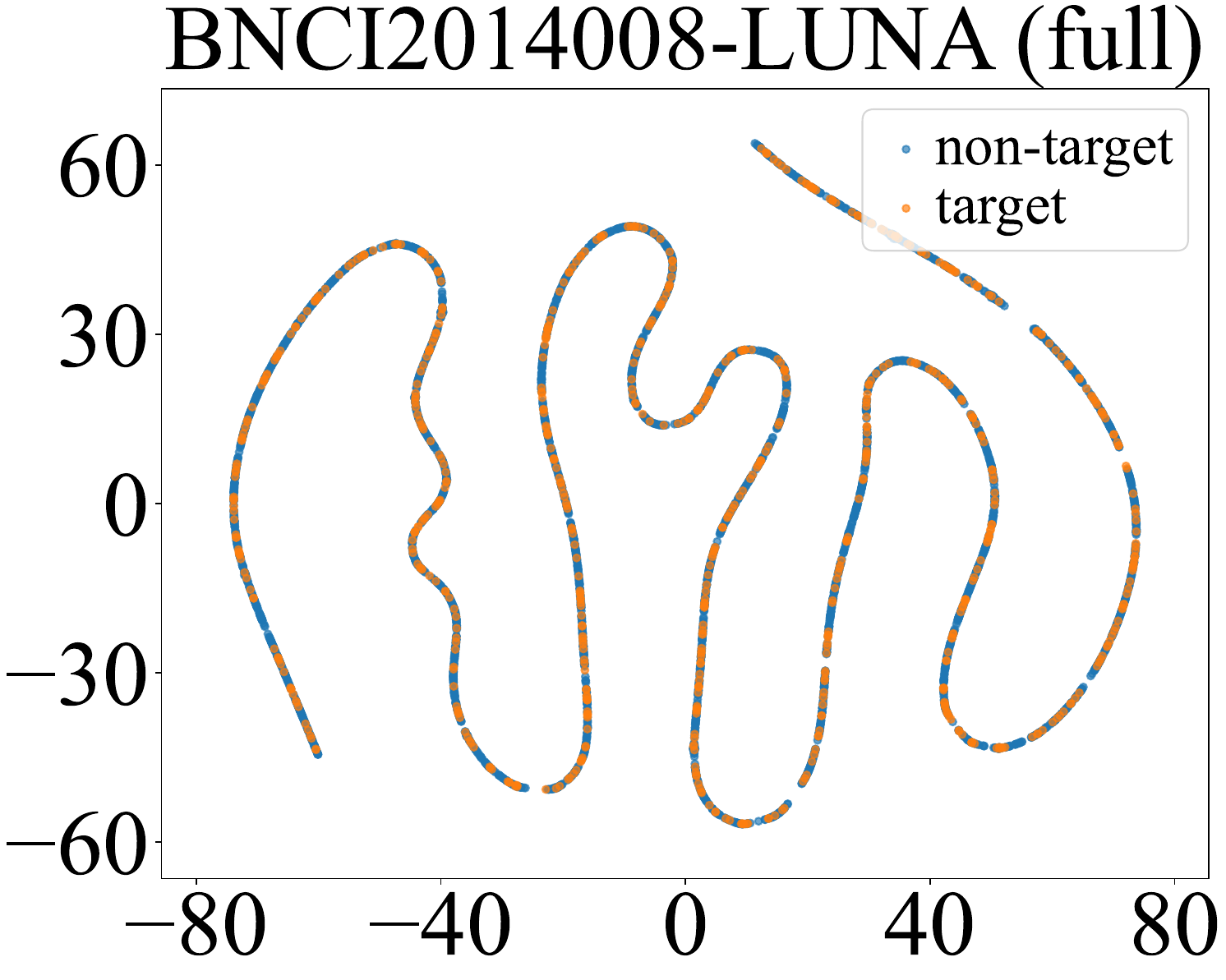}}
\subfigure[]{\includegraphics[width=0.19\linewidth,clip]{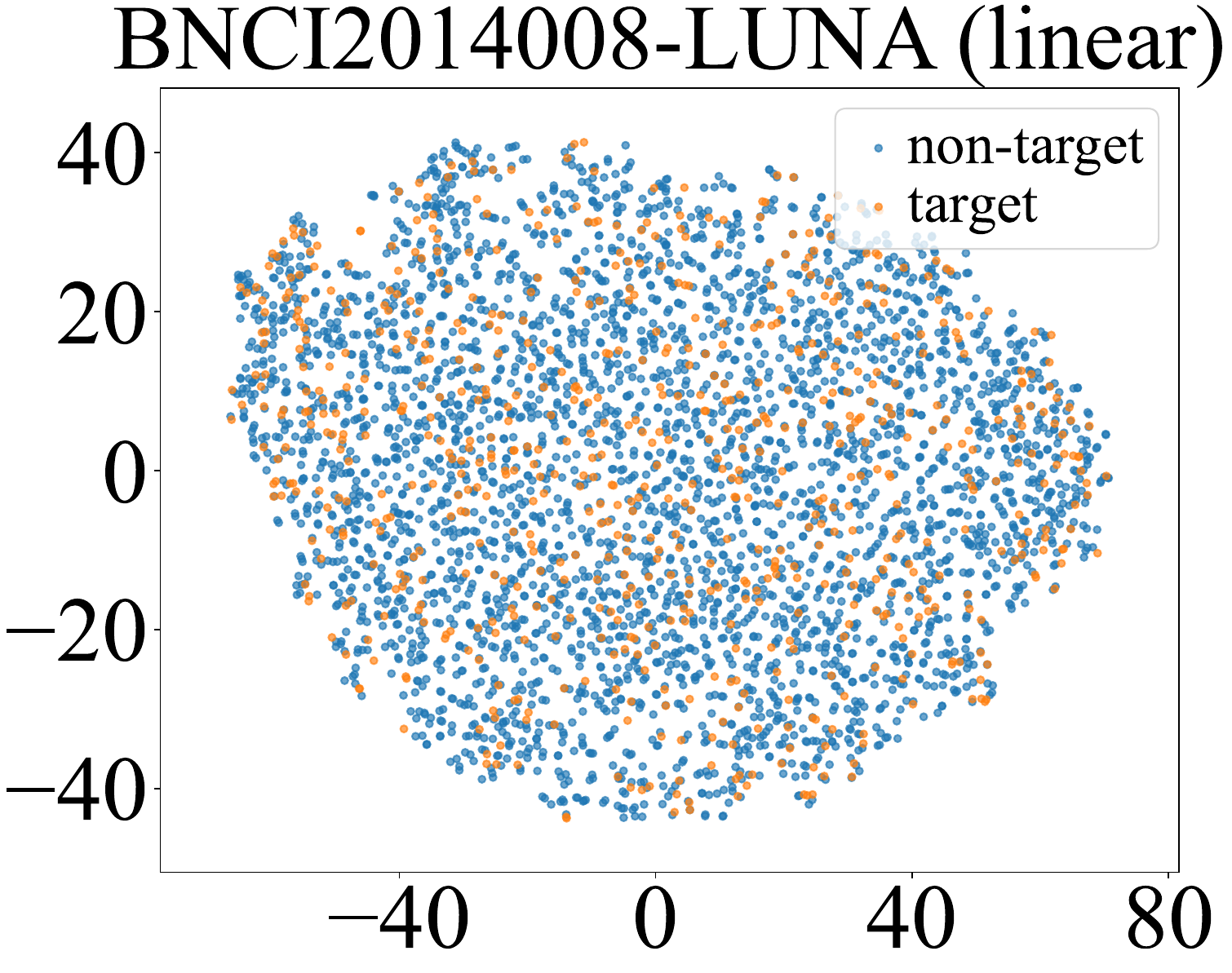}}
\subfigure[]{\includegraphics[width=0.19\linewidth,clip]{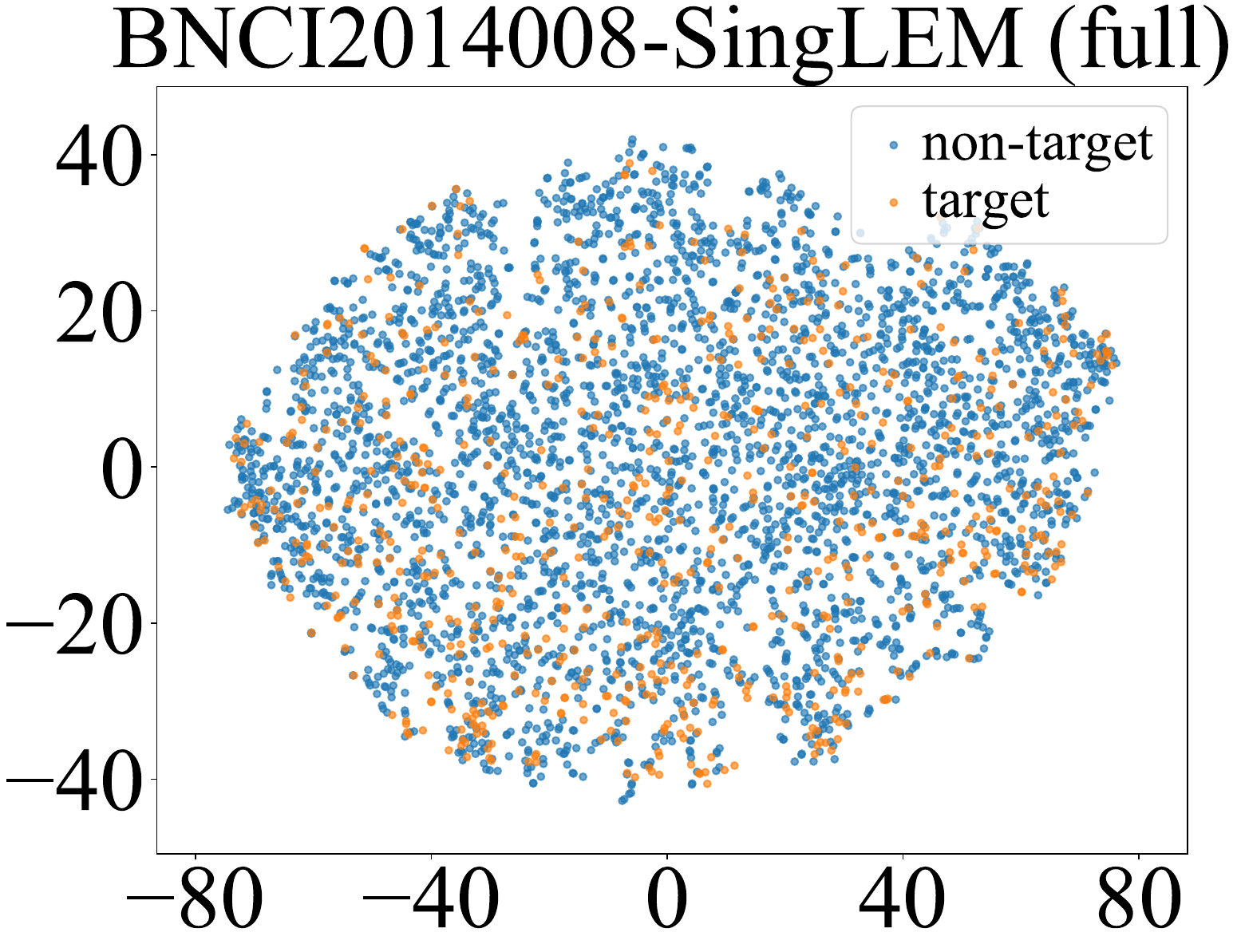}}
\subfigure[]{\includegraphics[width=0.19\linewidth,clip]{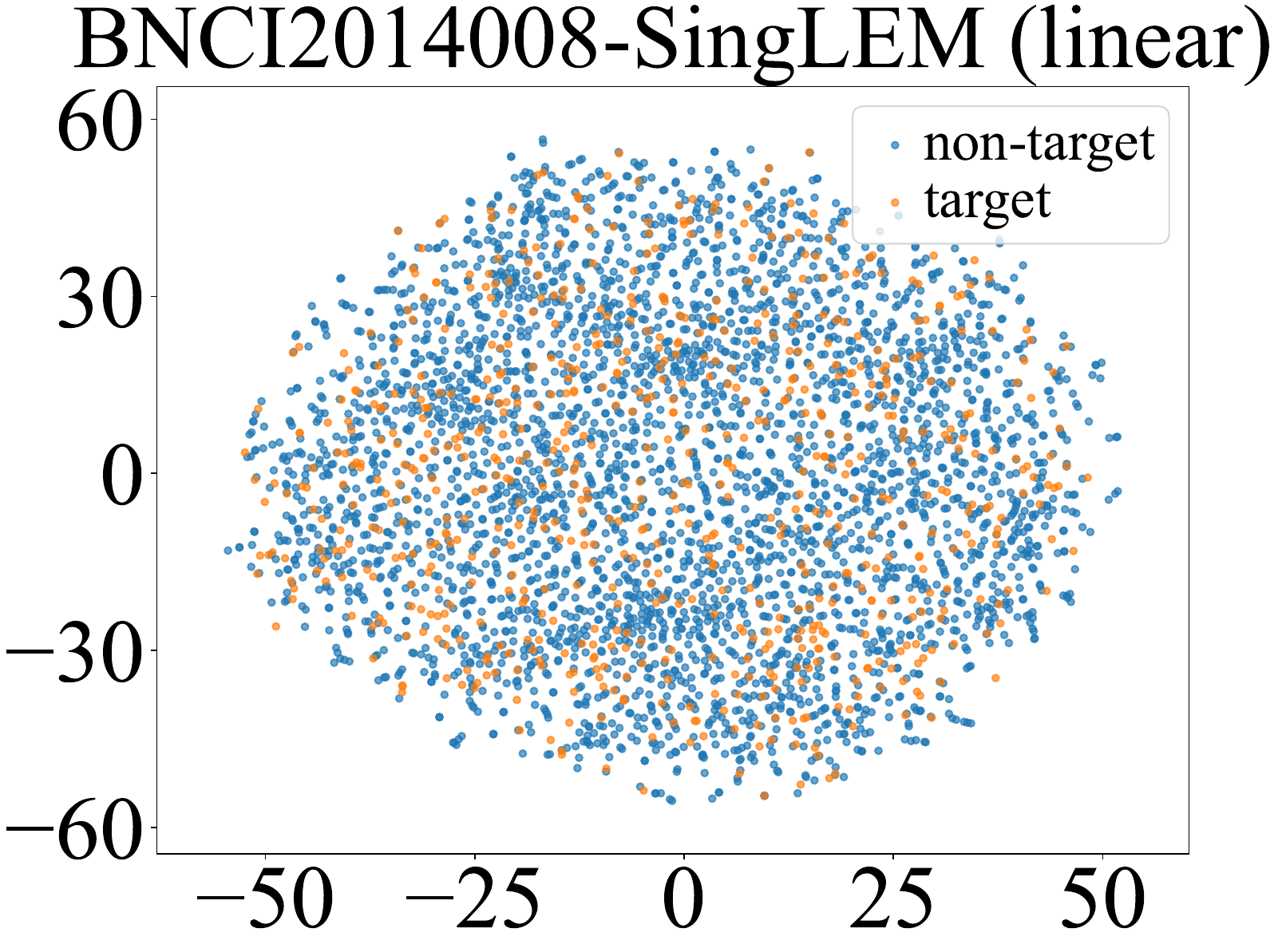}} \\
\caption{$t$-SNE visualization of the BNCI2014008 datasets.} \label{fig:tsne_rep3_appd}
\end{figure*}

Figs.~\ref{fig:tsne_rep1_appd}--\ref{fig:tsne_rep3_appd} provide the complete $t$-SNE visualizations of the learned representations on the SEED, TUAB, and BNCI2014008 datasets, respectively. For each dataset, we visualize seven specialist models and nine EEG FMs; for every FM, both full-parameter fine-tuning (full) and linear probing (linear) are shown.

\subsection{EXISTING SURVEYS AND BENCHMARKS}
\label{appendix:existing_work_comparison}

Table~\ref{tab:comparison_existing_works} compares our work with representative surveys, reviews, and benchmarks on EEG and brain foundation models published over the past two years. The comparison covers six aspects: the type of contribution, the number of EEG FMs included, the scope of benchmark experiments, the evaluation protocols, whether specialist baselines are included, and the main distinction of each work. Our work differs from prior efforts in two main respects. First, it surveys 55 EEG FMs, substantially more than any existing review or survey. Second, it provides a unified empirical benchmark of 12 open-source EEG FMs against 7 specialist baselines on 13 datasets covering 9 BCI paradigms, evaluated under both the LOSO and within-subject few-shot scenarios with both full-parameter fine-tuning and linear probing. To our knowledge, this is the largest EEG FM survey and the most comprehensive benchmark to date that systematically contrasts FMs with specialist baselines under unified evaluation protocols.

\begin{table*}[!t]
\centering
\renewcommand{\arraystretch}{1.0}
\caption{Comparison of our work with representative surveys, reviews, and benchmarks on EEG and brain foundation models. ``--'' indicates that the corresponding work does not provide benchmark experiments or the relevant component is not applicable.}
\label{tab:comparison_existing_works}
\scalebox{0.62}{
\begin{tabular}{c|c|c|c|c|c|c}
\toprule
\textbf{Work} & \textbf{Type} & \textbf{\# EEG FMs} & \makecell{\textbf{Benchmark}\\\textbf{coverage}} & \makecell{\textbf{Evaluation}\\\textbf{protocols}} & \makecell{\textbf{Specialist}\\\textbf{baselines}} & \makecell{\textbf{Main}\\\textbf{distinction}} \\
\midrule

\multirow{2}{*}{Lai et al.~\cite{lai2025simple}} & \multirow{2}{*}{Review} & \multirow{2}{*}{13} & \multirow{2}{*}{--} & \multirow{2}{*}{--} & \multirow{2}{*}{--} & Concise EEG-FM \\
& & & & & & review \\
\midrule

\multirow{2}{*}{Altaheri et al.~\cite{altaheri2025bridging}} & \multirow{2}{*}{Survey} & \multirow{2}{*}{27} & \multirow{2}{*}{--} & \multirow{2}{*}{--} & \multirow{2}{*}{--} & SSL-focused \\
& & & & & & brain-signal survey \\
\midrule

\multirow{2}{*}{Wu et al.~\cite{wu2025adabrain}} & \multirow{2}{*}{Benchmark} & \multirow{2}{*}{4} & 13 datasets; & cross-subj.; multi-subj.; & \multirow{2}{*}{Yes} & BCI transfer \\
& & & 7 BCI tasks & few-shot; full / linear & & benchmark \\
\midrule

\multirow{2}{*}{Zhou et al.~\cite{zhou2025brain}} & \multirow{2}{*}{Survey} & \multirow{2}{*}{15} & \multirow{2}{*}{--} & \multirow{2}{*}{--} & \multirow{2}{*}{--} & Broad BFM \\
& & & & & & framework \\
\midrule

\multirow{2}{*}{Kuruppu et al.~\cite{kuruppu2026eeg}} & \multirow{2}{*}{Review} & \multirow{2}{*}{10} & \multirow{2}{*}{--} & \multirow{2}{*}{--} & \multirow{2}{*}{--} & Early EEG-FM \\
& & & & & & critical review \\
\midrule

\multirow{2}{*}{Gu et al.~\cite{gu2025foundation}} & \multirow{2}{*}{Survey} & \multirow{2}{*}{8} & \multirow{2}{*}{--} & \multirow{2}{*}{--} & \multirow{2}{*}{--} & Biosignal-FM \\
& & & & & & roadmap \\
\midrule

\multirow{2}{*}{Xiong et al.~\cite{xiong2025eeg}} & \multirow{2}{*}{Benchmark} & \multirow{2}{*}{7} & 14 datasets; & full / frozen / LoRA; & \multirow{2}{*}{Partial} & Systematic EEG-FM \\
& & & 10 paradigms & single-/multi-task & & benchmark \\
\midrule

\multirow{2}{*}{Yao et al.~\cite{yao2025foundation}} & \multirow{2}{*}{Review} & \multirow{2}{*}{17} & 6 datasets; & Unified & \multirow{2}{*}{Yes} & EEG-FM progress \\
& & & 4 task categories & comparison & & review + comparison \\
\midrule

\multirow{2}{*}{Lee et al.~\cite{lee2025comprehensive}} & \multirow{2}{*}{Survey} & \multirow{2}{*}{29} & \multirow{2}{*}{--} & \multirow{2}{*}{--} & \multirow{2}{*}{--} & Comprehensive \\
& & & & & & biosignal-FM review \\
\midrule

\multirow{2}{*}{Kwon et al.~\cite{kwon2026foundation}} & \multirow{2}{*}{Review} & \multirow{2}{*}{3} & \multirow{2}{*}{--} & \multirow{2}{*}{--} & \multirow{2}{*}{--} & EEG-centered \\
& & & & & & perspective \\
\midrule

\multirow{2}{*}{Shen et al.~\cite{shen2026brain4fms}} & \multirow{2}{*}{Benchmark} & \multirow{2}{*}{15} & 18 EEG/iEEG datasets; & cross-subj.; & \multirow{2}{*}{Partial} & EEG / iEEG SSL \\
& & & 11 tasks & group-wise CV & & taxonomy + benchmark \\

\midrule

\multirow{2}{*}{\textbf{Ours}} & \multirow{2}{*}{\textbf{Survey + benchmark}} & \multirow{2}{*}{\textbf{55}} & \textbf{12 open-source FMs;} & \textbf{LOSO; few-shot;} & \multirow{2}{*}{\textbf{Yes}} & \textbf{Largest EEG-FM} \\
& & & \textbf{13 datasets; 9 paradigms} & \textbf{full / linear} & & \textbf{survey + benchmark} \\

\bottomrule
\end{tabular}
}
\end{table*}

\subsection{SYMBOLS AND NOTATION}
\label{appendix:symbols}

Tables~\ref{tab:symbols_main} and~\ref{tab:symbols_loss} summarize the principal mathematical symbols and loss functions used throughout this paper. Model-specific auxiliary losses that appear only in individual models are not exhaustively enumerated here; the complete list is provided in Table~\ref{tab:pretrain_info}.

\begin{table*}[!t]
\centering
\renewcommand{\arraystretch}{1.25}
\setlength{\tabcolsep}{4pt}
\caption{Main mathematical symbols used in the manuscript.}
\label{tab:symbols_main}
\scalebox{0.8}{
\begin{tabular}{c|>{\centering\arraybackslash}p{8.2cm}|>{\centering\arraybackslash}p{4.7cm}}
\toprule
\textbf{Symbol} & \textbf{Meaning} & \textbf{Usage} \\
\midrule

$X\in\mathbb R^{C\times T}$ & Raw EEG trial with $C$ channels and $T$ samples & EEG input representation \\

$\widetilde X=\mathcal G(X)$ & Preprocessed EEG signal & Data standardization \\

$\mathcal G(\cdot)$ & Preprocessing operator (e.g., $z$-score, CAR, EA, EMA) & Preprocessing pipeline \\

$C$ & Number of EEG channels & Input shape \\

$T$ & Number of temporal samples per trial & Input shape \\

$f_s$ & Sampling rate & Preprocessing \\

$M$ & Temporal patch length & Patch construction \\

$S$ & Patch stride & Patch construction \\

$O=M-S$ & Patch overlap & Patch construction \\

$N_p=\lfloor (T-M)/S \rfloor+1$ & Number of temporal patches & Patch construction \\

$\mathcal S(\cdot)$ & Patching operator & Tokenization \\

$P=\mathcal S(\widetilde X)$ & Patch tensor of shape $N_p\times C\times M$ & Pre-training input \\

$p_{k,c}$ & $k$-th patch of channel $c$ & Masked reconstruction \\

$\Omega$ & Set of masked patch indices & Masked modeling \\

$\mathcal M_x(\cdot)$ & Masking operator on raw-signal patches & Raw-signal reconstruction \\

$\mathcal M_z(\cdot)$ & Masking operator on embedded tokens & Token reconstruction \\

$P_{\mathrm{msk}}$ & Masked raw-signal patches & Masked reconstruction \\

$\mathcal E_{\theta}(\cdot)$ & Encoder with parameters $\theta$ & FM backbone \\

$\mathcal D_{\phi}(\cdot)$ & Decoder or reconstruction head with parameters $\phi$ & Pre-training head \\

$\mathcal T_{\psi}(\cdot)$ & Tokenizer or patch-embedding module with parameters $\psi$ & Tokenization \\

$Z=\mathcal T_{\psi}(P)\in\mathbb R^{N_p\times d}$ & Embedded EEG tokens & Token reconstruction / contrastive learning \\

$z_{k,c}$ & Embedded token corresponding to patch $(k,c)$ & Token-level objectives \\

$d$ & Token embedding dimension & Model architecture \\

$\widehat X,\widehat P,\widehat Z$ & Reconstructions of $X$, $P$, and $Z$ & Reconstruction objectives \\

$h_i,h_i^+$ & Anchor and positive representations & Contrastive learning \\

$\mathcal N_i$ & Negative set for anchor $h_i$ & Contrastive learning \\

$\operatorname{sim}(\cdot,\cdot)$ & Similarity function (e.g., cosine similarity) & Contrastive learning \\

$\tau$ & Temperature parameter & Contrastive learning \\

$\mathcal F_r(\cdot)$ & Spectral transform yielding target type $r$ & Frequency-domain reconstruction \\

$S^r=\mathcal F_r(P)$ & Spectral target of EEG patches & Frequency-domain reconstruction \\

$r$ & Spectral target type, $r\in\{\mathrm{spec},\mathrm{sa},\mathrm{bp},\mathrm{freq}\}$ & Frequency-domain reconstruction \\

$\mathcal Q(\cdot)$ & Quantizer mapping embeddings to codebook indices & Codebook objectives \\

$K$ & Codebook size & Codebook objectives \\

$I\in\{1,\ldots,K\}^{L}$ & Sequence of discrete codebook indices & Codebook prediction \\

$\widehat{\pi}_{k,c}$ & Predicted categorical distribution over codebook entries or pseudo-labels & Codebook / pseudo-label prediction \\

$u_{1:L}$ & Ordered token sequence for autoregressive modeling & Autoregressive pre-training \\

$L$ & Length of token or index sequence & Sequence modeling \\

$f_\Theta$ & Pre-trained EEG FM with parameters $\Theta$ & Model definition \\

$\Theta^\star$, $\Theta_j^\star$ & Pre-training optimum and task-$j$ fine-tuning optimum & Model optimization \\

$\mathcal D_{pre}$ & Pre-training corpus & Pre-training data \\

$\mathcal T=\{\tau_j\}_{j=1}^{J}$ & Set of downstream tasks; $\tau_j$ denotes task $j$ & Downstream evaluation \\

$\mathcal D_{task}^{(j)},\mathcal D_{ft}^{(j)},\mathcal D_{te}^{(j)}$ & Full, fine-tuning, and test splits of task $\tau_j$ & Downstream evaluation \\

$N_j,n_j$ & Total trials and fine-tuning trials in task $\tau_j$ & Downstream evaluation \\

$\mathcal C_j$ & Number of classes for task $\tau_j$ & Downstream classification \\

$y_i,\widehat y_i$ & Ground-truth label and predicted distribution for sample $i$ & Classification \\

$\mathbf 1(\cdot)$ & Indicator function & Evaluation metric \\

$B$ & Mini-batch size & Optimization \\

$\lambda$ & Weighting coefficient for multi-term losses & Hybrid objectives \\

\bottomrule
\end{tabular}
}
\end{table*}

\begin{table*}[!t]
\centering
\renewcommand{\arraystretch}{1.6}
\setlength{\tabcolsep}{4pt}
\caption{Common loss symbols used in the manuscript. Model-specific losses that occur only in individual models are summarized in Table~\ref{tab:pretrain_info}.}
\label{tab:symbols_loss}
\scalebox{0.9}{
\begin{tabular}{c|>{\centering\arraybackslash}p{8.4cm}|>{\centering\arraybackslash}p{4.2cm}}
\toprule
\textbf{Symbol} & \textbf{Meaning} & \textbf{Usage} \\
\midrule

$\mathcal L_{pre}$ & Overall pre-training objective, typically a weighted combination of reconstruction, prediction, contrastive, or auxiliary terms & General pre-training \\

$\mathcal L_{mse}^{rs}$ & Mean squared error for raw-signal reconstruction & Raw EEG waveform \\

$\mathcal L_{mse}^{nrs}$ & Mean squared error for denoised raw-signal reconstruction & Denoised EEG waveform \\

$\mathcal L_{s\text{-}l1}^{rs},\;\mathcal L_{huber}^{rs}$ & Smooth-$\ell_1$ or Huber-style robust reconstruction loss for raw signals & Robust waveform reconstruction \\

$\mathcal L_{mae}^{os}$ & Masked-autoencoding reconstruction loss on original-signal patches & Used by SAMBA \\

$\mathcal L_{mse}^{et}$ & Mean squared error for embedded-token reconstruction & Embedded EEG tokens \\

$\mathcal L_{mse}^{emb}$ & Mean squared error for latent-embedding reconstruction (equivalent to $\mathcal L_{mse}^{et}$ in some models, e.g., EEGPT) & Latent embeddings \\

$\mathcal L_{ce}$ & Standard cross-entropy loss & Building block for $\mathcal L_{cls}^{ci}$, $\mathcal L_{cls}^{skl}$, etc. \\

$\mathcal L_{cl}$ & Generic contrastive loss, typically InfoNCE-style & Representation alignment \\

$\mathcal L_{cl}^{et}$ & Contrastive loss on embedded EEG tokens & Token-level contrastive \\

$\mathcal L_{cl}^{emb}$ & Contrastive loss on latent embeddings & Embedding-level contrastive \\

$\mathcal L_{cl}^{glob}$ & Global contrastive loss for discriminative structure & Global representation \\

$\mathcal L_{mse}^{spec}$ & Mean squared error for spectrogram (time-frequency) reconstruction & Spectrogram target \\

$\mathcal L_{mse}^{sa}$ & Mean squared error for spectral-amplitude reconstruction & Spectral amplitude \\

$\mathcal L_{mse}^{bp}$ & Mean squared error for band-power reconstruction & Band-power features \\

$\mathcal L_{mse}^{freq}$ & Generic frequency-domain reconstruction loss & Fourier-domain target \\

$\mathcal L_{\ell_1}^{stft},\;\mathcal L_{\ell_2}^{stft}$ & Multi-scale STFT $\ell_1$ / squared-$\ell_2$ reconstruction loss & Time-frequency codec \\

$\mathcal L_{cls}^{skl}$ & Cross-entropy loss for spectrogram-derived $k$-means pseudo-label prediction & Spectral pseudo-labels \\

$\mathcal L_{cls}^{ci}$ & Cross-entropy loss for masked codebook-index prediction & Discrete neural codes \\

$\mathcal L_{nll}^{ci}$ & Negative log-likelihood for autoregressive codebook-index prediction & Autoregressive neural codes \\

$\mathcal L_{mse}^{cbe}$ & Mean squared error for codebook-embedding reconstruction & Codebook embeddings \\

$\mathcal L_{cls}$ & Supervised classification loss; task-specific instances denoted $\mathcal L_{cls}^{(j)}$ & Downstream classification \\

$\mathcal L_{aux}$ & Model-specific auxiliary regularization (alignment, MoE balancing, decomposition, or consistency-style terms) & Auxiliary objectives \\

\bottomrule
\end{tabular}
}
\end{table*}

\end{appendix}


\bibliographystyle{nsr}
\bibliography{nsr_sample}

\end{document}